\documentclass{article}
\newcommand{\arxivroot}{}
\usepackage{preprint}
\usepackage{times}

\usepackage[utf8]{inputenc}
\usepackage[T1]{fontenc}
\usepackage[breaklinks,colorlinks,
            linkcolor=black,
            anchorcolor=blue,
            citecolor=blue]{hyperref}
\usepackage{url}
\usepackage{booktabs}
\usepackage{amsfonts}
\usepackage{amssymb}
\usepackage{amsmath}
\usepackage{amsthm}
\usepackage{bm}
\usepackage{dsfont}
\usepackage{nicefrac}
\usepackage{microtype}
\usepackage{xcolor}
\usepackage{graphicx}
\usepackage{wrapfig}
\usepackage{subcaption}
\usepackage{multirow}
\usepackage{xfrac}
\usepackage[nameinlink,capitalize]{cleveref}

\newtheorem{theorem}{Theorem}[section]

\newtheorem{lemma}{Lemma}[section]

\newtheorem{assumption}{Assumption}[section]
\newtheorem{condition}{Condition}[section]

\crefname{theorem}{Theorem}{Theorems}
\crefname{lemma}{Lemma}{Lemmas}
\crefname{proposition}{Proposition}{Propositions}
\crefname{remark}{Remark}{Remarks}
\crefname{corollary}{Corollary}{Corollaries}
\crefname{definition}{Definition}{Definitions}
\crefname{assumption}{Assumption}{Assumptions}
\crefname{condition}{Condition}{Conditions}
\crefname{example}{Example}{Examples}
\newcommand{\E}{\mathbb{E}}
\newcommand{\R}{\mathbb{R}}
\newcommand{\cL}{\mathcal{L}}

\usepackage{amsmath,amsfonts,bm}

\def\1{\bm{1}}

\def\vtheta{{\bm{\theta}}}


\def\vx{{\bm{x}}}

\def\mB{{\bm{B}}}

\def\mG{{\bm{G}}}
\def\mH{{\bm{H}}}

\def\mM{{\bm{M}}}

\def\mSigma{{\bm{\Sigma}}}

\DeclareMathAlphabet{\mathsfit}{\encodingdefault}{\sfdefault}{m}{sl}
\SetMathAlphabet{\mathsfit}{bold}{\encodingdefault}{\sfdefault}{bx}{n}

\usepackage{enumitem}

\newcommand{\cD}{\mathcal{D}}

\newcommand{\cF}{\mathcal{F}}

\newcommand{\cJ}{\mathcal{J}}

\newcommand{\cN}{\mathcal{N}}

\newcommand{\half}{\frac{1}{2}}

\newcommand{\diag}{\mathrm{diag}}

\renewcommand{\d}{\mathop{}\!\mathrm{d}}

\newcommand{\bias}{\mathrm{bias}}
\newcommand{\var}{\mathrm{var}}
\title{Data Schedules and Learning Rate Schedules \\ Should Be Designed Together}
\author{%
Haodong Wen\textsuperscript{1,}\thanks{Equal contribution.}\quad
Kairong Luo\textsuperscript{1,*}\quad
Tingkai Yan\textsuperscript{1,*}\quad
Wenguang Chen\textsuperscript{1,2}\quad
Kaifeng Lyu\textsuperscript{1}\\
{\normalfont\textsuperscript{1}Tsinghua University}\\
{\normalfont\textsuperscript{2}Peng Cheng Laboratory}\\
{\normalfont\ttfamily\{whd25,luokr24,ytk26\}@mails.tsinghua.edu.cn}\\
{\normalfont\ttfamily\{cwg,klyu\}@mail.tsinghua.edu.cn}
}
\hypersetup{
  pdftitle={Data Schedules and Learning Rate Schedules Should Be Designed Together},
  pdfauthor={Haodong Wen, Kairong Luo, Tingkai Yan, Wenguang Chen, Kaifeng Lyu}
}
\begin{document}

\maketitle
\fancyhead{}
\renewcommand{\headrulewidth}{0pt}

\begin{abstract}
Mid-training (MT) is a training phase in LLM training pipelines that continues pretraining (PT) on smaller datasets of higher value.
Taking the PT and MT phases as a whole, training usually starts with only PT data and then transitions completely to MT data.
In this work, we view this complete transition from PT to MT data as a special case of a broader class of data mixing schedules that specify how the proportions of PT and MT tokens in each batch change over training.
Under fixed total training compute and MT token budgets, we ask: \textit{Is this complete transition to MT data optimal for minimizing final MT loss?}
We show that the answer depends on the learning rate (LR) schedule: one data schedule may outperform another under one LR schedule, but changing only the LR schedule can reverse the outcome.
Specifically, when the LR decays throughout continued training, introducing MT data earlier and mixing it with PT data lowers the final MT loss compared with placing all MT data at the end of training, but this result is reversed if we use constant LR.
We therefore study how to jointly choose when to introduce MT data and when to decay the LR through ablation studies, and find that it is better to decay the LR only after MT data enters training.
Finally, we develop a theoretical model that explains why the optimal data schedule for a fixed MT token budget depends on the LR schedule.
\end{abstract}

\section{Introduction}
\label{sec:introduction}
Pretraining (PT) and mid-training (MT) are two consecutive phases in large language model (LLM) training pipelines. PT uses broad web corpora, while MT uses smaller, curated mixtures enriched with mathematics, code, and other targeted data~\citep{deepseekv3tech,qwen3tech,olmo2furious,phi4tech,glm45arc,mimo2025,liu2025k2}. In the simplest form of this pipeline, training starts with PT data alone and then transitions completely to MT data.

However, reserving all MT tokens for the final phase is not the only way to allocate the MT token budget over training. From a data mixing perspective, the two phases described above form a particular \textit{data schedule}. Let $\gamma(t)$ denote the fraction of MT tokens in the batch at step $t$, which we call the \textit{data schedule function}. The complete transition to MT data as described above corresponds to $\gamma(t)=\1\{t>\tau_0\}$, where $\tau_0$ is the transition time. In general, different allocations of MT tokens over training correspond to different data schedules $\gamma(t)$. This motivates a natural question: \emph{with the total training compute and MT token budget fixed, is this complete transition to MT data optimal for minimizing final MT loss?}

\begin{figure}[t]

    \centering
    \captionsetup{skip=3pt}
    \captionsetup[subfigure]{skip=1pt}
    \begin{subfigure}[t]{\textwidth}
        \centering
        \includegraphics[width=\linewidth]{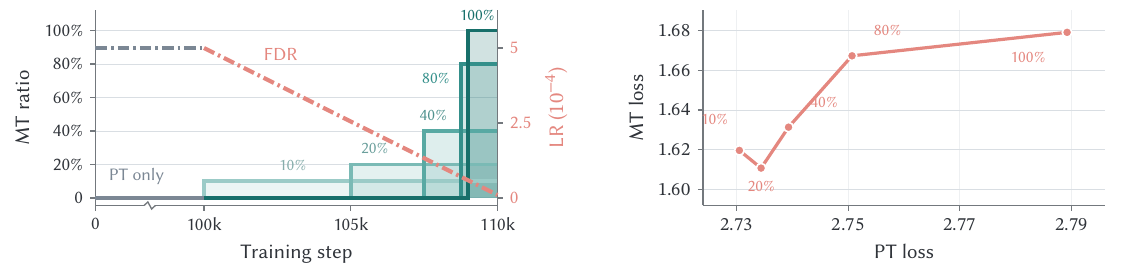}
        \caption{\textbf{Left:} Linear Decay LR $\&$ \textbf{T}wo-\textbf{S}tage (TS) data schedule. \textbf{Right:} PT-MT losses under different \textbf{T}wo-\textbf{S}tage (TS) data schedules with FDR.}
        \label{fig:intro-fdr}
    \end{subfigure}

    \par

    \begin{subfigure}[t]{\textwidth}
        \centering
        \includegraphics[width=\linewidth]{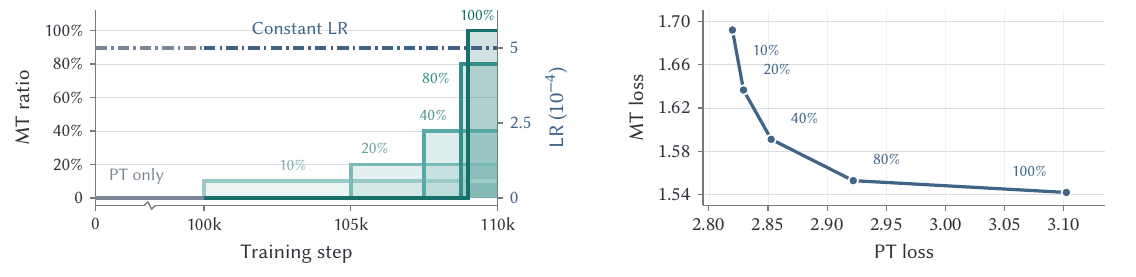}
        \caption{\textbf{Left:} Constant LR $\&$ \textbf{T}wo-\textbf{S}tage (TS) data schedule. \textbf{Right:} PT-MT losses under different \textbf{T}wo-\textbf{S}tage (TS) data schedules with constant LR.}
        \label{fig:intro-const}
    \end{subfigure}
     \caption{\textbf{Replaying PT data can be both beneficial and harmful under different LR schedules.} The compute and MT data budgets are fixed (the shaded areas in the left panels show the MT budgets). With \textbf{F}ull LR \textbf{D}ecay \textbf{R}ange throughout continued training (FDR, top), mixing PT and MT data in the second TS stage gives lower final MT loss than using MT data alone. With \textbf{constant LR} (Constant LR, bottom), the pure MT stage instead gives the lowest tested MT loss. Results use the dense model at a 10\% MT budget. Training-step coordinates in the schedule panels are schematic: the PT checkpoint is drawn at 100k steps, rather than its actual step 95,368.}
    \label{fig:intro-fdr-const}

\end{figure}

We study this question by comparing two schedule families that introduce MT data either abruptly or gradually.
Both schedule families begin with PT data alone and end at a chosen MT ratio $\gamma_{\mathrm{end}}\in(0,1]$, but they differ in how MT data enters training.
The \emph{Two-Stage (TS)} schedule jumps directly to this ratio and holds it fixed until training ends. The \emph{Zero-then-Linear (ZL)} schedule instead increases the ratio gradually:
\[
    \gamma(t) =
    \begin{cases}
        0, & 0 \le t \le \tau_1, \\
        \gamma_{\mathrm{end}}\dfrac{t-\tau_1}{T-\tau_1}, & \tau_1 < t \le T,
    \end{cases}
    \qquad 0 \le \tau_1 < T.
\]
Here, $T$ is the number of steps in the continued training period after the initial PT checkpoint, and $\tau_1$ is the time when MT data first enters training. At a fixed batch size, the area $\int_0^T\gamma(t)\,\mathrm{d}t$ is proportional to the number of MT tokens used. We hold this area fixed across schedules to compare different ways of allocating the same MT budget over time.

We first revisit the debate over whether we should replay PT data during continued training or transition to MT data alone.
At first glance,
introducing MT data earlier while replaying PT data
could produce larger updates while the learning rate (LR) is still high, but it also leaves more subsequent training that may weaken its effect. This tradeoff suggests that whether replaying PT data is beneficial may depend on the LR schedule, and we test this by comparing the same set of data schedules under different LR schedules.

We start by considering LR decay throughout the continued training period. Our results support the findings of \citet{kotha2026replaying}, who found that mixing PT and MT data in the second stage lowers final MT loss compared with using MT data alone under the same compute budget. In our experiments, we fix both the total training compute and MT token budget. Mixing PT data into the second stage therefore requires introducing MT data earlier and spreading it over more steps. We find that this earlier mixing yields lower final MT loss than placing all MT data in a final pure MT stage, consistently across dense and MoE models, multiple MT budgets, and different lengths of continued training (\cref{fig:exp-reversal-budgets,fig:exp-dense-moe}).

However, we find that \textbf{replaying PT data may not be beneficial if we change the LR schedule.}
Specifically, holding the data schedules and all other settings fixed, if we replace LR decay with a constant LR, then the same PT replay now increases the final MT loss. Among TS schedules, placing all MT data in a final stage containing only MT data gives the lowest tested MT loss. Figure~\ref{fig:intro-fdr-const} illustrates this phenomenon at a 10\% MT data budget, showing the data and LR schedules alongside their resulting PT and MT losses. The reversal persists across all our experimental setups (\cref{sec:empirical}). These results show that the preferred data schedule cannot be determined independently of the LR schedule.  In general, we identify the following \textit{ranking reversal phenomenon} of data schedules: the ranking of data schedules by final MT loss can shift or even reverse under different LR schedules.

The ranking reversal phenomenon above suggests that \textbf{the data and LR schedules should be designed together.} We therefore seek joint scheduling of LR and data
that minimizes final MT loss under fixed compute and MT token budgets. To explore what the optimal joint schedule of LR and data looks like, we hold each data schedule fixed and vary when LR decay begins relative to the time when MT data enters training, which we call the MT entry time.

We find that \textbf{LR should decay after MT data enters training.} Across all $19$ sweeps for different LR and data schedules, the best tested data and LR schedule pairs follow the above pattern and achieve lower final MT loss than constant LR or decay beginning at MT entry (\cref{sec:exp-decay}). For the dense model, the best tested LR decay covers the last $40\%$ of the MT data window for TS and the last $20\%$ for ZL (\cref{fig:delayed-lr-decay}). See Appendix~\ref{app:exp-decay} for the complete dense and MoE sweeps.

Finally, \textbf{we theoretically explain why the optimal data schedule depends on the LR schedule.} We develop a tractable model of SGD on noisy quadratic objectives and optimize the data schedule under a fixed MT budget. In this model, placing all MT data at the end is optimal under a constant LR, but can lose optimality under sufficiently strong linear decay. Under additional spectral assumptions and for sufficiently long training, we derive a unique optimum that begins with PT data, passes through a mixture of PT and MT data, and ends with MT data (\cref{sec:theory}).

\section{Data Schedules and Learning Rate Schedules Should Be Designed Together}
\label{sec:empirical}

In this section, we first reproduce the counterintuitive finding, inspired by \citet{kotha2026replaying}, that introducing MT data earlier and replaying PT data later during mid-training can improve MT loss. We then diagnose the role of the learning rate (LR) in this phenomenon. In particular, when using a constant LR during mid-training, placing MT data later instead leads to better MT loss. This observation aligns with \citet{luo2026bestdata} and motivates us to design data and LR schedules jointly. We conduct over $250$ runs across dense and MoE models. Across our LR decay sweeps, we obtain the lowest final MT loss by keeping the LR high after MT data enters and decaying only near the end of training (\cref{sec:exp-decay}).

\subsection{Experimental Setup}
\label{sec:exp-setup}

In the following experiments, we use DCLM-Baseline as the PT data~\citep{exp_dclm2024} and the finemath-4plus subset of FineMath as the MT data~\citep{exp_smollm2025} to construct our data schedules. We conduct the data schedule experiments on both dense and MoE models. Our dense model has 630M parameters and mainly follows the Qwen2.5-0.5B architecture, but uses untied input and output embeddings~\citep{exp_qwen252024}. It is first pretrained from scratch on $200$B DCLM tokens during the PT phase and then trained on another $20.97$B tokens during the MT phase. In the MT phase, MT tokens account for $10\%$, $25\%$, or $47.75\%$ of the total tokens ($2.10$B, $5.24$B, or $10.01$B MT tokens), while replayed PT tokens make up the remainder. Since FineMath contains approximately $10$B tokens in total, the MT data is consumed for at most one epoch. Our MoE model follows the OLMoE design~\citep{exp_olmoe2024}, with $2.99$B total parameters, $64$ experts per layer, and top-$8$ routing. The PT phase consists of $40$B PT tokens, while the MT phase contains either $4$B or $10$B tokens in two experimental settings. More setup details are provided in Appendix~\ref{app:exp-setup}.

\paragraph{Data and LR schedule configurations.}
To study the interplay between data and LR schedules during MT, we consider two representative data schedules under a fixed MT token budget.
Specifically, let $T$ denote the number of continual pretraining steps, with $t=0$ corresponding to a fixed PT checkpoint. Let $\gamma(t)$ be the fraction of MT tokens in the training batch at training step $t$. We fix the overall MT token fraction $\alpha$ and the batch size, so the budget constraint on MT tokens is $\int_0^T\gamma(t)\,\mathrm{d}t=\alpha T$.

\textbf{Two Stage (TS)} uses only PT data until $\tau_{\mathrm{TS}}$, then switches to a fixed mixture with MT ratio $\gamma_{\mathrm{end}}$:
\[
\gamma_{\scriptscriptstyle\mathrm{TS}}(t)=
\begin{cases}
0, & 0\le t<\tau_{\mathrm{TS}},\\
\gamma_{\mathrm{end}}, & \tau_{\mathrm{TS}}\le t\le T,
\end{cases}
\qquad
\tau_{\mathrm{TS}}=T\left(1-\frac{\alpha}{\gamma_{\mathrm{end}}}\right).
\]
\textbf{Zero-then-Linear (ZL)} uses only PT data until $\tau_{\mathrm{ZL}}$, then increases the MT ratio linearly from zero to $\gamma_{\mathrm{end}}$:
\[
\gamma_{\scriptscriptstyle\mathrm{ZL}}(t)=
\begin{cases}
0, & 0\le t<\tau_{\mathrm{ZL}},\\
\gamma_{\mathrm{end}}\dfrac{t-\tau_{\mathrm{ZL}}}{T-\tau_{\mathrm{ZL}}}, & \tau_{\mathrm{ZL}}\le t\le T,
\end{cases}
\qquad
\tau_{\mathrm{ZL}}=T\left(1-\frac{2\alpha}{\gamma_{\mathrm{end}}}\right).
\]
The budget requires $\gamma_{\mathrm{end}}\ge\alpha$ for TS and $\gamma_{\mathrm{end}}\ge2\alpha$ for ZL.

When comparing different data schedules, we consider three training regimes: \textbf{Const LR regime}, where the peak LR remains constant throughout MT; \textbf{Full LR Decay Range (FDR) regime}, where the LR decays throughout the entire continual training period regardless of the data schedule; \textbf{Matching the Decay Range (MDR) regime}, where the LR decay starts when MT data is introduced and reaches the final LR at the end of MT. Figure~\ref{fig:exp-schedule-pairs} illustrates all six combinations of LR and data schedules. Appendix~\ref{app:exp-setup} gives the exact definitions and experimental details. Specifically, in this paper, all the learning rate decays are linear.

\begin{figure}[t]

    \centering
    \includegraphics[width=\linewidth]{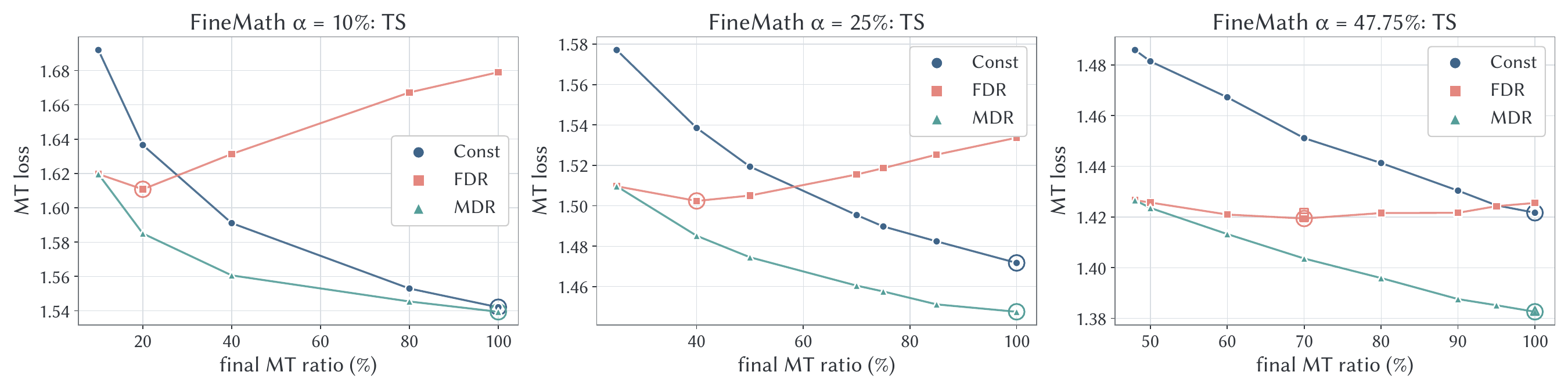}
    \caption{\textbf{Across all three MT budgets, replaying PT data improves MT loss under FDR, but not under Const or MDR.} The three panels show final MT loss as the final MT ratio in the second TS stage varies. Under the FDR regime, the lowest tested loss occurs when this stage mixes PT and MT data. Under the Const and MDR regimes, it occurs with pure MT data (100\%). Large circles mark these minima, while other hollow markers show repeated runs. See Section~\ref{sec:exp-reversal} and Appendix~\ref{app:exp-dense} for the setup and full results.}
    \label{fig:empirical}

    \label{fig:exp-reversal-budgets}
\end{figure}

\subsection{Different LR Schedules Prefer Different Data Schedules}
\label{sec:exp-reversal}

In practice, a TS data schedule typically sets $\gamma_{\mathrm{end}} = 1$, which places all MT data at the end of training. A more general TS, however, allows $\gamma_{\mathrm{end}} < 1$, which replays some PT data in the second stage. In the following paragraphs, we show that replaying PT data can improve MT loss in the FDR regime, but that this advantage disappears in the Const LR regime. The above ranking reversal phenomenon indicates that the timing of LR decay is crucial for MT loss, and that the benefit of a data schedule cannot be assessed independently of the LR schedule.

\paragraph{Replaying PT data can improve MT loss in the FDR regime.} Modern training pipelines commonly place most MT data late in training to make better use of it~\citep{deepseekv3tech,qwen3tech,olmo2furious,phi4tech}. However, some recent work shows that retaining PT data in later stages can improve MT loss~\citep{kotha2026replaying}. We observe that this counterintuitive pattern also appears across our experimental setups. As shown in \Cref{fig:intro-fdr}, we adopt a TS data schedule and vary $\gamma_{\mathrm{end}}$. As $\gamma_{\mathrm{end}}$ increases, more MT data is placed later, and the range over which MT data is used narrows. We find that both PT loss and MT loss generally increase with $\gamma_{\mathrm{end}}$. A similar trend also holds in \Cref{fig:app-dense10-grid,fig:app-dense25-grid,fig:app-dense4775-grid,fig:exp-dense-moe}, across architectures and MT budgets. These results confirm that the observation in prior work holds across a range of settings~\citep{kotha2026replaying}.

\paragraph{Replaying PT data is not beneficial in the Const LR regime.}
As shown in \Cref{fig:intro-const}, the advantage is reversed when we use a constant LR with the same type of TS data schedule. As $\gamma_{\mathrm{end}}$ increases, MT loss gradually decreases while PT loss gradually increases. This observation reveals that small learning rates in the later steps of training lead to underutilization of MT data, consistent with the observation in prior work on data curricula~\citep{luo2026bestdata}. These ranking reversal phenomena are also observed in \Cref{fig:exp-dense-moe,fig:app-dense10-grid,fig:app-dense25-grid}, across architectures and MT budgets; details are given in Appendices~\ref{app:exp-dense} and~\ref{app:exp-moe}. Therefore, different LR schedules favor different data schedules, and the benefit of a data schedule cannot be separated from the LR schedule.

\paragraph {The timing of LR decay is crucial for MT loss.}
Since small learning rates after LR annealing can lead to underutilization of MT data, the timing of LR decay is crucial for MT loss. For example, matching the LR decay range to the MT data range (MDR) can improve MT data efficiency over a full-range LR decay (FDR).
As shown in \Cref{fig:empirical,fig:exp-dense-moe}, MDR can improve MT loss at the same $\gamma_{\mathrm{end}}$ across both dense and MoE models and MT data budgets. As $\gamma_{\mathrm{end}}$ increases, MT data is placed later, and the benefit of matching LR decay also increases. This observation shows that the LR schedule regime is important for optimizing the data schedule:
even after tuning $\gamma_{\mathrm{end}}$ under FDR, its final MT loss is higher than that under MDR at the same $\gamma_{\mathrm{end}}$.
In addition, our results show that MDR lowers final MT loss relative to Const at all three dense budgets but raises it in all three main MoE settings (\Cref{fig:exp-reversal-budgets,fig:exp-dense-moe}). This motivates delaying LR decay to maintain a higher average LR over the MT data range. We next test whether keeping the LR high after MT data is introduced, followed by a shorter final cooldown, can improve on both Const and MDR.

\begin{figure}[!t]

    \centering
    \includegraphics[width=\linewidth]{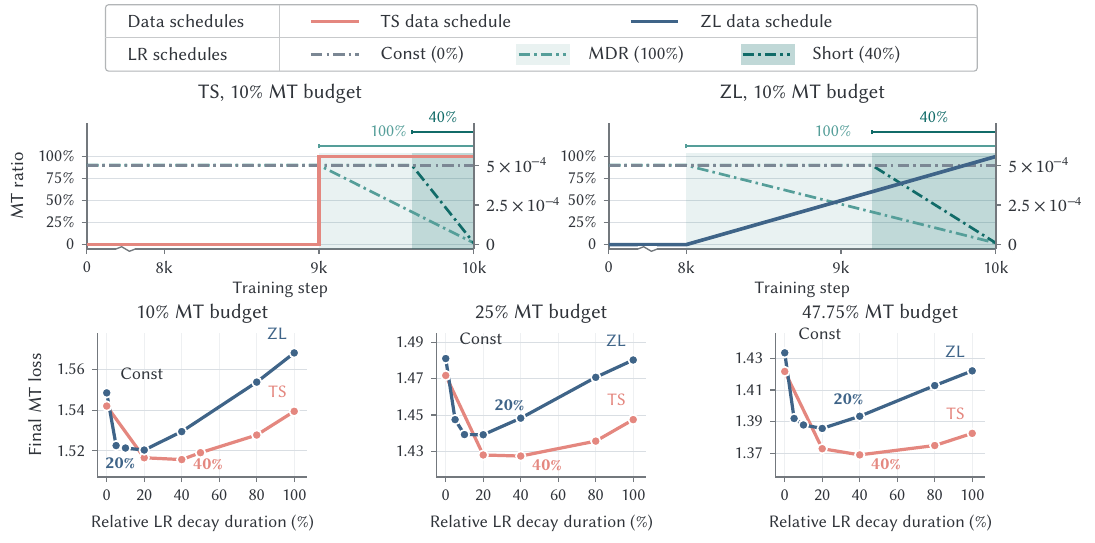}
    \caption{\textbf{LR should decay well after MT enters training for TS and ZL data schedules.} The bottom row shows how final MT loss changes with relative LR decay duration for TS and ZL across three MT budgets. ZL reaches its lowest MT loss at 20\%, and TS at 40\%, improving on both Const (0\%) and MDR (100\%). The top row illustrates TS (left) and ZL (right) at a 10\% MT budget, paired with LR schedules of different \textit{relative LR decay durations}, measured as percentages of the time from MT entry to training completion.
    See Section~\ref{sec:exp-decay} and Appendix~\ref{app:exp-decay} for the setup and further discussion.}

    \label{fig:delayed-lr-decay}
    \label{fig:exp-delayed-decay}
\end{figure}

\subsection{Learning Rate Decay Should Begin Well After MT Data Enters Training}
\label{sec:exp-decay}

Section~\ref{sec:exp-reversal} shows that the data and LR schedules must be chosen together, and matching the LR decay range to the MT data range can improve MT loss. Moreover, the timing of LR decay affects MT performance, and starting LR decay at MT entry time is not always beneficial. We now ask how long LR decay should last to minimize final MT loss for each fixed data schedule.

\paragraph{Delaying LR decay until after MT entry improves model performance over both Const and MDR.}
Keeping the LR high allows larger updates from MT data, while a final LR decay can reduce the effect of gradient noise. To balance these effects, we keep each data schedule fixed and vary when the LR begins to decrease. The LR remains at its peak before decaying linearly to the same final value. We define the \textit{relative LR decay duration} as the number of LR decay steps divided by the number of steps from MT entry time to training completion, expressed as a percentage. It is zero for Const and 100\% for MDR. Across all 19 scans, the best tested relative LR decay durations lie between 10\% and 40\%, giving lower MT loss than both baselines for dense and MoE models (Appendix~\ref{app:exp-decay}). Figure~\ref{fig:exp-delayed-decay} shows this pattern for dense models: the lowest losses occur when the LR stays high through most of the MT data window and decreases only near the end. This delayed LR decay design achieves a better balance and outperforms both Const and MDR.

\paragraph{The optimal relative LR decay duration for ZL is about half that for TS.}
The results above show that a delayed LR decay gives lower MT loss. This holds for both TS and ZL, and the two data schedules favor different relative LR decay durations. Additionally, we find an interesting result that across dense and MoE experiments, the best tested relative LR decay duration for ZL is about half that for TS (\Cref{fig:exp-delayed-decay,fig:app-decay-moe4,fig:app-decay-moe10,fig:app-moe4775-decay}). Each schedule uses its own MT data window, from MT entry time to training completion, to measure the relative LR decay duration. TS uses a fixed MT ratio during this interval, while ZL increases it gradually from zero. To use the same MT budget and reach the same final MT ratio, ZL therefore needs an interval twice as long. Its smaller relative LR decay duration therefore corresponds to approximately the same number of LR decay steps as TS.
We also repeat the LR decay sweep for dense TS schedules with different final MT ratios and find that the best tested relative LR decay duration remains unchanged (\Cref{fig:app-decay-dense10,fig:app-decay-dense4775-ts}).

\paragraph{The preferred relative LR decay duration is stable within the model family.} We further conduct extensive sweeps on the optimal relative LR decay duration for the MoE model. We find that the above optimal values ($10\%$ for ZL and $20\%$ for TS) remain unchanged across different setups. Specifically, in MoE, the best tested relative LR decay durations remain the same across continued training lengths, and TS retains its preference at earlier PT checkpoints (\Cref{fig:app-moe-horizons,fig:app-moe-pt-length}). We also conduct an ablation on the MT dataset. We find that replacing FineMath with StarCoderData~\citep{exp_starcoder2023} preserves the preferred relative LR decay durations for both TS and ZL (\Cref{fig:app-moe-mt-domain}). Appendix~\ref{app:exp-moe} describes the dataset and comparison setup. Appendix~\ref{app:exp-decay} summarizes all 19 scans.

\section{Theoretical Analysis}

In this section, we provide a theoretical analysis to understand how the ranking reversal phenomenon between data schedules for different learning rate schedules might arise.
Specifically, we consider a simple setting where SGD optimizes a block diagonal quadratic loss function. Though this model does not fully capture the complexity of data mixing and learning rate schedules, we believe it offers insight into how these two can be coupled together in the optimization landscape. Further, under a certain spectral assumption used in previous scaling law analyses, we can give a closed form of the optimal data schedule under practical learning rate schedules with decay.

\label{sec:theory}
\subsection{Setup}
\label{sec:theory-setup}

Given a data distribution $\cD$, we consider a quadratic loss $\cL_{\cD}(\vtheta)=\frac{1}{2}\vtheta^{\top}\mM\vtheta$,
where $\vtheta\in\R^d$ is the trainable parameter vector, and $\mM$ is the covariance matrix of the distribution $\cD$. During training, we consider a mixture of PT and MT data, drawn from the distributions $\cD_{\text{PT}}$ and $\cD_{\text{MT}}$, respectively. We denote their covariance matrices by $\mG=\E_{\cD_{\text{PT}}}\left[\vx\vx^{\top}\right]$ and $\mH=\E_{\cD_{\text{MT}}}\left[\vx\vx^{\top}\right]$. We evaluate our model using the MT loss $\cL(\vtheta)=\frac{1}{2}\vtheta^{\top}\mH\vtheta$.

\paragraph{PT and MT data.}

In practice, PT and MT data share some foundational knowledge, while each also contains knowledge that is specific to its own domain. For example, broad web PT data and mathematical MT data both contain general language patterns, while covering different specialized features.
Motivated by this structure, we model shared and domain specific features through block diagonal covariance matrices, with a structure similar to those studied in transfer learning and data mixing theory~\citep{ju2023generalization,mallinar2024minimum,pan2026understanding,dai2026explaining} as $\mH=\diag\bigl(\mH_0,\mH_1,\mathbf{0}_{d_2\times d_2}\bigr), \mG=\diag\bigl(\mH_0,\mathbf{0}_{d_1\times d_1},\mH_2\bigr).$ Here $\mH_0$, $\mH_1$, and $\mH_2$ represent the covariance blocks for shared, MT-specific, and PT-specific features, respectively, similar to \citet{dai2026explaining}.
For $l\in\{0,1,2\}$, we assume
$\mH_l=\diag(\lambda_1^{(l)},\ldots,\lambda_{d_l}^{(l)})$
with $\lambda_i^{(l)}>0$ and $d=d_0+d_1+d_2$.

\paragraph{Data mixing and SGD training dynamics.}
We analyze SGD on mixtures of PT and MT data using a continuous-time
SDE approximation \citep{jastrzkebski2017three,li2017stochastic,li2021validity,malladi2022sdes}.
Let $T$ denote the training horizon in continuous time, and let
$\gamma(t)\in[0,1]$ denote the probability of sampling MT data at time
$t\in[0,T]$.
At time $t$, the training objective is given by $\cL_t(\vtheta)
    =\half\vtheta^{\top}
    \bigl(\gamma(t)\mH+(1-\gamma(t))\mG\bigr)\vtheta.$
We model each stochastic gradient as $\nabla\cL_t(\vtheta)$ plus independent additive noise with mean zero and covariance $\mSigma$.
The continuous-time approximation describes the average gradient update through a deterministic term and the accumulated gradient noise through Brownian motion.
For a learning rate schedule $\eta(t)$, the training dynamics of $\vtheta_t$ are given by the following SDE:
\begin{equation}
    \d\vtheta_t
    =-\eta(t)\nabla\cL_t(\vtheta_t)\d t
      +\eta(t)\sqrt{h\mSigma}\d\mB_t.
    \label{equ:sde-dynamics}
\end{equation}
Here $\mB_t$ is a standard $d$-dimensional Brownian motion, and $h>0$ is the interval in continuous time between consecutive SGD updates. $\mSigma$ is the covariance matrix of the additive gradient noise.
See Appendix~\ref{app:theory-setup} for the discrete SGD update and the derivation of its SDE approximation.
Appendix~\ref{app:theory-simulation-sde-check} compares the final MT loss predicted by the SDE with exact calculations and Monte Carlo simulations of discrete SGD.

In our analysis, we focus on a regime where the quantity of MT data is constrained:

\begin{condition}
\label{asp:constrained-cpt-data}
For some fixed $\alpha\in(0,1]$, MT data accounts for at most
an $\alpha$ fraction of the total training data.
\end{condition}
Under this budget constraint, we seek a data schedule that minimizes the expected final MT loss for a fixed training horizon $T$ and LR schedule $\eta$:
\begin{equation}
    \min_{\gamma:[0,T]\to[0,1]}
    \E[\cL(\vtheta_T)]
    \qquad\text{s.t.}\qquad
    \frac{1}{T}\int_0^T\gamma(t)\d t\leq\alpha.
    \label{aim:data-schedule-minimization}
\end{equation}
The solution is unique up to changes on a set of measure zero, as established in the last paragraph of \Cref{app:terminal-loss-derivation}. We therefore refer to this solution as \emph{the optimal data schedule}.
For additional theoretical settings, see \Cref{app:theory-setup} for more details.

\subsection{Optimal Data Schedules Depend on the LR Schedule}
\label{sec:opt-ds-without-mpl}
\begin{figure}[t]

    \centering
    \includegraphics[width=\linewidth]{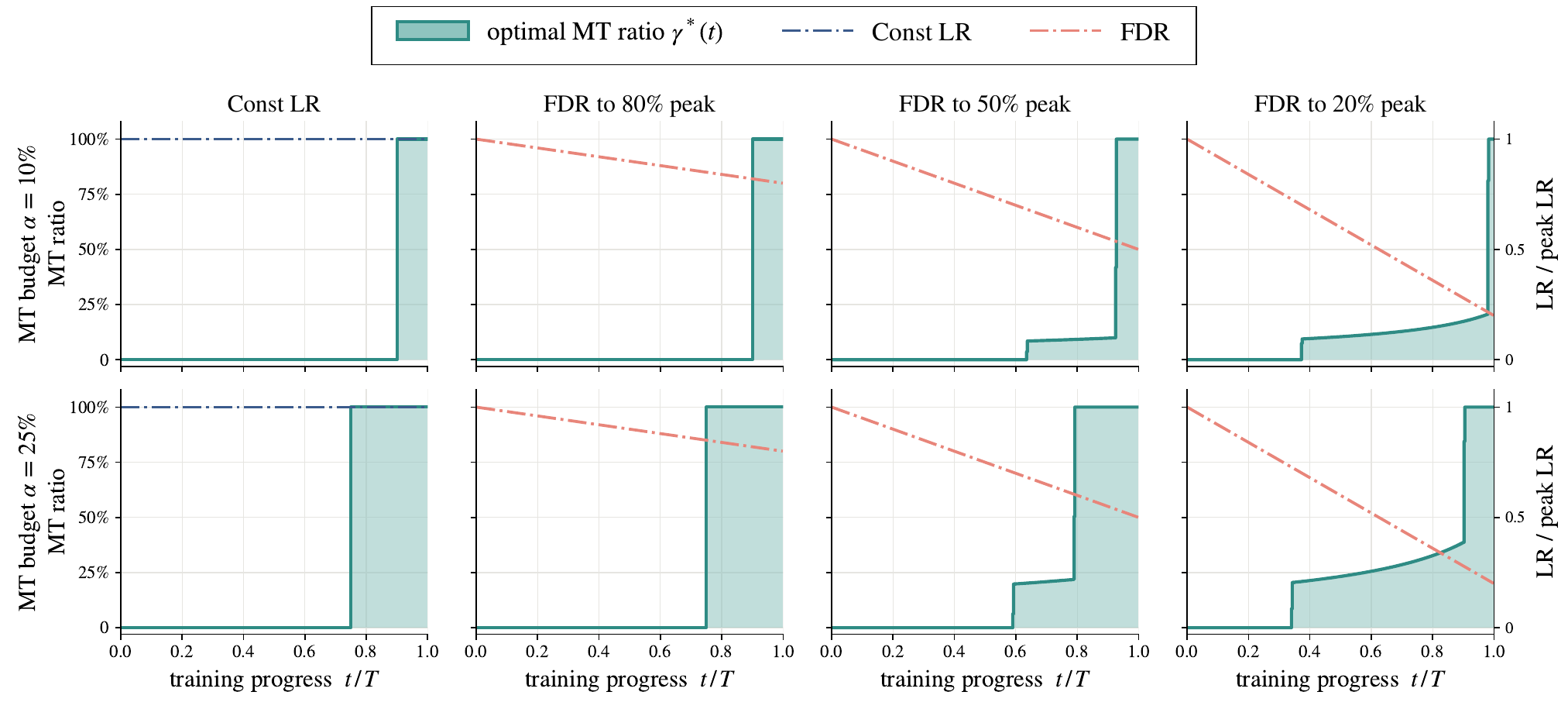}
    \caption{\textbf{The optimal data schedule can have three stages under LR decay.} The first two columns show an abrupt switch from pure PT to pure MT. With stronger LR decay in the last two columns, the teal curves instead include a mixing stage in which the MT ratio increases continuously. This change occurs at both MT budgets shown. See \Cref{thm:informal} for the theoretical result and Appendix~\ref{app:theory-simulation} for the numerical setup.}
    \label{fig:app-optimal-schedule-decay}

\end{figure}

In \Cref{sec:exp-reversal}, placing all MT data in a final stage where only MT data is used gives the lowest MT loss among TS schedules under Const, whereas FDR favors earlier MT entry with PT replay.
We now recover this LR dependence in theory. Furthermore, under some power-law spectral assumptions~\citep{lin2024scaling,luo2025a,yan2026larger}, we give a closed form of the optimal data schedule under the FDR regime.

\begin{theorem}[Informal] \label{thm:informal}
Under \Cref{asp:constrained-cpt-data}, let $\eta_{\max}>0$. The following statements hold:
\begin{enumerate}[label=(\roman*),leftmargin=*,nosep]
    \item \textbf{Const.}
    If $\eta(t)\equiv\eta_{\max}$, the optimal data schedule $\gamma^*$ is given by $\gamma^*(t) = \bar{\gamma}(t):=\mathds{1}_{\{t\geq (1-\alpha)T\}}$, namely the TS schedule with $\gamma_{\mathrm{end}}=1$. \label{thm:informal-const-lr}
    \item \textbf{Linear decay (FDR).}
    If $\eta(t)=\eta_{\max}-ct/T$, where $(1-4^{-1/3})\eta_{\max}<c<\eta_{\max}, 0<\alpha<1-(1-4^{-1/3})\eta_{\max}/c,$
    then $\Bar{\gamma}$ is not optimal. \label{thm:informal-lr-decay}
    \item \textbf{Three stage optimum under FDR.} In the FDR regime, under an additional power-law spectral assumption and for sufficiently large $T$, the optimal data schedule $\gamma^*$ is given by:
    \begin{equation*}
        \gamma^*(t)=
        \begin{cases}
            0, & 0\leq t<t_1, \\[2pt]
            C\,\eta(t)^{-\frac{2(b-1)}{b+1}},
                & t_1\leq t<t_2, \\[2pt]
            1, & t_2\leq t\leq T,
        \end{cases}
    \end{equation*}
where $0<t_1<t_2<T$, $C>0$ is constant in time, and $b>1$ is determined by the power-law spectral assumption. \label{thm:informal-mpl}
\end{enumerate}
\end{theorem}

\Cref{thm:informal} shows that different learning rate schedules induce different optimal data schedules, and the optimal data schedule under the FDR regime has a time range when both PT and MT data are used in training under the power-law spectral assumption. We then give some intuition on how this theorem works. The formal statements of our theoretical results are given in \Cref{app:theory-statement}, and proofs are given in \Cref{app:proof-opt-ds-const-lr,app:proof-opt-ds-linear-decay-lr,app:proof-opt-ds-mpl}.

\paragraph{Interpretation of \Cref{thm:informal}.}
For a fixed LR schedule, only the contribution from features specific to MT depends on the data schedule, since PT and MT train the shared features in the same way. For these MT features, training may leave a nonzero loss even without gradient noise; we call this the \emph{bias term}. Gradient noise introduces an additional contribution to the expected final loss, which we call the \emph{variance term}. The linear SDE and quadratic loss allow us to separate these contributions exactly.  Writing $R(s)=\int_s^T\eta(t)\gamma(t)\,\d t$ for the cumulative LR of MT updates after time $s$, \Cref{eq:app-terminal-loss} gives
\[
    \begin{aligned}
    \E[\cL(\vtheta_T)]
    = A
    +\underbrace{\frac12\sum_{i=1}^{d_1}\lambda_i^{(1)}
        \theta_i^{(1)}(0)^2
        e^{-2\lambda_i^{(1)}R(0)}}_{\text{bias term}}+\underbrace{\frac h2\sum_{i=1}^{d_1}
        \lambda_i^{(1)}\nu_i^{(1)}
        \int_0^T\eta(s)^2
        e^{-2\lambda_i^{(1)}R(s)}\,\d s}_{\text{variance term}},
    \end{aligned}
\]
where $A$ is the expected final loss from shared features, independent of the data schedule, and $\nu_i^{(1)}$ is the gradient noise variance for the corresponding MT feature. The bias term decreases with $R(0)$. For noise introduced at time $s$, the factor $e^{-2\lambda_i^{(1)}R(s)}$ describes how subsequent MT updates reduce its contribution to the final loss.

\textbf{With a constant LR}, using the full MT budget fixes $R(0)$, so all such schedules have the same bias term. Placing all MT data in a final pure MT stage maximizes $R(s)$ at every time $s$ and therefore minimizes the variance term. This explains the optimality of the abrupt switch in part~(i).

\textbf{Under LR decay}, consider moving a small amount of MT data from a later time to an earlier one while keeping the MT budget fixed. The larger LR at the earlier time increases $R(0)$, reducing the bias term, and more strongly suppresses noise introduced before that time. However, removing the corresponding later MT updates leaves less MT training to suppress noise introduced between the two times, increasing its contribution to the variance term. Under the conditions of part~(ii), moving some MT data earlier lowers the final loss, so the constant LR optimum is no longer optimal.

\textbf{To determine the optimal data schedule}, we compare the benefit of using the same amount of MT data at different times. If moving a small amount of MT data earlier or later lowers the final loss, the original allocation cannot be optimal. The optimal schedule must therefore balance the benefits of earlier and later MT updates so that no such reallocation can further reduce
the loss. Under the assumptions of part~(iii), solving for this balance gives the three-stage schedule in the theorem. During the middle stage, the optimal MT ratio satisfies $\gamma^*(t)=C\eta(t)^{-2(b-1)/(b+1)}$.
Since $b>1$, this ratio increases as the LR decreases: the optimal mixture contains progressively more MT data before the final pure MT stage. See Appendix~\ref{app:proof-opt-ds-mpl} for the derivation.

\subsection{LR Should Decay After MT Data Enters Training}\label{app:lr-setup}

In \Cref{sec:exp-decay}, we observed that to minimize the final MT loss, the optimal learning rate decay should begin after the MT data enters training. Under additional conditions, we also recover this finding theoretically.

\paragraph{Setup.} We use the SDE in \Cref{equ:sde-dynamics} and fix the data schedule to TS with a final MT ratio of $1$:
\begin{equation*}
    \gamma(t)=\mathds{1}_{\{t\geq T-L\}},\qquad L=\alpha T.
\end{equation*}
Thus, training uses only PT data before $(1-\alpha)T$ and only MT data during the remaining interval of length $L$. We consider stable-decay (SD) LR schedules, which keep the LR at $M$ before decreasing it linearly to $m$ over the final $W$ units of time:
\begin{equation*}
    \eta_W(t)=
    \begin{cases}
        M, & 0\leq t<T-W,\\
        m+\Delta(T-t)/W, & T-W\leq t\leq T,
    \end{cases}
    \qquad 0<W\leq T.
\end{equation*}
Here $0<m<M$ are fixed and $\Delta:=M-m$. We include Const by defining $\eta_0(t)=M$; $W=L$ gives MDR and $W=T$ gives FDR. As in part~(iii) of \Cref{thm:informal}, we average the final MT loss over a power-law spectral model as well as training noise, and denote it by $J_T(W)$. Our goal is to minimize $J_T(W)$ over $W\in[0,T]$.

Next we state the main theorem, which shows that the optimal learning rate decay begins after the MT data enters training.

\begin{theorem}[Informal]
\label{thm:lr-delayed-decay}
    For the dynamics in \Cref{equ:sde-dynamics}, consider TS and the SD LR family with $$\gamma(t)=\mathds{1}_{\{t\geq(1-\alpha)T\}},\qquad
        0<m<M.$$
    Under suitable power-law spectral conditions, for every sufficiently small fixed $\alpha>0$ and sufficiently large $T$, any optimal decay duration $W^*$ satisfies $ 0<W^*<\alpha T.$ Thus, optimal LR decay begins strictly after MT entry.
\end{theorem}
See \Cref{app:lr-schedule} for the formal statement (\Cref{thm:lr-delayed-decay-formal}) and proof.

\paragraph{Interpretation of \Cref{thm:lr-delayed-decay}.} Unlike changing the data schedule, changing the LR affects both the shared and MT-specific contributions to the final loss. Under the power-law spectral model, the contribution of earlier gradient noise decreases as a power of the cumulative LR of subsequent updates. The exponents are $b_0$ for shared features and $b_1$ for MT-specific features; a smaller exponent means slower decay. Thus, in our analysis, we focus on the scenario $b_0\geq b_1$, which describes a regime in which earlier noise in MT-specific features decreases no faster in the long run than in shared features. This captures a natural difficulty in MT adaptation: the new domain requires training specialized features, while the shared features already receive updates throughout PT.

\paragraph{Keeping the LR high after MT entry helps reduce the variance accumulated during PT.} Under the training dynamics in \Cref{equ:sde-dynamics}, gradient noise affects MT-specific features even during pure PT training. PT updates do not train these features, so only MT updates can reduce the contribution of this noise to the final variance term. The contribution of noise introduced during pure PT training to the variance term of the final MT loss is
\[
    V_1\frac{\displaystyle\int_0^{T-L}\eta_W(s)^2\,\d s}
                 {[g_1+R(0)]^{b_1}},
\]
where $R(0)$ is the cumulative LR of MT updates defined above, and $V_1,g_1>0$ are fixed constants from the spectral model (\Cref{eq:app-mpl-final-loss-fix-ds}). Starting LR decay before MT entry can reduce the noise accumulated during PT, represented by the numerator, but also weakens the MT updates that suppress it, represented by the denominator. With a small MT budget, noise accumulates over a long PT stage and must be reduced within a short MT stage. Keeping the LR high for the initial MT updates increases $R(0)$, reducing this variance contribution as well as the bias term.

\paragraph{A final LR decay reduces the variance term.} Keeping the LR high until the end would continue to introduce noise with little time for later updates to reduce it. Under $b_0\geq b_1$, a sufficiently small MT budget, and a sufficiently long training horizon, the benefit of stronger initial MT updates outweighs the noise reduction from starting decay at or before MT entry. Decaying later still reduces the final variance compared with Const. Together, these effects place the optimal decay strictly inside the MT stage.

\begin{figure}[t]

    \centering
    \includegraphics[width=\linewidth]{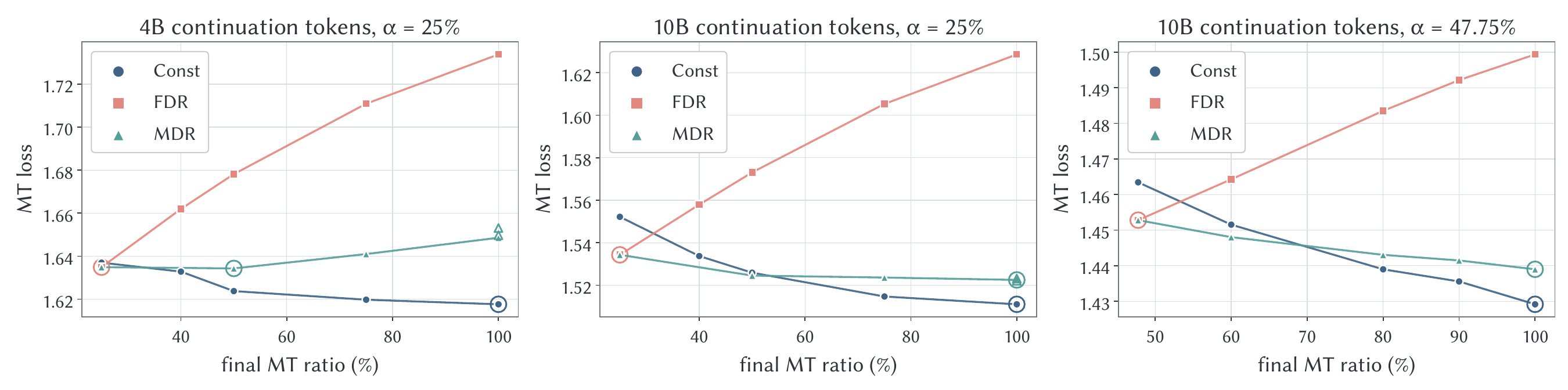}
    \caption{\textbf{LR-dependent data-schedule preferences also occur in MoE.} Left to right: 4B continuation tokens with a 25\% MT budget; 10B tokens with 25\% MT; and 10B tokens with 47.75\% MT. All three settings use the same 2.99B-parameter model. At a final MT ratio of 100\%, Const can outperform MDR. See Section~\ref{sec:exp-reversal} and Appendix~\ref{app:exp-moe} for the setup and full results.}
    \label{fig:exp-dense-moe}

\end{figure}

\section{Additional Experiments}
\label{sec:exp-moe}

In this section, we present additional experiments following the discussion in \Cref{sec:empirical}.

\subsection{Joint scheduling also shapes the PT--MT loss frontier}

In addition to MT loss, PT loss also matters in practice, and we usually need to balance the two. As discussed in \Cref{sec:exp-reversal} and shown in \Cref{fig:app-dense10-grid,fig:app-dense25-grid}, with FDR, both MT and PT loss decrease as the ending MT ratio decreases under both TS and ZL data schedules. However, in most cases where LR decay does not suppress learning from MT data, including Constant LR and MDR, PT loss usually increases as MT loss decreases. Therefore, we need to strike a balance between the PT and MT loss in these cases. From this perspective, we find that \textbf{the ZL data schedule usually yields a narrower range of MT and PT losses} across different ending MT ratios, \textbf{whereas the TS data schedule exhibits much larger variations, especially in PT loss.} Thus, in practice, using TS requires more careful tuning. In addition, given a TS or ZL data schedule, we find that the MDR regime dominates both Const and FDR regimes for dense models mostly (\Cref{fig:app-dense10-grid,fig:app-dense25-grid}). Moreover, under MDR, we find that TS dominates ZL in terms of both PT and MT loss (\Cref{fig:app-dense10-grid,fig:app-dense25-grid,fig:app-moe10-grid,fig:app-moe4-grid}). But under Constant LR, ZL exhibits advantages over TS for the MoE model, which may be attributed to the load-balancing benefits of ZL in the MoE setting discussed in the following section.

\subsection{Continuous data schedules reduce load imbalance caused by distribution shifts}

\begin{figure}[tb]

    \centering
    \includegraphics[width=0.8\linewidth]{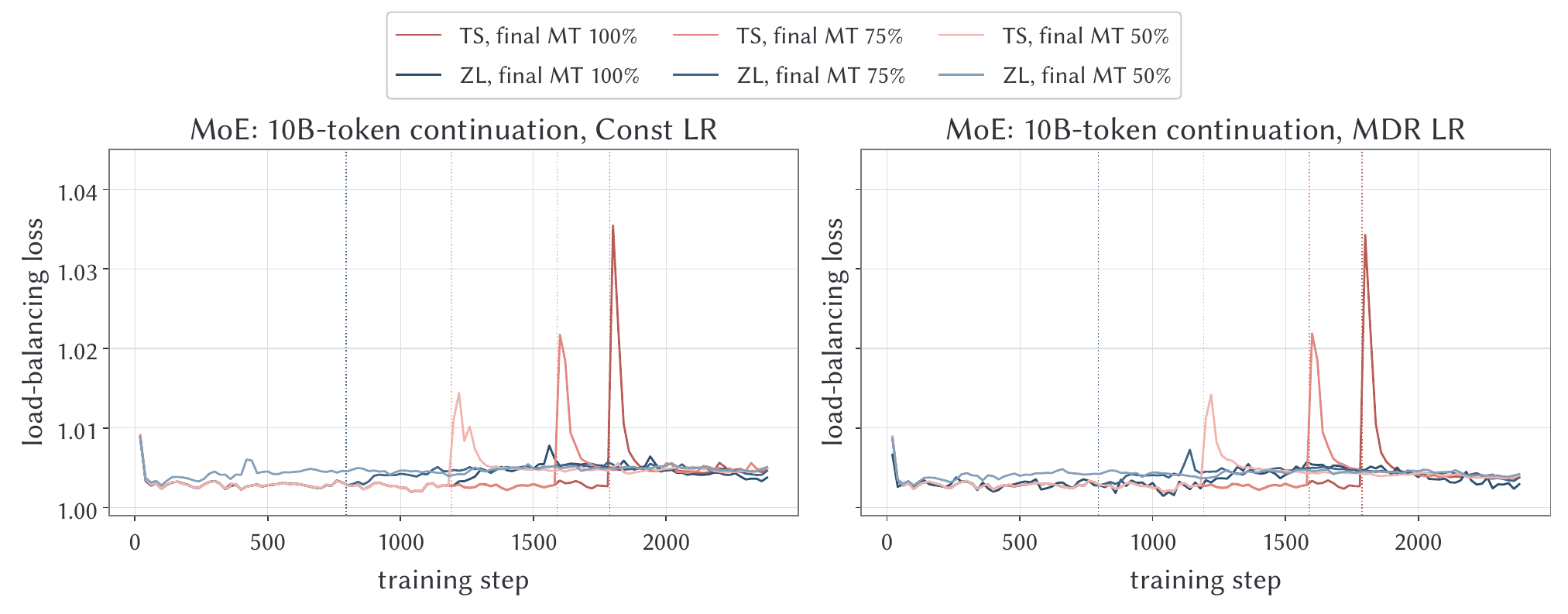}
    \caption{\textbf{Gradual MT entry reduces router imbalance under data distribution shift.} TS produces sharp spikes when the MT ratio jumps, while ZL avoids comparable spikes. Under Const (left), both schedules use the same LR trajectory, so this difference comes from the data schedule. MDR (right) shows a similar pattern, although LR decay starts at different times for TS and ZL. These MoE runs use 10B continuation tokens and a fixed 25\% MT budget. Dotted lines mark MT entry. See Appendix~\ref{app:exp-router} for setup and diagnostic details.}
    \label{fig:exp-router-smoothing}
\end{figure}

\paragraph{Continuous data schedules suppress the load balancing loss spike.} One additional training requirement for MoE models compared with dense models is that MoE must also balance the training tokens assigned to different experts. A common approach is to add an auxiliary load balancing term to the training loss, discouraging the router from sending most tokens to a few experts while leaving others underused~\citep{shazeer2017outrageously,lepikhin2021gshard,fedus2022switch}.
In our experiments, the abrupt data distribution shift under TS produces a sharp spike in the load balancing loss, indicating routing instability during the data shift. However, introducing MT data gradually through a continuous schedule such as ZL largely avoids this spike (\Cref{fig:exp-router-smoothing}). For the $10$B runs with a final MT ratio of $100\%$, ZL also ends with a slightly lower load balancing loss than TS under both Const and MDR. We also provide further ablations across training horizons and LR schedules in Appendix~\ref{app:exp-router} (\Cref{fig:app-router-lr-control}).

\textbf{ZL also better preserves the rankings of expert loads across the data shift.} Beyond the loss, we further rank experts by how many MT tokens they receive in held-out evaluations before MT entry time and at the end of training. Under ZL, experts that receive more tokens before the shift tend to remain more heavily used afterwards.
Under TS, expert loads become more balanced after the switch, but some experts that received more tokens before the switch drop in the load ranking, while previously less used experts move up. Figure~\ref{fig:app-router-rebalancing} compares both the final load balance and these ranking changes under TS+Const and ZL+MDR. This substantial reshuffling of expert loads may make subsequent training less stable. Appendix~\ref{app:exp-router} gives the full comparison, with the ranking statistics reported in Table~\ref{tab:app-router-comparison}.

\section{Conclusion}

In this work, we show that data schedules and LR schedules should be designed together by combining experiments on dense and MoE models with a quadratic theoretical model. There are several promising directions for future work: 1) Our results provide an initial understanding of this coupling, but determining the form of jointly optimal schedules under fixed compute and MT token budgets remains theoretically challenging.
2) We also focus primarily on minimizing final MT loss.
Future work should consider PT and MT losses jointly and potentially incorporate downstream task metrics when evaluating and choosing schedules.

\section{Acknowledgement}
We thank Jianxun Zhou and Chenyi Dang for their early efforts on this project. We would like to thank Zixuan Wang, Kaiyue Wen, Huaqing Zhang, Zhixuan Pan, Binghui Li, and Zhenbo Sun for helpful discussions and suggestions. This work was supported by the Major Key Project of PCL under Grant PCL2025A12.

\subsection*{AI use statement}
In this work, we used generative AI tools to assist with checking mathematical claims, theoretical derivations, and proofs, and with analyzing and interpreting experimental results.
We did not use generative AI tools to design experiments or implement methods.
Additionally, we used these tools for manuscript drafting and editing, figure preparation, literature searches, and \LaTeX{} formatting.
We reviewed all AI-assisted work, checking the text for accuracy, comparing experimental analyses with the original results, and examining theoretical derivations and proofs for correctness.
We take responsibility for the final content of this work, including text, claims, and artifacts produced with the aid of generative AI.
\clearpage
\bibliography{bibliography/references}
\bibliographystyle{preprint}

\newpage
\tableofcontents
\newpage
\appendix
\crefalias{section}{appendix}
\crefalias{subsection}{appendix}
\crefalias{subsubsection}{appendix}

\section{Related Works}
\label{sec:related-work}

\paragraph{Midtraining and continual pretraining.}
Modern language model recipes increasingly include a later training phase that
shifts the data distribution from broad pretraining corpora toward more curated
or capability-focused data. Recent technical reports describe this pattern
through late stage training on curated mixtures and multistage data
curricula~\citep{olmo2furious,phi4tech}. In Qwen3, the reasoning stage both
raises the proportions of STEM, coding, reasoning, and synthetic data and
accelerates learning rate decay~\citep{qwen3tech}. This practice couples the
data and learning rate schedules but does not establish whether that coupling
is optimal. A closely
related literature studies continual pretraining, where a model adapts to a new
distribution while retaining performance on its source distribution. Existing
work examines source data replay, learning rate rewarming and redecay, and
mixture ratios for balancing adaptation and
retention~\citep{ibrahim2024simple,que2024dcpt,wang2025learning}.
Recent controlled studies make the timing question more explicit.
\citet{emmy2025bridge} find that the introduction time and mixture weight of
specialized data interact strongly in midtraining. \citet{kotha2026replaying}
show that source data replay can improve target data efficiency and that its
benefit depends on how much target data is already present during pretraining.
They also study LR cooldown and report that cosine decay can make uniform
mixing outperform placing target data last. \citet{baek2026finetuner} find that
repeating domain data during pretraining can improve domain performance and
preserve general capabilities after finetuning.
A separate line of work studies how loss changes when a finite corpus is reused
over multiple epochs~\citep{muennighoff2023scaling,xue2023repeat,
yan2026larger}. These studies vary the amount of reuse rather than the temporal
allocation of a fixed target data budget.
Together, prior results establish that replay, timing, and reuse all matter.
We build on these observations by comparing the same data schedules across LR
schedules while fixing total training compute and the MT token budget.

\paragraph{Learning rate scheduling in LLM training.}
Cosine decay is widely used in language model pretraining, but it
ties the decay horizon to a predetermined training
length~\citep{loshchilov2017sgdr,luo2025a}. Warmup Stable Decay separates a
long stable phase from a final cooldown, which
supports continued training and reusable decay branches~\citep{hu2024minicpm,hagele2024scaling}.
A constant LR with exponential weight averaging can also match the optimization
efficiency of cosine decay and Warmup Stable Decay, particularly at large batch
sizes~\citep{zhang2025criticalbatch}.
Several studies model how the full learning rate trajectory affects loss rather
than treating the schedule as a fixed experimental detail. Schedule aware
scaling laws quantify the effects of accumulated learning rate and annealing on
loss curves~\citep{tissue2025annealing,luo2025a}. Complementary analyses study
the optimization dynamics of the stable and cooldown phases
directly~\citep{wen2025river,dremov2025cooldown}. In continual pretraining,
learning rate rewarming and redecay are often coupled with source data
replay~\citep{ibrahim2024simple}, while \citet{wang2025learning} jointly model
distribution shift and learning rate annealing in the loss curve. Most directly
related to our motivation, \citet{luo2026bestdata} show that strong decay can
reduce the benefit of high quality examples placed late in a curriculum ordered
by data quality. Their remedies use a milder final decay or model averaging. We
instead vary the timing of domain mixtures under a fixed target data budget and
ask how the preferred data schedule changes across learning rate schedules.

\paragraph{Data mixture ratios and temporal scheduling.}
Prior work on data mixing primarily optimizes how much data to sample from each
domain, whereas our question concerns how a fixed target data budget should be
allocated over training time. Proxy based methods learn mixture weights through
robust optimization or regression over smaller training
runs~\citep{xie2023doremi,liu2025regmix}. Scaling law methods predict
performance at unseen mixture ratios and data
scales~\citep{bimix2025,ye2025data,que2024dcpt,gu2024cmr}. In particular,
\citet{gu2024cmr} characterize the tradeoff between general and domain
performance during continual pretraining, but treat the mixture ratio as a
variable for the full training run rather than as a temporal schedule. Other
methods allow the proportions to change during training.
\citet{albalak2023online} use a bandit algorithm to adapt domain sampling
probabilities online, while learned and adaptive frameworks optimize the ratios
directly or use a learned policy~\citep{chen2025aioli,yang2026datamixingagent}.
Recent frameworks address efficient mixture search and mixture reuse as the
available domains change~\citep{wang2026mergemix,chen2026olmix}. These methods
broaden how mixture proportions can be selected and adapted. Our comparisons
keep both total training compute and the cumulative MT token budget fixed,
then vary the LR schedule to test whether the preferred temporal allocation
changes.

\paragraph{Theory of data scheduling under distribution shift.}
A complementary theoretical literature studies how prescribed task sequences
affect transfer and forgetting. Analyses of overparameterized linear regression relate forgetting to task geometry, task order, and repeated
exposure~\citep{evron2022catastrophic,goldfarb2023analysis,
goldfarb2024joint}. Teacher student and feature learning models show that task similarity and representation dynamics shape interference~\citep{lee2021continual,graldi2025lazy}. Repeated gradient descent on linear classification tasks can converge to a joint solution under suitable conditions~\citep{jung2025convergence}. Closer to our setting,
\citet{wu2022power} derive target risk bounds when online SGD trains first on a source distribution and then on a target distribution under covariate shift. \citet{ding2024understanding} analyze noisy linear regression across multiple tasks and show that covariance spectra, task order, and step size jointly shape forgetting. These works prescribe the task sequence and then evaluate transfer, forgetting, or target risk. Our analysis instead treats the within training mixture as the decision variable, fixes its cumulative target data budget, and characterizes how the optimal temporal allocation changes with the learning rate schedule.

\paragraph{Comparison with \citet{kotha2026replaying}.}
A closely related work is \citet{kotha2026replaying}, which demonstrates the benefits of replaying pretraining data during mid-training and fine-tuning. They also investigate how pretraining data and target data should be scheduled. However, they focus on a two-stage training scenario consisting of pretraining followed by mid-training or fine-tuning, with a fixed mixing ratio between the two data sources within each stage. In contrast, we use a stepwise data schedule within MT, which allows the proportions of PT and MT data to vary over the training process. More importantly, while \citet{kotha2026replaying} observe that the learning rate schedule can affect the effectiveness of data ordering, we systematically study the interaction between the learning rate schedule and the data schedule. Across a wide range of settings, we show that changing only the LR schedule can reverse which data schedule achieves lower final MT loss. Our results therefore show that the effectiveness of a data schedule cannot be considered independently of the LR schedule. Furthermore, we provide a simple guideline for jointly designing the LR schedule and the data schedule: LR decay should begin only after MT data enters training.

\section{Training Setup and Schedule Definitions}
\label{app:exp-setup}

This section describes the training configuration and explicitly defines the data and LR schedules used in \Cref{sec:empirical}.

\paragraph{Losses and plot conventions.} PT loss denotes DCLM validation loss, and MT loss denotes validation loss on FineMath, or on StarCoderData Python in the code-domain ablation.
In the data-schedule and LR-decay scans, curves connect primary runs, and rings mark their lowest tested losses. Hollow overlays show repeated runs using different subsets or orders of training examples (\Cref{app:exp-replicas}).
The data schedule comparison figures plot final PT loss horizontally and final MT loss vertically, with point labels giving the final MT ratio. Top panels separate TS and ZL; the remaining panels group the same results by LR rule.

\paragraph{Data and preprocessing.}
PT uses DCLM-Baseline~\citep{exp_dclm2024}, and MT uses the \texttt{finemath-4plus} subset of FineMath released with SmolLM2~\citep{exp_smollm2025}.
We tokenize the same source corpora for the dense model and the MoE model using their respective tokenizers.
An EOS token is appended to each document, and best-fit packing produces fixed-length rows of 4,097 tokens, padded when necessary and shuffled within each shard.
The first 4,096 tokens are inputs and the last 4,096 are prediction targets.
Attention is causal across the packed row, without a document-level attention mask.
The loss mask excludes positions whose \emph{input} token equals EOS.
The tokenized pools contain approximately 206B DCLM and 10.0B FineMath tokens for the dense model, and 212.6B DCLM and 9.75B FineMath tokens for the MoE model.
DCLM data used for continued training comes from shards not consumed during PT, and both validation sets are held out from training.

\paragraph{Models and optimization.}
\Cref{tab:app-dense-setup,tab:app-moe-setup} give the full configurations of the dense model and MoE model used in this paper.
Both models are pretrained on DCLM before any continued training comparison. The dense model uses the layer dimensions of Qwen2.5-0.5B~\citep{exp_qwen252024}, with 630M parameters under the specified untied embedding configuration. The MoE model scales an OLMoE configuration~\citep{exp_olmoe2024} to 2.99B total parameters. Within a model family, all continued training runs start from the same final PT weights, with a freshly initialized optimizer and no LR warmup. The implementation for the dense model follows the original MindSpore optimizer convention of omitting Adam bias correction and applying weight decay to every parameter. MoE training uses Megatron-LM v0.15.3~\citep{exp_megatron2019}, whose default decay exclusions leave normalization weights unregularized. These conventions are held fixed within each family.

The final PT checkpoint has DCLM and FineMath validation losses of 2.8147 and 2.1472 for the dense model, and 2.6824 and 2.1625 for the MoE model.

\begin{table}[!htbp]
\centering
\small
\setlength{\tabcolsep}{5pt}
\begin{tabular}{@{}p{0.38\linewidth}p{0.57\linewidth}@{}}
\toprule
Hyperparameter & Value \\
\midrule
\multicolumn{2}{@{}l}{\textit{Architecture}} \\
Reference architecture & Qwen2.5-0.5B \\
Total parameters & 630M \\
Transformer layers & 24 \\
Hidden size & 896 \\
Attention heads (query / KV) & 14 / 2 \\
Attention head dimension & 64 \\
Q/K/V biases & Enabled \\
FFN activation / width & SwiGLU / 4,864 \\
Normalization & RMSNorm, $\epsilon=10^{-6}$ \\
Positional encoding & RoPE, base 10,000 \\
Vocabulary size & 151,936 \\
Input/output embeddings & Untied \\
Tokenizer & Qwen2.5 \\
EOS / pad token IDs & 151643 / 151656 \\
\midrule
\multicolumn{2}{@{}l}{\textit{Optimization}} \\
Optimizer & AdamW, $\beta=(0.9,0.95)$, $\epsilon=10^{-8}$ \\
Adam bias correction & Disabled \\
Weight decay & 0.1 on all parameters \\
Gradient-norm clipping & 1.0 \\
Precision & BF16 autocast; FP32 master weights and optimizer states \\
Weight initialization & $\mathcal{N}(0,0.01^2)$ \\
Bias / norm initialization & Zero / one \\
\midrule
\multicolumn{2}{@{}l}{\textit{Training budgets and learning rates}} \\
Sequence length & 4,096 tokens \\
Global batch & 512 sequences ($2{,}097{,}152$ tokens/step) \\
PT budget & 95,368 steps; approximately 200B DCLM tokens \\
PT learning rate & 5,000-step linear warmup to $5\times10^{-4}$, then constant \\
Continuation budget & 10,000 steps; $20.97152$B tokens \\
MT fraction & 10\%, 25\%, or 47.75\% \\
MT token budget & 2.10B, 5.24B, or 10.01B, respectively \\
Continuation peak / final LR & $5\times10^{-4}$ / $10^{-5}$ \\
\bottomrule
\end{tabular}
\caption{Architecture and training settings for the 630M dense model. The final LR applies to schedules with decay; Const retains the peak LR.}
\label{tab:app-dense-setup}
\end{table}

\begin{table}[!htbp]
\centering
\small
\setlength{\tabcolsep}{5pt}
\begin{tabular}{@{}p{0.38\linewidth}p{0.57\linewidth}@{}}
\toprule
Hyperparameter & Value \\
\midrule
\multicolumn{2}{@{}l}{\textit{Architecture}} \\
Reference architecture & OLMoE \\
Total parameters & 2.99B \\
Active non-embedding parameters & 454M \\
Input/output embedding parameters & 155M combined \\
Transformer layers & 12 \\
Hidden size & 1,536 \\
Attention heads & 12, multi-head attention \\
Attention head dimension & 128 \\
Linear biases & Disabled \\
Experts per layer / active per token & 64 / 8; dropless, no shared experts \\
Expert FFN activation / width & SwiGLU / 768 \\
Normalization & RMSNorm and QK-norm \\
Positional encoding & RoPE, base 10,000 \\
Vocabulary size & 50,304 (padded) \\
Input/output embeddings & Untied \\
Tokenizer & OLMoE-1B-7B-0924 \\
EOS / pad token IDs & 50279 / 1 \\
\midrule
\multicolumn{2}{@{}l}{\textit{Optimization}} \\
Optimizer & AdamW, $\beta=(0.9,0.95)$, $\epsilon=10^{-8}$ \\
Weight decay & 0.1, excluding one-dimensional parameters \\
Gradient-norm clipping & 1.0 \\
Precision & BF16 \\
Weight initialization & $\mathcal{N}(0,0.02^2)$ \\
Router & Softmax in FP32; balancing coefficient 0.01; z-loss coefficient 0.001 \\
Implementation & Megatron-LM v0.15.3; local transformer, FlashAttention, PyTorch norms, sequential experts \\
\midrule
\multicolumn{2}{@{}l}{\textit{Training budgets and learning rates}} \\
Sequence length & 4,096 tokens \\
Global batch & 1,024 sequences ($4{,}194{,}304$ tokens/step) \\
PT budget & 9,536 steps; approximately 40B DCLM tokens \\
PT learning rate & 238-step warmup to $4\times10^{-4}$, then constant \\
Continuation budget & 954 steps (4B tokens) or 2,384 steps (10B tokens) \\
MT fraction & 25\% in the base configuration \\
Continuation peak / final LR & $4\times10^{-4}$ / $8\times10^{-6}$ \\
\bottomrule
\end{tabular}
\caption{Architecture and training settings for the 2.99B-parameter MoE model. Unlike the original OLMoE recipe, PT retains a constant LR after warmup, and normalization weights have no weight decay. The final LR applies only to schedules with decay.}
\label{tab:app-moe-setup}
\end{table}

\paragraph{Fixed data budgets.}
Let $T$ be the number of continued training steps, $\alpha$ the fraction of its token budget allocated to MT, and $\gamma_{\mathrm{end}}$ the final MT ratio, as in the main text.
Write $\gamma(t)$ for the MT fraction, with $\int_0^T\gamma(t)\,\mathrm{d}t=\alpha T$.
For Two-Stage (TS),
\begin{equation*}
\gamma_{\mathrm{TS}}(t)=
\begin{cases}
0, & 0\leq t<t_0,\\
\gamma_{\mathrm{end}}, & t_0\leq t\leq T,
\end{cases}
\qquad
\frac{t_0}{T}=1-\frac{\alpha}{\gamma_{\mathrm{end}}},\quad \alpha\leq\gamma_{\mathrm{end}}\leq1.
\end{equation*}
Zero-then-Linear (ZL) linearly increases the MT fraction after an initial PT-only interval:
\begin{equation*}
\gamma_{\mathrm{ZL}}(t)=
\begin{cases}
0, & 0\leq t<t_0,\\
\gamma_{\mathrm{end}}\dfrac{t-t_0}{T-t_0}, & t_0\leq t\leq T,
\end{cases}
\qquad
\frac{t_0}{T}=1-\frac{2\alpha}{\gamma_{\mathrm{end}}}.
\end{equation*}
ZL is feasible when $\alpha\leq\gamma_{\mathrm{end}}/2$, equivalently $2\alpha\leq\gamma_{\mathrm{end}}\leq1$.
At a shared $(\alpha,\gamma_{\mathrm{end}})$, its MT data window $L:=T-t_0$ is twice the TS window.
Changing $\gamma_{\mathrm{end}}$ therefore changes both the final mixture and the time when MT data first enter.
At the respective minimum feasible final MT ratio, $t_0=0$ and the initial PT-only interval disappears.

To determine each batch's composition, we calculate the total number of PT sequences prescribed by the schedule up to that batch and round it up. The increase from the previous batch gives the PT count for the current batch, with the remaining sequences drawn from MT data. Sequences are assigned to workers in a fixed repeating order. This makes batch composition deterministic, avoids random fluctuations in the mixing ratio, and allows the allocation to be reconstructed from the current training step. Reported switch times and LR decay durations use the actual integer step counts.

\paragraph{Learning rate windows.}
For a positive decay length $W$, LR stays at $\eta_{\max}$ until $T-W$ and then decreases linearly to $\eta_{\min}=\eta_{\max}/50$:
\begin{equation*}
\eta_W(t)=
\begin{cases}
\eta_{\max}, & 0\leq t<T-W,\\
\eta_{\min}+(\eta_{\max}-\eta_{\min})\dfrac{T-t}{W}, & T-W\leq t\leq T.
\end{cases}
\end{equation*}
\textbf{Const} keeps $\eta_{\max}$ throughout; it is plotted at zero window length and does not decay to $\eta_{\min}$.
\textbf{Full-Decay-Range (FDR)} uses $W=T$.
\textbf{Matched-Decay-Range (MDR)} uses $W=L=T-t_0$, beginning decay when MT data first enter.
The scans vary $r:=W/L$ with the data schedule fixed.
The denominator is the entire interval \emph{from first MT exposure to training end}, including the low-MT-density part of a ZL ramp; it is neither $T$ nor the MT budget expressed in training steps, $\alpha T$.
A 20\% window means $W=0.2L$, and a 120\% window begins decay before MT exposure.
When $t_0=0$, MDR and FDR coincide and use one run.

\begin{figure}[!htbp]
    \centering
    \includegraphics[width=\linewidth]{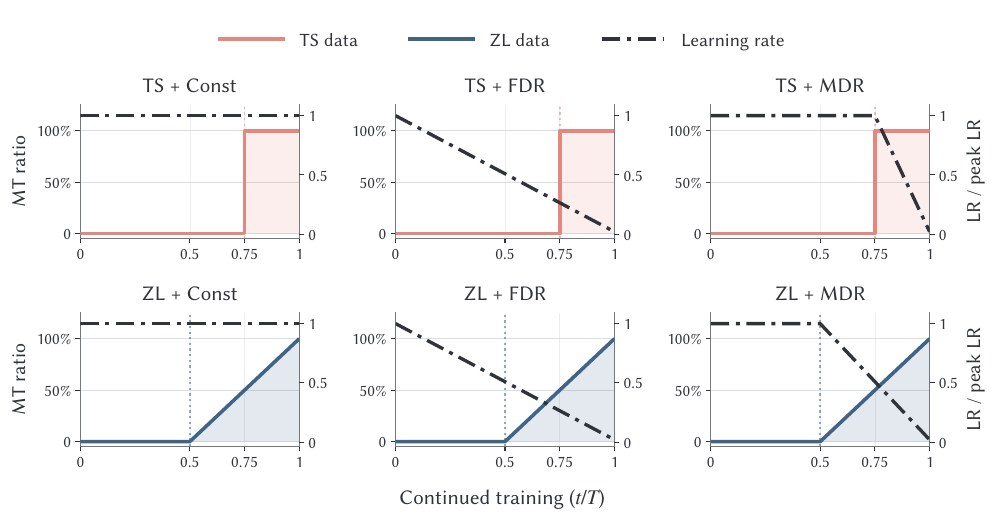}
    \caption{\textbf{Illustration of different LR schedules and data schedules.} Rows show TS and ZL, and columns show Const, FDR, and MDR. Solid lines and shading show the MT ratio on the left axes. Dash dotted lines show LR relative to its peak on the right axes. Both data schedules use a 25\% MT budget and end at 100\% MT, so ZL introduces MT earlier. MDR therefore begins LR decay earlier for ZL than for TS. Time starts at the shared PT checkpoint.}
    \label{fig:exp-schedule-pairs}
\end{figure}

\paragraph{Validation and comparability.}
All final PT/MT loss comparisons use the average cross entropy per token on the validation set, evaluated at the final checkpoint.
Final evaluation for the dense model uses the full held-out sets, approximately 560M DCLM and 160M FineMath tokens; intermediate evaluations use 32M tokens every 500 steps.
MoE final evaluation uses a separate skip-training pass on each domain, with $32\times1024\times4096=134{,}217{,}728$ tokens per domain.
Because the two families use different tokenizers and training implementations, their absolute loss values are not directly comparable.
Cross-family comparisons concern schedule ordering and the shape of the response to LR changes.

\section{Complete Data Schedule Results for the Dense Model}
\label{app:exp-dense}

\Cref{tab:app-dense-grid} lists all tested final MT ratios, each crossed with Const, FDR, and MDR, with identical boundary schedules run only once.
The continued training token budget remains $20.97152$B across all three MT fractions.
The largest fraction is 47.75\%, rather than 50\%, to use approximately one pass through FineMath: 2,444,800 of 2,445,076 packed rows.
At this fraction, TS requires $\gamma_{\mathrm{end}}\geq47.75\%$, and ZL requires $\gamma_{\mathrm{end}}\geq95.5\%$.
For ZL, we test five final MT ratios across its narrow feasible range, from 95.5\% to 100\%.

\begin{table}[!htbp]
\centering
\small
\setlength{\tabcolsep}{5pt}
\begin{tabular}{@{}lll@{}}
\toprule
MT budget & TS final MT ratios (\%) & ZL final MT ratios (\%) \\
\midrule
10\% & 10, 20, 40, 80, 100 & 20, 40, 60, 80, 100 \\
25\% & 25, 40, 50, 70, 75, 85, 100 & 50, 60, 70, 80, 90, 100 \\
47.75\% & 48, 50, 60, 70, 80, 90, 95, 100 & 95.5, 97, 98, 99, 100 \\
\bottomrule
\end{tabular}
\caption{Complete data schedule runs for the dense model. Total and MT token budgets are fixed within each group.}
\label{tab:app-dense-grid}
\end{table}

\begin{figure}[!htbp]
\centering
\includegraphics[width=\linewidth]{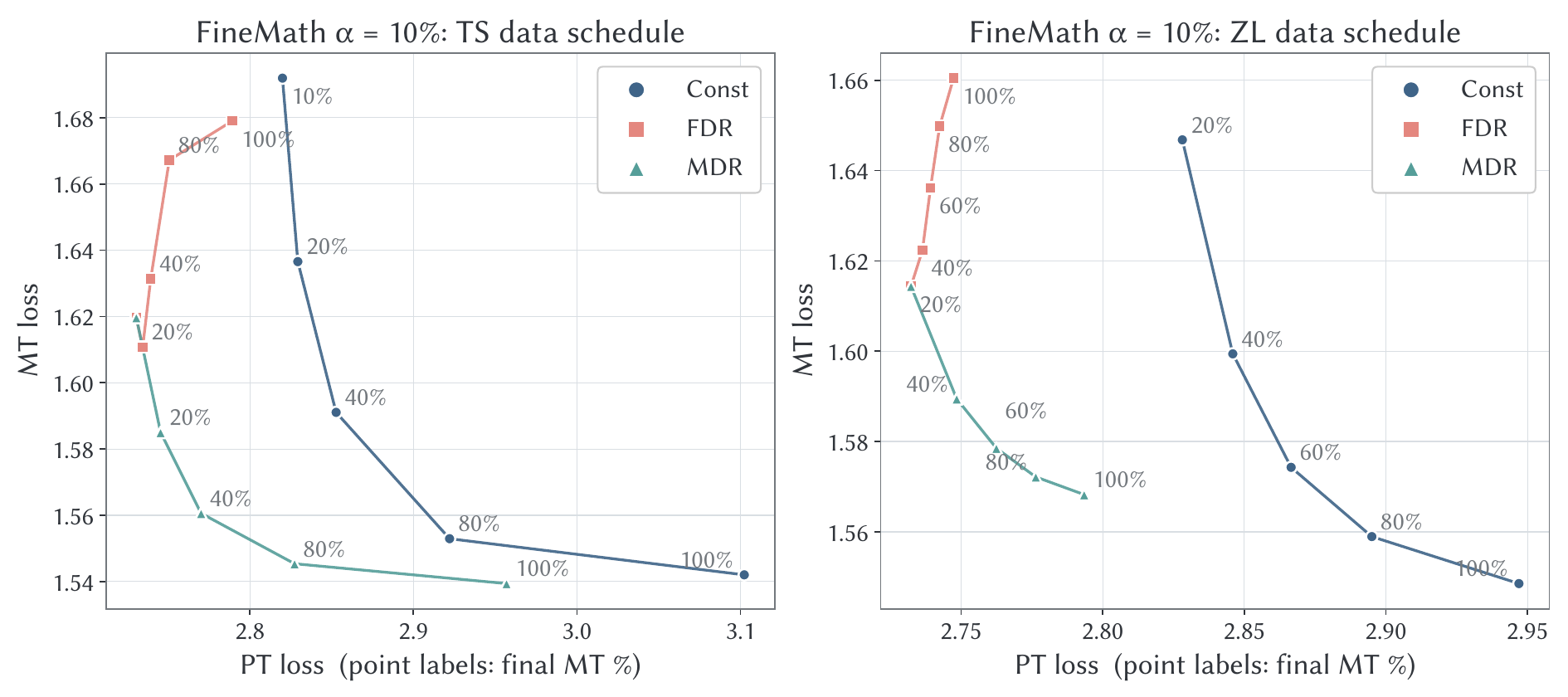}
\par
\includegraphics[width=0.49\linewidth]{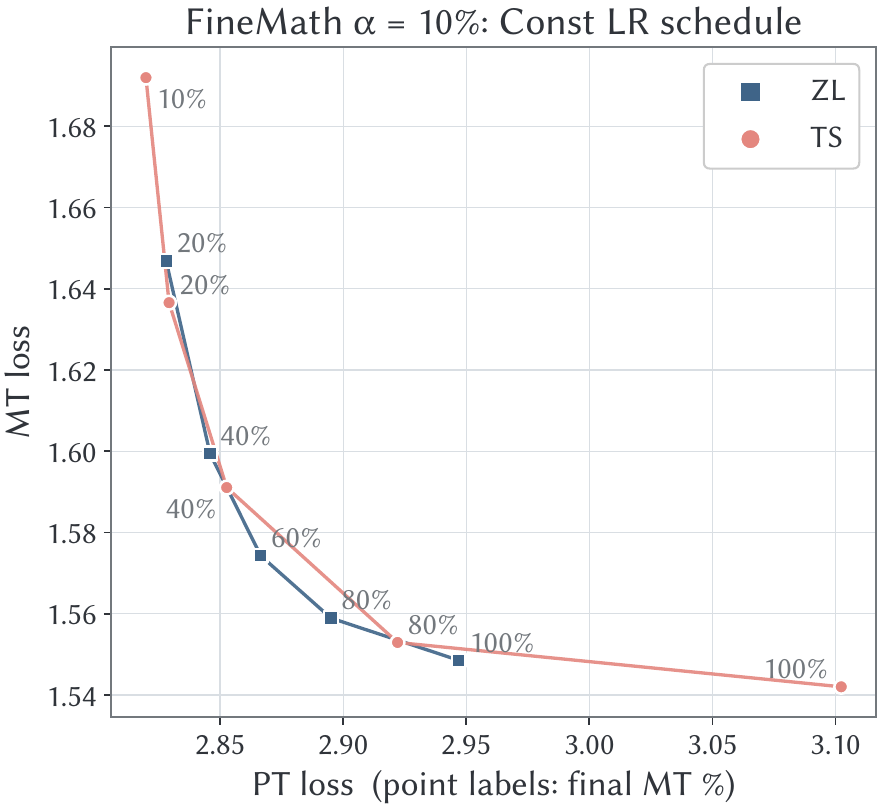}
\hfill
\includegraphics[width=0.49\linewidth]{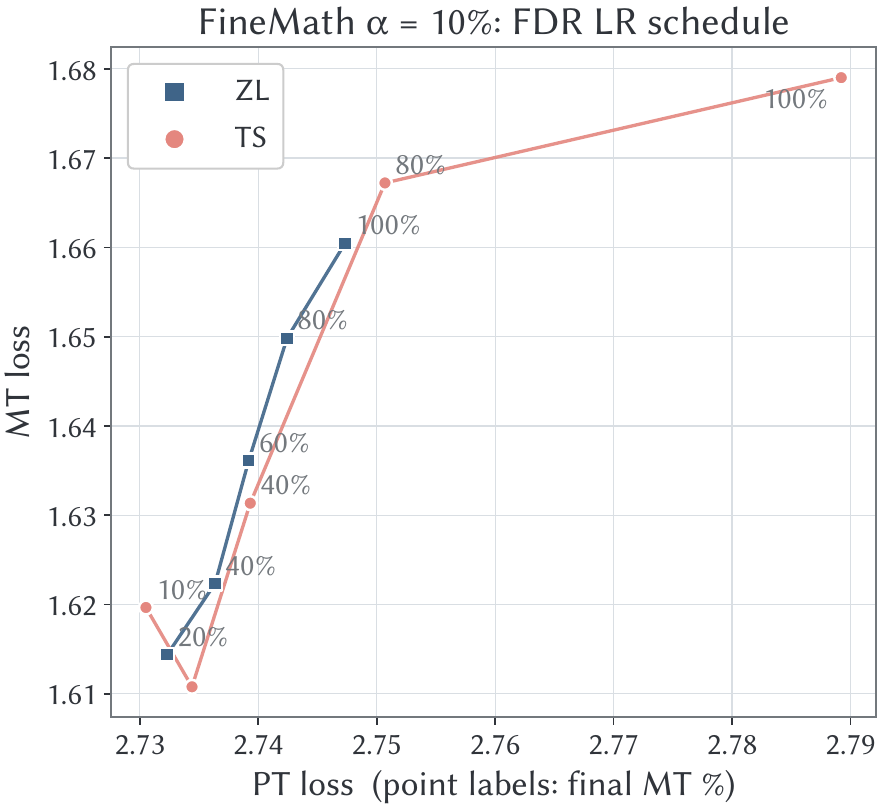}
\par
\includegraphics[width=0.49\linewidth]{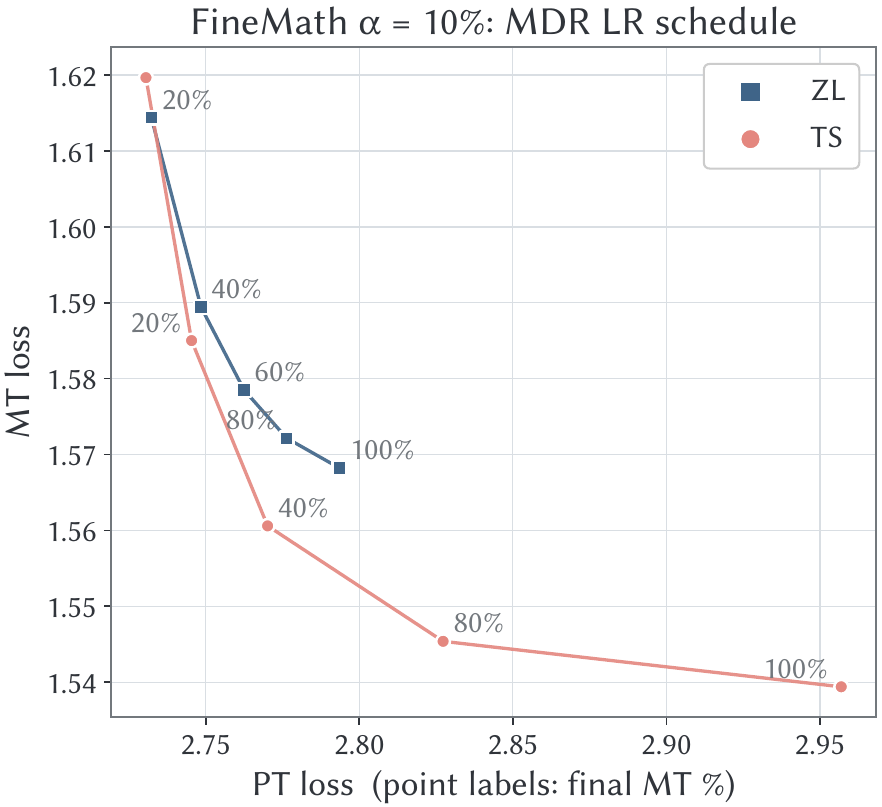}
\caption{\textbf{FDR favors earlier MT mixing, while Const and MDR favor ending with MT data alone.} For both TS and ZL, the lowest tested MT loss occurs at a final MT ratio of 20\% under FDR and 100\% under Const and MDR. These dense model runs use a 10\% MT budget over 20.97B continued training tokens, including 2.10B FineMath tokens.}
\label{fig:app-dense10-grid}
\end{figure}

\begin{figure}[!htbp]
\centering
\includegraphics[width=\linewidth]{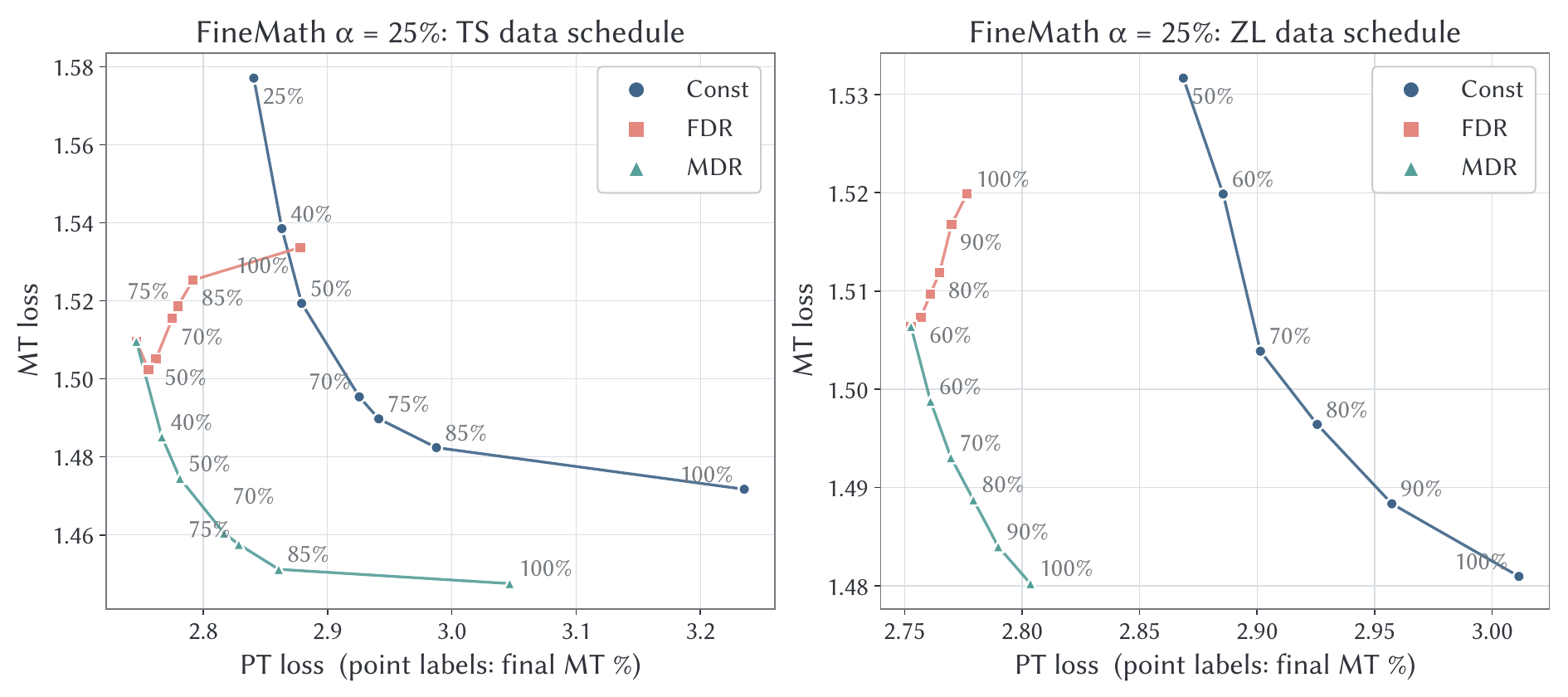}
\par
\includegraphics[width=0.49\linewidth]{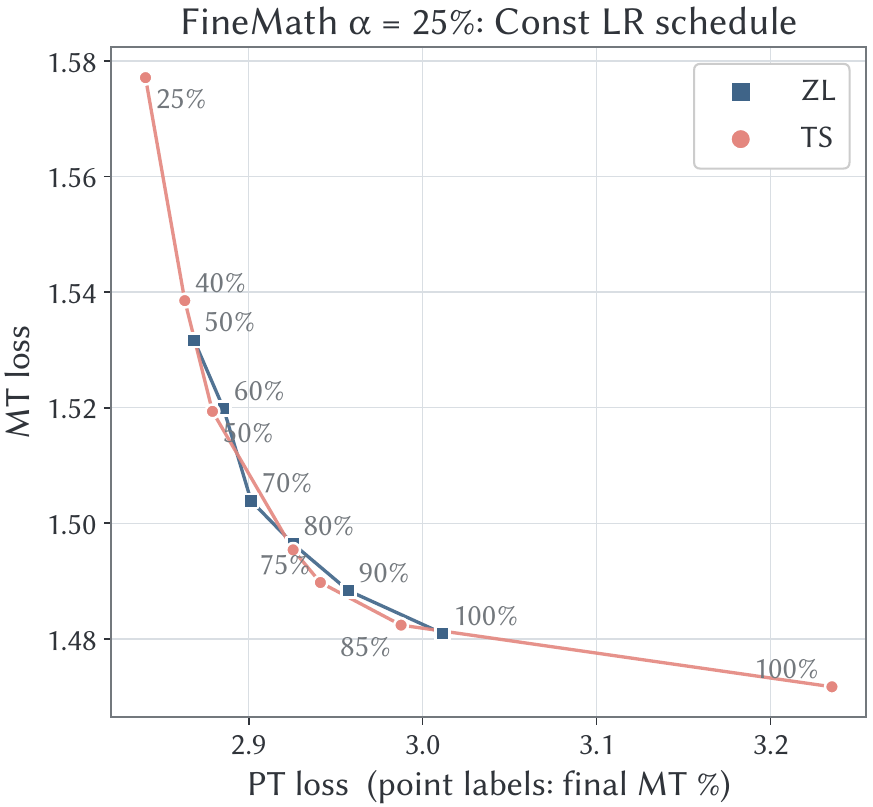}
\hfill
\includegraphics[width=0.49\linewidth]{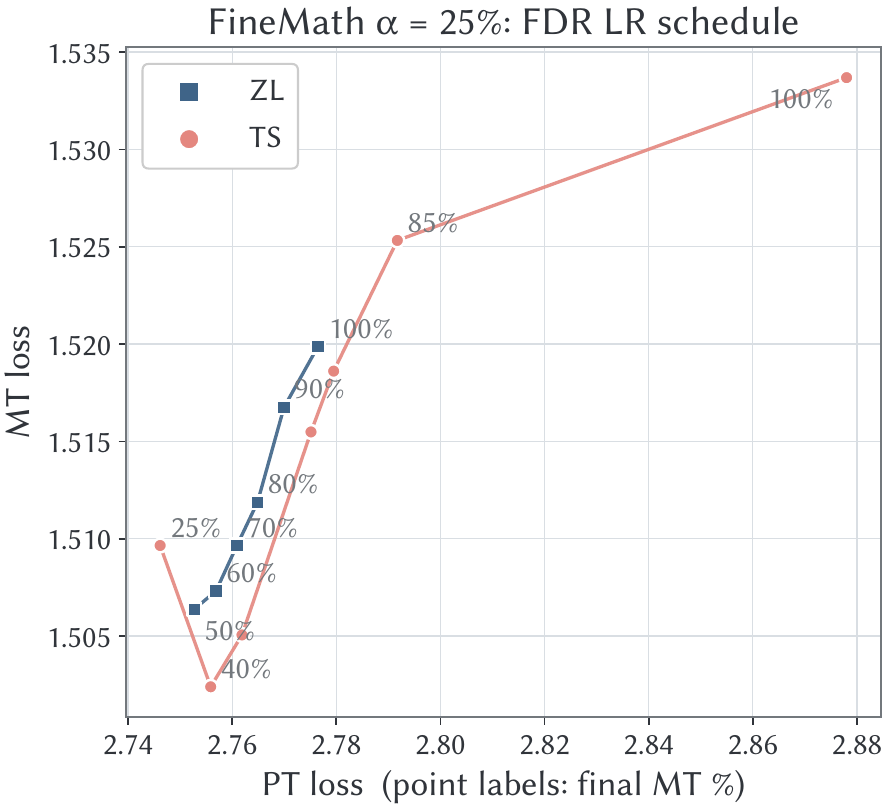}
\par
\includegraphics[width=0.49\linewidth]{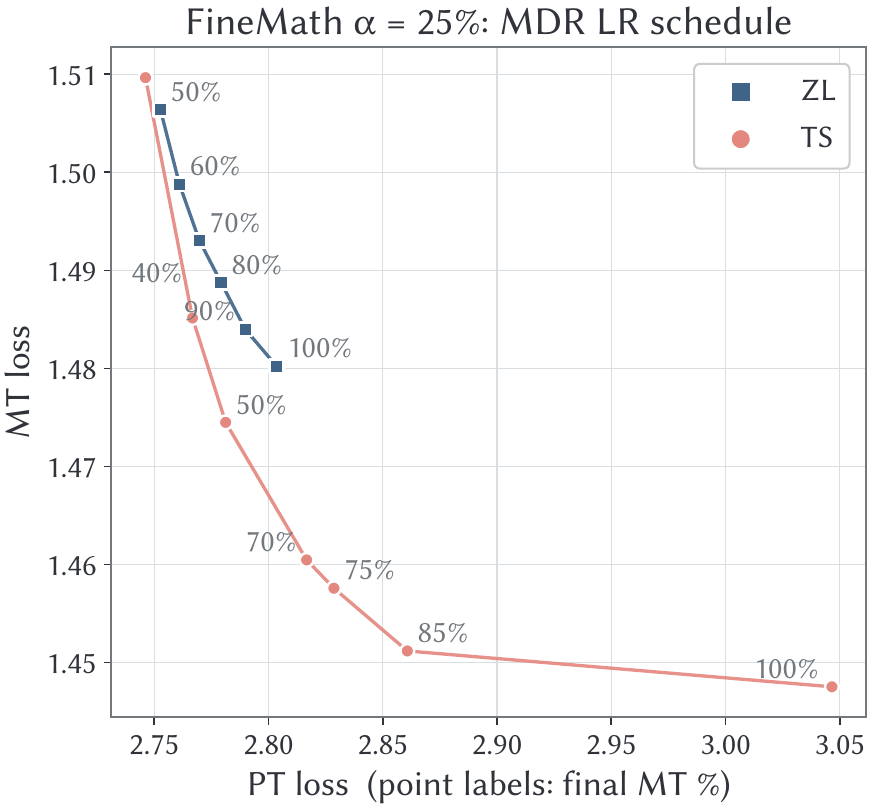}
\caption{\textbf{Changing the LR schedule changes the preferred final MT ratio.} FDR gives the lowest tested MT loss at 40\% MT for TS and 50\% for ZL; Const and MDR favor 100\% for both. These dense model runs use a 25\% MT budget over 20.97B continued training tokens, including 5.24B FineMath tokens.}
\label{fig:app-dense25-grid}
\end{figure}

\begin{figure}[!htbp]
\centering
\includegraphics[width=\linewidth]{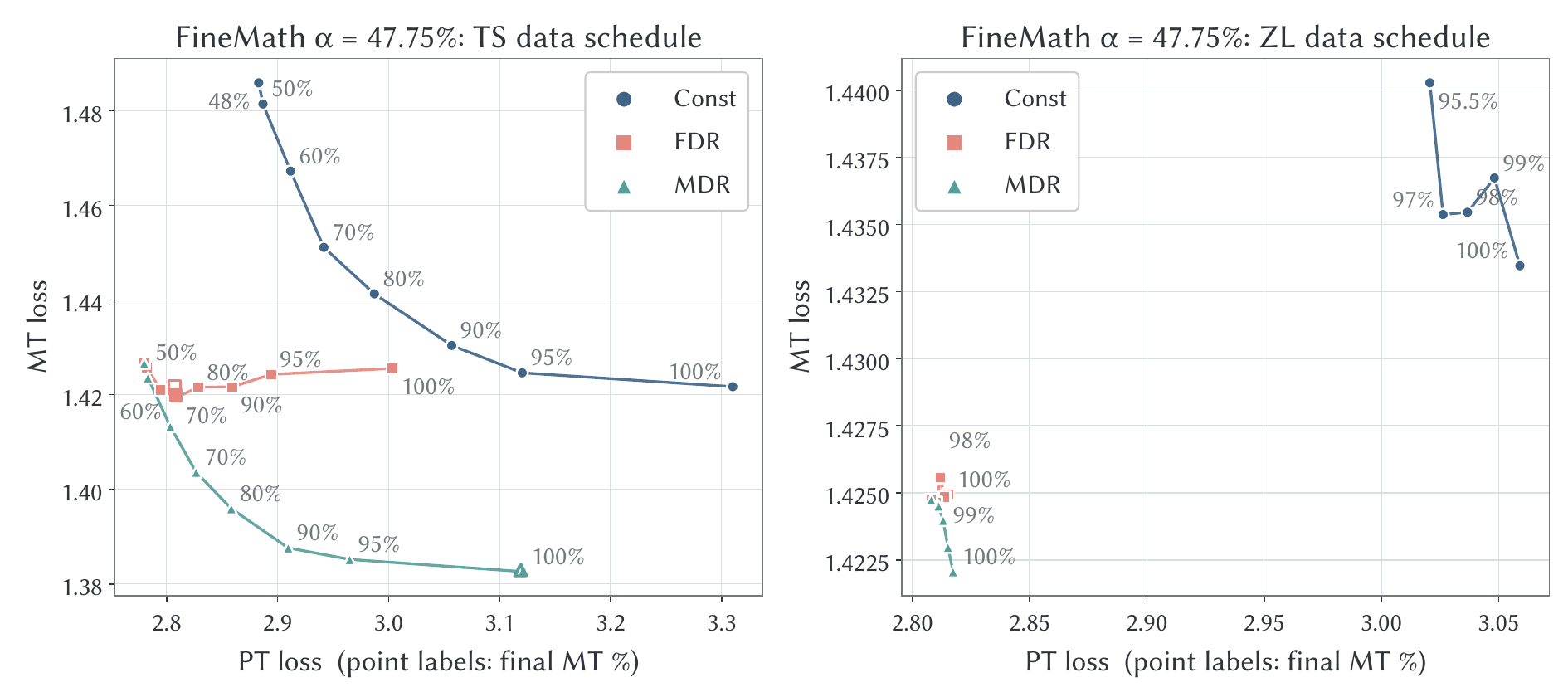}
\par
\includegraphics[width=0.49\linewidth]{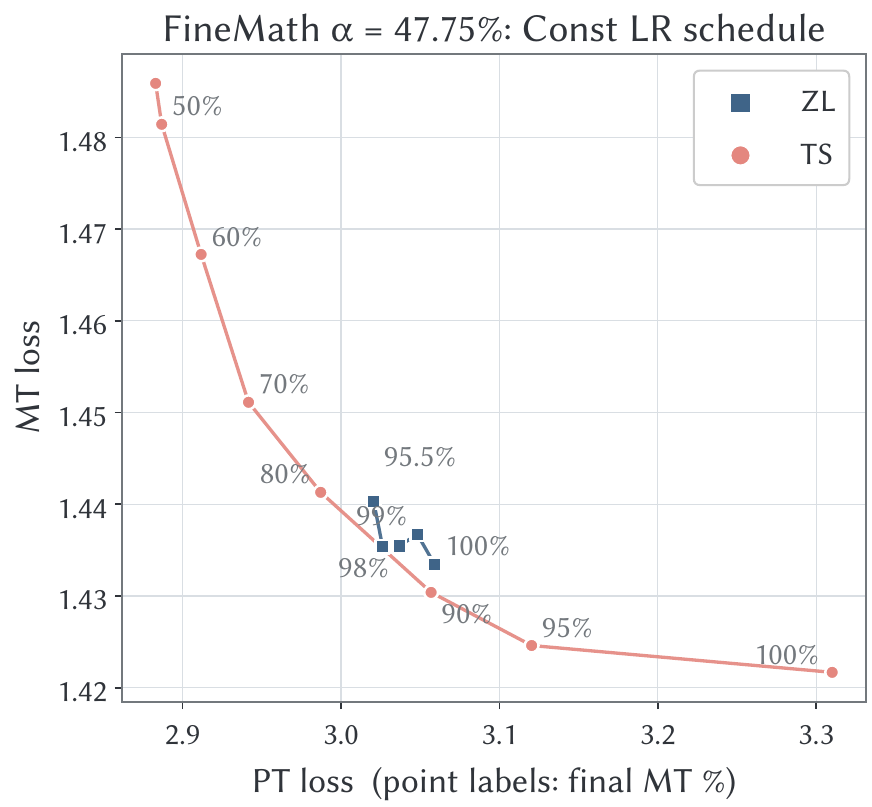}
\hfill
\includegraphics[width=0.49\linewidth]{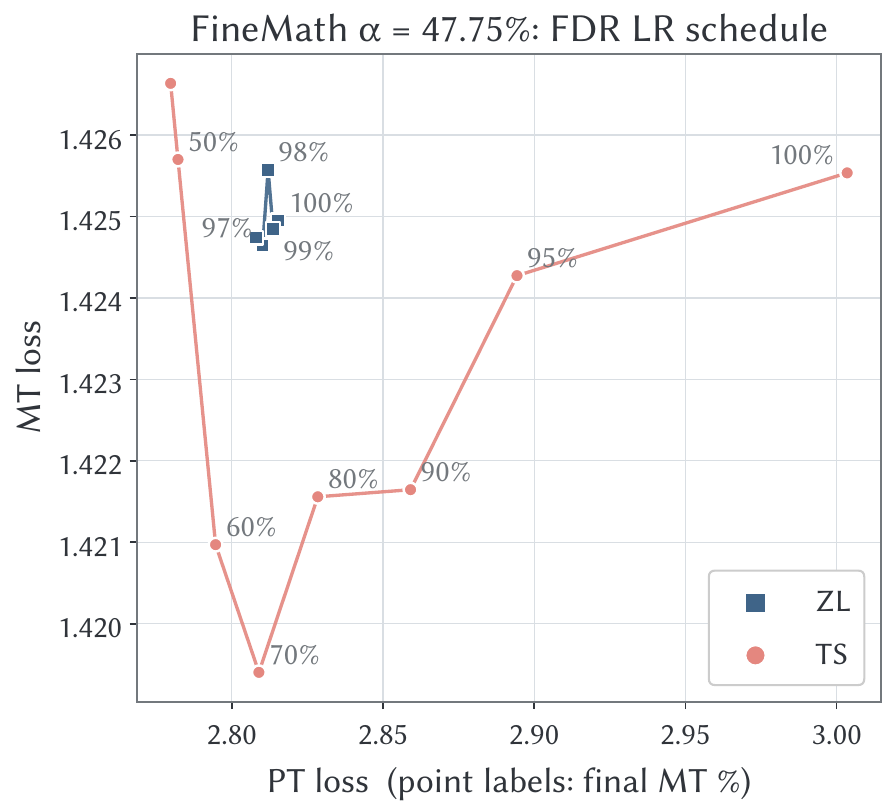}
\par
\includegraphics[width=0.49\linewidth]{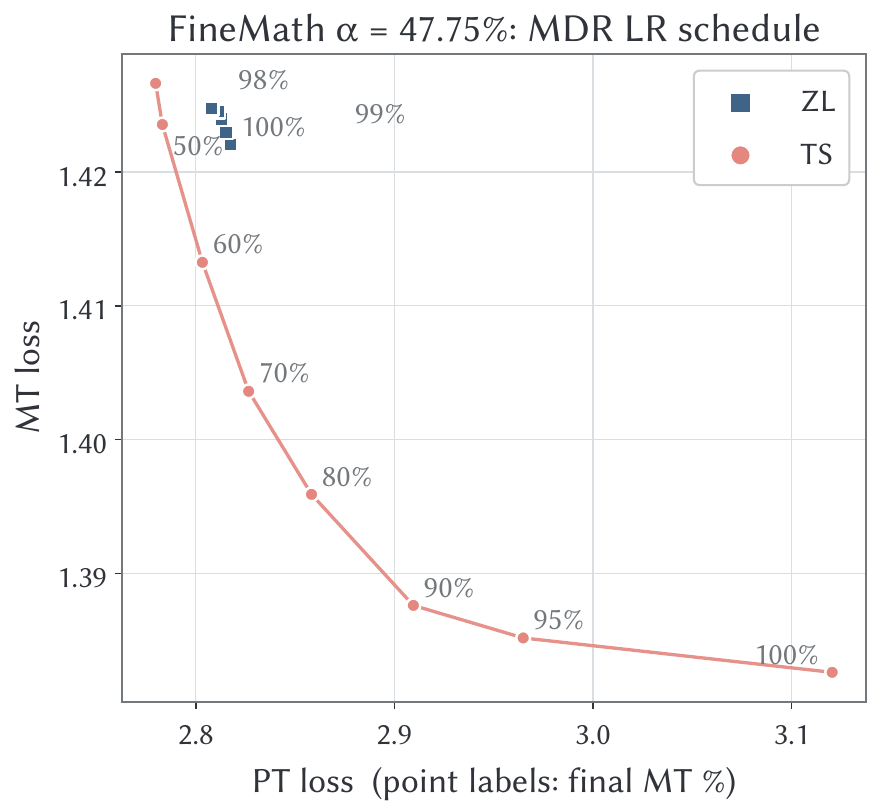}
\caption{\textbf{ZL losses under FDR change little across its tested final MT ratios.} They vary by only 0.001, while Const and MDR favor a final MT ratio of 100\%. TS covers eight final MT ratios; ZL covers five from 95.5\% to 100\%. These dense model runs use a 47.75\% MT budget over 20.97B continued training tokens, including 10.01B FineMath tokens. Hollow overlays show selected data sampling repeats.}
\label{fig:app-dense4775-grid}
\end{figure}

\Cref{fig:app-dense10-grid,fig:app-dense25-grid,fig:app-dense4775-grid} show the final PT and MT losses for all tested configurations.
\Cref{tab:app-dense-minima} gives the best MT loss in each sampled data/LR family.
For TS, the best tested final MT ratio under Const and MDR is 100\% at every budget, whereas FDR selects 20\%, 40\%, and 70\% as the MT budget increases.
The FDR loss range contracts from approximately 0.068 to 0.031 to 0.007.
Lowering $\gamma_{\mathrm{end}}$ does not always improve the FDR loss: at 10\%, loss rises from 1.6108 at $\gamma_{\mathrm{end}}=20\%$ to 1.6197 at $\gamma_{\mathrm{end}}=10\%$; at 25\%, it rises from 1.5024 at $\gamma_{\mathrm{end}}=40\%$ to 1.5097 at $\gamma_{\mathrm{end}}=25\%$.
At 47.75\%, the FDR minimum at $\gamma_{\mathrm{end}}=70\%$ is only 0.0016 below the loss at $\gamma_{\mathrm{end}}=60\%$, so the broad flattening is more informative than a precise preferred endpoint.

\begin{table}[!htbp]
\centering
\small
\setlength{\tabcolsep}{6pt}
\begin{tabular}{@{}llccc@{}}
\toprule
MT budget & Data schedule & Const & FDR & MDR \\
\midrule
10\% & TS & 1.5420 (100) & 1.6108 (20) & 1.5394 (100) \\
 & ZL & 1.5485 (100) & 1.6144 (20) & 1.5682 (100) \\
25\% & TS & 1.4717 (100) & 1.5024 (40) & 1.4475 (100) \\
 & ZL & 1.4810 (100) & 1.5064 (50) & 1.4802 (100) \\
47.75\% & TS & 1.4217 (100) & 1.4194 (70) & 1.3826 (100) \\
 & ZL & 1.4335 (100) & 1.4246 (97) & 1.4221 (100) \\
\bottomrule
\end{tabular}
\caption{Lowest final MT loss among the tested final MT ratios for each MT budget, data schedule, and LR schedule. Parentheses give the final MT ratio $\gamma_{\mathrm{end}}$ in percent.}
\label{tab:app-dense-minima}
\end{table}

\paragraph{TS versus ZL depends on the LR rule and final ratio.}
\Cref{tab:app-dense-shape-difference} compares the shapes at the same final MT ratio of 100\%.
Here ZL gives lower observed MT loss under FDR, while TS gives lower loss under MDR and Const.
This differs from selecting each shape's best final ratio: TS then gives lower FDR loss at all three MT budgets for the dense model (\Cref{tab:app-dense-minima}).
Selecting each shape's best final ratio also favors TS under FDR in all three main MoE settings (\Cref{tab:app-moe-minima,fig:app-moe4775-grid}).
The points labeled 100\% in \Cref{fig:app-dense10-grid,fig:app-dense25-grid} show the first comparison, while each curve's minimum gives the second.
Under MDR, changing from TS to ZL also starts LR decay earlier, so this comparison does not isolate data shape at a fixed LR trajectory.
MT gains can also come at a PT cost. At the 25\% budget under Const, TS has lower MT loss than ZL but substantially higher PT loss.

To quantify the LR applied to MT data, define $E=\int_0^T\eta(t)\gamma(t)\,\mathrm{d}t$ for an LR trajectory $\eta(t)$.
This quantity weights each MT token by the LR at which it is used.
For TS and ZL with the same budget $\alpha$ and final MT ratio $\gamma_{\mathrm{end}}$, let $u=\alpha/\gamma_{\mathrm{end}}$ and $\rho=\eta_{\min}/\eta_{\max}$.
Integrating the schedules in Section~\ref{sec:exp-setup} gives
\[
\frac{E_{\mathrm{ZL}}}{E_{\mathrm{TS}}}=
\begin{cases}
1, & \mathrm{Const},\\[2pt]
\dfrac{\rho+\frac{2}{3}(1-\rho)u}{\rho+\frac{1}{2}(1-\rho)u}, & \mathrm{FDR},\\[6pt]
\dfrac{2(1+2\rho)}{3(1+\rho)}, & \mathrm{MDR}.
\end{cases}
\]
Both models use $\rho=0.02$, giving an MDR ratio of approximately 0.68 and FDR ratios of approximately 1.24 to 1.31 across settings where TS and ZL have the same MT budget and final MT ratio.
These are continuous time calculations, with small differences possible from discrete step rounding.
They describe the LR assigned to MT tokens, but omit the effects of later updates and optimization noise and do not by themselves predict final loss.
In particular, equal exposure under Const does not imply equal loss, and ZL can have higher FDR loss despite greater exposure, as in the dense model at a 10\% MT budget with $\gamma_{\mathrm{end}}=20\%$.
The role of LR in learning from late MT data is consistent with \citet{luo2026bestdata}, who find that strong LR decay can reduce the benefit of valuable examples encountered late in training.

\begin{table}[!htbp]
\centering
\small
\begin{tabular}{@{}lrrr@{}}
\toprule
MT budget & Const & FDR & MDR \\
\midrule
10\% & $+0.0065$ & $-0.0185$ & $+0.0288$ \\
25\% & $+0.0093$ & $-0.0138$ & $+0.0327$ \\
47.75\% & $+0.0118$ & $-0.0006$ & $+0.0395$ \\
\bottomrule
\end{tabular}
\caption{MT loss difference in the dense model, ZL minus TS, with both schedules ending at 100\% MT. Negative values favor ZL. Differences use the reported final losses and are not significance estimates.}
\label{tab:app-dense-shape-difference}
\end{table}

\clearpage

\section{Learning Rate Decay Duration Scans}
\label{app:exp-decay}

Following Section~\ref{sec:exp-decay}, we keep each data schedule fixed and vary when LR decay begins.
The LR remains at the same peak until decay begins, then decreases linearly to the same final value at training completion.
The \emph{relative LR decay duration} expresses this decay duration as a percentage of the MT data window, from MT entry time to training completion.
For example, a relative duration of 20\% means that LR decay covers the final 20\% of this interval. Const corresponds to 0\%, and MDR to 100\%. Values above 100\% start LR decay before MT entry time.

\Cref{tab:app-decay-grid} lists the tested relative durations for all $19$ scans, and \Cref{tab:app-decay-results} summarizes their lowest observed losses.
Each scan includes Const and MDR as baselines. We also test FDR, which decays the LR throughout continued training, except for StarCoder ZL.

\paragraph{A delayed LR decay gives the lowest tested loss in all nineteen scans.}
For the dense model, the best tested relative durations are 40\% for TS and 20\% for ZL. For the MoE model, they are 20\% for TS and 10\% for ZL.
They improve on both Const and MDR, with reductions in MT loss relative to MDR ranging from 0.0094 to 0.0660.
These results support keeping the LR high after MT entry time and delaying its decay, as discussed in Section~\ref{sec:exp-decay}.
The preferred relative durations refer to the tested values. In particular, the scans from the two earlier MoE PT checkpoints test only 20\% and 40\% in addition to Const, MDR, and FDR, so they do not locate the optimum more precisely.

\begin{table}[!htbp]
\centering
\small
\setlength{\tabcolsep}{4pt}
\begin{tabular}{@{}llrrl@{}}
\toprule
MT budget & Schedule & $\gamma_{\mathrm{end}}$ & \shortstack{MT window\\(steps)} & \shortstack[l]{Additional relative LR\\decay durations (\%)} \\
\midrule
\multicolumn{5}{@{}l}{\textit{Dense: 200B-token PT; 20.97B-token continuation; FineMath}} \\
10\% & TS & 100\% & 1,000 & 20, 40, 50, 80, 120 \\
 & TS & 80\% & 1,250 & 20, 40, 80, 90, 110, 120 \\
 & ZL & 100\% & 2,000 & 5, 10, 20, 40, 80, 90, 110, 120 \\
25\% & TS & 100\% & 2,500 & 20, 40, 80, 120 \\
 & ZL & 100\% & 5,000 & 5, 10, 20, 40, 80 \\
47.75\% & TS & 100\% & 4,775 & 20, 40, 80, 120 \\
 & TS & 80\% & 5,969 & 20, 40, 80, 120 \\
 & TS & 60\% & 7,958 & 20, 40, 80 \\
 & ZL & 100\% & 9,550 & 5, 10, 20, 40, 80 \\
\midrule
\multicolumn{5}{@{}l}{\textit{MoE: 40B-token PT; 4B-token continuation; FineMath}} \\
25\% & TS & 100\% & 238 & 5, 10, 20, 40, 80, 120 \\
 & ZL & 100\% & 477 & 5, 10, 20, 40, 80 \\
\midrule
\multicolumn{5}{@{}l}{\textit{MoE: 40B-token PT; 10B-token continuation; FineMath}} \\
25\% & TS & 100\% & 596 & 5, 10, 20, 40, 80, 120 \\
 & ZL & 100\% & 1,192 & 5, 10, 20, 40, 80 \\
47.75\% & TS & 100\% & 1,138 & 5, 10, 20, 40, 80, 120 \\
 & ZL & 100\% & 2,277 & 5, 10, 20, 40, 80 \\
\midrule
\multicolumn{5}{@{}l}{\textit{MoE: 20B-token PT; 4B-token continuation; FineMath}} \\
25\% & TS & 100\% & 238 & 20, 40 \\
\midrule
\multicolumn{5}{@{}l}{\textit{MoE: 30B-token PT; 4B-token continuation; FineMath}} \\
25\% & TS & 100\% & 238 & 20, 40 \\
\midrule
\multicolumn{5}{@{}l}{\textit{MoE: 40B-token PT; 10B-token continuation; StarCoderData Python}} \\
25\% & TS & 100\% & 596 & 5, 10, 20, 40, 80, 120 \\
 & ZL & 100\% & 1,192 & 10, 20, 40, 80 \\
\bottomrule
\end{tabular}
\caption{LR decay scan settings. Relative LR decay durations are percentages of each schedule's MT data window. Const (0\%) and MDR (100\%) are tested in every scan; FDR is also tested except for StarCoderData ZL.}
\label{tab:app-decay-grid}
\end{table}

\begin{table}[!htbp]
\centering
\small
\setlength{\tabcolsep}{4pt}
\begin{tabular}{@{}llrrrrr@{}}
\toprule
MT budget & Schedule, $\gamma_{\mathrm{end}}$ & \shortstack{Best relative LR\\decay duration} & Best loss & Const & MDR & \shortstack{Gain over\\MDR} \\
\midrule
\multicolumn{7}{@{}l}{\textit{Dense: 200B-token PT; 20.97B-token continuation; FineMath}} \\
10\% & TS, 100\% & 40\% & 1.5156 & 1.5420 & 1.5394 & 0.0238 \\
 & TS, 80\% & 40\% & 1.5215 & 1.5529 & 1.5454 & 0.0239 \\
 & ZL, 100\% & 20\% & 1.5203 & 1.5485 & 1.5682 & 0.0479 \\
25\% & TS, 100\% & 40\% & 1.4275 & 1.4717 & 1.4475 & 0.0200 \\
 & ZL, 100\% & 20\% & 1.4392 & 1.4810 & 1.4802 & 0.0410 \\
47.75\% & TS, 100\% & 40\% & 1.3690 & 1.4217 & 1.3826 & 0.0136 \\
 & TS, 80\% & 40\% & 1.3831 & 1.4413 & 1.3959 & 0.0128 \\
 & TS, 60\% & 40\% & 1.4039 & 1.4672 & 1.4133 & 0.0094 \\
 & ZL, 100\% & 20\% & 1.3857 & 1.4335 & 1.4221 & 0.0364 \\
\midrule
\multicolumn{7}{@{}l}{\textit{MoE: 40B-token PT; 4B-token continuation; FineMath}} \\
25\% & TS, 100\% & 20\% & 1.6080 & 1.6176 & 1.6486 & 0.0406 \\
 & ZL, 100\% & 10\% & 1.6000 & 1.6131 & 1.6586 & 0.0586 \\
\midrule
\multicolumn{7}{@{}l}{\textit{MoE: 40B-token PT; 10B-token continuation; FineMath}} \\
25\% & TS, 100\% & 20\% & 1.4853 & 1.5110 & 1.5225 & 0.0372 \\
 & ZL, 100\% & 10\% & 1.4872 & 1.5101 & 1.5508 & 0.0636 \\
47.75\% & TS, 100\% & 20\% & 1.4007 & 1.4292 & 1.4390 & 0.0383 \\
 & ZL, 100\% & 10\% & 1.4110 & 1.4424 & 1.4770 & 0.0660 \\
\midrule
\multicolumn{7}{@{}l}{\textit{MoE: 20B-token PT; 4B-token continuation; FineMath}} \\
25\% & TS, 100\% & 20\% & 1.6562 & 1.6666 & 1.6969 & 0.0407 \\
\midrule
\multicolumn{7}{@{}l}{\textit{MoE: 30B-token PT; 4B-token continuation; FineMath}} \\
25\% & TS, 100\% & 20\% & 1.6260 & 1.6409 & 1.6673 & 0.0413 \\
\midrule
\multicolumn{7}{@{}l}{\textit{MoE: 40B-token PT; 10B-token continuation; StarCoderData Python}} \\
25\% & TS, 100\% & 20\% & 0.9317 & 0.9468 & 0.9594 & 0.0277 \\
 & ZL, 100\% & 10\% & 0.9332 & 0.9476 & 0.9762 & 0.0430 \\
\bottomrule
\end{tabular}
\caption{Final MT losses in all 19 scans. Best loss and its relative LR decay duration refer to the lowest-loss primary run among the tested schedules. Gain over MDR is MDR minus best loss, calculated from the displayed values. Group headings specify the model, PT budget, continuation budget, and MT dataset.}
\label{tab:app-decay-results}
\end{table}

\Cref{tab:app-decay-robustness} reports every sampled MT loss for four further MoE configurations: TS and ZL over 10B continuation tokens with a 47.75\% MT budget, and TS over 4B continuation tokens with a 25\% MT budget starting from 20B-token or 30B-token PT checkpoints.
The best tested relative LR decay durations remain 20\% for TS and 10\% for ZL.
The two earlier PT checkpoints confirm that delaying LR decay improves on Const and MDR, although testing only two additional durations leaves the shape of the loss curve near its minimum uncertain.

\begin{table}[!htbp]
\centering
\small
\setlength{\tabcolsep}{8pt}
\begin{tabular}{@{}lrrrr@{}}
\toprule
 & \multicolumn{2}{c}{\shortstack{MoE: 10B continuation tokens\\47.75\% MT}} & \multicolumn{2}{c}{\shortstack{MoE: 4B continuation tokens\\25\% MT, TS}} \\
\cmidrule(lr){2-3}\cmidrule(l){4-5}
\shortstack[l]{Relative LR decay\\duration or rule} & TS & ZL & PT 20B & PT 30B \\
\midrule
Const (0\%) & 1.4292 & 1.4424 & 1.6666 & 1.6409 \\
5\% & 1.4091 & 1.4131 & --- & --- \\
10\% & 1.4046 & \textbf{1.4110} & --- & --- \\
20\% & \textbf{1.4007} & 1.4137 & \textbf{1.6562} & \textbf{1.6260} \\
40\% & 1.4040 & 1.4268 & 1.6578 & 1.6303 \\
80\% & 1.4252 & 1.4609 & --- & --- \\
MDR ($100\%$) & 1.4390 & 1.4770 & 1.6969 & 1.6673 \\
120\% & 1.4534 & --- & --- & --- \\
FDR & 1.4994 & 1.4806 & 1.7899 & 1.7552 \\
\bottomrule
\end{tabular}
\caption{Final MT losses at every tested relative LR decay duration for the four additional scans, including the reference LR schedules. All data schedules end at 100\% MT. The 10B-token continuations start from the 40B-token PT checkpoint, while the last two columns use earlier checkpoints from the same PT trajectory. Bold entries mark the lowest tested loss in each column. Dashes indicate durations that were not tested.}
\label{tab:app-decay-robustness}
\end{table}

\begin{figure}[!htbp]
\centering
\includegraphics[width=\linewidth]{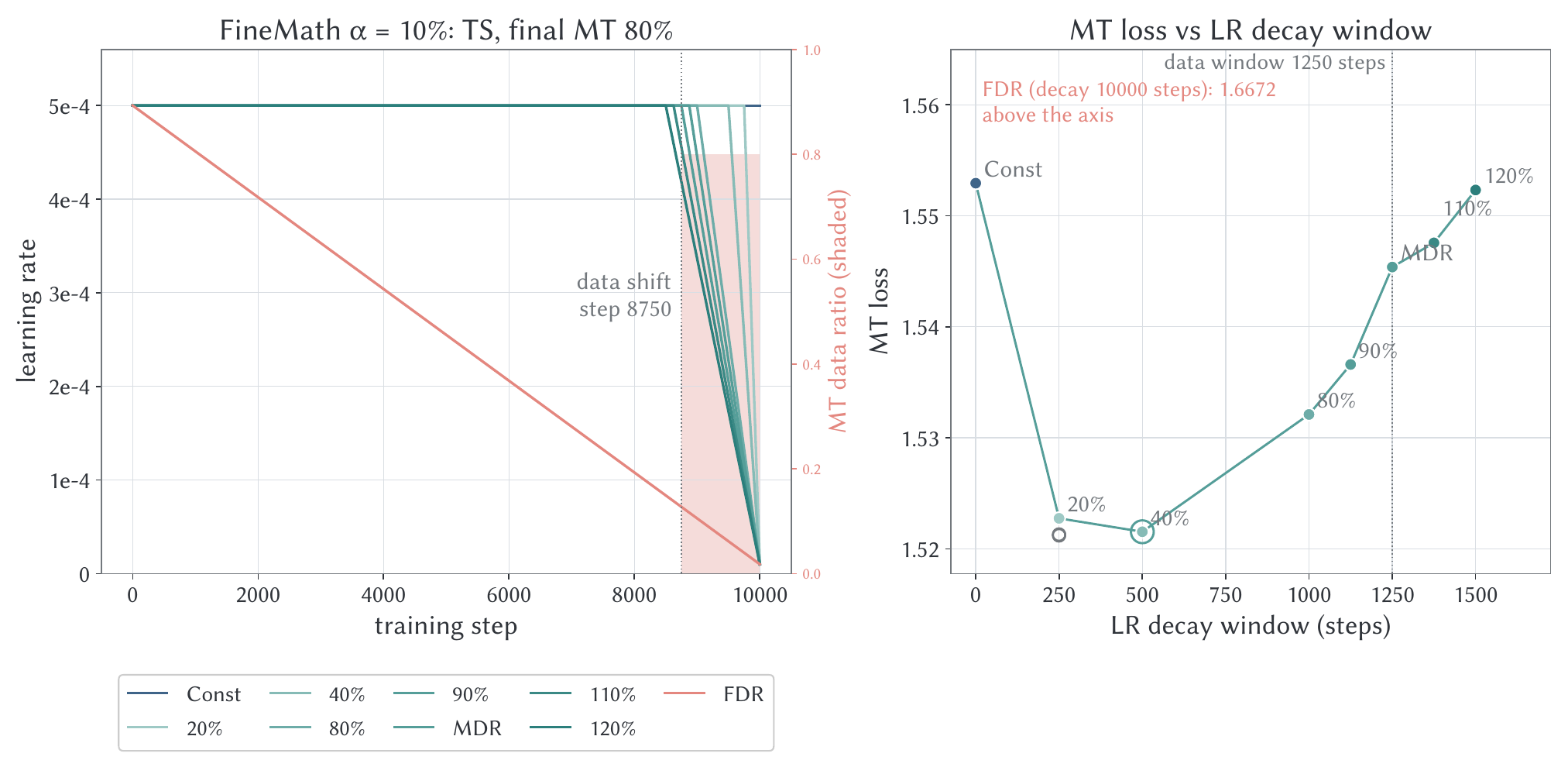}
\par
\includegraphics[width=\linewidth]{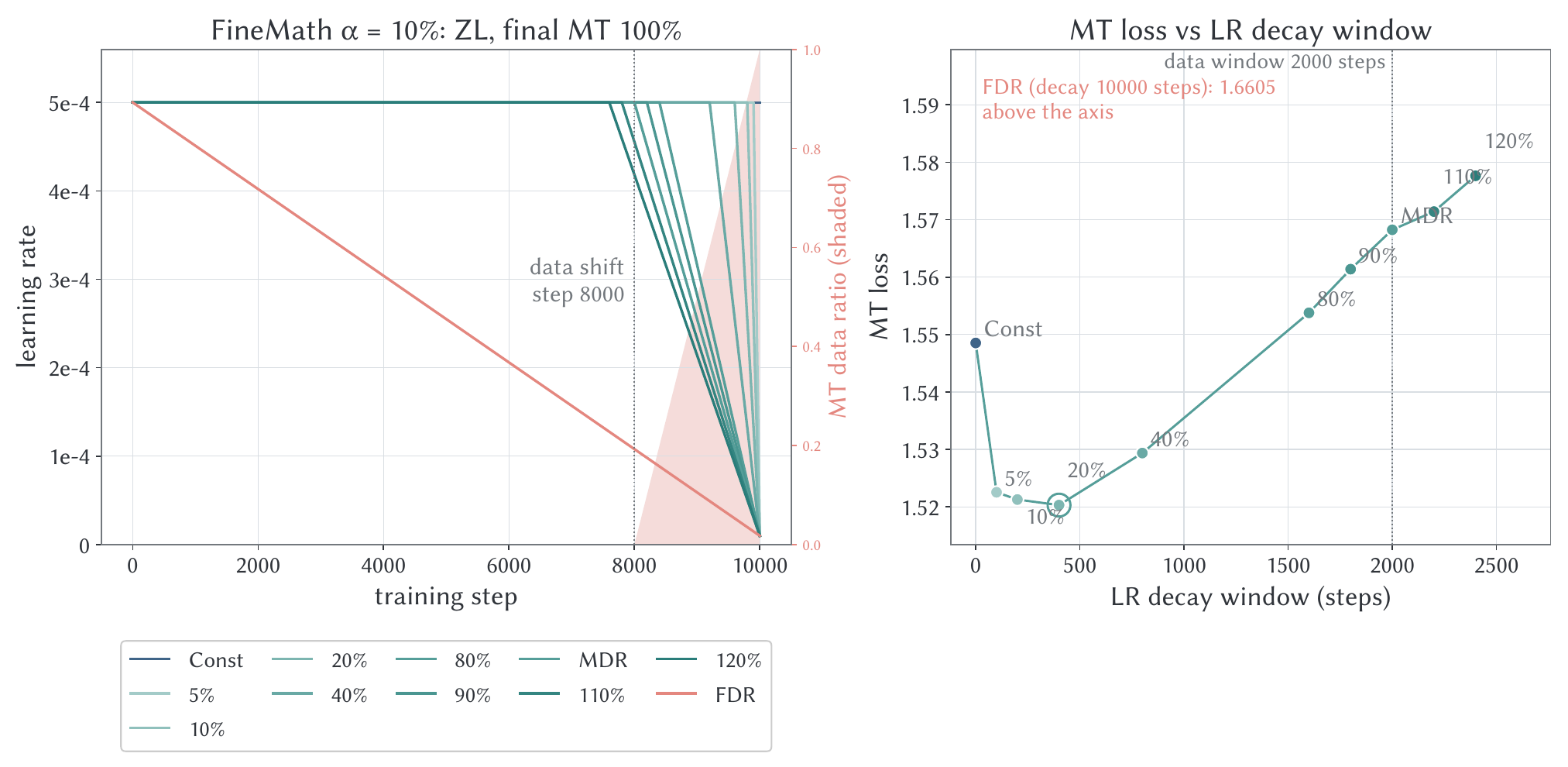}
\caption{\textbf{A short final LR decay improves on Const and MDR.} The lowest tested MT losses are 1.5215 at a relative LR decay duration of 40\% for TS (top) and 1.5203 at 20\% for ZL (bottom). These dense model runs use a 10\% MT budget. TS ends at 80\% MT with a 1,250-step MT window; ZL ends at 100\% MT with a 2,000-step window. Hollow TS markers show repeats at 20\%; the 40\% minimum was not repeated.}
\label{fig:app-decay-dense10}
\end{figure}

\begin{figure}[!htbp]
\centering
\includegraphics[width=\linewidth]{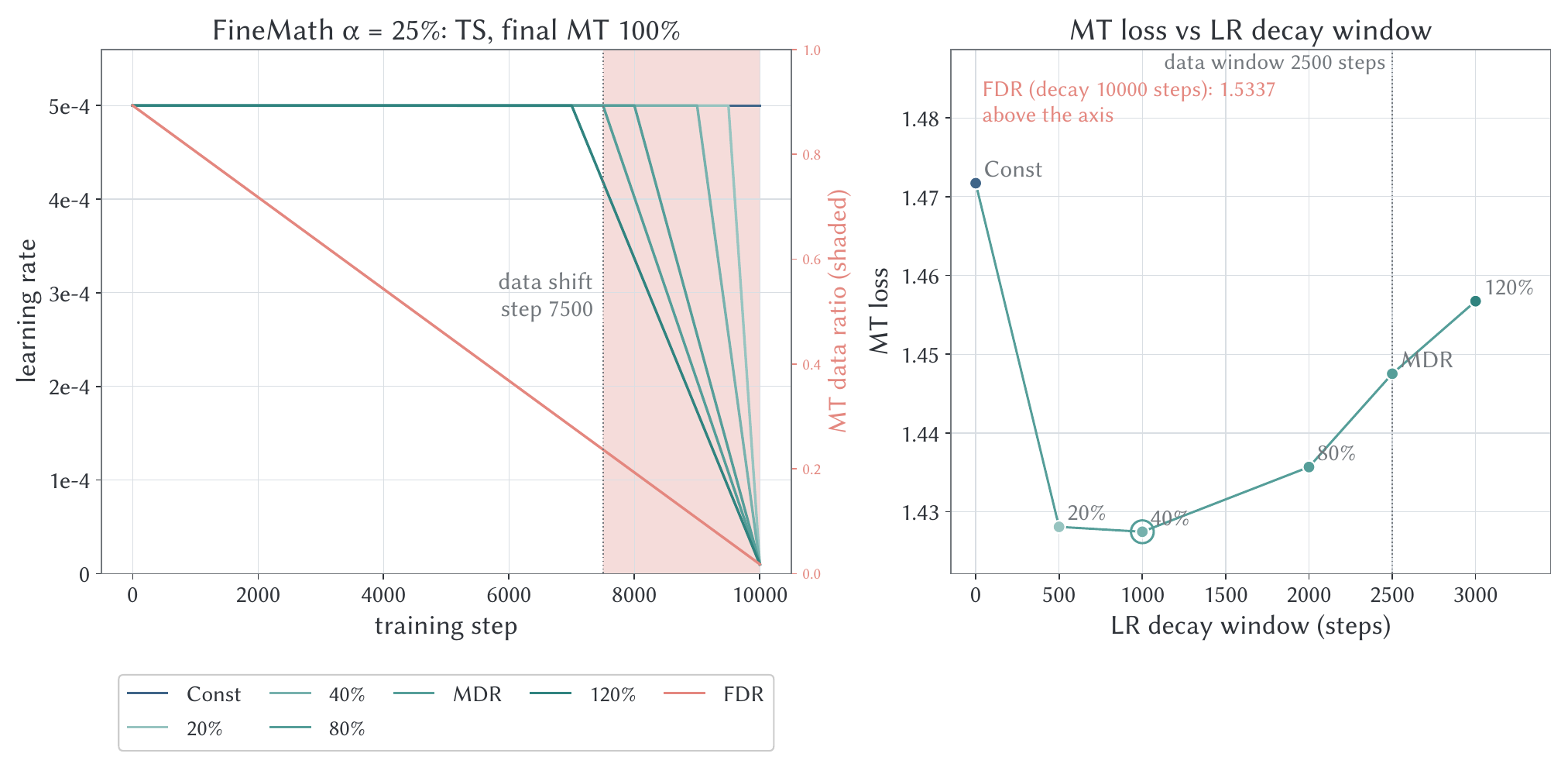}
\par
\includegraphics[width=\linewidth]{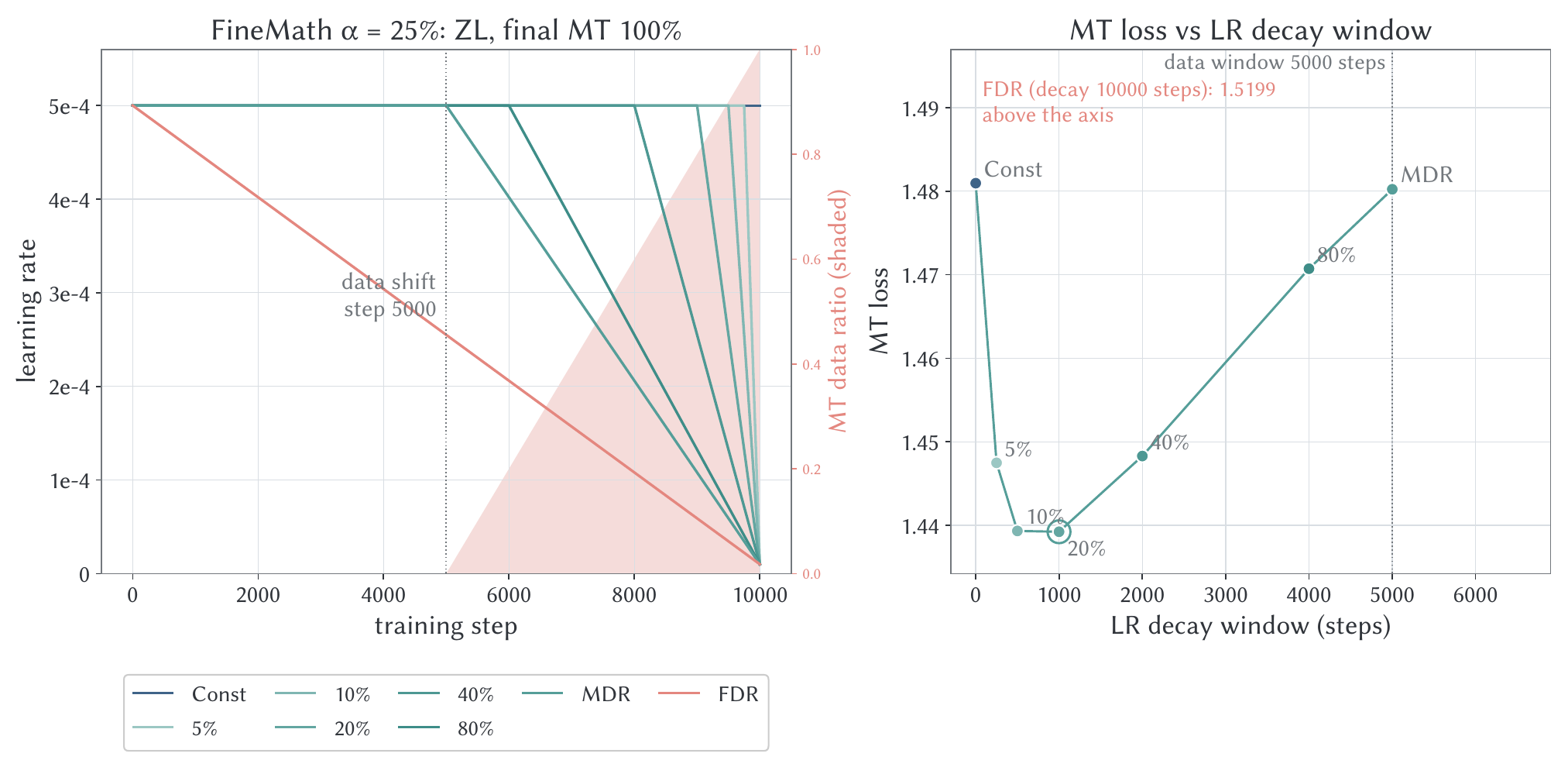}
\caption{\textbf{TS and ZL favor different relative LR decay durations.} TS (top) reaches its lowest tested loss of 1.4275 at 40\%, and ZL (bottom) reaches 1.4392 at 20\%. Nearby short decay durations give similar losses. Both dense model runs use a 25\% MT budget and end at 100\% MT, with MT windows of 2,500 steps for TS and 5,000 for ZL. These configurations have no repeated runs.}
\label{fig:app-decay-dense25}
\end{figure}

\begin{figure}[!htbp]
\centering
\includegraphics[width=\linewidth]{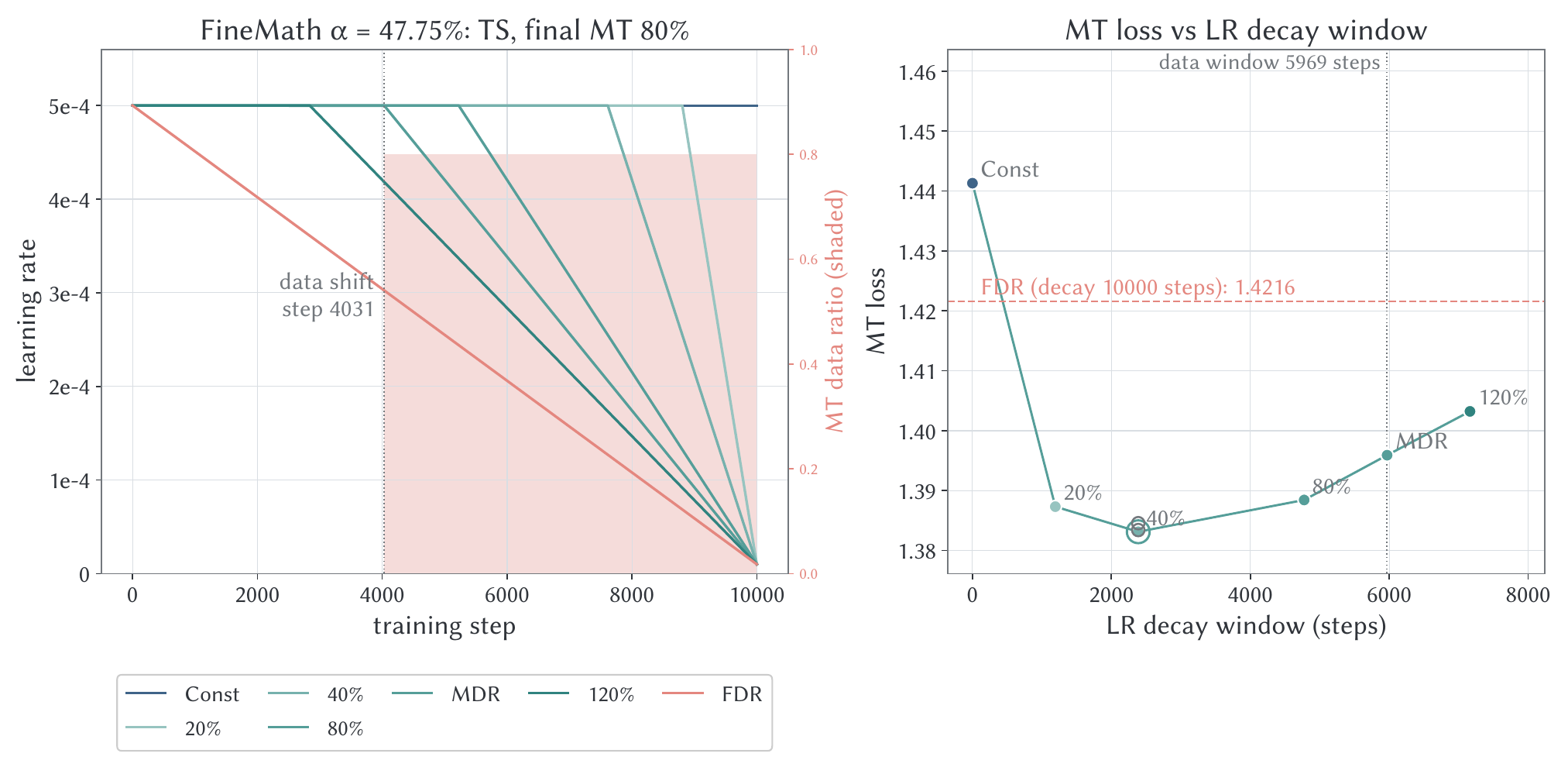}
\par
\includegraphics[width=\linewidth]{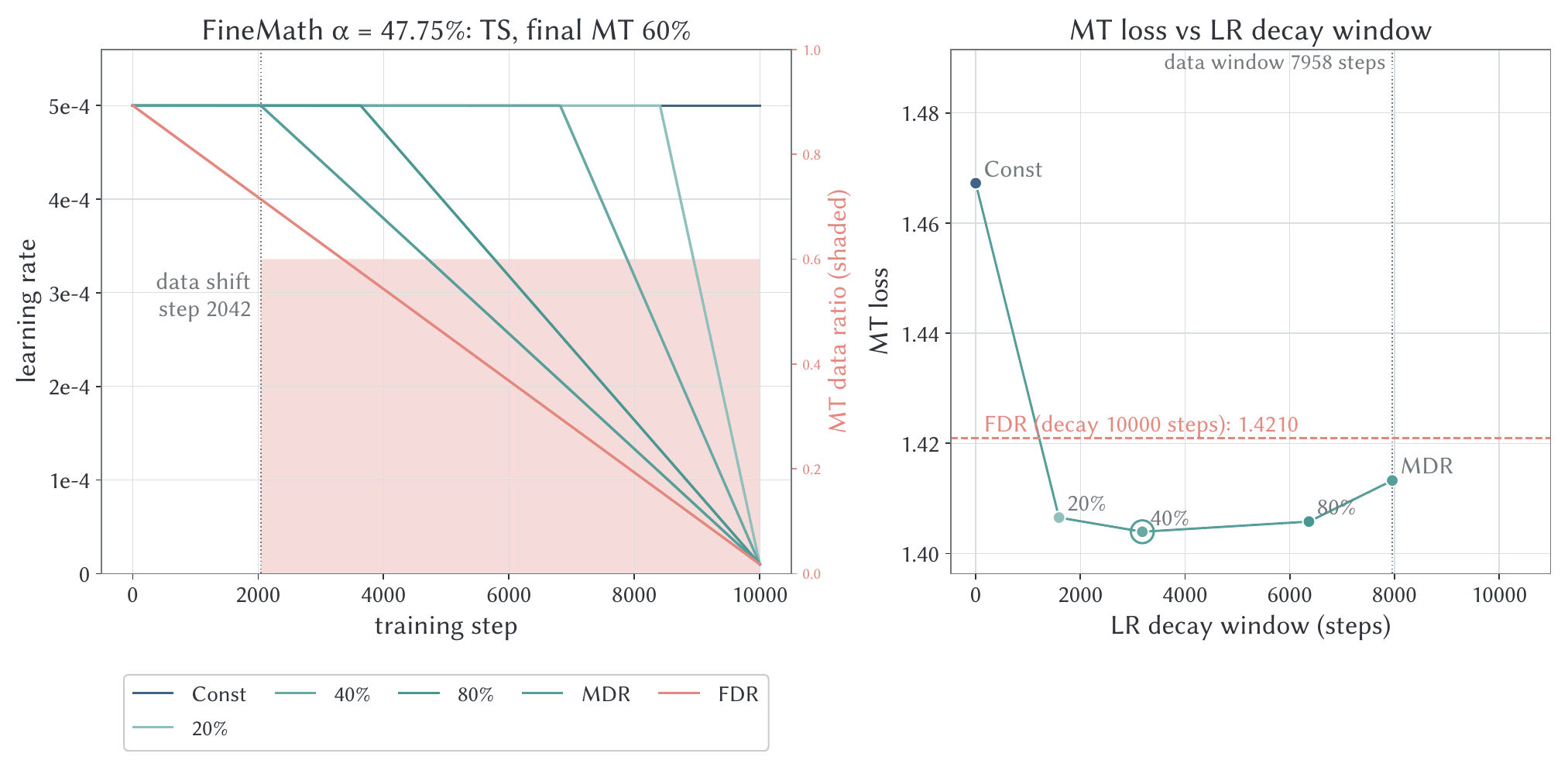}
\caption{\textbf{The best tested relative LR decay duration remains 40\% when TS ends at lower MT ratios.} At a final MT ratio of 80\% (top), this gives a loss of 1.3831, with repeats at 1.3834 and 1.3846. At 60\% MT (bottom), it gives 1.4039, with similar losses at relative LR decay durations between 20\% and 80\%. These dense model runs use a 47.75\% MT budget and MT windows of 5,969 and 7,958 steps, respectively. The scan ending at 100\% MT appears in \Cref{fig:exp-delayed-decay}.}
\label{fig:app-decay-dense4775-ts}
\end{figure}

\begin{figure}[!htbp]
\centering
\includegraphics[width=\linewidth]{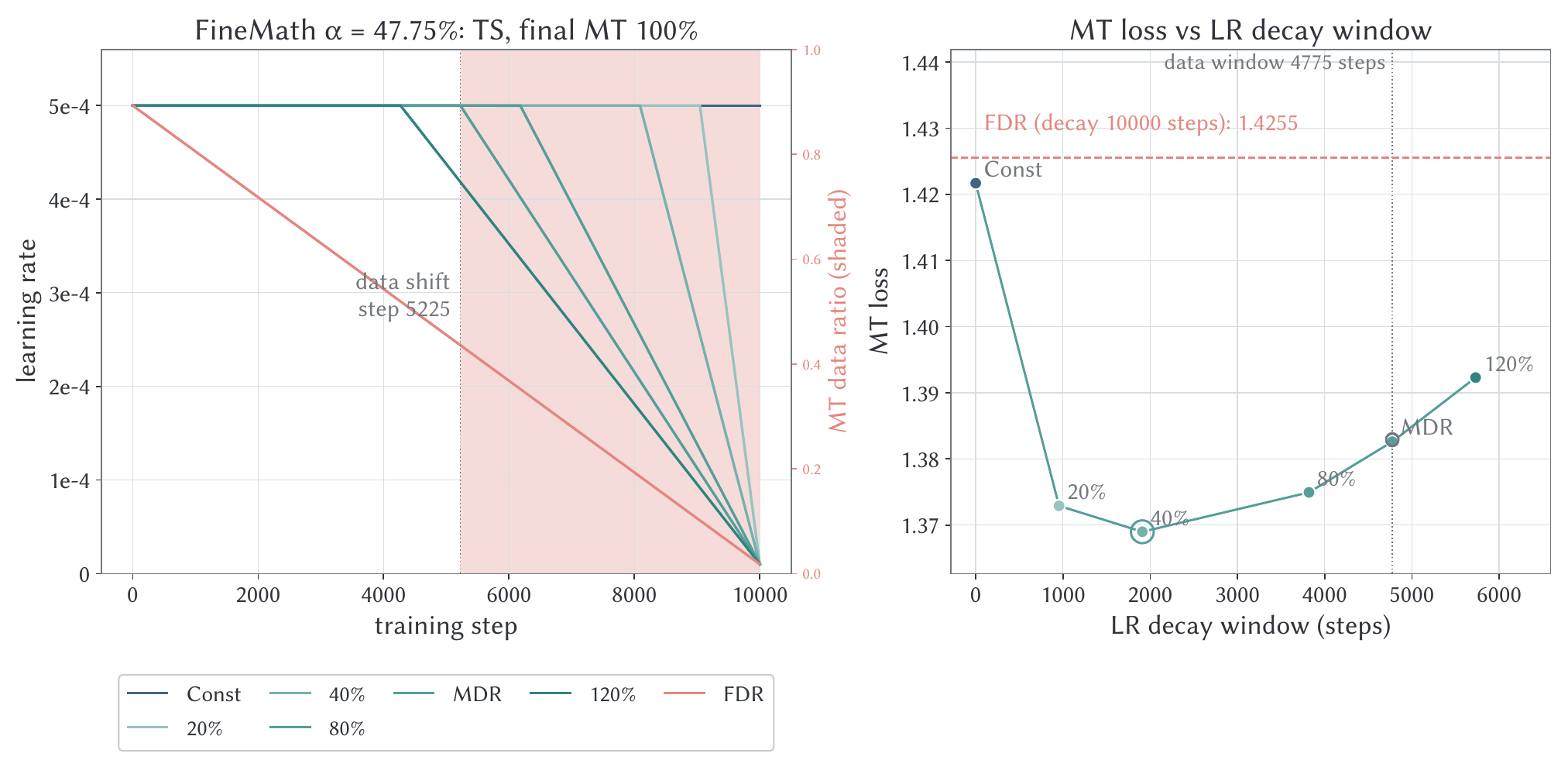}
\par
\includegraphics[width=\linewidth]{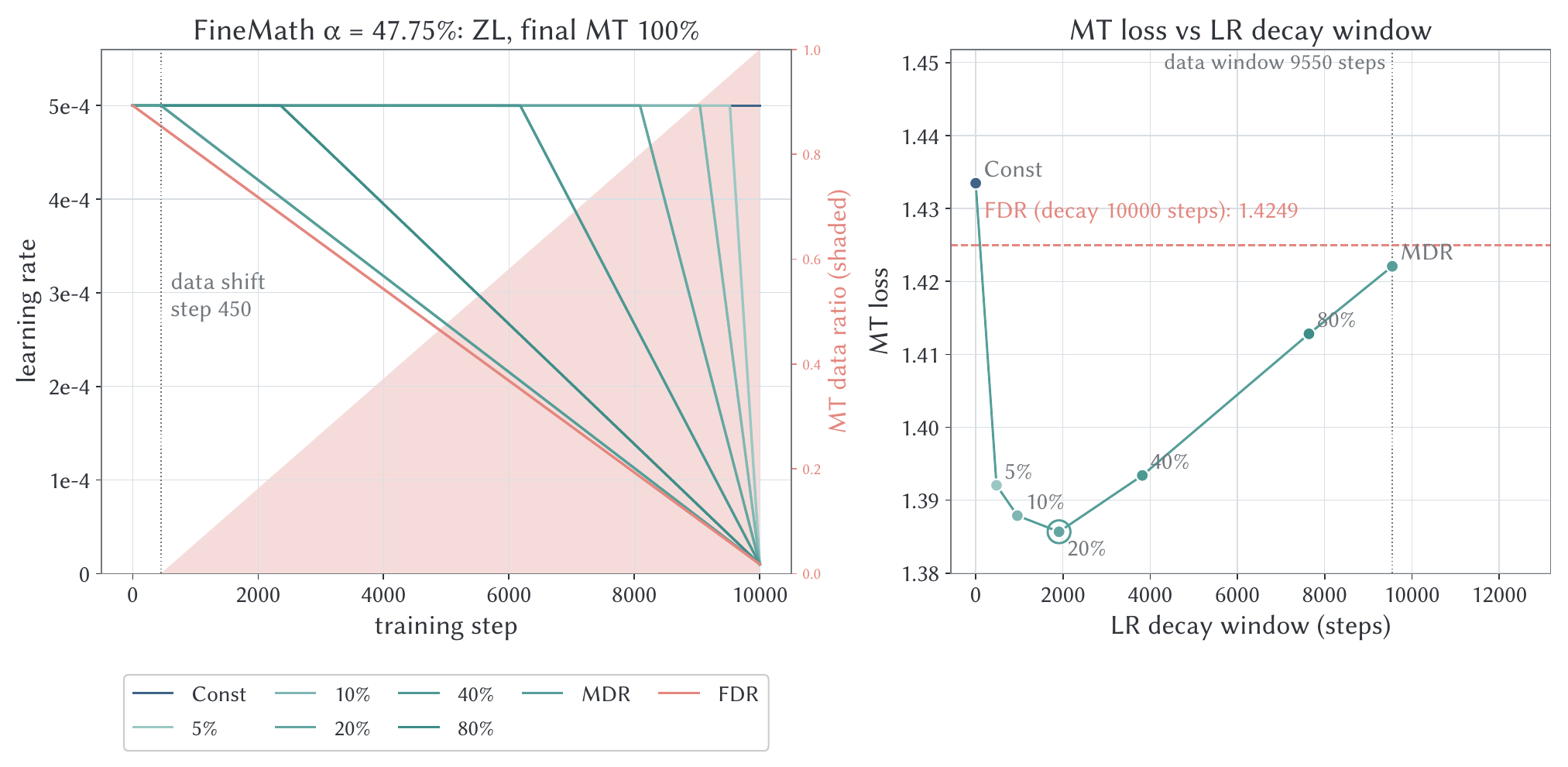}
\caption{\textbf{TS favors a relative LR decay duration of 40\%, while ZL favors 20\%.} Their lowest tested losses are 1.3690 for TS (top) and 1.3857 for ZL (bottom). For ZL, MDR and FDR differ by only 450 decay steps and yield similar losses of 1.4221 and 1.4249, respectively. Both dense model runs use a 47.75\% MT budget and end at 100\% MT, with MT windows of 4,775 steps for TS and 9,550 for ZL. This ZL configuration has no repeated runs.}
\label{fig:app-decay-dense4775-sd}
\end{figure}

\begin{figure}[!htbp]
\centering
\includegraphics[width=\linewidth]{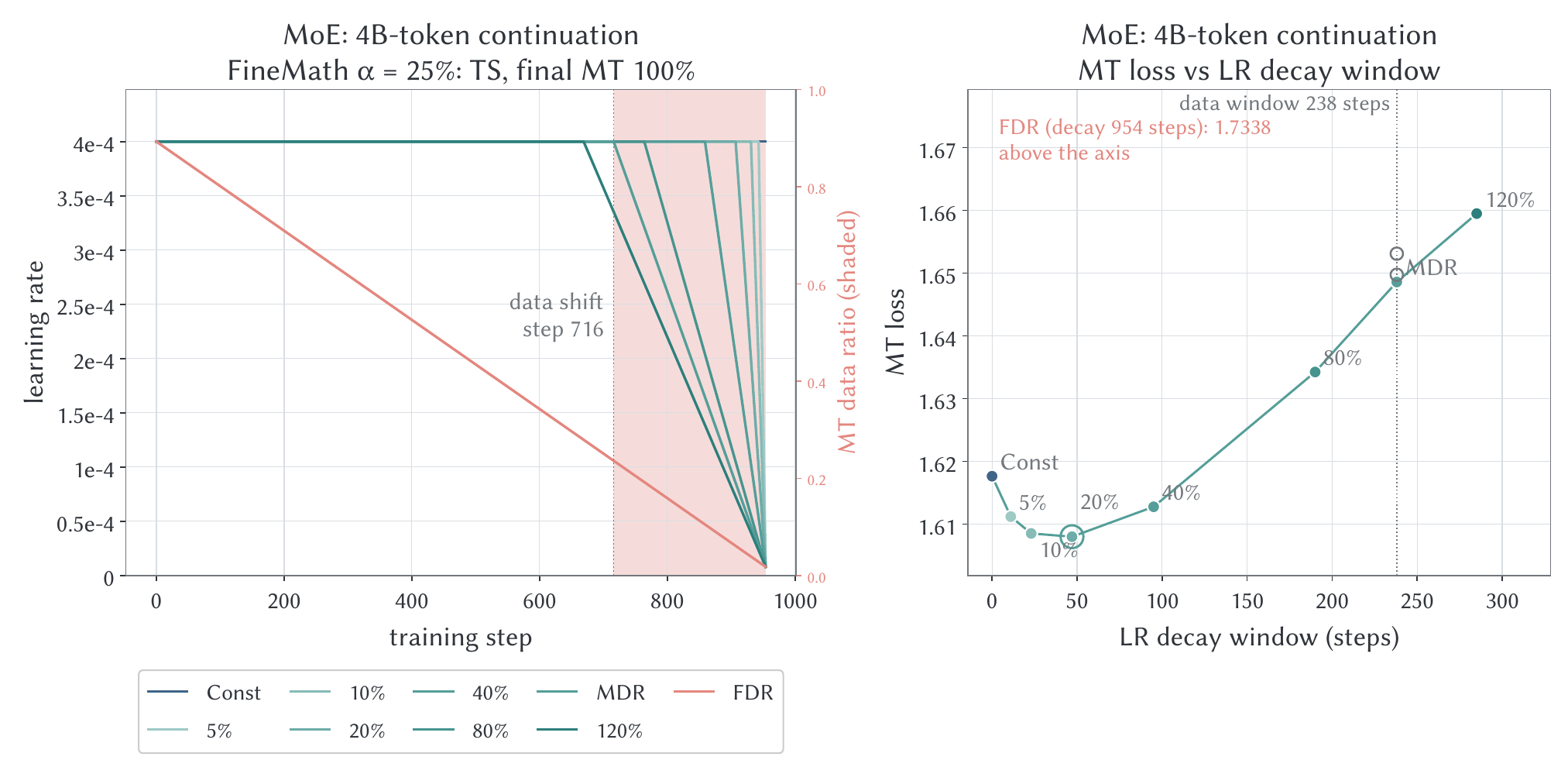}
\par
\includegraphics[width=\linewidth]{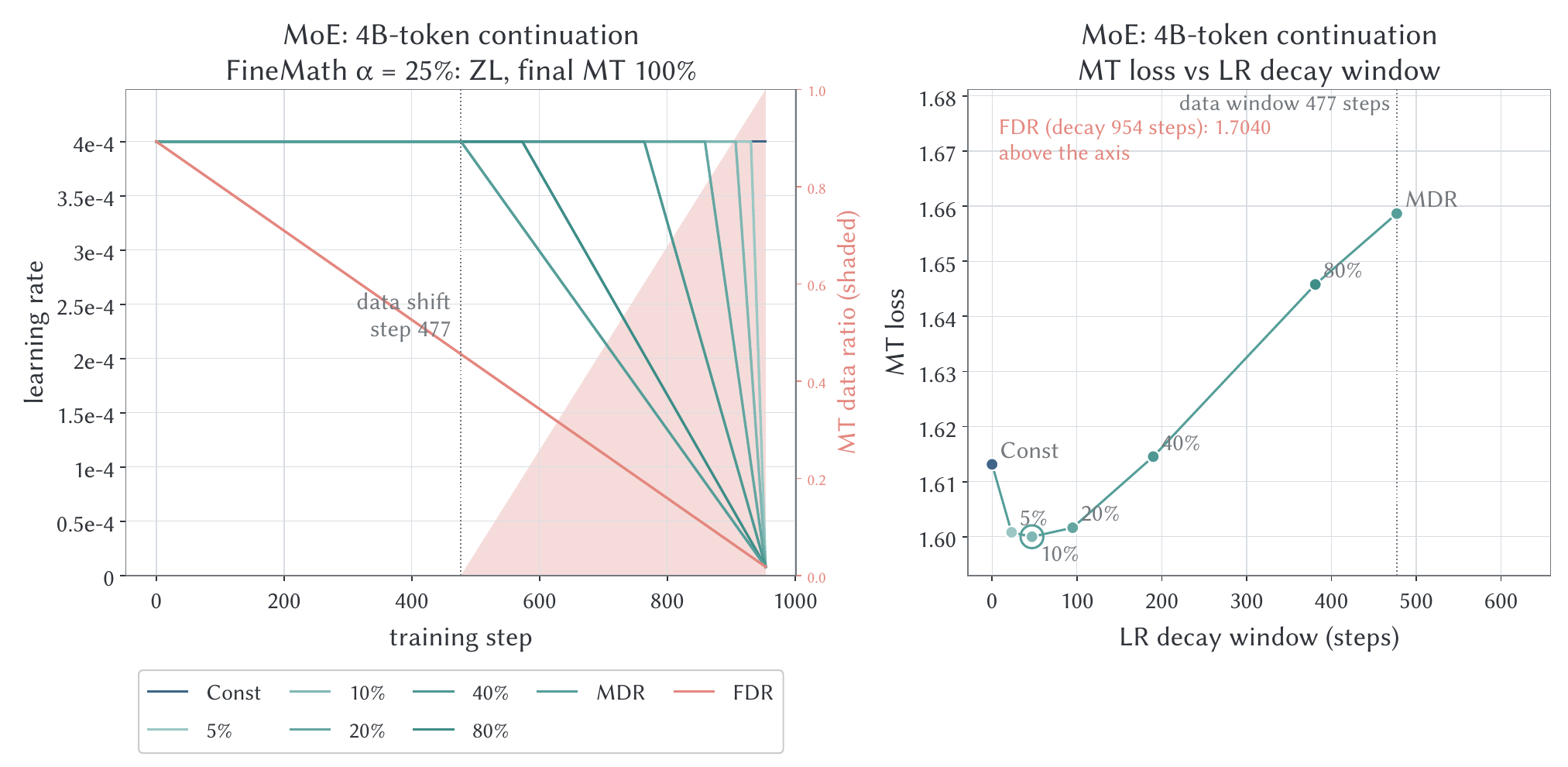}
\caption{\textbf{The best tested TS and ZL decays span nearly the same number of training steps.} TS (top) reaches 1.6080 at a relative LR decay duration of 20\% (47 steps), while ZL (bottom) reaches 1.6000 at 10\% (48 steps). These MoE runs use 4B continued training tokens, a 25\% MT budget, and a final MT ratio of 100\%. Their MT windows last 238 and 477 steps, respectively. Hollow TS markers repeat MDR; neither minimum was repeated.}
\label{fig:app-decay-moe4}
\end{figure}

\begin{figure}[!htbp]
\centering
\includegraphics[width=\linewidth]{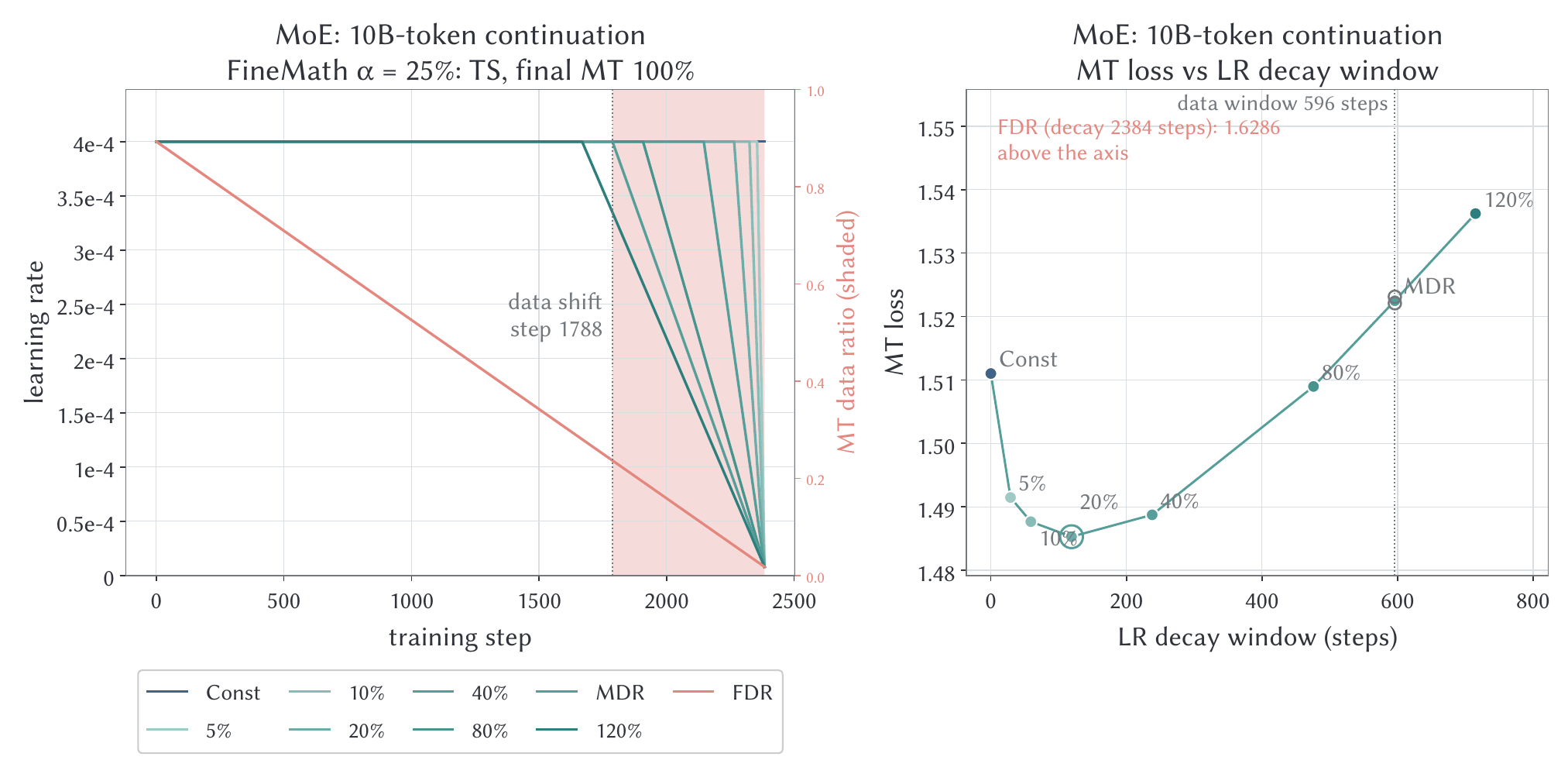}
\par
\includegraphics[width=\linewidth]{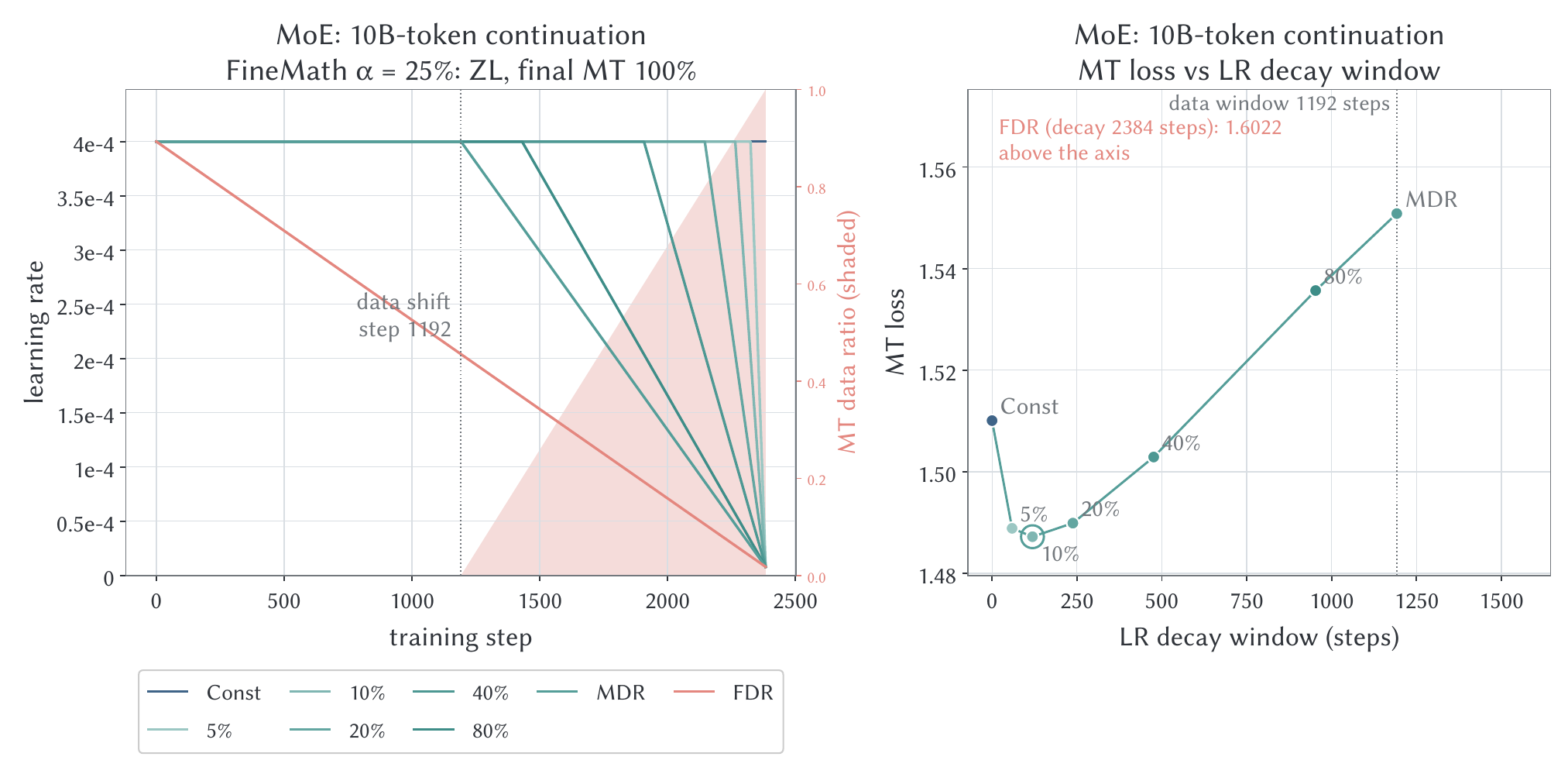}
\caption{\textbf{The best tested TS and ZL decays both last 119 steps.} TS (top) reaches 1.4853 at a relative LR decay duration of 20\%, while ZL (bottom) reaches 1.4872 at 10\%. These MoE runs use 10B continued training tokens, a 25\% MT budget, and a final MT ratio of 100\%. Their MT windows last 596 and 1,192 steps, respectively. Hollow TS markers repeat MDR; the best TS and ZL configurations have no matched data repetitions.}
\label{fig:app-decay-moe10}
\end{figure}

\Cref{fig:app-decay-dense10,fig:app-decay-dense25,fig:app-decay-dense4775-ts,fig:app-decay-dense4775-sd,fig:app-decay-moe4,fig:app-decay-moe10} show the LR trajectories alongside the final MT loss curves.
In each paired scan, the left panel shows the fixed data schedule and candidate LR trajectories. The right panel plots final MT loss against decay duration in steps; point labels give the duration as a percentage of that schedule's MT data window. Lines connect evaluated configurations.
These figures cover the twelve primary scans, including the complete LR/data and loss panels for TS at a 47.75\% MT budget in the dense model, as highlighted in the main text.
For the dense model, the additional TS scan at a 10\% MT budget and a final MT ratio of 100\% appears in Figure~\ref{fig:exp-delayed-decay}. Its MT data window lasts 1,000 steps, and the lowest tested loss is 1.5156 at a relative LR decay duration of 40\%. The 120\% relative duration, outside the displayed range, gives 1.5512, while FDR gives 1.6790.
\Cref{tab:app-decay-robustness} supplies the complete sampled losses for the four further configurations.
Their LR trajectories and loss curves are shown in \Cref{fig:app-moe4775-decay,fig:app-moe-pt-decay} in Appendix~\ref{app:exp-moe}.
The two StarCoder scans are compared with their FineMath counterparts in \Cref{fig:app-moe-mt-domain}.

\clearpage

\section{Data Stream Replicas and Noise Estimates}
\label{app:exp-replicas}

For six selected configurations, we repeat training twice using different training examples or a different order of the same examples (Table~\ref{tab:app-replicas}).
All three runs start from the same PT checkpoint and use the same training setup.
For the dense model, the two repeats begin one third and two thirds of the way through each dataset, respectively, returning to the beginning when they reach the end.
At the 47.75\% MT budget, each run uses almost the entire FineMath dataset, so the runs mainly differ in the order of MT examples.
Each repeat shares about 36\% of its DCLM examples with the original run.
For the MoE model, the three runs instead use separate subsets of both PT and MT data, with no overlap between runs.

\begin{table}[!htbp]
\centering
\small
\setlength{\tabcolsep}{4pt}
\begin{tabular}{@{}lrlrrrr@{}}
\toprule
MT budget & $\gamma_{\mathrm{end}}$ & \shortstack{Relative LR decay\\duration or rule} & Primary & Repeat 1 & Repeat 2 & $s_3$ \\
\midrule
\multicolumn{7}{@{}l}{\textit{Dense: 200B-token PT; 20.97B-token continuation}} \\
10\% & 80\% & 20\% & 1.5227 & 1.5213 & 1.5212 & 0.000706 \\
47.75\% & 100\% & MDR & 1.3826 & 1.3828 & 1.3830 & 0.000149 \\
47.75\% & 70\% & FDR & 1.4194 & 1.4217 & 1.4198 & 0.001008 \\
47.75\% & 80\% & 40\% & 1.3831 & 1.3834 & 1.3846 & 0.000620 \\
\midrule
\multicolumn{7}{@{}l}{\textit{MoE: 40B-token PT; 4B-token continuation}} \\
25\% & 100\% & MDR & 1.6486 & 1.6531 & 1.6497 & 0.001916 \\
\midrule
\multicolumn{7}{@{}l}{\textit{MoE: 40B-token PT; 10B-token continuation}} \\
25\% & 100\% & MDR & 1.5225 & 1.5231 & 1.5221 & 0.000421 \\
\bottomrule
\end{tabular}
\caption{Final MT losses for the six repeated TS configurations, all using FineMath. The descriptive dispersion $s_3=\sqrt{\frac{1}{3}\sum_{j=1}^3(\ell_j-\bar\ell)^2}$ is computed from the original losses before rounding; its denominator is three, rather than the denominator-two sample standard deviation.}
\label{tab:app-replicas}
\end{table}

Repeated runs are shown as hollow points. We use only the original runs to fit the curves and identify the best tested configuration.
These repeats show how the loss changes when we change the training examples or their order at the six selected configurations.
All runs start from the same PT checkpoint, so they do not capture variation across independently pretrained models. The repeats for the dense model also share many training examples.
Repeating an MDR run does not tell us how much the loss would vary at other relative LR decay durations or with other data schedules.
We therefore do not treat the small loss differences between nearby relative LR decay durations for ZL at a 25\% MT budget for the dense model, or between the best TS and ZL MoE settings with a 10B-token continuation, as evidence that one setting reliably outperforms another.

\section{MoE Results and Additional Ablations}
\label{app:exp-moe}

Our main MoE experiments compare approximately 4B and 10B tokens of continued training, with MT data accounting for 25\% in both settings.
These runs last 954 and 2,384 steps and use approximately 1.0B and 2.5B MT tokens, respectively, corresponding to 244,224 and 610,304 packed FineMath sequences.
The 10B-token continuation uses approximately 25.6\% of the FineMath dataset. In both settings, continued training draws DCLM data from shards not consumed during PT. \Cref{tab:app-moe-grid} lists all tested final MT ratios and the training step at which MT data first enters for each schedule.
Figure~\ref{fig:exp-dense-moe} also shows results for 10B tokens of continued training with a 47.75\% MT budget.

\begin{table}[!htbp]
\centering
\small
\setlength{\tabcolsep}{5pt}
\begin{tabular}{@{}llrr@{}}
\toprule
Data schedule & Final MT ratio & \shortstack{First MT step\\4B-token continuation} & \shortstack{First MT step\\10B-token continuation} \\
\midrule
TS & 100\% & 716 & 1,788 \\
 & 75\% & 636 & 1,589 \\
 & 50\% & 477 & 1,192 \\
 & 40\% & 358 & 894 \\
 & 25\% & 0 & 0 \\
\midrule
ZL & 100\% & 477 & 1,192 \\
 & 75\% & 318 & 795 \\
 & 50\% & 0 & 0 \\
\bottomrule
\end{tabular}
\caption{Tested data schedules for the MoE model at a 25\% MT budget. The first MT step denotes the TS switch or ZL ramp start.  The TS $\gamma_{\mathrm{end}}=100\%$ data windows are 238 and 596 steps.}
\label{tab:app-moe-grid}
\end{table}

\begin{table}[!htbp]
\centering
\small
\setlength{\tabcolsep}{6pt}
\begin{tabular}{@{}llccc@{}}
\toprule
Continuation tokens & Data schedule & Const & FDR & MDR \\
\midrule
4B & TS & 1.6176 (100) & 1.6349 (25) & 1.6343 (50) \\
 & ZL & 1.6131 (100) & 1.6548 (50) & 1.6548 (50) \\
10B & TS & 1.5110 (100) & 1.5343 (25) & 1.5225 (100) \\
 & ZL & 1.5101 (100) & 1.5548 (50) & 1.5508 (100) \\
\bottomrule
\end{tabular}
\caption{Lowest final MT loss among the tested final MT ratios for each training duration, data schedule, and LR schedule in the MoE model at a 25\% MT budget. Parentheses give the final MT ratio $\gamma_{\mathrm{end}}$ in percent. Both MDR minima after 4B continued training tokens occur at $\gamma_{\mathrm{end}}=50\%$, so MDR does not select $\gamma_{\mathrm{end}}=100\%$ at both continued training token budgets.}
\label{tab:app-moe-minima}
\end{table}

\begin{figure}[!htbp]
\centering
\includegraphics[width=\linewidth]{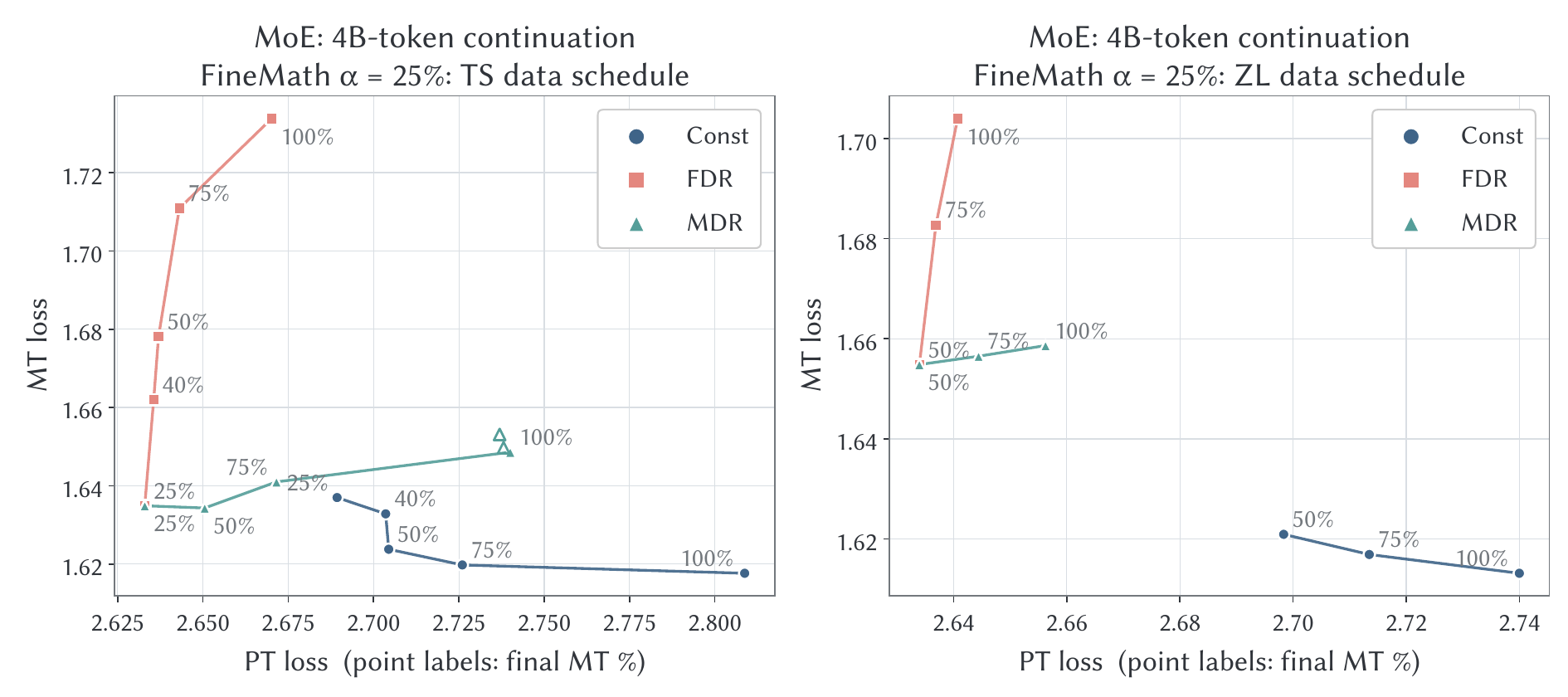}
\par
\includegraphics[width=0.49\linewidth]{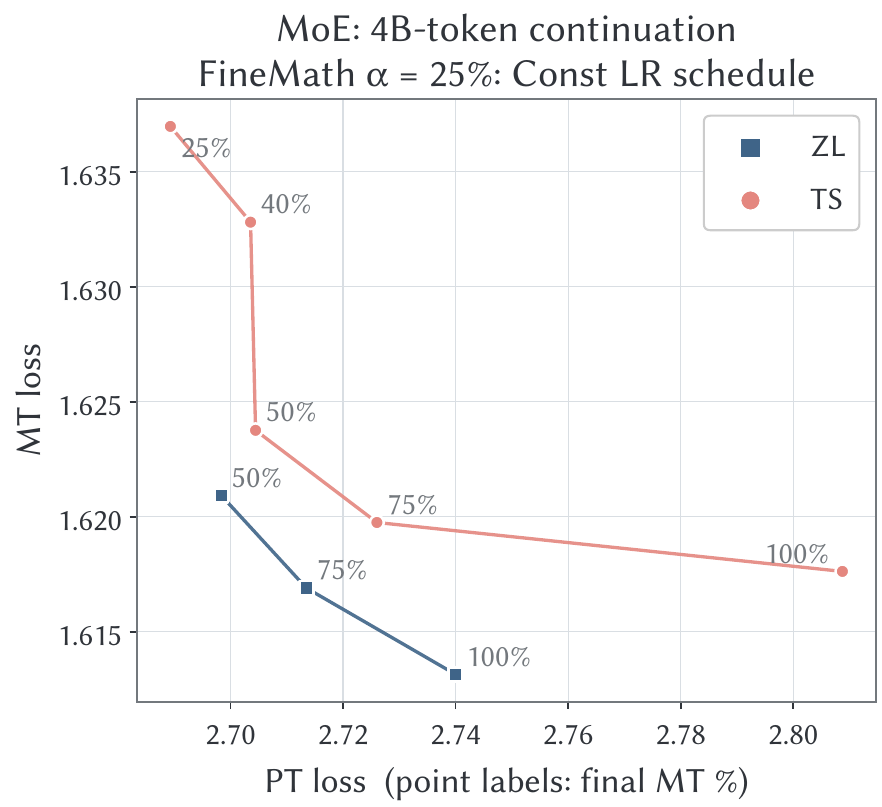}
\hfill
\includegraphics[width=0.49\linewidth]{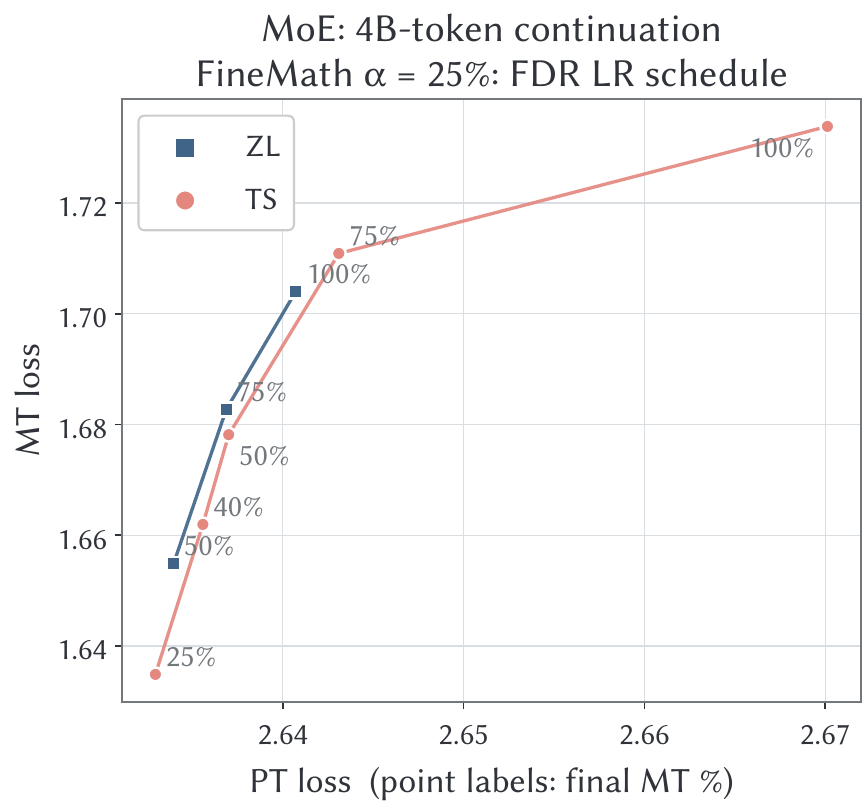}
\par
\includegraphics[width=0.49\linewidth]{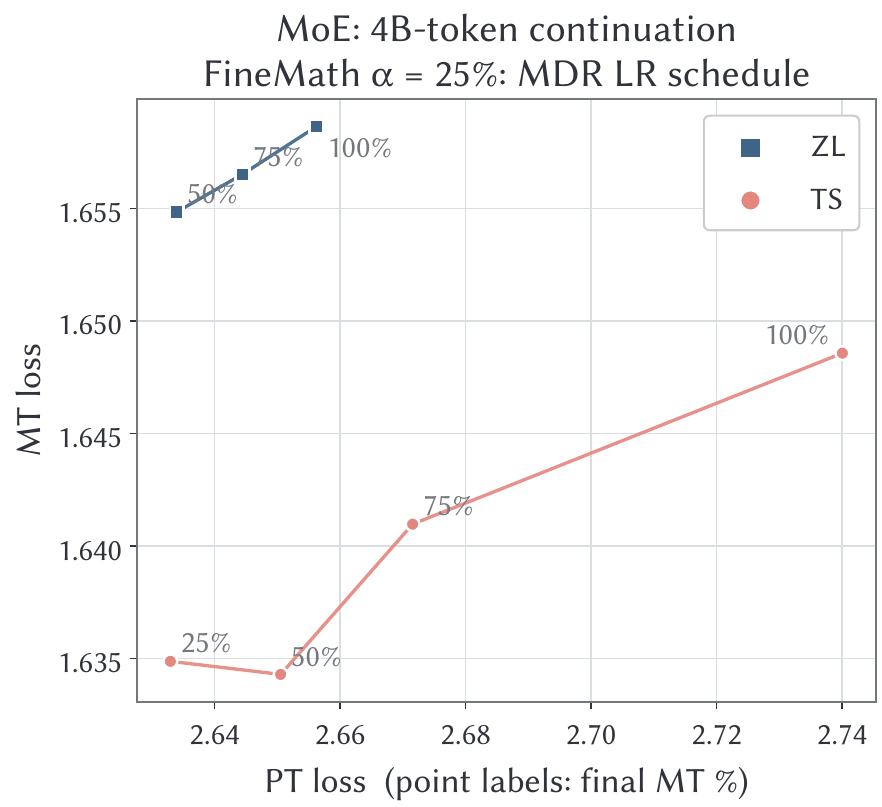}
\caption{\textbf{Const gives the lowest tested MT loss among the three baseline LR schedules.} MDR favors a final MT ratio of 50\% for both TS and ZL. These MoE runs use 4B continued training tokens with a 25\% MT budget. Hollow overlays repeat TS+MDR at a final MT ratio of 100\%.}
\label{fig:app-moe4-grid}
\end{figure}

\begin{figure}[!htbp]
\centering
\includegraphics[width=\linewidth]{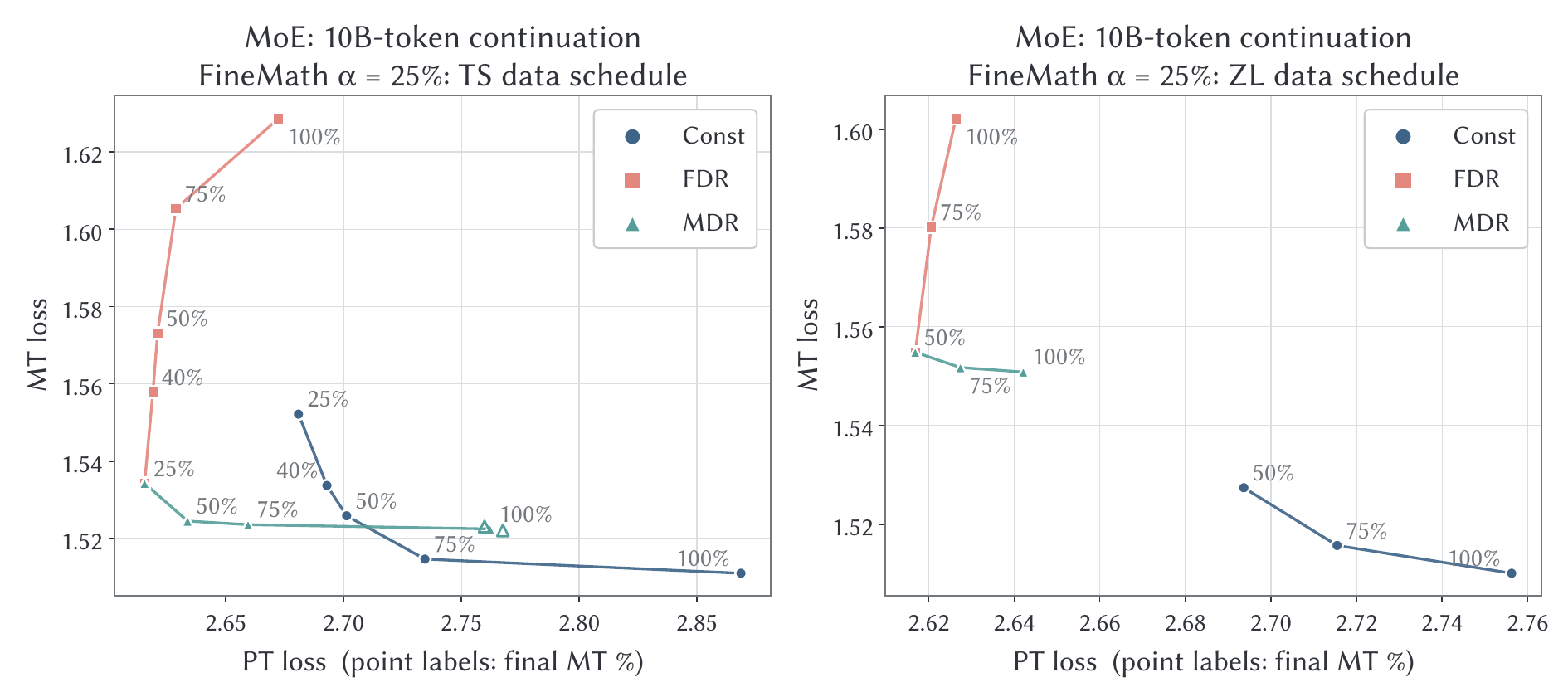}
\par
\includegraphics[width=0.49\linewidth]{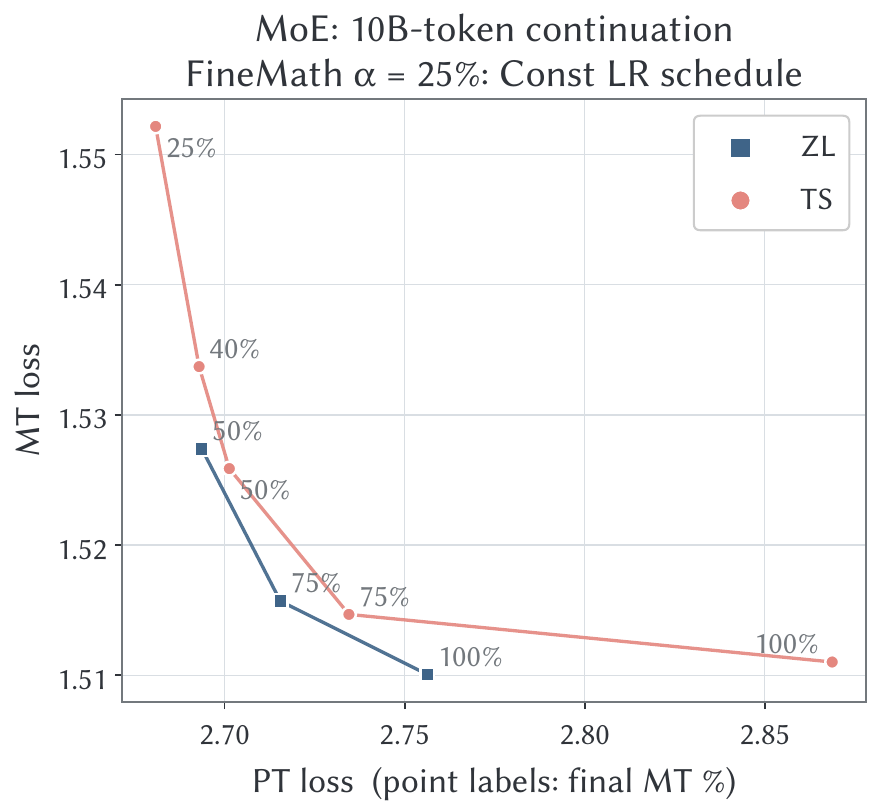}
\hfill
\includegraphics[width=0.49\linewidth]{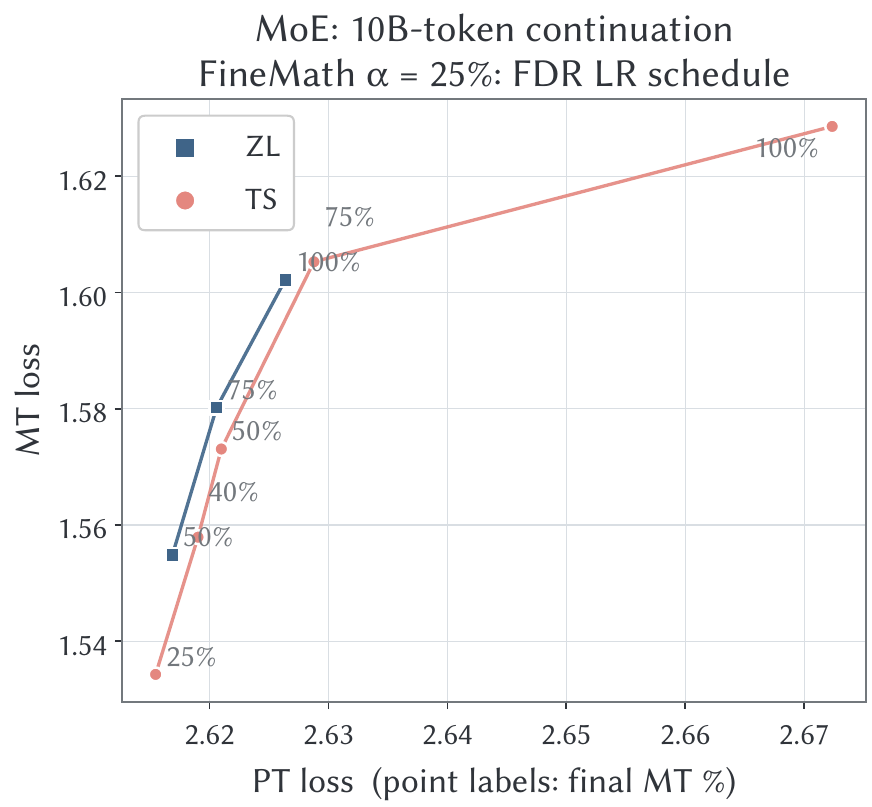}
\par
\includegraphics[width=0.49\linewidth]{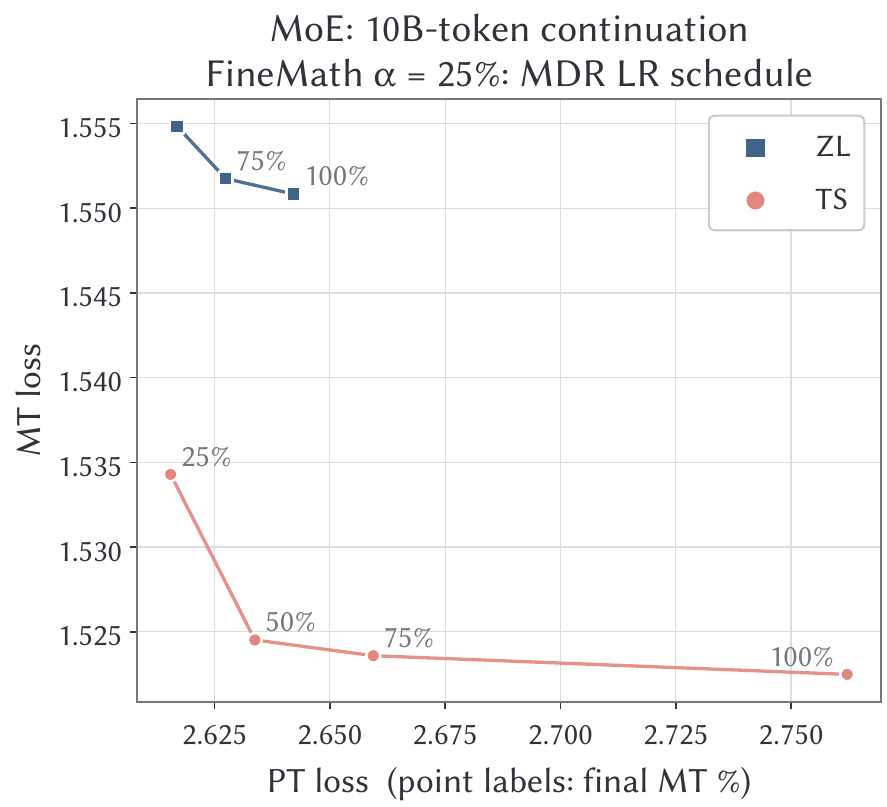}
\caption{\textbf{Longer MoE continued training narrows the TS loss gap between MDR and Const.} At a final MT ratio of 100\%, TS+MDR is closer to TS+Const than after 4B tokens, while ZL+MDR remains worse than ZL+Const. These runs use 10B continued training tokens with a 25\% MT budget. Hollow overlays repeat TS+MDR at 100\% MT.}
\label{fig:app-moe10-grid}
\end{figure}

\paragraph{LR dependent preferences persist, but MDR is not uniformly better than Const.}
\Cref{fig:app-moe4-grid,fig:app-moe10-grid} show the final PT and MT losses for all tested configurations.
At both horizons, TS+Const favors $\gamma_{\mathrm{end}}=100\%$, whereas TS+FDR favors $\gamma_{\mathrm{end}}=25\%$.
At 4B, the best TS+MDR loss is worse than the best TS+Const loss, and MDR's own minimum occurs at $\gamma_{\mathrm{end}}=50\%$.
This does not assert pointwise dominance: at the $\gamma_{\mathrm{end}}=25\%$ boundary, MDR and FDR coincide.
At 10B, MDR is nearly flat for the non-boundary endpoints: its loss is 1.5225 at $\gamma_{\mathrm{end}}=100\%$, 1.5236 at 75\%, and 1.5245 at 50\%.
The MDR-minus-Const difference at $\gamma_{\mathrm{end}}=100\%$ shrinks from $+0.0310$ at 4B to $+0.0115$ at 10B; at $\gamma_{\mathrm{end}}=50\%$ in the 10B experiments, MDR is 0.0014 below Const.
The approximately 63\% reduction of the gap at 100\% MT is consistent with sensitivity to adaptation time, but extending the horizon increases total PT and MT tokens together and does not isolate a single mechanism.

\begin{figure}[!htbp]
\centering
\includegraphics[width=\linewidth]{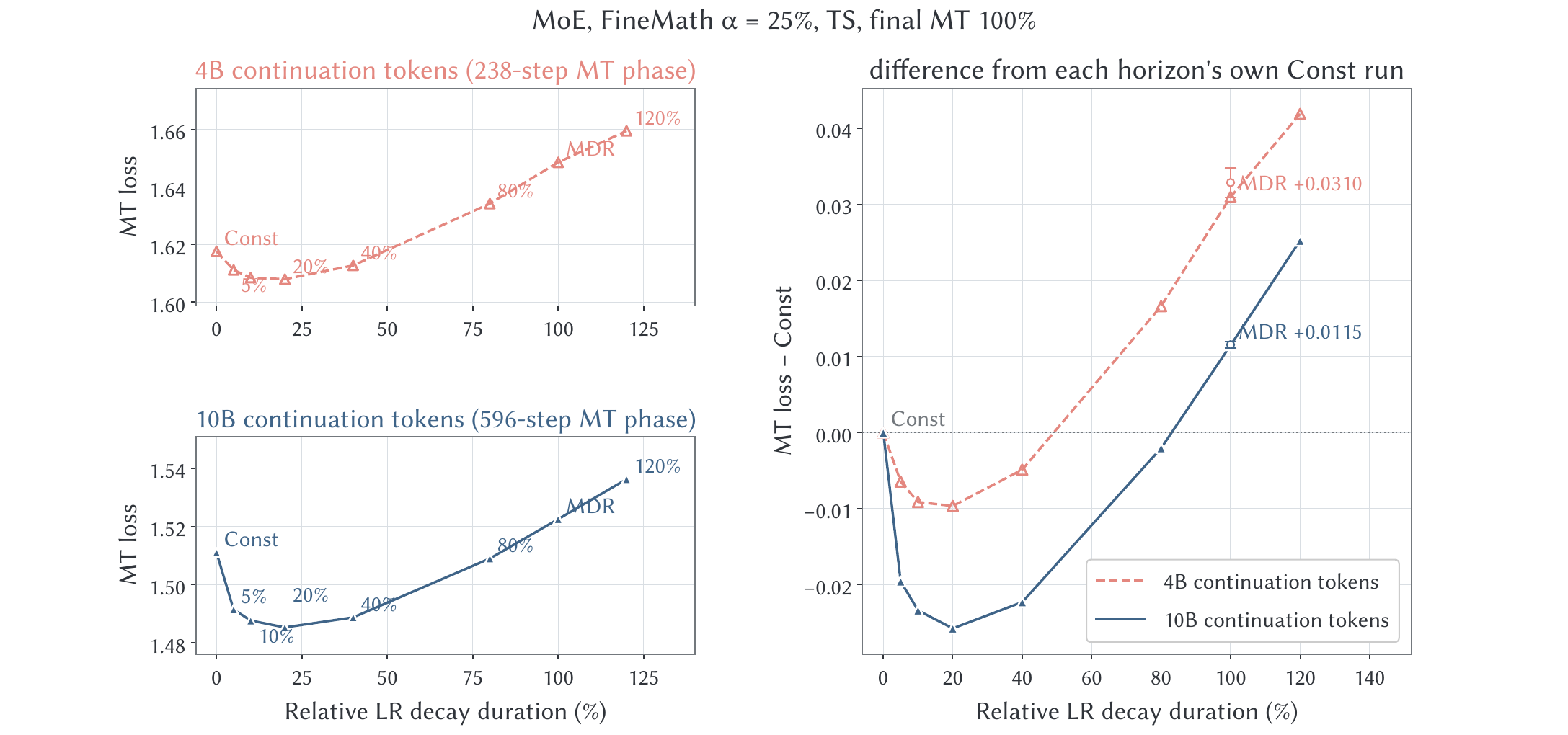}
\par
\includegraphics[width=\linewidth]{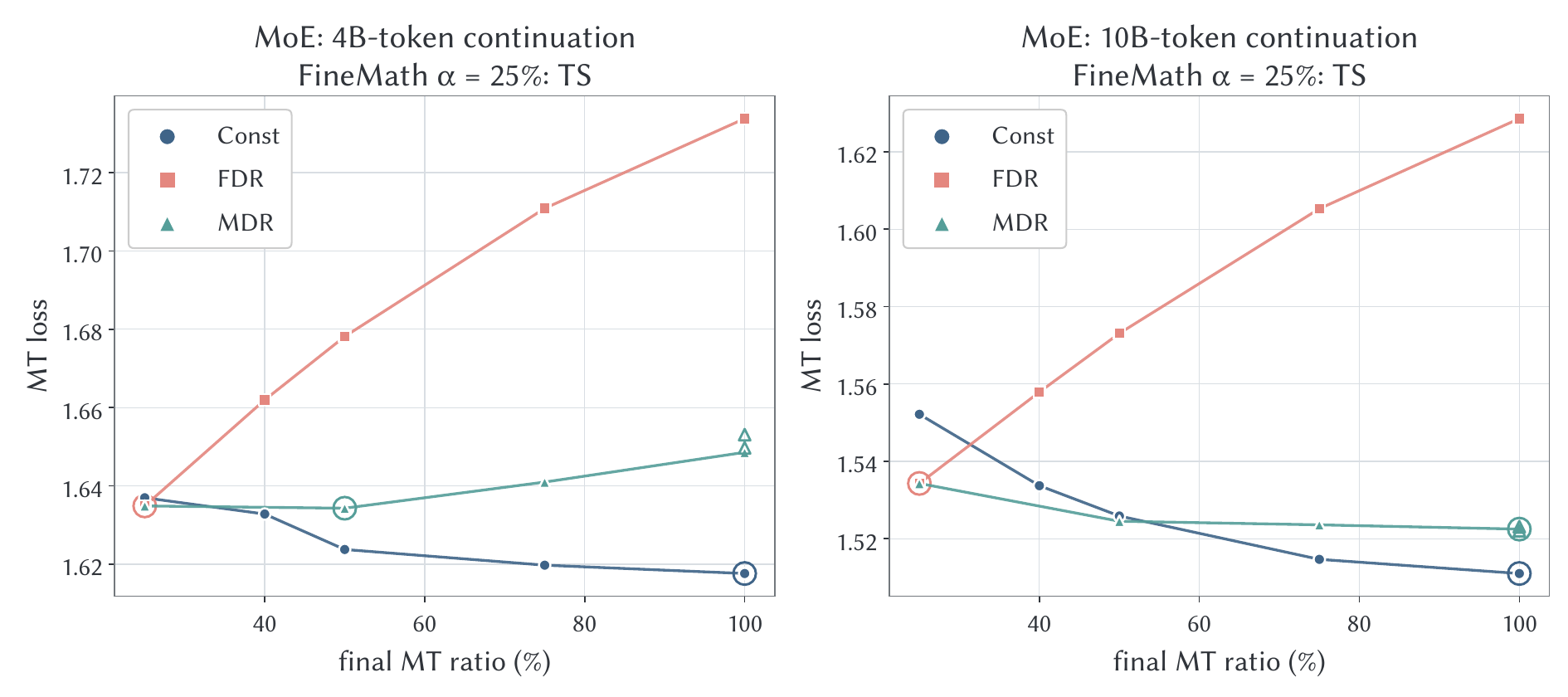}
\caption{\textbf{Longer MoE continued training narrows the MDR--Const loss gap.} For TS, extending training from 4B to 10B tokens reduces the primary-run gap from 0.0310 to 0.0115 at a final MT ratio of 100\%, while the best tested relative LR decay duration remains 20\%. The upper panels show final MT loss and its change from each horizon's Const baseline; the lower panels vary the final MT ratio. All runs use a 25\% MT budget. In the difference panel, circles and local error bars at MDR show the three-run mean and $\pm s_3$, computed from unrounded losses relative to the primary Const loss; they do not estimate uncertainty at other durations or in the Const baseline.}
\label{fig:app-moe-horizons}
\end{figure}

\paragraph{TS versus ZL.}
At matched endpoints, ZL improves MT loss under FDR and worsens it under MDR in both horizons (\Cref{tab:app-moe-shape-difference}).
Const gives smaller differences, with the 4B advantage of ZL largely disappearing at 10B.
After separately selecting each shape's best scanned window, ZL gives 1.6000 versus TS's 1.6080 at 4B, whereas TS gives 1.4853 versus ZL's 1.4872 at 10B.
This ordering change prevents a general claim that either shape is preferable for MoE training.
The scans appear in \Cref{app:exp-decay}; \Cref{fig:app-moe-horizons} compares the TS horizon response directly.

\begin{table}[!htbp]
\centering
\small
\setlength{\tabcolsep}{5pt}
\begin{tabular}{@{}lrrr@{}}
\toprule
\shortstack[l]{Continuation tokens\\and LR rule} & $\gamma_{\mathrm{end}}=100\%$ & $\gamma_{\mathrm{end}}=75\%$ & $\gamma_{\mathrm{end}}=50\%$ \\
\midrule
4B Const & $-0.0045$ & $-0.0029$ & $-0.0028$ \\
4B FDR & $-0.0298$ & $-0.0282$ & $-0.0233$ \\
4B MDR & $+0.0101$ & $+0.0155$ & $+0.0206$ \\
\midrule
10B Const & $-0.0010$ & $+0.0010$ & $+0.0015$ \\
10B FDR & $-0.0264$ & $-0.0251$ & $-0.0183$ \\
10B MDR & $+0.0283$ & $+0.0282$ & $+0.0303$ \\
\bottomrule
\end{tabular}
\caption{MoE MT loss difference, ZL minus TS, at matched final MT ratios. Differences are computed from unrounded losses and reported to four decimal places. Negative values favor ZL. Each comparison uses one primary run per schedule, without significance testing.}
\label{tab:app-moe-shape-difference}
\end{table}

\clearpage

\paragraph{A larger MT budget at the 10B horizon.}
We increase the MT budget from 25\% to 47.75\% while retaining the 40B PT checkpoint and the 10B continuation horizon of 2,384 steps.
This setting uses approximately 4.77B FineMath tokens.
The optimizer, data sources, batch size, and peak and final LRs are unchanged.
TS is tested at final MT ratios of 100\%, 90\%, 80\%, 60\%, and 47.75\%.
ZL is tested at $\gamma_{\mathrm{end}}=100\%$.
The group contains 28 runs across the basic LR rules and the additional decay windows.
At $\gamma_{\mathrm{end}}=100\%$, the TS switch occurs at step 1,246 and the ZL ramp starts at step 107, giving MT data windows of 1,138 and 2,277 steps, respectively.

\begin{figure}[!htbp]
\centering
\includegraphics[width=\linewidth]{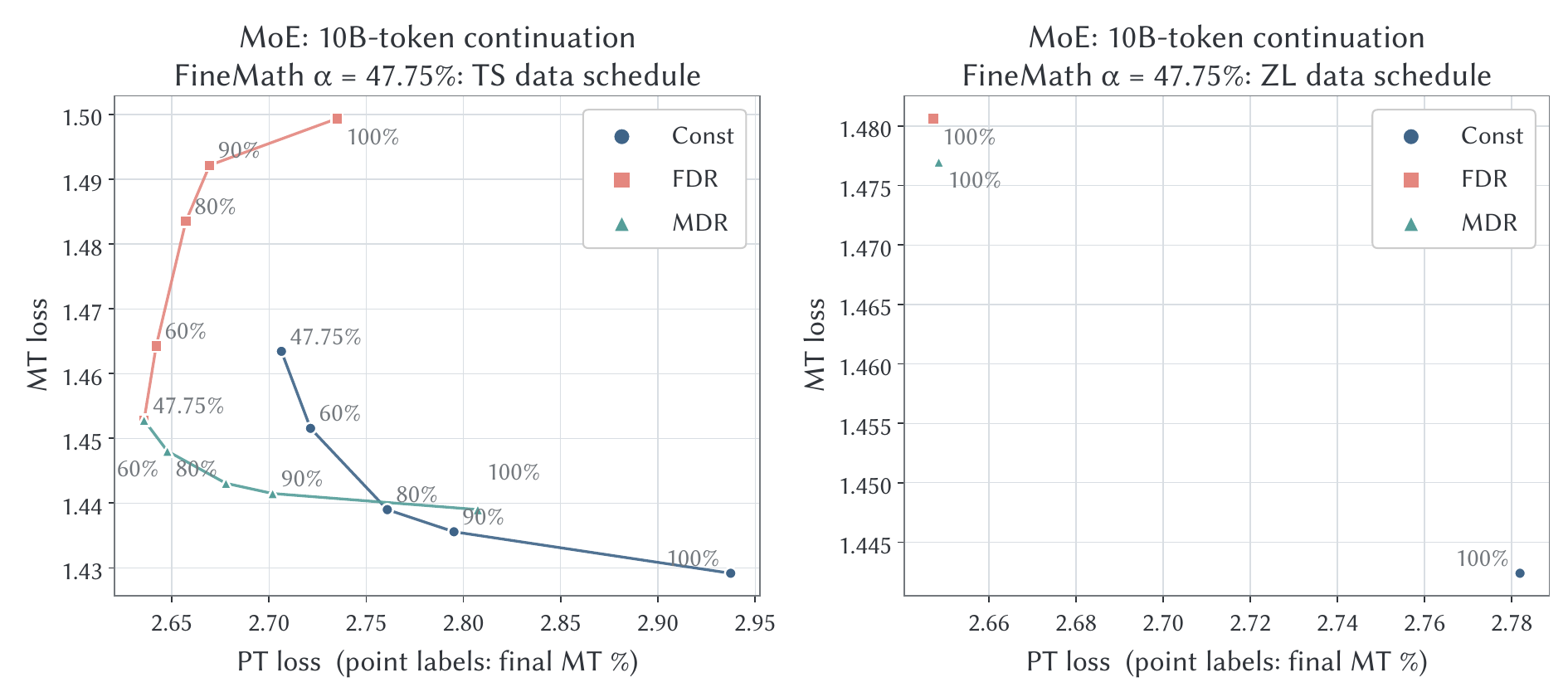}
\par
\includegraphics[width=0.49\linewidth]{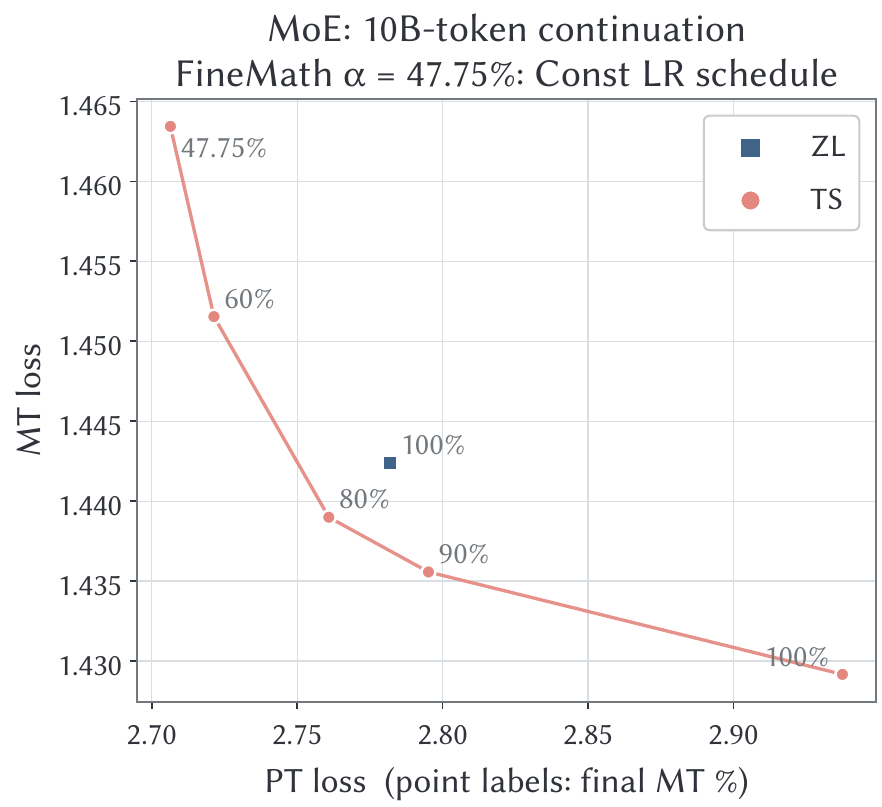}\hfill
\includegraphics[width=0.49\linewidth]{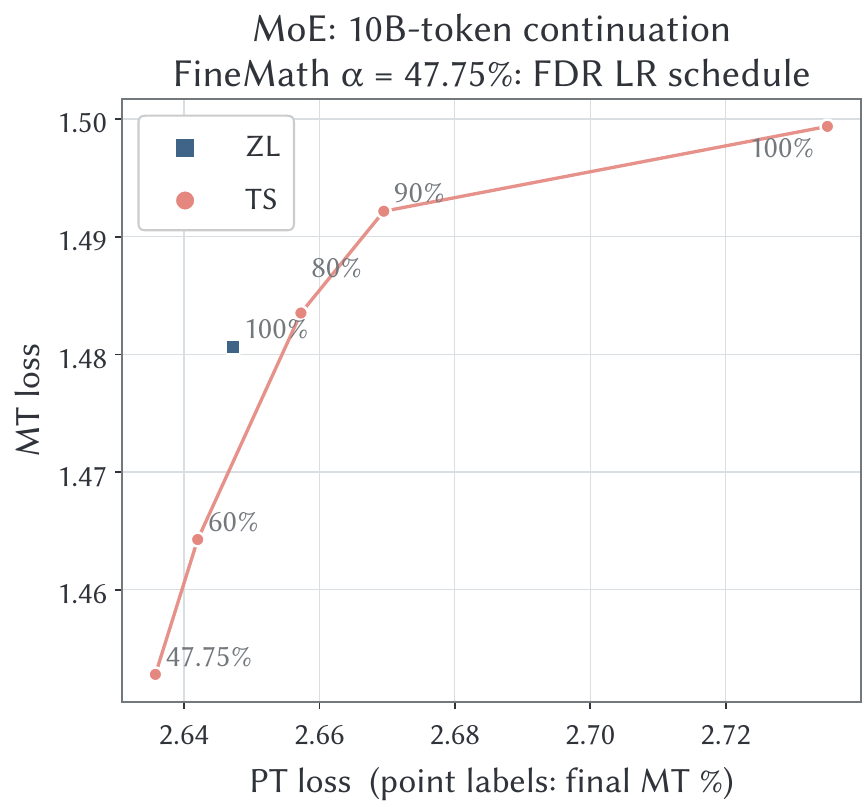}
\par
\includegraphics[width=0.49\linewidth]{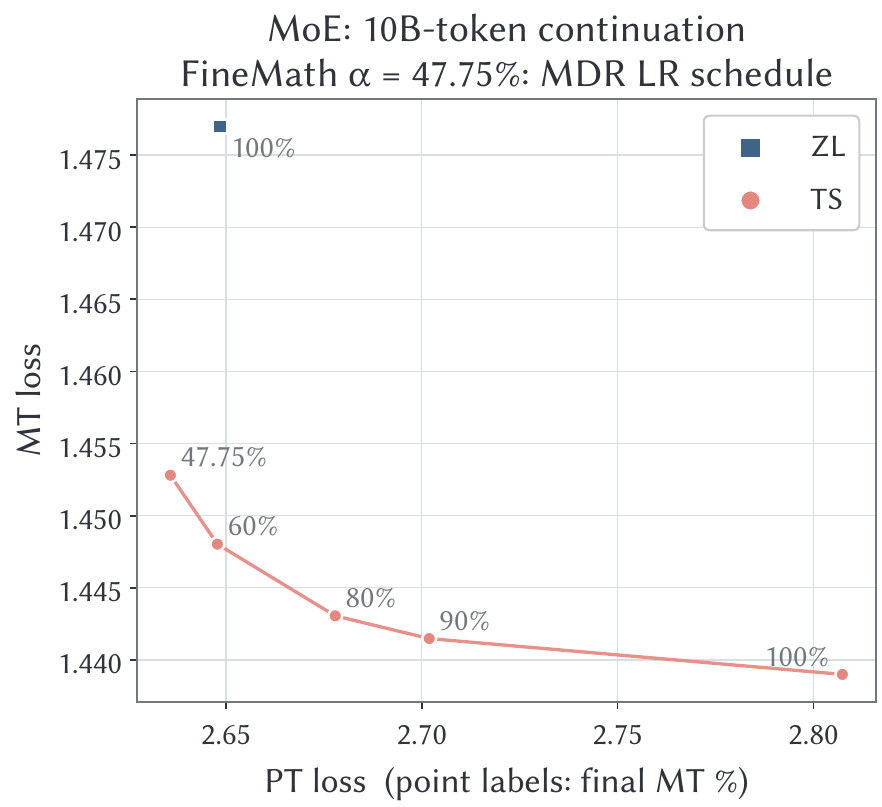}
\caption{\textbf{Const and FDR favor opposite TS endpoints at a larger MoE MT budget.} Const favors a final MT ratio of 100\%, while FDR favors 47.75\%. At 100\% MT, ZL improves on TS under FDR but has higher MT loss under Const and MDR. These runs use 10B continued training tokens with a 47.75\% MT budget, starting from the 40B-token PT checkpoint. ZL is tested only at 100\% MT. Each point is one run.}
\label{fig:app-moe4775-grid}
\end{figure}

The ranking reversal phenomenon between Const and FDR persists at this larger budget (\Cref{fig:app-moe4775-grid}).
Const favors a final MT ratio of 100\%, with MT loss 1.4292, while FDR favors the lowest tested final MT ratio, 47.75\%, with loss 1.4528.
The range of FDR losses across TS endpoints decreases from 0.094 at the 25\% budget to 0.047 at the 47.75\% budget.
At $\gamma_{\mathrm{end}}=100\%$, MDR gives 1.4390, remaining 0.0098 above Const despite the longer MT data window.
At this endpoint, ZL improves on TS under FDR by 0.0188, but has higher MT loss under Const and MDR by 0.0132 and 0.0380, respectively.
Thus, increasing the MT budget preserves the dependence on the LR rule without making either MDR or a smooth data transition uniformly preferable.

\begin{figure}[!htbp]
\centering
\includegraphics[width=\linewidth]{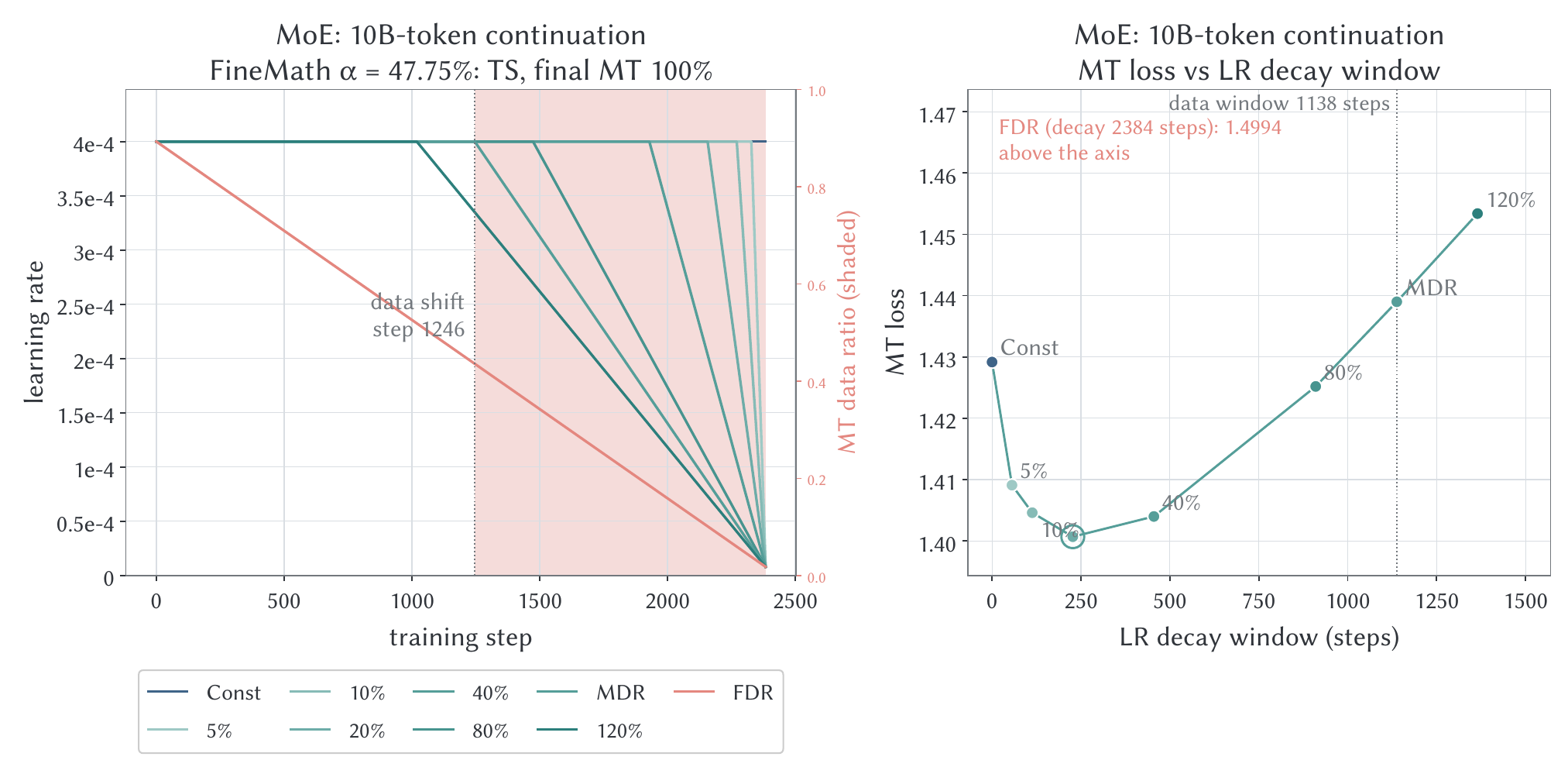}
\par
\includegraphics[width=\linewidth]{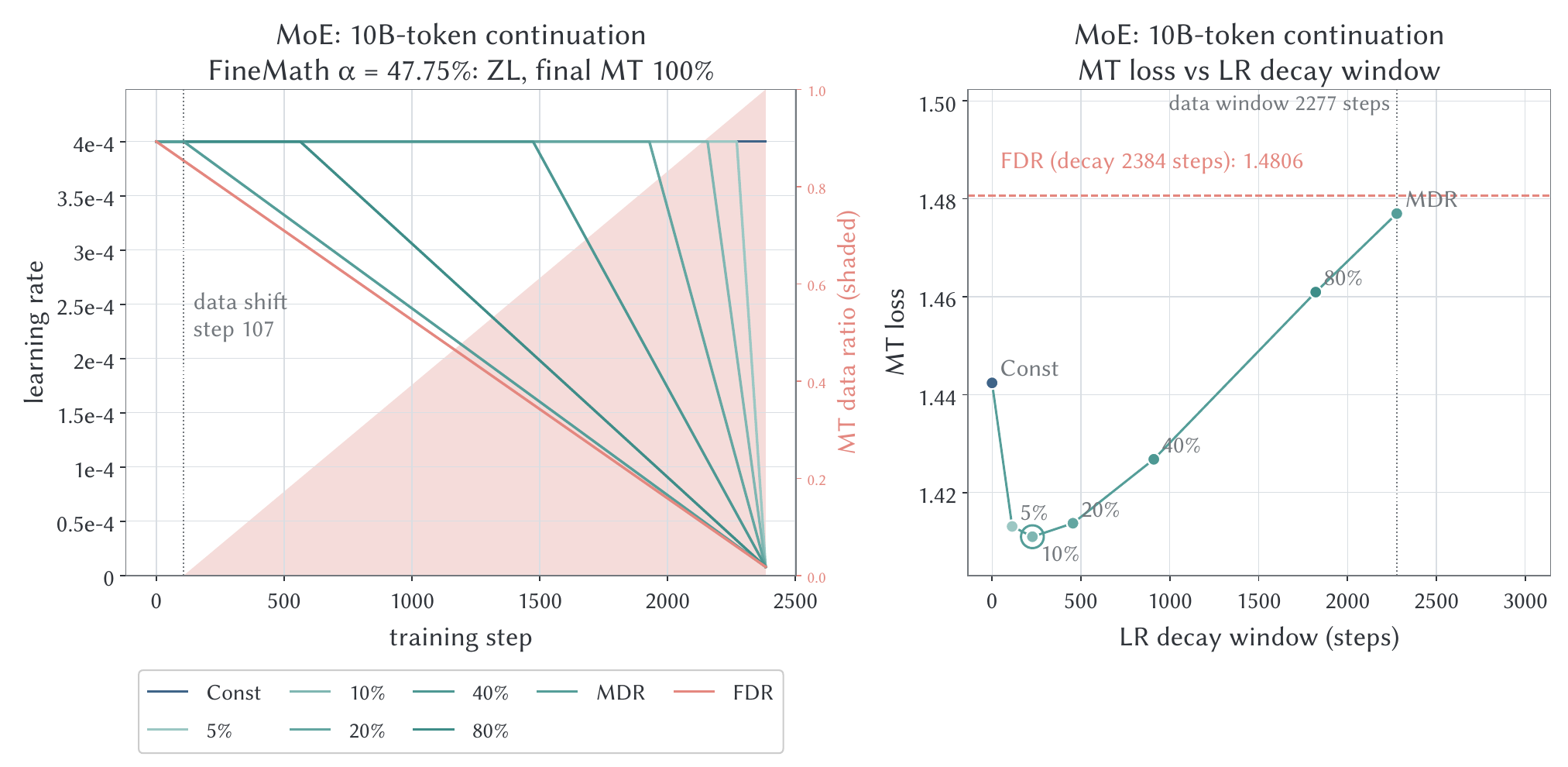}
\caption{\textbf{Increasing the MoE MT budget preserves the best tested relative LR decay durations.} TS (top) favors 20\%, and ZL (bottom) favors 10\%; both correspond to 228 decay steps. These runs use 10B continued training tokens with a 47.75\% MT budget and end at 100\% MT. Their MT windows last 1,138 and 2,277 steps, respectively. These configurations have no data stream repeats.}
\label{fig:app-moe4775-decay}
\end{figure}

A short final decay improves both data shapes (\Cref{fig:app-moe4775-decay}).
The best tested TS window covers 20\% of its MT data window and reaches 1.4007, while the best tested ZL window covers 10\% and reaches 1.4110.
Both selected windows span 228 steps because ZL has an MT data window approximately twice as long.
Relative to their own Const baselines, these losses are lower by 0.0285 and 0.0314, respectively.
\Cref{tab:app-decay-robustness} reports all sampled values, and \Cref{tab:app-decay-results} places these minima alongside the other sweeps.
No data stream replicas were run for this budget, so the comparisons describe the observed runs rather than uncertainty across repetitions.

\clearpage
\paragraph{Earlier checkpoints from the same PT trajectory.}
We repeat a subset of the 4B-token MoE continuations from checkpoints saved after 4,800 and 7,200 PT steps, corresponding to approximately 20.1B and 30.2B tokens.
We call these the 20B and 30B starting points and compare them with the main 40B checkpoint at step 9,536.
All three checkpoints come from the same PT run.
Each continuation uses 954 steps, a fixed 25\% MT budget, and the same training configuration as the main 4B-token MoE continuations, including optimizer reinitialization and no LR warmup.
At each earlier checkpoint, we test nine TS runs: Const and FDR at final MT ratios of 100\%, 50\%, and 25\%, plus MDR and 20\% and 40\% decay windows at a final MT ratio of 100\%.
At a final MT ratio of 25\%, mixing begins at step zero, so MDR and FDR share the same run.
These groups do not include ZL runs or data stream replicas.

\begin{figure}[!htbp]
\centering
\includegraphics[width=0.49\linewidth]{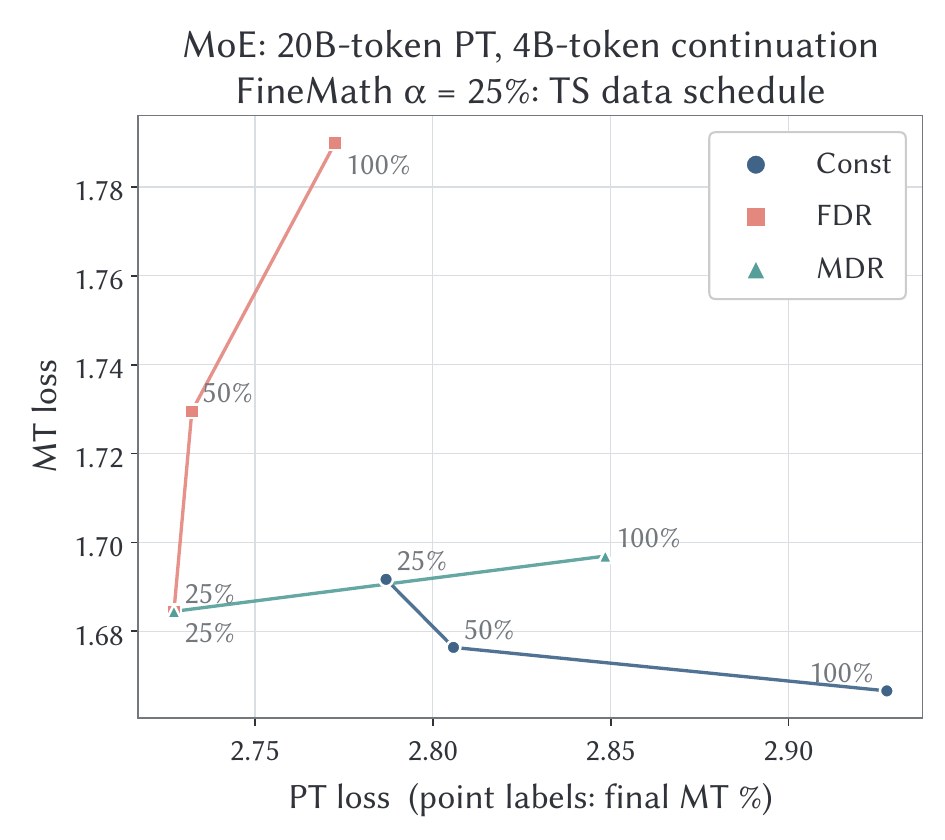}\hfill
\includegraphics[width=0.49\linewidth]{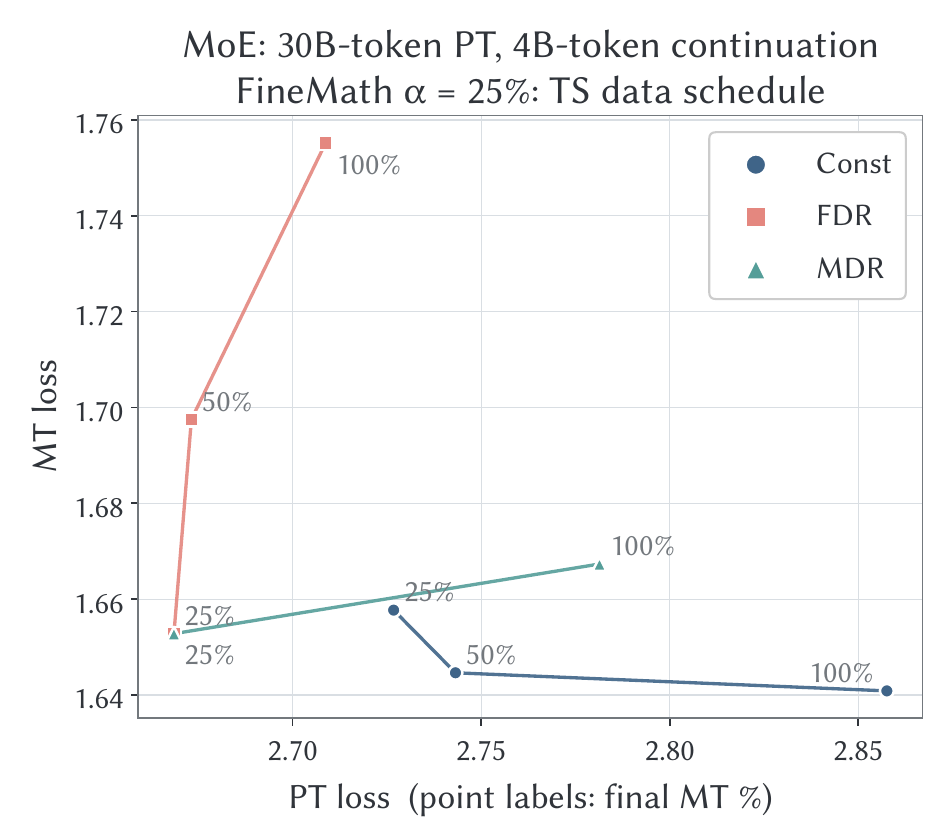}
\caption{\textbf{Const and FDR favor opposite TS endpoints even from earlier PT checkpoints.} After either 20B PT tokens (left) or 30B (right), Const gives its lowest tested MT loss when TS ends at 100\% MT, while FDR favors 25\%. Both use 4B continued training tokens and a 25\% MT budget. The checkpoints come from the same PT run.}
\label{fig:app-moe-pt-checkpoints}
\end{figure}

When we compare the same final MT ratios, changing the PT checkpoint preserves the ranking reversal phenomenon (\Cref{fig:app-moe-pt-checkpoints}).
At all three checkpoints, Const favors a final MT ratio of 100\%, while FDR favors 25\%.
Longer PT lowers the final MT loss at each shared configuration.
For example, Const at a final MT ratio of 100\% gives 1.6666, 1.6409, and 1.6176 after 20B, 30B, and 40B PT tokens, respectively.
The corresponding FDR losses exceed Const by 0.1233, 0.1143, and 0.1162.
At a final MT ratio of 25\%, FDR instead improves on Const by 0.0072, 0.0049, and 0.0021.
These comparisons show that the ranking reversal phenomenon is present across the sampled checkpoints, without requiring that the model first completes the full 40B PT run.

\begin{figure}[!htbp]
\centering
\includegraphics[width=\linewidth]{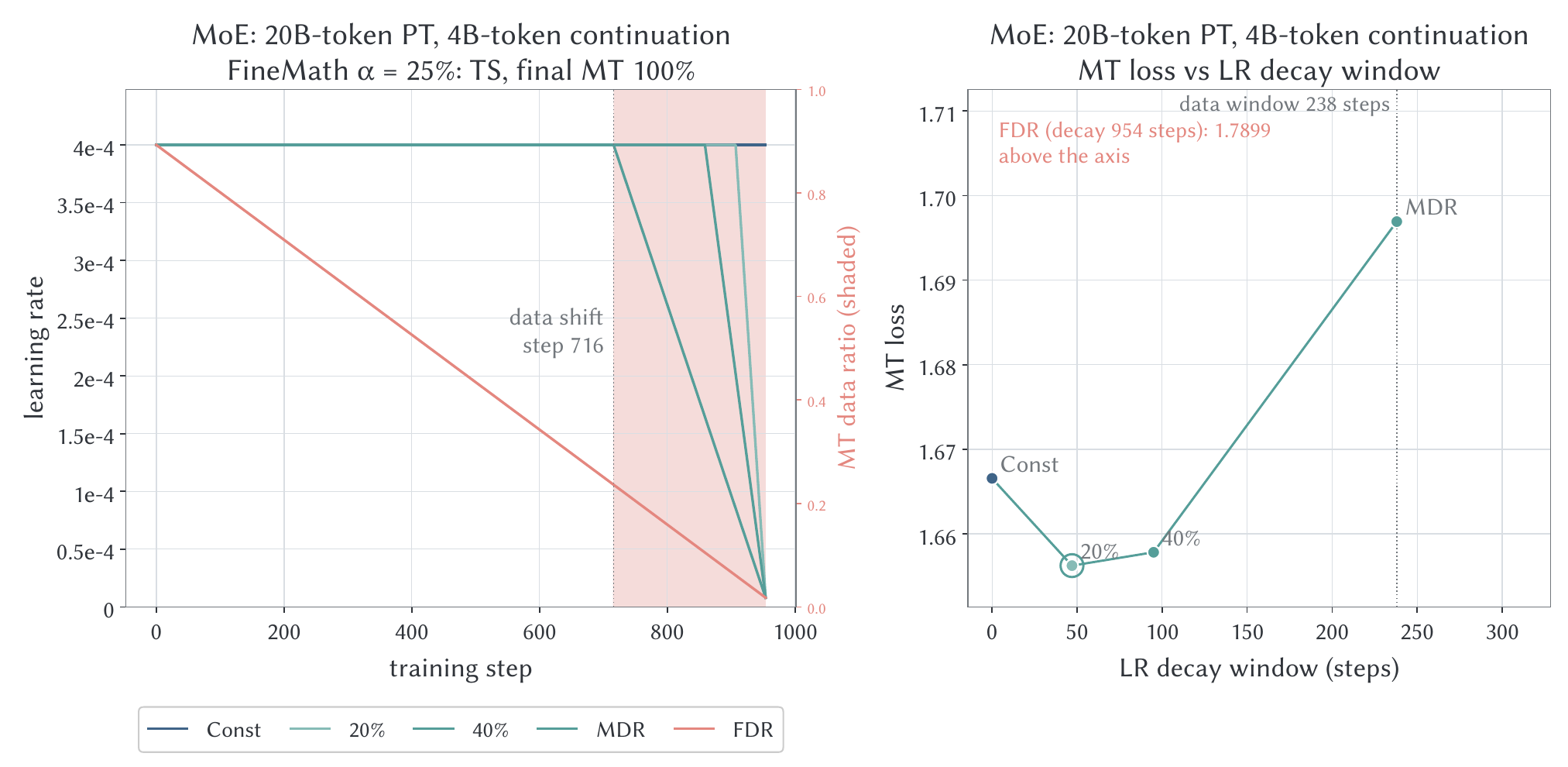}
\par
\includegraphics[width=\linewidth]{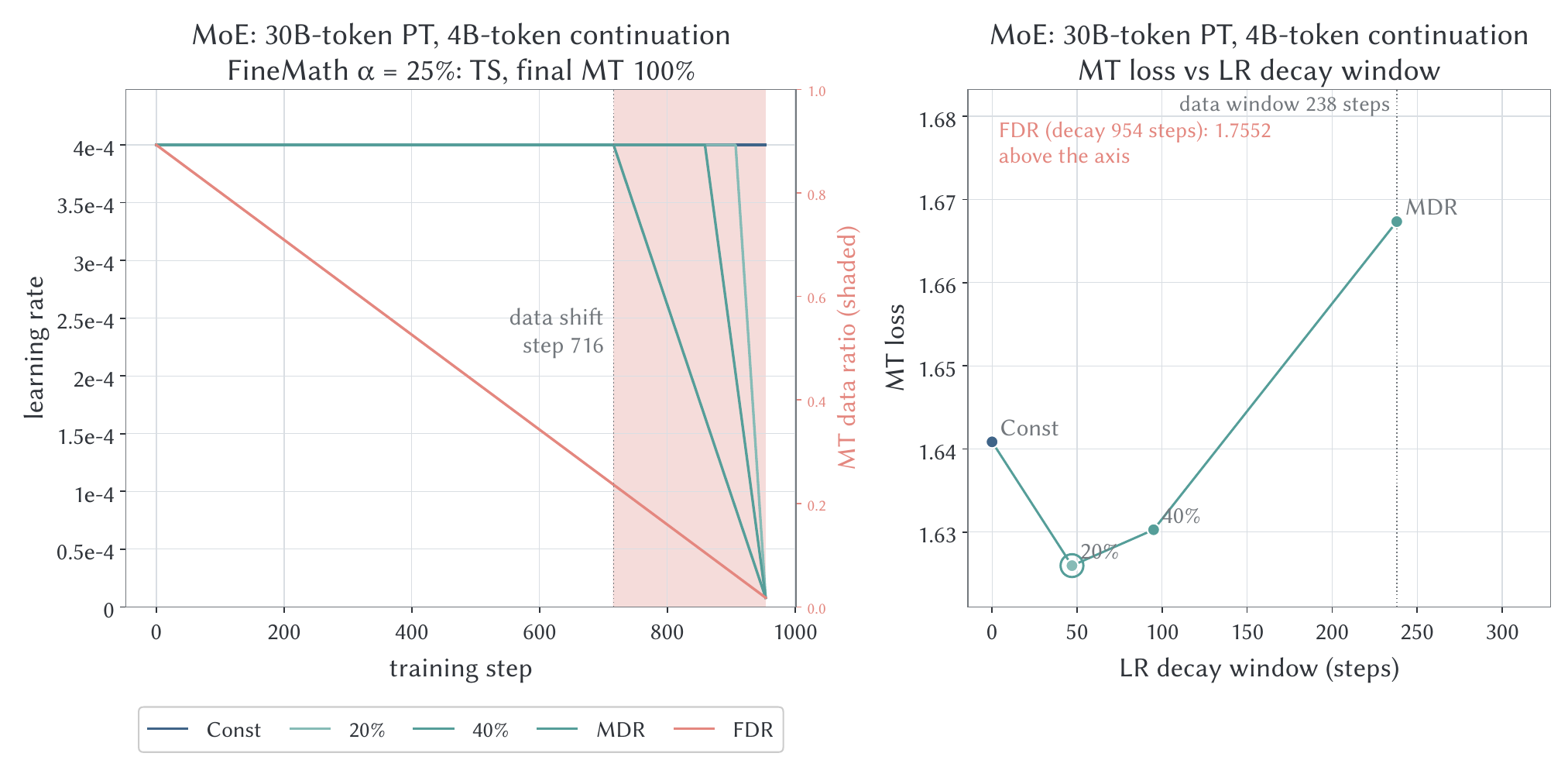}
\caption{\textbf{Delaying LR decay improves on Const and MDR from both earlier PT checkpoints.} The lowest tested MT losses are 1.6562 after 20B PT tokens (top) and 1.6260 after 30B (bottom), both at a relative LR decay duration of 20\% (47 steps). Each MoE TS run uses 4B continued training tokens, a 25\% MT budget, and a final MT ratio of 100\%, giving a 238-step MT window. Only 20\% and 40\% are tested beyond Const, MDR, and FDR.}
\label{fig:app-moe-pt-decay}
\end{figure}

\begin{figure}[!htbp]
\centering
\includegraphics[width=\linewidth]{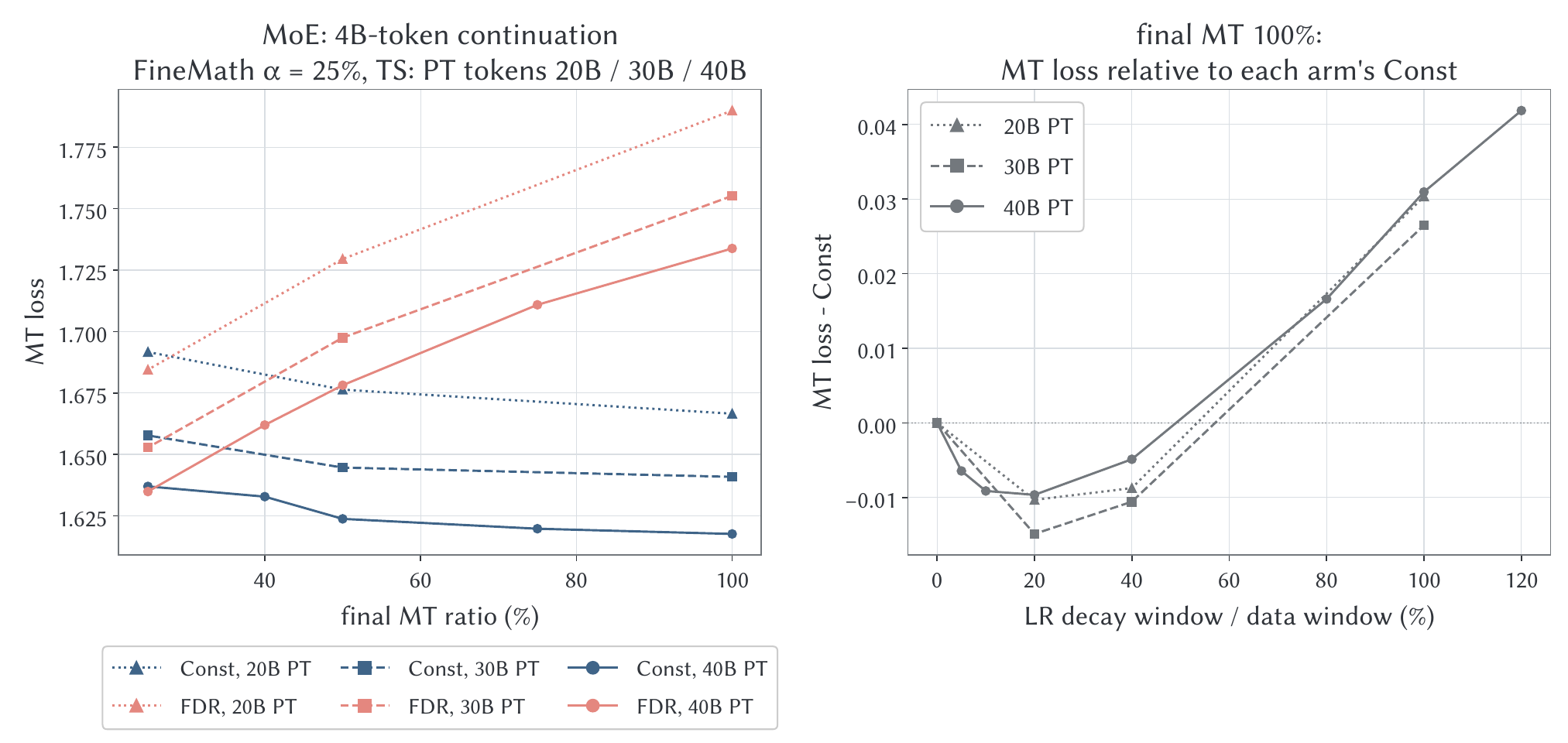}
\caption{\textbf{Changing the PT checkpoint preserves the preferred data and LR schedules among the shared configurations.} Const and FDR favor opposite TS endpoints (left). A relative LR decay duration of 20\% gives the lowest tested loss, while MDR remains worse than Const (right). The right panel shows loss changes relative to each checkpoint's own Const baseline at a final MT ratio of 100\%. These MoE runs start from 20B, 30B, and 40B checkpoints of the same PT trajectory and use 4B continued training tokens with a 25\% MT budget. The 40B checkpoint has additional sampled durations.}
\label{fig:app-moe-pt-length}
\end{figure}

The preference for a short final decay also persists (\Cref{fig:app-moe-pt-decay,fig:app-moe-pt-length}).
At a final MT ratio of 100\%, MT data enters at step 716 and occupies the last 238 steps.
The lowest tested MT loss occurs at a 20\% window for all three checkpoints, improving on Const by 0.0104, 0.0149, and 0.0096 for the 20B, 30B, and 40B starting points, respectively.
MDR instead increases MT loss by 0.0303, 0.0264, and 0.0310.
The earlier checkpoints test only the 20\% and 40\% windows in addition to Const, MDR, and FDR.
They therefore support the same ordering among these choices, but do not locate a continuous optimum or establish the shape of the curve near its minimum.
Full values for the earlier checkpoints appear in \Cref{tab:app-decay-robustness}.

\paragraph{Changing the MT dataset.}
StarCoderData is a collection of source code from GitHub released for training StarCoder~\citep{exp_starcoder2023}.
It is derived from The Stack, with additional cleaning, removal of near duplicates, and filtering of personal information.
We use its \href{https://huggingface.co/datasets/bigcode/starcoderdata}{Python subset} as MT data to test whether the preferred relative LR decay durations persist when the target domain changes from mathematics to code.

We replace FineMath with StarCoderData Python, starting from the same 40B PT checkpoint and continuing training for 10B tokens with a 25\% MT budget.
Both TS and ZL end at a final MT ratio of 100\%. PT data, tokenizer, batch size, and LR settings remain unchanged.
These two scans add 15 runs. Both include Const and MDR, while FDR is tested only for TS. ZL does not include a 5\% window, and neither scan includes data sampling replicas.

\begin{figure}[!htbp]
\centering
\includegraphics[width=\linewidth]{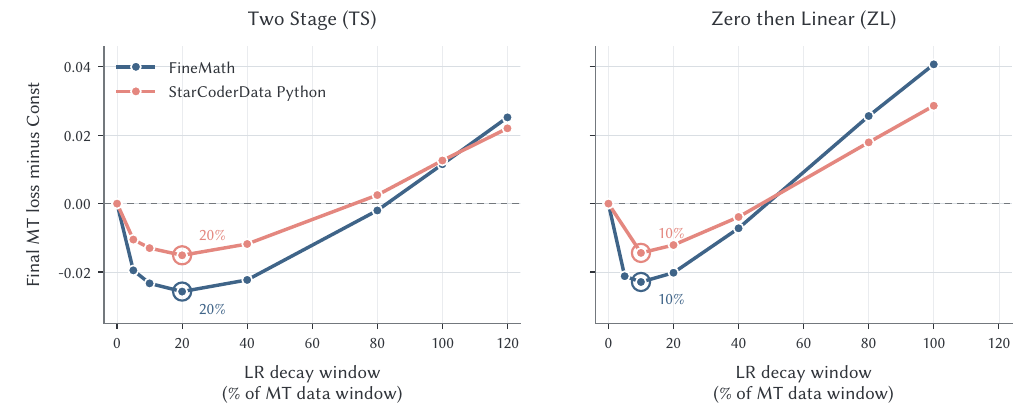}
\caption{\textbf{Changing MT data from mathematics to code preserves the best tested relative LR decay duration} for both TS (left) and ZL (right). Both datasets use the same MoE model and 10B-token continuation with a 25\% MT budget and final MT ratio 100\%. The horizontal axis gives decay duration as a percentage of each schedule's MT data window, with zero denoting Const and 100\% denoting MDR. The vertical axis gives final MT loss minus the corresponding Const loss for that dataset and data schedule. Points show measured values, and rings mark the lowest tested losses at 20\% for TS and 10\% for ZL. Lines connect tested values.}
\label{fig:app-moe-mt-domain}
\end{figure}

\textbf{Changing MT data from mathematics to code preserves the best tested relative LR decay duration.}
Figure~\ref{fig:app-moe-mt-domain} compares each scan with its FineMath counterpart after subtracting its own Const loss.
For both datasets, the lowest tested loss occurs at 20\% for TS and 10\% for ZL, corresponding to the same 119 steps of LR decay.
The improvement over Const differs between datasets, but the best relative durations among those tested remain unchanged.
MT loss is evaluated on each dataset separately, so absolute losses should not be compared across the two datasets.

\clearpage

\section{Router Dynamics and Interpretation}
\label{app:exp-router}

\begin{figure}[!htbp]
\centering
\includegraphics[width=\linewidth]{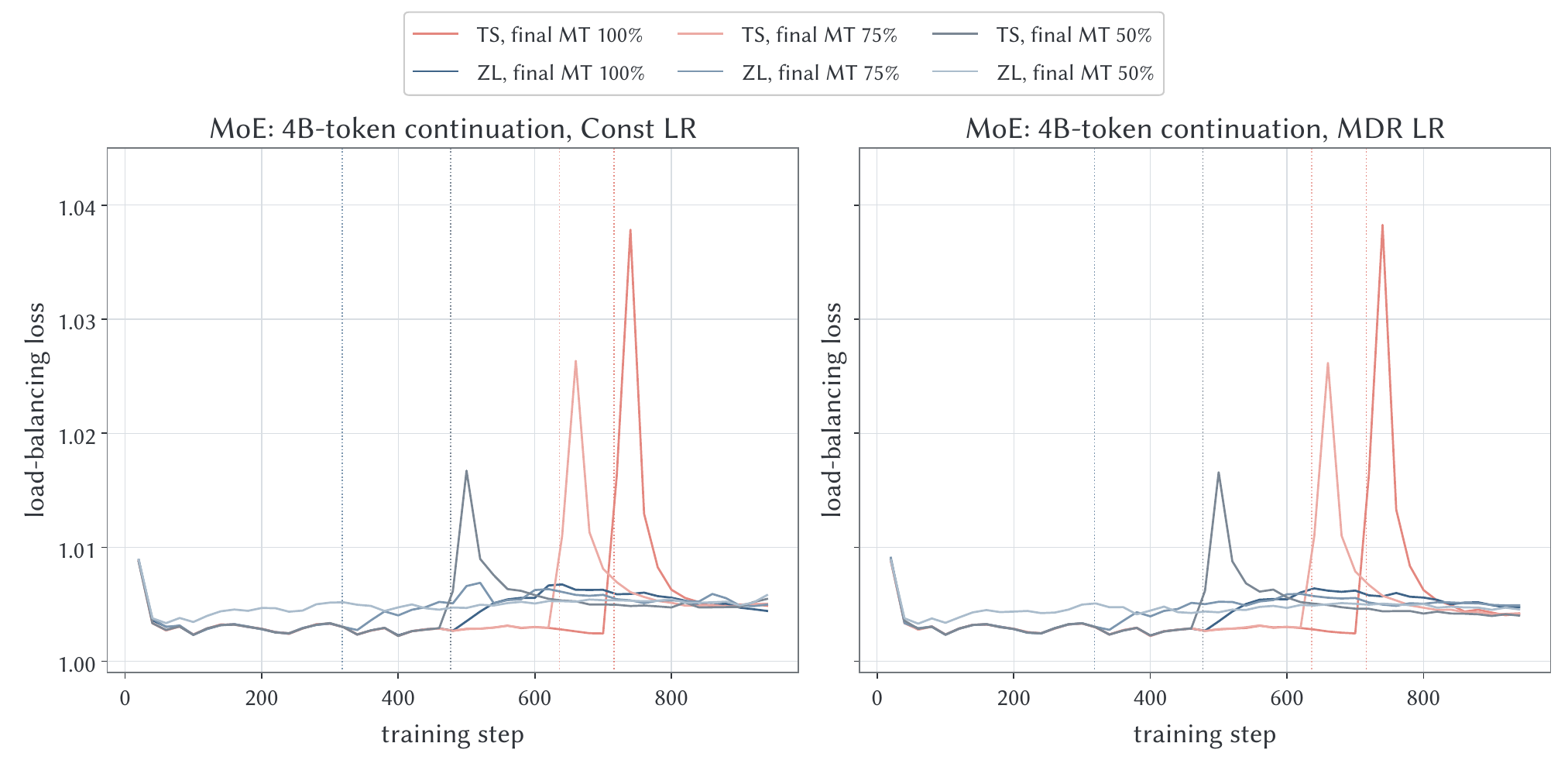}
\par
\includegraphics[width=\linewidth]{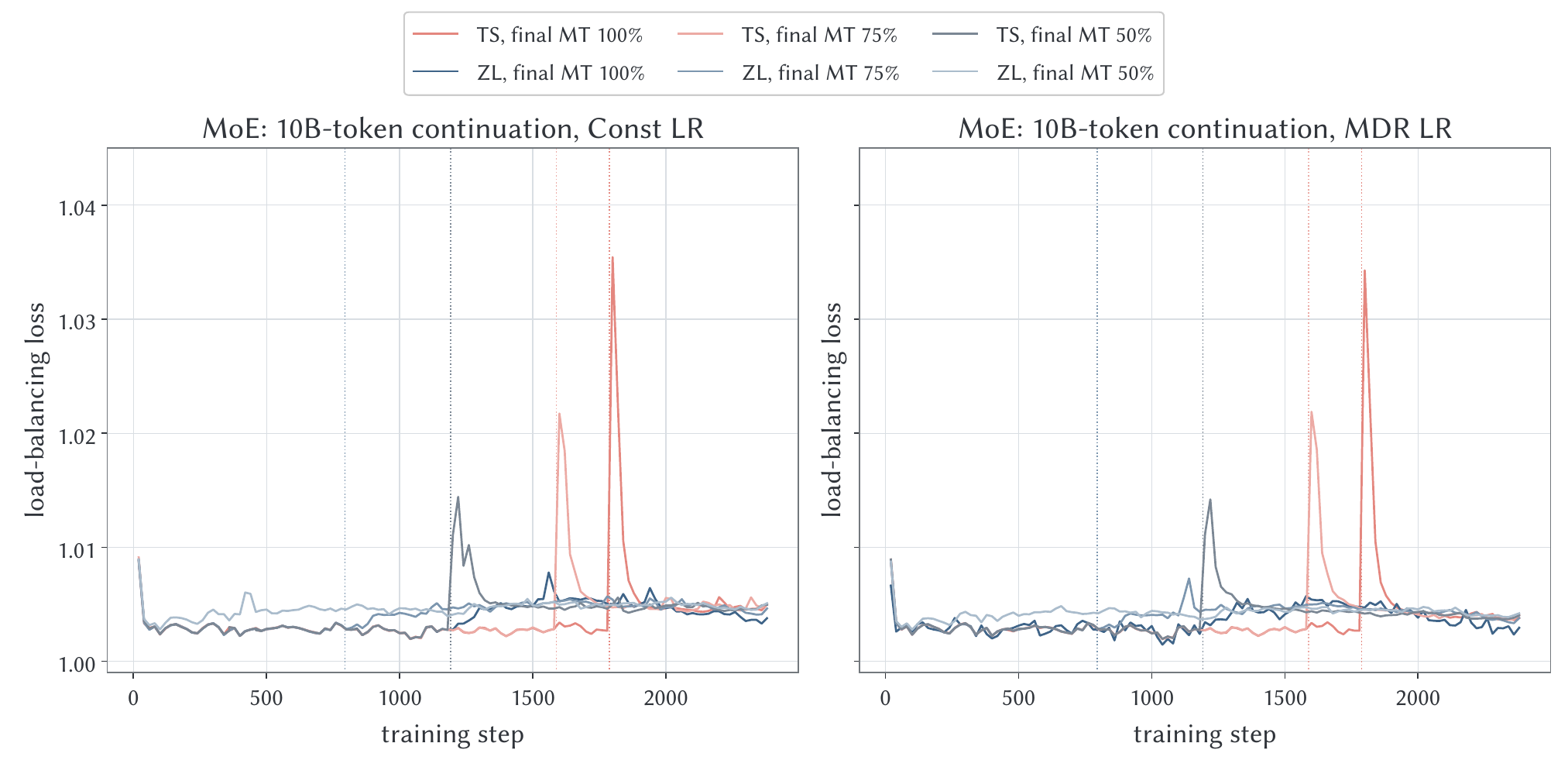}
\caption{\textbf{Gradual MT entry reduces load balancing spikes.} Under Const (left), TS and ZL use the same LR at every step, allowing us to compare the data schedules without also changing the LR. TS produces a sharp spike when the MT ratio suddenly increases, while ZL avoids a comparable spike by introducing MT data gradually. MDR (right) shows a similar pattern, although LR decay begins at different times for TS and ZL. Rows show MoE experiments with 4B (top) and 10B (bottom) tokens of continued training. Curves compare final MT ratios of 100\%, 75\%, and 50\%. Dotted lines mark when MT data first enters. The 10B results also appear in Figure~\ref{fig:exp-router-smoothing}.}
\label{fig:app-router-lr-control}
\end{figure}

\begin{figure}[!htbp]
\centering
\includegraphics[width=\linewidth]{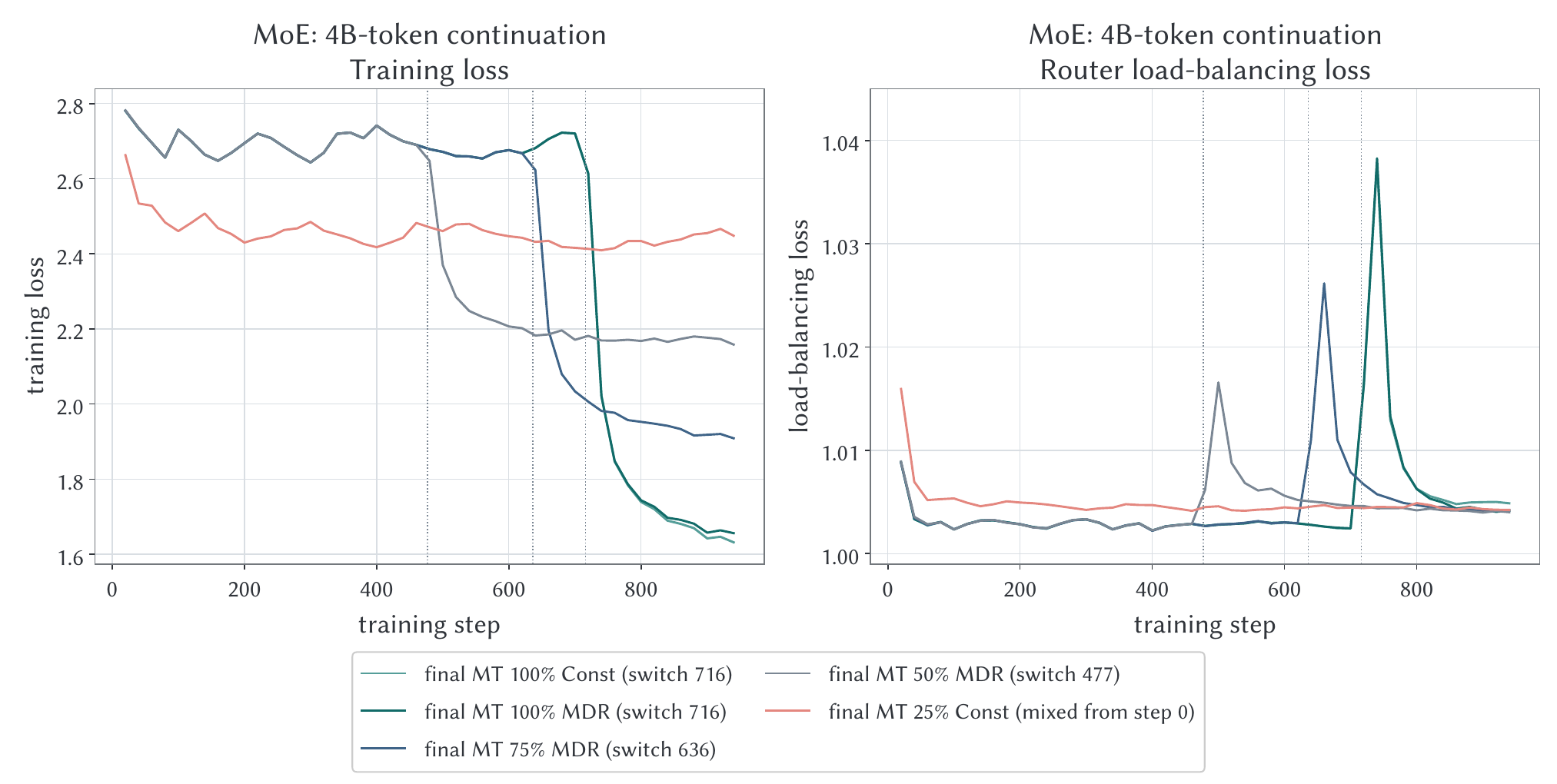}
\par
\includegraphics[width=\linewidth]{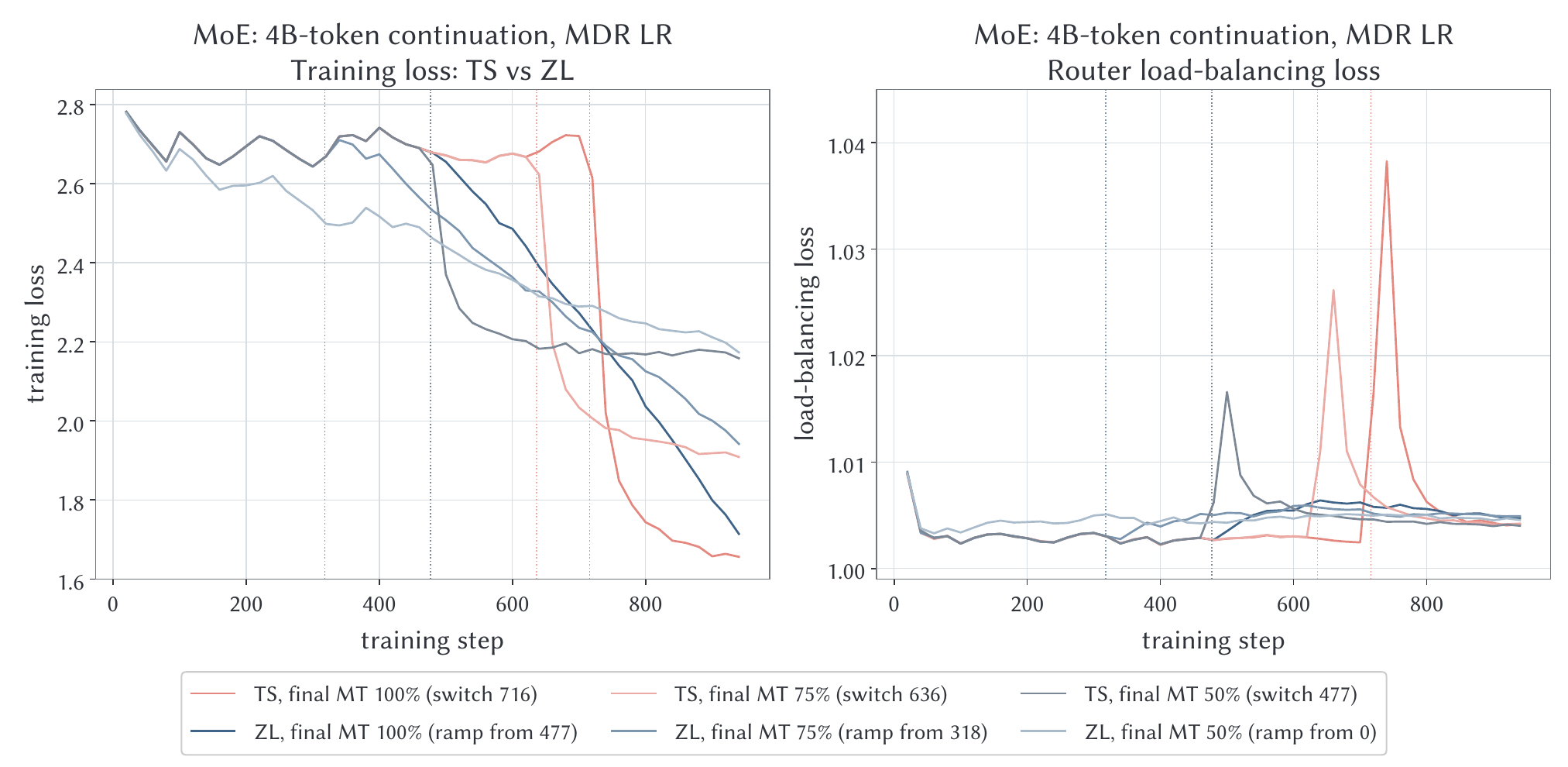}
\caption{\textbf{TS produces a transient load balancing spike, while gradual MT entry avoids a comparable peak.} The upper panels compare TS at several final MT ratios, including Const and MDR at 100\% MT; the lower panels compare TS and ZL under MDR at final MT ratios of 100\%, 75\%, and 50\%. These MoE runs use 4B continued training tokens. Training loss (left) and the logged load balancing statistic (right) are shown every 20 training steps.}
\label{fig:app-router-switch4}
\end{figure}

\begin{figure}[!htbp]
\centering
\includegraphics[width=\linewidth]{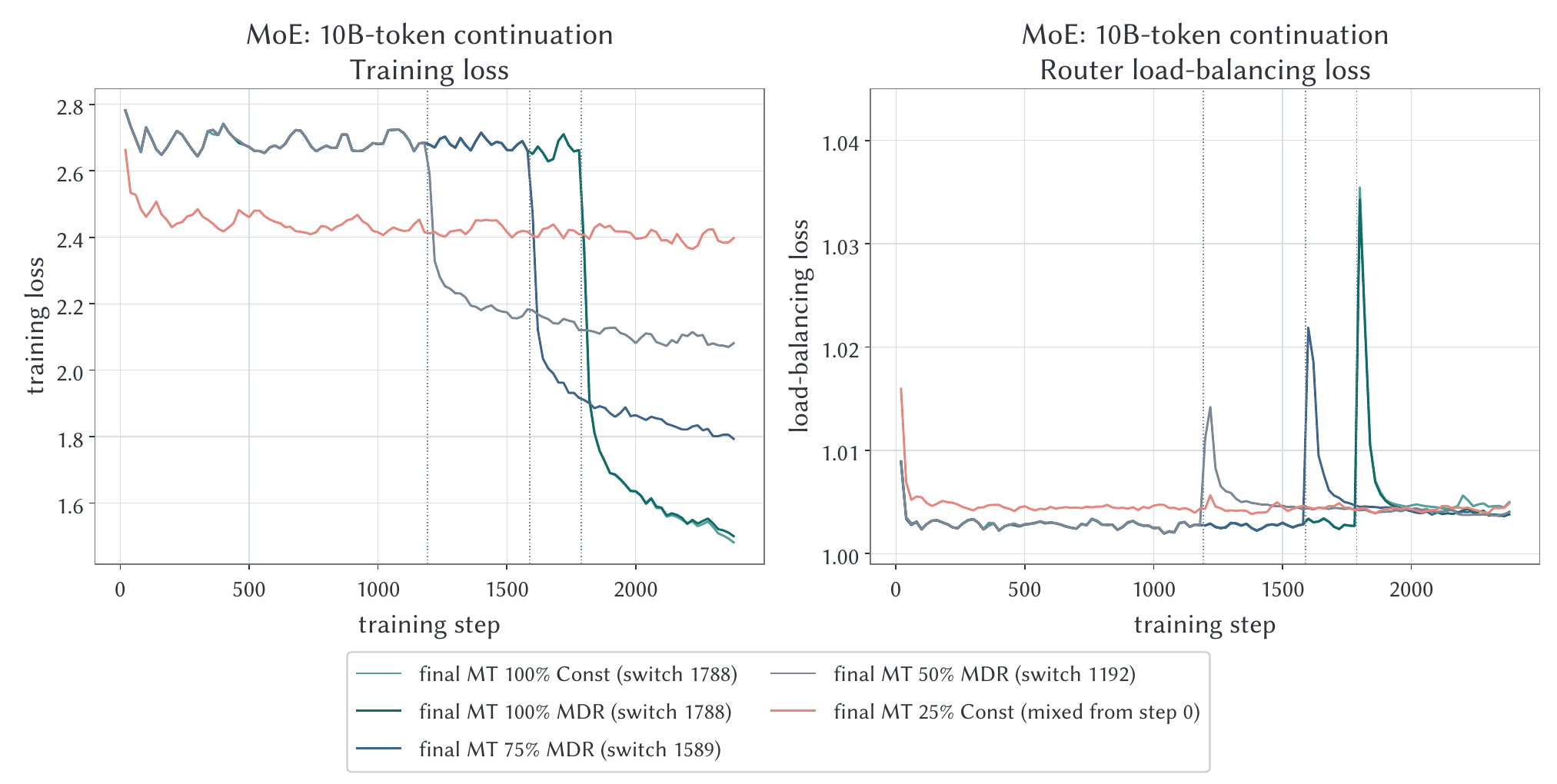}
\par
\includegraphics[width=\linewidth]{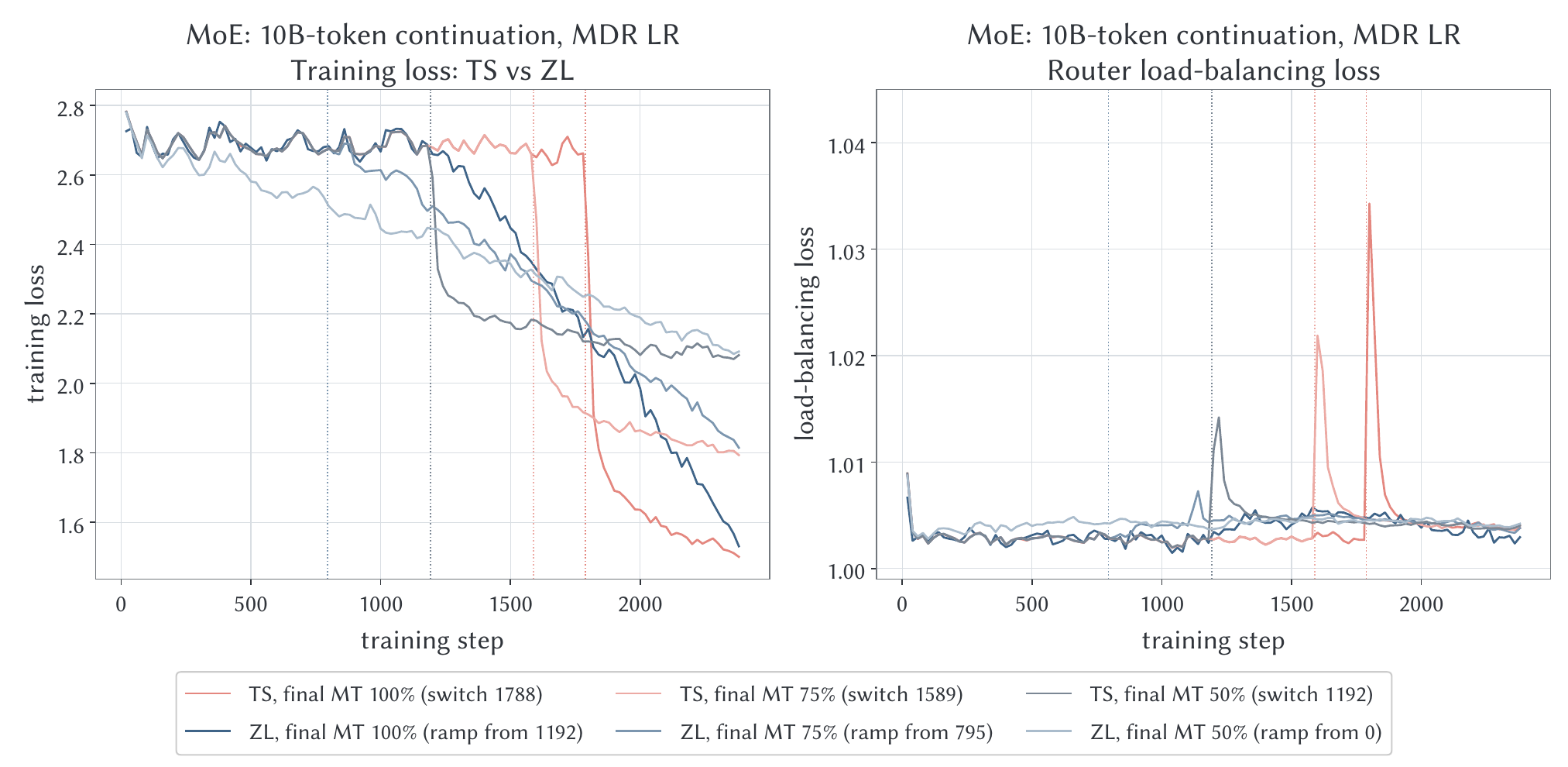}
\caption{\textbf{Smaller TS data shifts produce smaller load balancing spikes, while ZL avoids a comparable peak.} The upper panels compare TS switches at steps 1,788, 1,589, and 1,192 for final MT ratios of 100\%, 75\%, and 50\%; the lower panels compare TS and ZL under MDR at the same endpoints. These MoE runs use 10B continued training tokens. Training step is horizontal, with training loss on the left and logged load balancing on the right. Final validation losses are reported separately in \Cref{tab:app-moe-minima,tab:app-moe-shape-difference}.}
\label{fig:app-router-switch10}
\end{figure}

\paragraph{Measurement.}
We use training logs to track changes during training and separate checkpoint evaluations to examine individual expert loads. The training curves in \Cref{fig:exp-router-smoothing,fig:app-router-lr-control,fig:app-router-switch4,fig:app-router-switch10} show the auxiliary load balancing loss on the current mixture of PT and MT data, with its coefficient removed before logging. For a training microbatch, the value for layer $\ell$ is
\[
B_\ell=64\sum_{j=1}^{64}f_{\ell j}P_{\ell j},
\]
where $f_{\ell j}$ is the fraction of routing assignments sent to expert $j$, counting all eight assignments per token, and $P_{\ell j}$ is the mean router softmax probability for that expert before selecting the top eight experts. Uniform routing gives $B_\ell=1$. The training objective multiplies this loss by the auxiliary loss coefficient, but Megatron divides out the coefficient before recording \texttt{load\_balancing\_loss}. Each plotted value is the unweighted mean across the 12 MoE layers, also averaged over microbatches, data parallel workers, and the 20 training steps in the logging interval. We plot these logged values without further rescaling.

For the checkpoint evaluations, we measure expert loads directly from routing assignments on held-out DCLM and FineMath, using approximately $34$M tokens per domain at each checkpoint. For each of the 12 layers, let $p_{\ell j}$ be the number of assignments sent to expert $j$ divided by the total number of assignments in that layer, so that the fractions sum to one across the 64 experts. We measure \textbf{expert load imbalance} using
\begin{equation}
I_\ell=64\sum_{j=1}^{64}p_{\ell j}^{\,2}.
\label{eq:app-router-metrics}
\end{equation}
Uniform routing gives $I_\ell=1$, and larger values indicate greater expert load imbalance.
The heatmaps and regressions express expert shares relative to uniform routing as $64p_{\ell j}$, so a value of one denotes the uniform share.
The training statistic combines assignment fractions with router probabilities, whereas $I_\ell$ uses only assignment counts. The two quantities are therefore different, even though both equal one under uniform routing. We use the training curves to locate transient spikes and the checkpoint measurements to examine how expert loads change on each domain, without directly comparing their numerical values.

\paragraph{Abrupt data distribution shifts under TS produce transient imbalance.}
At both 4B and 10B continuation horizons, the TS data switch produces a sharp load balancing spike that subsides over approximately 80 steps (\Cref{fig:app-router-switch4,fig:app-router-switch10}).
Larger jumps in the MT ratio produce larger spikes, whereas ZL avoids a comparable peak as MT data enters gradually.
The effect is therefore concentrated around the change in training distribution.

\paragraph{ZL reduces load balancing spikes under the same LR schedule.}
To separate the effect of the data transition from that of LR decay, we compare TS and ZL under Const.
Both schedules use the same LR at every training step, yet TS produces a sharp load balancing spike while ZL avoids a comparable spike (\Cref{fig:exp-router-smoothing,fig:app-router-lr-control}).
This difference persists across both training lengths and all three tested final MT ratios, supporting the benefit of gradually introducing MT data.

\begin{figure}[!htbp]
\centering
\includegraphics[width=\linewidth]{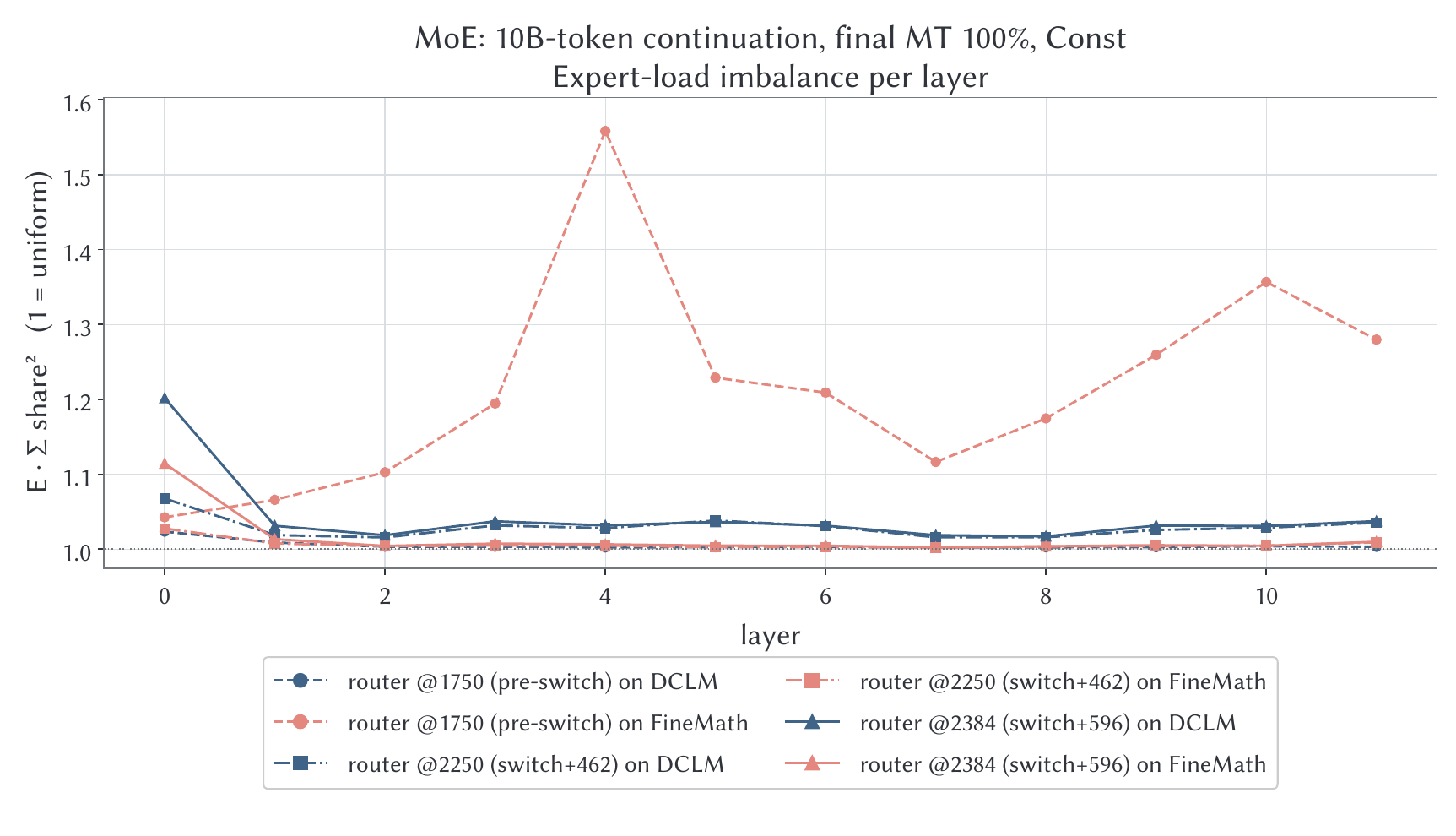}
\par
\includegraphics[width=\linewidth]{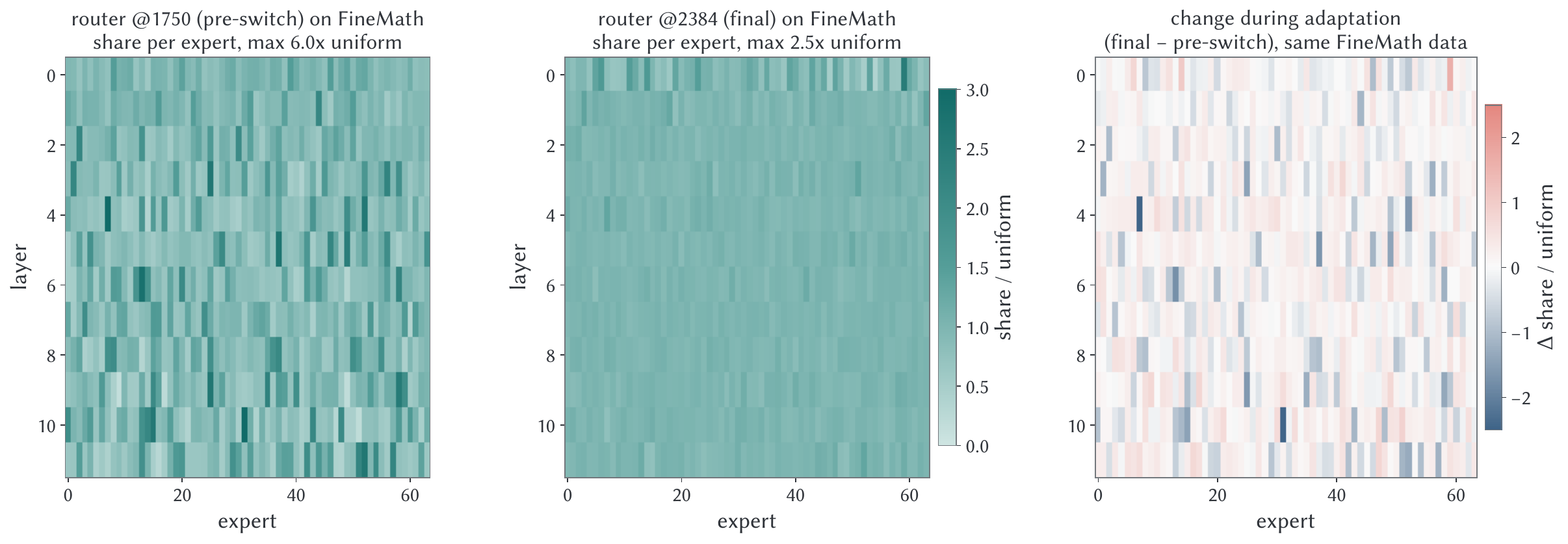}
\caption{\textbf{The router initially assigns MT tokens unevenly across experts and becomes more balanced during MT training.} Checkpoint routing diagnostics for the MoE TS+Const run with 10B continuation tokens and a final MT ratio of 100\%. Top: imbalance by layer on DCLM and FineMath at steps 1,750, 2,250, and 2,384. Bottom: FineMath expert shares before the switch, at training end, and their difference, with expert and layer indices as coordinates. Shares are relative to uniform routing. These diagnostics describe one run, without replica averaging.}
\label{fig:app-router-checkpoints}
\end{figure}

\paragraph{Expert loads change as the router adapts to MT data.}
To understand the load balancing spike under TS, we examine checkpoints before the data switch, during MT training, and at training completion in the MoE TS+Const run with 10B continuation tokens and a final MT ratio of 100\%.
At each checkpoint, we separately evaluate how the router distributes DCLM (PT) and FineMath (MT) tokens across experts.
Before the switch, expert loads are nearly balanced on DCLM. However, evaluating the same checkpoint on FineMath shows that a few experts receive many more tokens than the others, even though MT data has not yet entered training.
This suggests that the spike occurs because the routing learned on PT data initially assigns MT tokens unevenly across experts.
As MT training proceeds, the router distributes FineMath tokens more evenly, and the measured imbalance falls substantially (\Cref{tab:app-router-checkpoints}).
The FineMath heatmaps in \Cref{fig:app-router-checkpoints} show this change directly: experts that receive a large share of tokens before the switch become less dominant by training completion.
Meanwhile, expert loads on DCLM become somewhat less balanced, showing that adaptation to MT data also changes how the router distributes PT tokens.

\begin{table}[!htbp]
\centering
\small
\begin{tabular}{@{}lrr@{}}
\toprule
Checkpoint (step) & DCLM imbalance & FineMath imbalance \\
\midrule
1,750 (before switch) & 1.005 & 1.216 \\
2,250 (462 steps after switch) & 1.029 & 1.006 \\
2,384 (final) & 1.043 & 1.015 \\
\bottomrule
\end{tabular}
\caption{Expert load imbalance on DCLM and FineMath for the MoE TS+Const run with 10B continuation tokens and a final MT ratio of 100\%. Each value is the average of $I_\ell$ across the 12 layers, as defined in \Cref{eq:app-router-metrics}. The first checkpoint is 38 steps before the data switch. These measurements evaluate each domain separately on validation data, whereas the training logs summarize routing on the changing mixture of PT and MT data.}
\label{tab:app-router-checkpoints}
\end{table}

\begin{figure}[!htbp]
\centering
\begin{subfigure}[t]{0.49\linewidth}
\centering
\includegraphics[width=\linewidth]{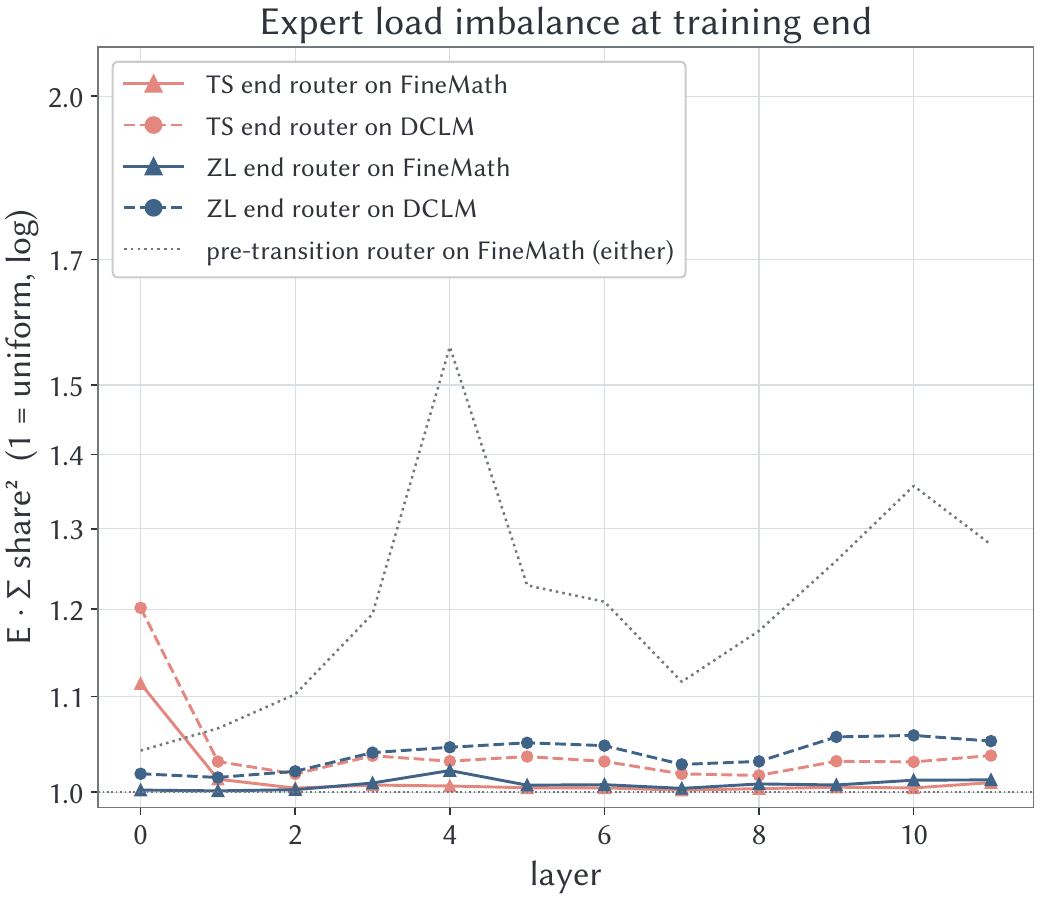}
\caption{Final expert load imbalance.}
\end{subfigure}
\hfill
\begin{subfigure}[t]{0.49\linewidth}
\centering
\includegraphics[width=\linewidth]{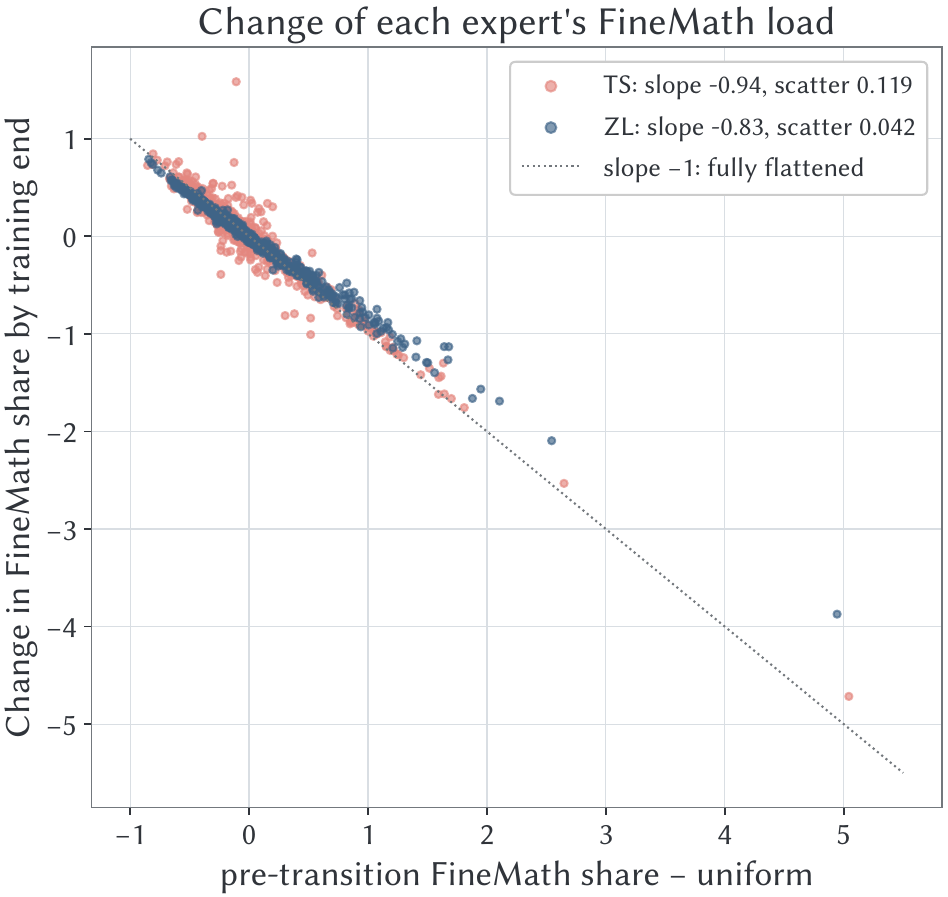}
\caption{Changes in the fraction of FineMath assignments sent to each expert.}
\end{subfigure}
\par
\begin{subfigure}[t]{0.49\linewidth}
\centering
\includegraphics[width=\linewidth]{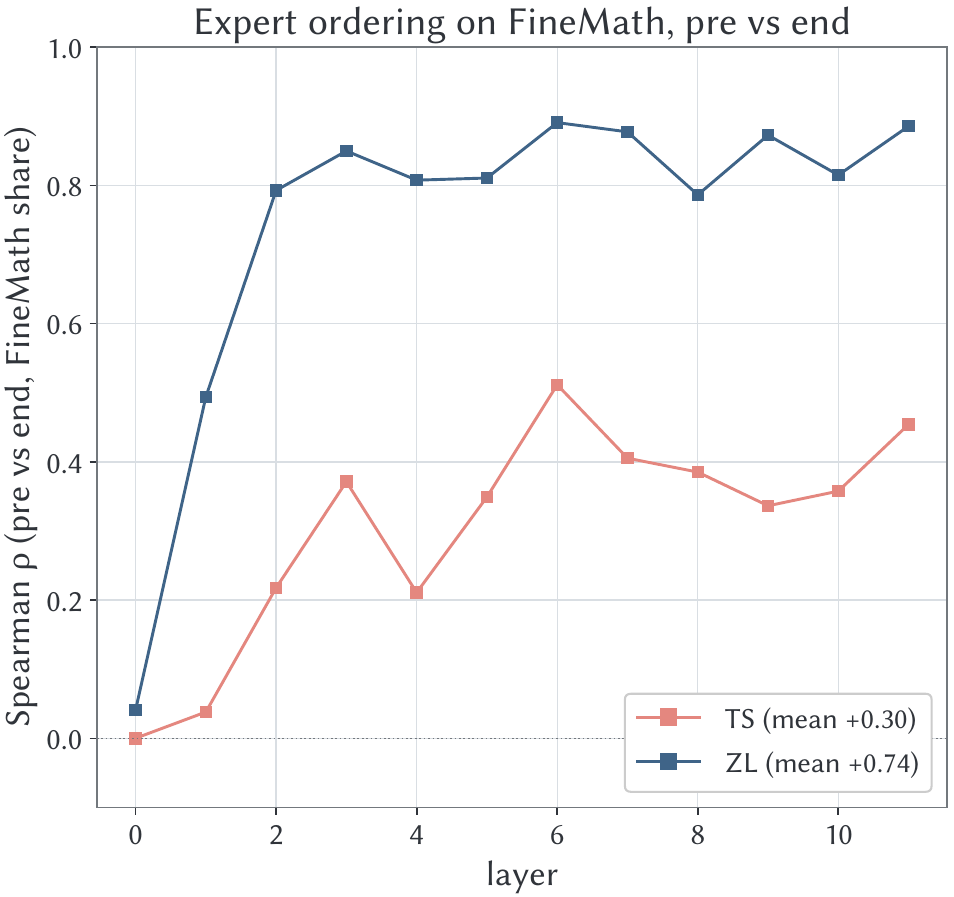}
\caption{Preservation of expert load rankings.}
\end{subfigure}
\caption{\textbf{ZL+MDR preserves expert load rankings better than TS+Const, although both reach nearly balanced MT loads.}
(a) Both schedules bring FineMath (MT) loads close to uniform in most layers; one denotes equal assignment shares across experts.
(b) Initially more heavily used experts lose share, while less used experts gain it.
Each point represents one expert in one layer. The horizontal axis shows its share before MT entry time minus the uniform share, and the vertical axis shows its final minus initial share, both in units of $1/64$. Points on the dotted line end at the uniform share.
(c) Despite similar final balance, the ordering of experts changes more under TS+Const. The higher rank correlations under ZL+MDR mean that experts receiving relatively more FineMath assignments initially tend to remain more heavily used at the end.
All checkpoint measurements use held-out data. These 10B-token MoE continuations use different LR schedules, so the comparison does not isolate the effect of the data schedule. See \Cref{tab:app-router-comparison} for the setup.}
\label{fig:app-router-rebalancing}
\end{figure}

\paragraph{How expert loads change under TS and ZL.}
The preceding analysis shows that expert loads become more balanced after the TS switch.
We next examine which experts receive more or fewer FineMath assignments after adaptation and compare these changes under TS and ZL. In panel (b) of Figure~\ref{fig:app-router-rebalancing}, the horizontal axis shows how much each expert's initial assignment share differs from the uniform share, and the vertical axis shows how its share changes by training completion.
The negative trend means that experts receiving more than the uniform share tend to receive less after adaptation, while those receiving less tend to receive more.
To compare TS and ZL, we examine both the size of these changes and whether experts that were initially more heavily used remain so afterwards.

We use three statistics to compare the above changes (\Cref{tab:app-router-comparison}). For routing shares $p$ before MT entry time and $q$ at training completion, \textbf{total variation (TV)} is
\begin{equation*}
\operatorname{TV}(p,q)=\frac{1}{12}\sum_{\ell=1}^{12}
\frac{1}{2}\sum_{j=1}^{64}|p_{\ell j}-q_{\ell j}|.
\end{equation*}
It measures the net change in the routing distribution, averaged across layers.
For example, TV of 0.15 means that matching the final shares requires redistributing at least 15\% of the assignment mass, on average across layers.
It does not measure cumulative movement during training.
The regression residual standard deviation measures how far each expert's share change deviates from the fitted relationship with its initial excess.
Spearman correlation compares the expert rankings by assignment share before and after adaptation within each layer, then averages across layers.

TS+Const and ZL+MDR show comparable net redistribution, but ZL+MDR has substantially less scatter around the fitted trend and preserves more of the initial expert ordering (\Cref{tab:app-router-comparison,fig:app-router-rebalancing}).
In other words, ZL+MDR shows a more consistent contraction of the initial load differences, while TS+Const shows more reshuffling of which experts receive the most assignments.
These checkpoint statistics compare different LR schedules and similar but nonidentical starting routers, so they describe the redistribution pattern rather than isolating the effect of data shape alone.

\begin{table}[!htbp]
\centering
\small
\setlength{\tabcolsep}{4pt}
\renewcommand{\arraystretch}{1.15}
\begin{tabular}{@{}lcccc@{}}
\toprule
Schedule & \shortstack{Final imbalance\\FineMath / DCLM} & \shortstack{Net routing change\\FineMath TV} & \shortstack{Regression\\residual std.} & \shortstack{Rank retention\\Spearman} \\
\midrule
TS+Const & 1.015 / 1.043 & 0.156 & 0.119 & 0.30 \\
ZL+MDR & 1.008 / 1.039 & 0.140 & 0.042 & 0.74 \\
\bottomrule
\end{tabular}
\caption{MoE routing comparison over 10B continuation tokens with a 25\% MT budget and final MT ratio 100\%. TS+Const compares steps 1,750 and 2,384, and ZL+MDR compares steps 1,150 and 2,384, from before each schedule's MT entry time to training completion. TV reports the net change in FineMath assignment shares. Regression residuals use shares relative to uniform routing, and Spearman correlations are averaged across layers. Figure~\ref{fig:exp-router-smoothing} provides the separate control with identical LR trajectories.}
\label{tab:app-router-comparison}
\end{table}

\clearpage
\section{Additional Theoretical Setup}
\label{app:theory-setup}

\paragraph{Original SGD training algorithm with data mixing.}
Let $K$ denote the total number of training steps, and at each training step $k \in \left[0, K-1\right]$,
we sample MT data with probability $\gamma(k)$ and sample PT data with probability $1-\gamma(k)$.
Thus, for the $k^{\text{th}}$ data point $\vx_k$, its covariance matrix is $\E\vx_k\vx_k^{\top}=\gamma(k)\mH+\left(1-\gamma(k)\right)\mG$.
At the $k^{\text{th}}$ step, we therefore have $\cL_k(\vtheta)=\frac{1}{2}\vtheta^{\top}\left(\gamma(k)\mH+\left(1-\gamma(k)\right)\mG\right)\vtheta$.
We train our model using online stochastic gradient descent (SGD) with learning rate schedule $E=\{\eta_0, \cdots, \eta_{K-1}\}$.
The update rule can be written as
\begin{equation}
    \vtheta_{k+1}=\vtheta_{k}-\eta_k\left(\nabla\cL_k(\vtheta_k)+\xi_k\right). \label{equ:sgd-dynamics}
\end{equation}
Here $\xi_k \sim \cN(0, \mSigma)$ is additional noise independent of the signal.
We let $\mSigma=\diag(\mSigma_0, \mSigma_1, \mSigma_2)$, where $\mSigma_l=\diag(\nu_{1}^{(l)},\ldots,\nu_{d_l}^{(l)})\in\R^{d_l\times d_l}$ for $l=0,1,2$. We assume that $\nu_{i}^{(l)} > 0$ for all $0<i\leq d_l$ and $l=0,1,2$.
Our goal is to minimize the expected validation loss $\E[\cL(\vtheta_K)]$, where the expectation is taken over the randomness in the gradient noise and the data sampling.

\paragraph{SDE approximation.}
Motivated by \citet{li2017stochastic,li2019stochastic,li2020intrinsic,li2021validity,pesme2021implicit,li2022zero,li2025functional}, we analyze the continuous time limit of \Cref{equ:sgd-dynamics}, which makes the analysis more tractable. We fix a discretization step size $h>0$, set
$t_k=kh$ for $k=0,\ldots,K$, and let $T=Kh$. We also define
$\eta(t_k)=\eta_k/h$ and $\gamma(t_k)=\gamma(k)$ as the continuous-time learning rate schedule and data schedule, respectively. The corresponding continuous-time loss function is defined as
\begin{equation*}
    \cL_t(\vtheta)
    :=\frac{1}{2}\vtheta^{\top}
    \left(\gamma(t)\mH+\left(1-\gamma(t)\right)\mG\right)\vtheta.
\end{equation*}
Using $\eta_k=h\eta(t_k)$, \Cref{equ:sgd-dynamics} becomes
\begin{equation*}
    \vtheta_{k+1}-\vtheta_k
    =-\eta(t_k)\nabla\cL_{t_k}(\vtheta_k)h
      -\eta(t_k)h\xi_k.
\end{equation*}
Following \citet{li2019stochastic,orvieto2019continuous}, we approximate the iteration above with the It\^o SDE given in \Cref{equ:sde-dynamics}.
The minimization of the expected validation loss $\E[\cL(\vtheta_K)]$ is then approximated by the minimization of $\E[\cL(\vtheta_T)]$, where the expectation is taken over the randomness in the Brownian motion and the data sampling.

\section{Formal Statements of Theoretical Results} \label{app:theory-statement}
In this section, we provide the formal statement of \Cref{thm:informal} and its related assumptions. We restate \Cref{thm:informal} as \Cref{thm:opt-ds-const-lr,thm:opt-ds-linear-decay-lr,thm:opt-ds-multi-power-law-spectra}, corresponding to parts \ref{thm:informal-const-lr}, \ref{thm:informal-lr-decay} and \ref{thm:informal-mpl}, respectively. The proofs of \Cref{thm:opt-ds-const-lr,thm:opt-ds-linear-decay-lr,thm:opt-ds-multi-power-law-spectra} are given in \Cref{app:proof-opt-ds-const-lr,app:proof-opt-ds-linear-decay-lr,app:proof-opt-ds-mpl} respectively.

We assume that $\sum_{i=1}^{d_1}\theta_i^{(1)}(0)^2>0$ in subsequent results. We first show the optimal data schedule under a constant learning rate.

\begin{theorem}[Optimal Data Schedule under a Constant Learning Rate] \label{thm:opt-ds-const-lr}
    Suppose that $\eta(t)\equiv\eta_{\max}>0$.
    Under \Cref{asp:constrained-cpt-data}, the optimal data schedule
    is given by the following schedule:
    \begin{align*}
        \gamma^*(t) = \begin{cases}
            0, & t \in [0, (1-\alpha)T), \\
            1, & t \in [(1-\alpha)T, T],
        \end{cases}
    \end{align*}
    which we refer to as the 0-1 data schedule.
\end{theorem}

We next replace the constant learning rate with a learning rate schedule with linear decay, while
keeping all other settings fixed. The following theorem gives
sufficient conditions on the decay strength of the learning rate and the MT budget, under which
the 0-1 data schedule in \Cref{thm:opt-ds-const-lr} is no longer optimal.

\begin{theorem}[Optimal Schedule under Linear Learning Rate Decay] \label{thm:opt-ds-linear-decay-lr}
    Suppose that the learning rate decays linearly according to
    \begin{equation*}
        \eta(t)=\eta_{\max}-\frac{c}{T}t,
        \qquad (1-4^{-\frac{1}{3}})\eta_{\max}<c<\eta_{\max}.
    \end{equation*}
    Then, under \Cref{asp:constrained-cpt-data}, when $0<\alpha < 1-\frac{(1-4^{-\frac{1}{3}})\eta_{\max}}{c}$, $\gamma^*(t)$ in \Cref{thm:opt-ds-const-lr}
     is not the optimal schedule.
\end{theorem}

\Cref{thm:opt-ds-linear-decay-lr} does not characterize the exact form of the optimal data schedule under a linearly decaying learning rate schedule. However, it proves that changing only the learning rate schedule from constant to linear decay can change the corresponding optimal data schedule with everything else fixed.

\Cref{thm:opt-ds-const-lr} and \Cref{thm:opt-ds-linear-decay-lr} show that different learning rate schedules induce different optimal data schedules. Nevertheless, the optimal data schedule under a linear decay learning rate is not explicitly solvable. Motivated by the multi-power law in \citet{luo2025a}, we introduce a power-law spectral model that allows us to derive the optimal data schedule under a linear decay learning rate in closed form.

\begin{assumption}[Power-Law Spectral Model]
\label{asp:multi-power-law-spectra}
    For each block $l\in\{0,1,2\}$, assume that the pairs
    $\{(\lambda_i^{(l)},\nu_i^{(l)})\}_{i=1}^{d_l}$ are independent and
    identically distributed across $i$, with
    \begin{align*}
        \lambda_i^{(l)}
        &\sim p_l(\lambda)
        =D_l\lambda^{\alpha_l}\exp(-K_l\lambda),
        \qquad \lambda\in[0,\infty),\qquad i=1,\ldots,d_l,\\
        \nu_i^{(l)}\mid\lambda_i^{(l)}
        &\sim q_l(\nu\mid\lambda_i^{(l)}),
        \qquad \nu\in[0,\infty),\qquad i=1,\ldots,d_l,
    \end{align*}
    where $-1<\alpha_l<0$, $K_l>0$, and
    \begin{equation*}
        D_l=\frac{K_l^{\alpha_l+1}}{\Gamma(\alpha_l+1)}.
    \end{equation*}
    Moreover, for some $\rho_l>0$,
    \begin{equation*}
        \E_{q_l(\cdot\mid\lambda)}[\nu\mid\lambda]
        =F_l\lambda^{\rho_l}\exp(-G_l\lambda),
    \end{equation*}
    where $F_l,G_l>0$ are constants independent of the learning rate
    schedule.
\end{assumption}

\paragraph{The spectral assumption describes a collection of directions that are
learned at different speeds.} A smaller
eigenvalue $\lambda$ means that the loss changes more weakly along the
corresponding direction, so the error in that direction is corrected more
slowly. The density $p_l$ describes how common different eigenvalues
are. Since $-1<\alpha_l<0$, the factor $\lambda^{\alpha_l}$ places
more mass near zero, representing many slowly learned directions,
while $e^{-K_l\lambda}$ makes very large eigenvalues increasingly rare.
Together, these factors give a Gamma density with shape $\alpha_l+1$
and rate $K_l$.
\citet{luo2025a} use an eigenvalue density with a power-law expression to explain how Multi-Power Law arises. The two models differ in how they describe large eigenvalues:
\citet{luo2025a} impose a finite upper bound on the eigenvalues, and our
model uses a density on $[0,\infty)$ with an exponentially decaying tail.

\paragraph{The noise assumption describes how the strength of the additional
gradient noise varies with eigenvalues.} Here, $\nu_i^{(l)}$ is the
variance of noise in direction $i$ of block $l$:
a larger value means stronger random perturbations to the gradient
along that direction. Thus,
$\mathbb{E}[\nu\mid\lambda]
=F_l\lambda^{\rho_l}e^{-G_l\lambda}$
specifies the average noise variance among directions with eigenvalue
$\lambda$. The parameter $F_l$ sets the overall noise scale.
In directions with small eigenvalues, where the exponential factor $e^{-G_l\lambda}$ is close to one,
$\rho_l>0$ describes weaker noise in those directions.
The factor $e^{-G_l\lambda}$ makes the average noise variance decay
in directions with very large eigenvalues.
This functional form follows \citet{luo2025a}, who use such a
relationship between eigenvalues and noise variance to derive the
additional loss reduction caused by learning rate decay in a
quadratic model.

Under \Cref{asp:multi-power-law-spectra}, we consider minimizing the expected validation loss, where the expectation is taken with respect to the training process and the distribution of the eigenvalues. Under this setup, we show the shape of the optimal data schedule under a linearly decaying learning rate in closed form.

\begin{theorem}[Optimal Data Schedule under the Power-Law Spectral Model]
\label{thm:opt-ds-multi-power-law-spectra}
    Suppose that $0<\rho_1<1$ and that
    $\eta(t)=\eta_{\max}-ct/T$, where $\eta_{\max}$ and $c$ are fixed constants satisfying $\left(1-4^{-1/3}\right)\eta_{\max}<c<\eta_{\max}$.
    Define $x_\star:=c^{-1}\left(4^{-1/3}\eta_{\max}-(\eta_{\max}-c)\right)\in(0,1)$,
    and suppose that the data fraction $0<\alpha<x_\star$ is fixed.
    Then, when $T$ is sufficiently large, the optimal data schedule under
    \Cref{asp:constrained-cpt-data}
    and \Cref{asp:multi-power-law-spectra} can be given by
    \begin{align*}
        \gamma^*(t) = \begin{cases}
            0, & t \in [0, t_1), \\
            \displaystyle
            \left(G_{\mathrm{eff}}+\int_{t_2}^{T}\eta(\tau)\d\tau\right)
            \frac{4c}{T(b+1)}
            \left(\eta(t_2)\right)^{-\frac{4}{b+1}}
            \left(\eta(t)\right)^{-\frac{2(b-1)}{b+1}},
                & t \in [t_1, t_2), \\
            1, & t \in [t_2, T],
        \end{cases}
    \end{align*}
    where $b:=\alpha_1+\rho_1+2$, $G_{\mathrm{eff}}:=\frac{K_1+G_1}{2}$, and $t_1, t_2$ are two time points that satisfy
    $0<t_1<t_2<T$, $t_1=\Theta(T)$ and $t_2=\Theta(T)$.
\end{theorem}

\section{\texorpdfstring{Proof of \Cref{thm:opt-ds-const-lr}}{Proof of Theorem \ref*{thm:opt-ds-const-lr}}}
\label{app:proof-opt-ds-const-lr}

In this section, we derive the evaluation loss from the SDE in
\Cref{equ:sde-dynamics} and then prove the optimality of the 0--1 data schedule
in \Cref{thm:opt-ds-const-lr}.

\subsection{Derivation of the Terminal Loss}
\label{app:terminal-loss-derivation}

Before we begin our proof, we first define some additional notation. We denote
\begin{equation}
    a_i^{(l)}(t):=
    \begin{cases}
        \lambda_i^{(0)}, & l=0, \\
        \gamma(t)\lambda_i^{(1)}, & l=1, \\
        \bigl(1-\gamma(t)\bigr)\lambda_i^{(2)}, & l=2,
    \end{cases}
    \qquad i=1,\ldots,d_l.
    \label{def:a_i^{(l)}}
\end{equation}
For $0\leq s\leq t\leq T$, we define
$R_i^{(l)}(s,t):=\exp\!\left(-\int_s^t\eta(u)a_i^{(l)}(u)\d u\right)$, and
$E_i^{(l)}(s,t):=R_i^{(l)}(s,t)^2=\exp\!\left(-2\int_s^t\eta(u)a_i^{(l)}(u)\d u\right)$.
Because $\mH$, $\mG$, and $\mSigma$ are diagonal, the $i$th coordinate of
\Cref{equ:sde-dynamics} satisfies
\begin{equation}
    \d\theta_i^{(l)}(t)
    =-\eta(t)a_i^{(l)}(t)\theta_i^{(l)}(t)\d t
      +\eta(t)\sqrt{h\nu_i^{(l)}}\d B_i^{(l)}(t).
    \label{eq:app-coordinate-sde}
\end{equation}
Solving \Cref{eq:app-coordinate-sde} gives
\begin{equation*}
    \theta_i^{(l)}(T)
    =R_i^{(l)}(0,T)\theta_i^{(l)}(0)
     +\sqrt{h\nu_i^{(l)}}\int_0^T
       \eta(s)R_i^{(l)}(s,T)\d B_i^{(l)}(s).
\end{equation*}
The stochastic integral has zero mean. Since $R_i^{(l)}(0,T)\theta_i^{(l)}(0)$ is
deterministic, its cross term with the stochastic integral vanishes after
taking expectations. The It\^o isometry then gives
\begin{equation}
    \E\!\left[\theta_i^{(l)}(T)^2\right]
    =\theta_i^{(l)}(0)^2E_i^{(l)}(0,T)
     +h\nu_i^{(l)}\int_0^T\eta(s)^2E_i^{(l)}(s,T)\d s.
    \label{eq:app-coordinate-second-moment}
\end{equation}

We denote $\cL_l(\vtheta)=\frac12\sum_{i=1}^{d_l}\lambda_i^{(l)}\bigl(\theta_i^{(l)}\bigr)^2$. Since the model is evaluated on the MT loss
$\cL(\vtheta)=\sum_{l=0}^{1}\cL_l(\vtheta)=\frac12\sum_{l=0}^1\sum_{i=1}^{d_l}\lambda_i^{(l)}\bigl(\theta_i^{(l)}\bigr)^2$, substituting
\Cref{eq:app-coordinate-second-moment} yields
\begin{align}
    \cJ(\gamma,\eta)
    &:=\E\!\left[\cL(\vtheta_T)\right]
    =\frac12\sum_{l=0}^1\sum_{i=1}^{d_l}\lambda_i^{(l)}
      \left[
        \theta_i^{(l)}(0)^2E_i^{(l)}(0,T)
        +h\nu_i^{(l)}\int_0^T\eta(s)^2E_i^{(l)}(s,T)\d s
      \right] \notag \\
    &=\frac12\sum_{l=0}^1\sum_{i=1}^{d_l}\lambda_i^{(l)}
        \theta_i^{(l)}(0)^2E_i^{(l)}(0,T)
        +\frac12\sum_{l=0}^1\sum_{i=1}^{d_l}\lambda_i^{(l)}h\nu_i^{(l)}\int_0^T\eta(s)^2E_i^{(l)}(s,T)\d s.
    \label{eq:app-terminal-loss}
\end{align}
We call the first term the bias term, which arises because the initialization of the parameters is different from the global optimum. We call
the second term the variance term, which is induced by the accumulation of gradient noise.
Note that for all $i,l$, and $s<t$, $E_i^{(l)}(s,t)$ is convex up to a set of measure zero. Thus $\cJ(\gamma,\eta)$ is a convex functional of the data schedule $\gamma(t)$ up to a set of measure zero. Furthermore, the feasible set $\{\gamma(t):[0,T]\to[0,1], \frac{1}{T}\int_{0}^{T}\gamma(t)\,\mathrm{d}t\leq\alpha\}$ is convex. Thus the optimization problem in \Cref{aim:data-schedule-minimization} is a convex optimization problem. The feasible set also satisfies the Slater condition, so the KKT conditions are necessary and sufficient for optimality.

Next we prove that if the optimal data schedule exists, then the optimal data schedule is unique up to a set of measure zero.
We proceed by contradiction and assume that there exist two different optimal data schedules $\gamma_1$ and $\gamma_2$, and we define $M=\cJ(\gamma_1,\eta)$. By the convexity of $\cJ(\gamma,\eta)$, we have
\[\cJ(x\gamma_1+(1-x)\gamma_2,\eta)\leq x\cJ(\gamma_1,\eta)+(1-x)\cJ(\gamma_2,\eta)=M.\]
Equality must hold; otherwise, $x\gamma_1+(1-x)\gamma_2$ would achieve a smaller objective value, contradicting the optimality of $\gamma_1$ and $\gamma_2$. Notice that the summands for $l=0$ are independent of the data schedule. Thus by the strict convexity of the exponential function, the equality holds if and only if for all $i$ and for all $0\leq s \leq T$,
\[-2\int_s^T\eta(u)\gamma_1(u)\lambda_i^{(1)}\d u=-2\int_s^T\eta(u)\gamma_2(u)\lambda_i^{(1)}\d u.\]
Taking the derivative, we have $\gamma_1(u)=\gamma_2(u), a.e.$, which is a contradiction.

\subsection{\texorpdfstring{KKT Condition for \Cref{aim:data-schedule-minimization}}{KKT Condition for the Data Schedule Optimization Problem}}
\label{app:proof-constant-lr-schedule}
We denote the bracketed term in
\Cref{eq:app-terminal-loss} by
\begin{equation*}
    Q_i^{(l)}(t):=\theta_i^{(l)}(0)^2E_i^{(l)}(0,T)
      +h\nu_i^{(l)}\int_0^t\eta(s)^2E_i^{(l)}(s,T)\d s.
\end{equation*}
Note that $\gamma(t)$ does not affect $Q_i^{(0)}(t)$, so we only need to minimize $\frac{1}{2}\sum_{i=1}^{d_1}\lambda_i^{(1)}Q_i^{(1)}(T)$. From now on, we omit the superscript $(1)$ for simplicity.

We rewrite the MT data constraint as
\begin{equation*}
    \int_0^T\gamma(t)\d t-\alpha T\leq0,
\end{equation*}
and define the Lagrangian functional
\begin{equation*}
    \mathcal{F}(\gamma,\kappa;\eta)
    :=\cJ(\gamma,\eta)
      +\kappa\left(\int_0^T\gamma(t)\d t-\alpha T\right).
\end{equation*}

Now we derive the first variation of $\cF(\gamma,\kappa;\eta)$. For a bounded perturbation $v(t)$, let $\gamma_\varepsilon=\gamma+\varepsilon v$. We
use a subscript $\varepsilon$ to denote quantities induced by the perturbed
schedule and define
\begin{align*}
    a_{i,\varepsilon}(u)
    &:=\lambda_i\gamma_\varepsilon(u)
      =a_i(u)+\varepsilon\lambda_i v(u), \\
    E_{i,\varepsilon}(s,T)
    &:=\exp\!\left(
        -2\int_s^T\eta(t)a_{i,\varepsilon}(t)\d t
      \right) \\
    &=E_i(s,T)
      \exp\!\left(
        -2\varepsilon\lambda_i\int_s^T\eta(t)v(t)\d t
      \right).
\end{align*}
In particular, $a_{i,0}=a_i$ and $E_{i,0}=E_i$, so the notation without an
$\varepsilon$ subscript always refers to the original schedule $\gamma$.
Differentiating the second identity at $\varepsilon=0$ gives
\begin{equation}
    \left.\frac{\d}{\d\varepsilon}
    E_{i,\varepsilon}(s,T)\right|_{\varepsilon=0}
    =-2\lambda_i E_i(s,T)
      \int_s^T\eta(t)v(t)\d t.
    \label{eq:app-propagator-variation}
\end{equation}
To differentiate the terminal loss, we substitute
\Cref{eq:app-propagator-variation} into \Cref{eq:app-terminal-loss} and obtain
\begin{align*}
    D\cJ(\gamma)[v]
    ={}&-\sum_{i=1}^{d_1}\lambda_i^2\theta_i(0)^2E_i(0,T)
       \int_0^T\eta(t)v(t)\d t \\
      &-h\sum_{i=1}^{d_1}\lambda_i^2\nu_i
       \int_0^T\eta(s)^2E_i(s,T)
       \left[\int_s^T\eta(t)v(t)\d t\right]\d s.
\end{align*}
The second term integrates over the triangular region
$\{(s,t):0\leq s\leq t\leq T\}$. By Fubini's theorem,
\begin{align*}
    &\int_0^T\eta(s)^2E_i(s,T)
       \left[\int_s^T\eta(t)v(t)\d t\right]\d s \\
    &\qquad=
       \int_0^T\eta(t)v(t)
       \left[\int_0^t\eta(s)^2E_i(s,T)\d s\right]\d t.
\end{align*}
Both parts of the objective variation can therefore be collected into a single
integral. Including the variation of the constraint gives
\begin{equation*}
    \begin{aligned}
    D_\gamma\mathcal{F}(\gamma,\kappa;\eta)[v]
    &=D\cJ(\gamma)[v]+\kappa\int_0^T v(t)\d t \\
    &=\int_0^T\bigl(\kappa-M_\gamma(t)\bigr)v(t)\d t.
    \end{aligned}
\end{equation*}
where the marginal value of MT data at time $t$ is
\begin{equation}
    M_\gamma(t)
    :=\eta(t)\sum_{i=1}^{d_1}\lambda_i^2
      \left[
        \theta_i(0)^2E_i(0,T)
        +h\nu_i\int_0^t\eta(s)^2E_i(s,T)\d s
      \right].
    \label{eq:app-switching-function}
\end{equation}
Consequently, the pointwise KKT conditions for the box constraint
$0\leq\gamma(t)\leq1$ are
\begin{equation}
    \begin{cases}
      M_{\gamma^\star}(t)\leq\kappa,
        & \gamma^\star(t)=0, \\
      M_{\gamma^\star}(t)=\kappa,
        & 0<\gamma^\star(t)<1, \\
      M_{\gamma^\star}(t)\geq\kappa,
        & \gamma^\star(t)=1,
    \end{cases} \label{original-kkt}
\end{equation}
together with primal feasibility, $\kappa\geq0$, and complementary slackness
\begin{equation*}
    \kappa\left(\int_0^T\gamma^\star(t)\d t-\alpha T\right)=0.
\end{equation*}

\subsection{\texorpdfstring{Proof of \Cref{thm:opt-ds-const-lr}}{Proof of Theorem \ref*{thm:opt-ds-const-lr}}}
\begin{proof}
Recall that
\begin{equation*}
    Q_i(t):=\theta_i(0)^2E_i(0,T)
      +h\nu_i\int_0^t\eta(s)^2E_i(s,T)\d s.
\end{equation*}
The fundamental theorem of calculus gives
$Q_i'(t)=h\nu_i\eta(t)^2E_i(t,T)$. Hence, for a differentiable learning-rate
schedule,
\begin{equation*}
    M_\gamma'(t)
    =\eta'(t)\sum_{i=1}^{d_1}\lambda_i^2Q_i(t)
     +h\eta(t)^3\sum_{i=1}^{d_1}
       \lambda_i^2\nu_iE_i(t,T).
\end{equation*}
Under a constant learning rate $\eta_{\max}$, we have $\eta'(t)=0$, and therefore
\begin{equation*}
    M_\gamma'(t)
    =h\eta_{\max}^3\sum_{i=1}^{d_1}
      \lambda_i^2\nu_i E_i(t,T)>0.
\end{equation*}
The strict inequality follows from $h>0$, $\eta_{\max}>0$, and $\lambda_i,\nu_i>0$ in our setup.

We now give a global comparison, which also proves uniqueness.  For a feasible
schedule $\gamma$, define its remaining MT mass after time $s$ by
\begin{equation*}
    C_\gamma(s):=\int_s^T\gamma(u)\d u.
\end{equation*}
Because $M_\gamma(t)>0$ for almost every $t>0$, adding MT mass on any
positive-measure set on which $\gamma<1$ strictly decreases the objective.
Thus, except for the immediate boundary cases $\alpha=0$ and $\alpha=1$, every
optimal schedule exhausts the budget:
\begin{equation}
    \int_0^T\gamma(t)\d t=\alpha T.
    \label{eq:app-constant-budget-active}
\end{equation}
For any schedule satisfying \eqref{eq:app-constant-budget-active}, the box
constraint gives
\begin{equation*}
    C_\gamma(s)\leq\min\{\alpha T,T-s\},
    \qquad 0\leq s\leq T.
\end{equation*}
The schedule $\bar\gamma$ in \Cref{thm:opt-ds-const-lr} attains this upper bound
for every $s$.  Moreover, under the constant learning rate,
\begin{equation*}
    E_i^\gamma(s,T)
    =\exp\!\left[-2\eta_{\max}\lambda_i C_\gamma(s)\right].
\end{equation*}
Consequently,
$E_i^{\bar\gamma}(s,T)\leq E_i^\gamma(s,T)$ for every $i$ and $s$.  The
initialization term in \Cref{eq:app-terminal-loss} is the same for all
schedules satisfying \eqref{eq:app-constant-budget-active}, while the variance
term of $\bar\gamma$ is no larger coordinate-wise.  It follows that
\begin{equation*}
    \cJ(\bar\gamma,\eta_{\max})\leq\cJ(\gamma,\eta_{\max}).
\end{equation*}
If equality holds, the positivity of the variance weights forces
$C_\gamma(s)=C_{\bar\gamma}(s)$ for almost every $s$.  Taking weak derivatives
then gives $\gamma=\bar\gamma$ almost everywhere.  The cases $\alpha=0$ and
$\alpha=1$ follow directly from the box constraint and the strict decrease of
the objective when MT mass is added.
\end{proof}

\section{\texorpdfstring{Proof of \Cref{thm:opt-ds-linear-decay-lr}}{Proof of Theorem \ref*{thm:opt-ds-linear-decay-lr}}}
\label{app:proof-opt-ds-linear-decay-lr}

\begin{proof}
We prove the theorem by contradiction. Let
\begin{equation*}
    t_\alpha:=(1-\alpha)T,
    \qquad
    \bar\gamma(t):=\mathbf{1}\{t\geq t_\alpha\}
\end{equation*}
denote the transition time and the 0-1 schedule from
\Cref{thm:opt-ds-const-lr}, respectively. Suppose that $\bar\gamma$ is
optimal under the linear decay learning rate schedule. The KKT conditions derived
above imply
\begin{equation*}
    M_{\bar\gamma}(t)\leq\kappa
      \quad (t<t_\alpha),
    \qquad
    M_{\bar\gamma}(t)\geq\kappa
      \quad (t>t_\alpha).
\end{equation*}
Because $M_{\bar\gamma}$ is continuously differentiable, it follows that
$M_{\bar\gamma}(t_\alpha)=\kappa$ and
\begin{equation}
    M_{\bar\gamma}'(t_\alpha)\geq0.
    \label{eq:app-linear-decay-kkt-slope}
\end{equation}

We now calculate this derivative explicitly. Define the integrated learning
rate over an interval by
\begin{equation*}
    \Phi(a,b):=\int_a^b\eta(u)\d u,
    \qquad 0\leq a\leq b\leq T.
\end{equation*}
The two intervals before and after the transition occur repeatedly below, so
we introduce the mnemonic abbreviations
\begin{equation*}
    \Phi_\alpha^{-}(s):=\Phi(s,t_\alpha),
    \qquad
    \Phi_\alpha^{+}:=\Phi(t_\alpha,T),
\end{equation*}
where the superscripts $-$ and $+$ indicate accumulation before and after
$t_\alpha$, respectively. For $\eta(t)=\eta_{\max}-ct/T$, these quantities are
\begin{align*}
    \eta_\alpha
    &:=\eta(t_\alpha)=\eta_{\max}-c(1-\alpha), \\
    \Phi_\alpha^{+}
    &=\eta_{\max}(T-t_\alpha)
      -\frac{c}{2T}\bigl(T^2-t_\alpha^2\bigr) \\
    &=\alpha T\left[\eta_{\max}-c\left(1-\frac{\alpha}{2}\right)\right], \\
    \Phi_\alpha^{-}(s)
    &=\eta_{\max}(t_\alpha-s)
      -\frac{c}{2T}\bigl(t_\alpha^2-s^2\bigr),
      \qquad 0\leq s\leq t_\alpha.
\end{align*}
Under $\bar\gamma$, coordinate $i$ has zero curvature before
$t_\alpha$ and curvature $\lambda_i$ afterwards. Let $\bar E_i$ denote the squared
propagator induced by this candidate schedule. Consequently,
\begin{align}
    \bar E_i(s,T)
    &=\exp\!\left(-2\lambda_i\Phi_\alpha^{+}\right),
      &&0\leq s\leq t_\alpha.
      \label{eq:app-linear-decay-propagator}
\end{align}
In particular, the quantity in \Cref{eq:app-switching-function} at the
transition point is
\begin{align*}
    \bar Q_i(t_\alpha)
    ={}&\theta_i(0)^2
       \exp\!\left(-2\lambda_i\Phi_\alpha^{+}\right) \\
      &+h\nu_i\exp\!\left(-2\lambda_i\Phi_\alpha^{+}\right)
       \int_0^{t_\alpha}\eta(s)^2\d s.
\end{align*}
Since $\eta'(t)=-c/T$, substituting this identity and
\Cref{eq:app-linear-decay-propagator} into the general expression
for $M_\gamma'(t)$ gives
\begin{align}
    M_{\bar\gamma}'(t_\alpha)
    ={}&-\frac{c}{T}\sum_{i=1}^{d_1}
       \lambda_i^2\theta_i(0)^2
       \exp\!\left(-2\lambda_i\Phi_\alpha^{+}\right) \notag\\
      &+h\sum_{i=1}^{d_1}\lambda_i^2\nu_i
       \exp\!\left(-2\lambda_i\Phi_\alpha^{+}\right)
       \left[
         \eta_\alpha^3
         -\frac{c}{T}\int_0^{t_\alpha}\eta(s)^2\d s
       \right] \notag\\
    ={}&-\frac{c}{T}\sum_{i=1}^{d_1}
       \lambda_i^2\theta_i(0)^2
       \exp\!\left(-2\lambda_i\Phi_\alpha^{+}\right) \notag\\
      &+h\sum_{i=1}^{d_1}\lambda_i^2\nu_i
       \exp\!\left(-2\lambda_i\Phi_\alpha^{+}\right)
       \left[\frac{\eta_\alpha}{(1-\alpha)T}\int_0^{t_\alpha}\eta_\alpha^2\d s
         -\frac{c}{T}\int_0^{t_\alpha}\eta(s)^2\d s
       \right]. \label{eq:app-linear-decay-switch-derivative}
\end{align}
Because $\eta$ is decreasing,
$\eta(s)\geq\eta_\alpha$ for $0\leq s\leq t_\alpha$.  Therefore, the bracket
in \Cref{eq:app-linear-decay-switch-derivative} satisfies
\begin{align*}
    \eta_\alpha^3-\frac{c}{T}\int_0^{t_\alpha}\eta(s)^2\d s
    &\leq
      \eta_\alpha^3-\frac{1}{3}\left(\eta_{\max}^3-\eta_\alpha^3\right)\\
    &=\frac{4}{3}\left(\eta_{\max}-c(1-\alpha)\right)^3-\frac{1}{3}\eta_{\max}^3<0,
\end{align*}
where the last inequality uses
$\alpha<1-\frac{(1-4^{-\frac{1}{3}})\eta_{\max}}{c}$.  The first term of
\Cref{eq:app-linear-decay-switch-derivative} is nonpositive, and the
positivity of $h$, $\lambda_i$, and $\nu_i$ makes the second contribution strictly negative.  Hence
$M_{\bar\gamma}'(t_\alpha)<0$, contradicting
\Cref{eq:app-linear-decay-kkt-slope}.
\end{proof}

\section{\texorpdfstring{Proof of \Cref{thm:opt-ds-multi-power-law-spectra}}{Proof of Theorem \ref*{thm:opt-ds-multi-power-law-spectra}}}
\label{app:proof-opt-ds-mpl}
\subsection{Derivations of the Minimization Objective}
\label{app:multi-power-law-final-loss}

We begin from the final loss expression in
\Cref{eq:app-terminal-loss}. Recall that only
the shared block $l=0$ and the MT-specific block $l=1$ contribute to the final loss. Since the
$l=0$ term is independent of $\gamma$,
we only need to minimize
\begin{align*}
    \cL_1(\vtheta)&=
    \underbrace{\frac12\sum_{i=1}^{d_1}\lambda_i^{(1)}
        \theta_i^{(1)}(0)^2
        \exp\!\left(
            -2\lambda_i^{(1)}
            \int_0^T\eta(t)\gamma(t)\d t
        \right)}_{\cL_1^{\bias}(\vtheta)}\\
        &+\underbrace{\frac12\sum_{i=1}^{d_1}\lambda_i^{(1)}h\nu_i^{(1)}\int_0^T\eta(s)^2
        \exp\!\left(
            -2\lambda_i^{(1)}
            \int_s^T\eta(t)\gamma(t)\d t
        \right)\d s}_{\cL_1^{\var}(\vtheta)}.
\end{align*}

We now average $\cL_1(\vtheta)$ over the i.i.d.\ spectral pairs
$\{(\lambda_i^{(1)},\nu_i^{(1)})\}_{i=1}^{d_1}$ specified in
\Cref{asp:multi-power-law-spectra}. In subsequent analysis, we will repeatedly use the
Gamma-integral identity
\begin{equation*}
    \int_0^\infty \lambda^{r-1}e^{-x\lambda}\d\lambda
    =\frac{\Gamma(r)}{x^r},
    \qquad r,x>0.
\end{equation*}
Because the $\lambda_i^{(1)}$ are identically distributed, the same spectral
expectation multiplies every initialization coefficient. Thus
\begin{align}
    \E\cL_1^{\bias}(\vtheta)&=\frac{D_1}{2}\left(\sum_{i=1}^{d_1}\theta_i^{(1)}(0)^2\right)\int_0^\infty
      \lambda^{\alpha_1+1}
      \exp\!\left(
        -\left(K_1+2\int_0^T\eta(t)\gamma(t)\d t\right)\lambda
      \right)\d\lambda \notag\\
    &=\frac{D_1}{2}\left(\sum_{i=1}^{d_1}\theta_i^{(1)}(0)^2\right)
      \frac{\Gamma(\alpha_1+2)}
      {\left(
        K_1+2\int_0^T\eta(t)\gamma(t)\d t
      \right)^{\alpha_1+2}}.
    \label{eq:app-mpl-initialization-term}
\end{align}
We next turn to the variance term. We define the remaining accumulated learning rate after time $s$ by
\begin{equation*}
    R_\gamma(s):=\int_s^T\eta(t)\gamma(t)\d t.
\end{equation*}
For a representative spectral pair $(\lambda,\nu)$, we first condition on
$\lambda$. Then
\Cref{asp:multi-power-law-spectra} gives
\begin{align*}
    &\E_{\nu\mid\lambda}\!\left[
        \lambda\nu e^{-2\lambda R_\gamma(s)}
      \,\middle|\,\lambda\right]=
      \lambda e^{-2\lambda R_\gamma(s)}
      \E_{\nu\mid\lambda}[\nu\mid\lambda]=
      F_1\lambda^{\rho_1+1}
      \exp\!\left[-\bigl(2R_\gamma(s)+G_1\bigr)\lambda\right].
\end{align*}
We then average over $\lambda$.  Using
$p_1(\lambda)=D_1\lambda^{\alpha_1}e^{-K_1\lambda}$, the law of total
expectation yields
\begin{align*}
    &\E_{\lambda,\nu}\!\left[
        \lambda\nu e^{-2\lambda R_\gamma(s)}
      \right] =
      \E_\lambda\!\left[
        \E_{\nu\mid\lambda}\!\left[
          \lambda\nu e^{-2\lambda R_\gamma(s)}
        \,\middle|\,\lambda\right]\right] \notag \\
    &=D_1F_1\int_0^\infty
        \lambda^{\alpha_1+\rho_1+1}
        \exp\!\left[-\bigl(2R_\gamma(s)+G_1+K_1\bigr)\lambda\right]
      \d\lambda.
\end{align*}
Tonelli's theorem permits exchanging the
expectation with respect to $\{(\lambda_i^{(1)},\nu_i^{(1)})\}_{i=1}^{d_1}$ and the time integral, so
\begin{align}
    \E\!\left[\cL_1^{\var}(\vtheta)\right]
    &=
      \frac{h d_1}{2}\int_0^T\eta(s)^2
      \E_{\lambda,\nu}\!\left[
        \lambda\nu e^{-2\lambda R_\gamma(s)}
      \right]\d s \notag\\
    &=
      \frac{h d_1D_1F_1
            \Gamma(\alpha_1+\rho_1+2)}
           {2^{\alpha_1+\rho_1+3}}
      \int_0^T
        \frac{\eta(s)^2}{
          \left(
            R_\gamma(s)+(K_1+G_1)/2
          \right)^{\alpha_1+\rho_1+2}}
      \d s.
    \label{eq:app-mpl-noise-term}
\end{align}
Here we require $\alpha_1+\rho_1+2>0$ so that the spectral integral is finite
at the origin. We define $C_0:=\E\cL_0(\vtheta)$,
\begin{equation*}
    C:=D_1\left(\sum_{i=1}^{d_1}\theta_i^{(1)}(0)^2\right)
       \Gamma(\alpha_1+2)2^{-\alpha_1-3},
    \qquad
    C':=h d_1D_1F_1\Gamma(\alpha_1+\rho_1+2)
       2^{-\alpha_1-\rho_1-3}.
\end{equation*}
Combining
\Cref{eq:app-mpl-initialization-term,eq:app-mpl-noise-term} gives
the following minimization objective:
\begin{equation}
    \begin{aligned}
    \E\!\left[\cJ(\gamma,\eta)\right]
    ={}&C_0+
    \frac{C}
    {\left(K_1/2+\int_0^T\eta(t)\gamma(t)\d t\right)^{\alpha_1+2}}\\
    &+C'\int_0^T
      \frac{\eta(s)^2}
      {\left(
        (K_1+G_1)/2+\int_s^T\eta(t)\gamma(t)\d t
      \right)^{\alpha_1+\rho_1+2}}
    \d s.
    \end{aligned}
    \label{eq:app-mpl-final-loss}
\end{equation}
Both schedule-dependent terms in \Cref{eq:app-mpl-final-loss} are nonincreasing when MT
data is added. Therefore, for $0<\alpha<1$, an optimal schedule exhausts the
MT data budget:
\begin{equation*}
    \int_0^T\gamma(t)\d t=\alpha T.
\end{equation*}
Equation \eqref{eq:app-mpl-final-loss}, together with this active
constraint and $0\leq\gamma(t)\leq1$, is the functional optimized in the next
subsection.

By a similar argument to that in \Cref{app:terminal-loss-derivation}, the KKT conditions are necessary and sufficient for optimality, and the optimal data schedule is unique if it exists.

\subsection{KKT Condition}

For compactness, define
\begin{equation*}
    a:=\alpha_1+2,\qquad
    b:=\alpha_1+\rho_1+2,\qquad
    G_{\mathrm{eff}}:=\frac{K_1+G_1}{2},
\end{equation*}
and introduce the two schedule-dependent accumulated learning rates
\begin{equation*}
    A_\gamma:=\frac{K_1}{2}+\int_0^T\eta(t)\gamma(t)\d t,
    \qquad
    B_\gamma(s):=G_{\mathrm{eff}}+\int_s^T\eta(t)\gamma(t)\d t.
\end{equation*}
Thus, after dropping terms independent of $\gamma$, the objective in
\Cref{eq:app-mpl-final-loss} is
\begin{equation}
    J(\gamma)
    =C A_\gamma^{-a}
     +C'\int_0^T\eta(s)^2B_\gamma(s)^{-b}\d s.
    \label{eq:app-mpl-compact-objective}
\end{equation}

Let $v\in L^\infty([0,T])$ be an admissible perturbation. Differentiating
\Cref{eq:app-mpl-compact-objective} in the direction $v$ gives
\begin{align*}
    D J(\gamma)[v]
    ={}&
    -\frac{aC}{A_\gamma^{a+1}}
       \int_0^T\eta(t)v(t)\d t \notag\\
    &-bC'\int_0^T
       \frac{\eta(s)^2}{B_\gamma(s)^{b+1}}
       \left(\int_s^T\eta(t)v(t)\d t\right)\d s.
\end{align*}
The second term is integrated over
$\{(s,t):0\leq s\leq t\leq T\}$. Applying Fubini's theorem therefore yields
\begin{equation*}
    D J(\gamma)[v]
    =-\int_0^T M_\gamma(t)v(t)\d t,
\end{equation*}
where the marginal value of MT data at time $t$ is
\begin{equation}
    M_\gamma(t)
    :=\eta(t)\left[
       \frac{aC}{A_\gamma^{a+1}}
       +bC'\int_0^t
          \frac{\eta(s)^2}{B_\gamma(s)^{b+1}}\d s
       \right].
    \label{eq:app-mpl-switching-function}
\end{equation}

We now solve the optimization problem. Let $\kappa\geq0$ be the
multiplier of
$\int_0^T\gamma(t)\d t-\alpha T\leq0$, and let
$u_0(t),u_1(t)\geq0$ be the pointwise multipliers of
$-\gamma(t)\leq0$ and $\gamma(t)-1\leq0$, respectively. The Lagrangian is
\begin{align*}
    \mathfrak{L}(\gamma,\kappa,u_0,u_1)
    :={}&J(\gamma)
       +\kappa\left(\int_0^T\gamma(t)\d t-\alpha T\right) \notag\\
       &-\int_0^T u_0(t)\gamma(t)\d t
       +\int_0^T u_1(t)\bigl(\gamma(t)-1\bigr)\d t.
\end{align*}
The KKT conditions are, almost everywhere on $[0,T]$,
\begin{align}
    &0\leq\gamma(t)\leq1,\qquad
      \int_0^T\gamma(t)\d t\leq\alpha T, \notag\\
    &\kappa\geq0,\qquad u_0(t)\geq0,\qquad u_1(t)\geq0,
      \notag\\
    &-M_\gamma(t)+\kappa-u_0(t)+u_1(t)=0,
      \label{eq:app-mpl-kkt-stationarity}\\
    &\kappa\left(\int_0^T\gamma(t)\d t-\alpha T\right)=0,
      \quad
      u_0(t)\gamma(t)=0,
      \quad
      u_1(t)\bigl(1-\gamma(t)\bigr)=0.
      \label{eq:app-mpl-kkt-complementarity}
\end{align}
Because $M_\gamma(t)>0$ whenever $\eta(t)>0$, adding MT data strictly decreases the objective. Hence, for
$0<\alpha<1$, the optimal data schedule must satisfy
\begin{equation*}
    \int_0^T\gamma^\star(t)\d t=\alpha T.
\end{equation*}
Eliminating $u_0(t)$ and $u_1(t)$ from
\Cref{eq:app-mpl-kkt-stationarity,eq:app-mpl-kkt-complementarity} gives the
equivalent conditions
\begin{equation*}
    \begin{cases}
        M_{\gamma^\star}(t)\leq\kappa,
            & \gamma^\star(t)=0,\\
        M_{\gamma^\star}(t)=\kappa,
            & 0<\gamma^\star(t)<1,\\
        M_{\gamma^\star}(t)\geq\kappa,
            & \gamma^\star(t)=1.
    \end{cases}
\end{equation*}
Since $x\mapsto x^{-a}$ and $x\mapsto x^{-b}$ are convex on
$(0,\infty)$, $J$ is convex in $\gamma$. The feasible set is also convex, so
these KKT conditions are sufficient as well as necessary.

\subsection{\texorpdfstring{The Shape of the Data Schedule when $M_{\gamma}(t)$ is a Constant}{The Shape of the Data Schedule when M\_gamma(t) Is Constant}}

In this section, we assume that $\eta'(t)<0$ for $t\in[0,T]$, and $M_{\gamma}(t)$ is constant for $t\in [t_1,t_2]$. Differentiation of
\Cref{eq:app-mpl-switching-function} gives
\begin{equation}
    M_\gamma'(t)
    =\frac{\eta'(t)}{\eta(t)}M_\gamma(t)
     +\frac{bC'\eta(t)^3}{B_\gamma(t)^{b+1}}.
    \label{eq:app-m-gamma-derivative}
\end{equation}
Thus when $M_{\gamma}(t)=\kappa$ is constant,
\begin{equation*}
    \kappa\eta'(t)
    +\frac{bC'\eta(t)^4}{B_{\gamma}(t)^{b+1}}=0,
\end{equation*}
which implies
\begin{equation*}
    B_{\gamma}(t)
    =\left(
        -\frac{bC'\eta(t)^4}{\kappa\eta'(t)}
      \right)^{1/(b+1)}.
\end{equation*}
When $\eta(t)=\eta_{\max}-\frac{c}{T}t$, $\eta'(t)=-c/T$ is constant. Thus for a fixed time $t_2$, we have
\begin{align*}
  \frac{B_{\gamma}(t)}{B_{\gamma}(t_2)}=\left(
        \frac{\eta(t)^4}{\eta(t_2)^4}
      \right)^{1/(b+1)},
\end{align*}
and thus
\begin{align}
  B_{\gamma}(t)=B_{\gamma}(t_2)\left(
        \frac{\eta(t)^4}{\eta(t_2)^4}
      \right)^{1/(b+1)}. \label{eq:app-mpl-interior-state}
\end{align}
We define
\begin{equation*}
    B_2:=B_{\gamma}(t_2)
       =G_{\mathrm{eff}}+\int_{t_2}^T\eta(\tau)\gamma(\tau)\d\tau,
    \qquad
    K_2:=B_2\eta(t_2)^{-4/(b+1)},
    \qquad
    \kappa_2:=\frac{bC'T}{cK_2^{b+1}}.
\end{equation*}
A direct corollary is that
\begin{equation*}
    B_{\gamma_{t_1,t_2}}(t)=K_2\eta(t)^{4/(b+1)}.
\end{equation*}
Then we take the derivative of \Cref{eq:app-mpl-interior-state} and substitute $B_{\gamma}'(t)=-\eta(t)\gamma(t)$. We get
\begin{align}
  \gamma(t)=\frac{4cK_2}{T(b+1)}\eta(t)^{-\frac{2(b-1)}{b+1}}. \label{equ:gamma-int-expression}
\end{align}

\subsection{\texorpdfstring{Proof of \Cref{thm:opt-ds-multi-power-law-spectra}}{Proof of Theorem \ref*{thm:opt-ds-multi-power-law-spectra}}}
\begin{proof}
We define a family of data schedules $\gamma_{t_1,t_2}(t)$ parameterized by two time steps $0<t_1<t_2<T$. The expression of $\gamma_{t_1,t_2}(t)$ is given by
\begin{equation*}
    \gamma_{t_1,t_2}(t):=
    \begin{cases}
        0, & 0\leq t<t_1,\\
        \displaystyle
        \frac{4cK_2}{T(b+1)}\eta(t)^{-\frac{2(b-1)}{b+1}},
            & t_1\leq t<t_2,\\
        1, & t_2\leq t\leq T.
    \end{cases}
\end{equation*}

We first need to verify that $\gamma_{t_1,t_2}(t)$ is a valid data schedule, i.e., $0\leq\gamma_{t_1,t_2}(t)\leq 1$.
\begin{lemma} \label{lem:gamma-leq-1}
    There exists a constant $T_0>0$ such that for any $T>T_0$, $\gamma_{t_1,t_2}(t)$ is a valid data schedule for any $0<t_1<t_2<T$.
\end{lemma}
\begin{proof}
The assumptions $-1<\alpha_1<0$ and $0<\rho_1<1$ imply
$1<b<3$. Therefore, $\gamma_{t_1,t_2}(t)$ is monotonically increasing in $t\in(t_1,t_2)$. So we only need to verify that the left limit satisfies $\gamma_{t_1,t_2}(t_2^-)\leq 1$. By definition, we have
\begin{equation*}
    \gamma_{t_1,t_2}(t_2^-)
    =\frac{4cB_2}{T(b+1)\eta(t_2)^2}.
\end{equation*}

\begin{lemma} \label{lem:B-upper-bound}
    When $t\geq t_2$, we have
    \[B_{\gamma_{t_1,t_2}}(t) \leq G_{\mathrm{eff}} + \frac{T\eta(t)^2}{2c}.\]
\end{lemma}
\begin{proof}
We conclude the proof through a direct calculation:
\begin{align*}
    B_{\gamma_{t_1,t_2}}(t)&=G_{\mathrm{eff}}+\int_{t}^T\eta(\tau)\d\tau
    =G_{\mathrm{eff}}+\int_{t}^T\left(\eta_{\max}-\frac{c}{T}\tau\right)\d\tau \\
    &\leq G_{\mathrm{eff}}+\int_{t}^{c^{-1}T\eta_{\max}}\left(\eta_{\max}-\frac{c}{T}\tau\right)\d\tau \\
    &\leq G_{\mathrm{eff}}+\frac{T\eta(t)^2}{2c}.
\end{align*}
\end{proof}

By \Cref{lem:B-upper-bound}, we have
\begin{equation*}
    B_2=B_{\gamma_{t_1,t_2}}(t_2)\leq G_{\mathrm{eff}}+\frac{T\eta(t_2)^2}{2c}.
\end{equation*}
Substituting the upper bound on $B_2$ into the expression for $\gamma_{t_1,t_2}(t_2^-)$ gives an upper bound on $\gamma_{t_1,t_2}(t_2^-)$:
\begin{equation*}
    \gamma_{t_1,t_2}(t_2^-)
    \leq \frac{4c}{T(b+1)\eta(t_2)^2}\cdot\left(\frac{T\eta(t_2)^2}{2c}+G_{\mathrm{eff}}\right)\leq\frac{2}{b+1}+\frac{4cG_{\mathrm{eff}}}{T(b+1)\eta(T)^2}.
\end{equation*}
Thus a sufficient condition for $\gamma_{t_1,t_2}(t_2^-)\leq 1$ is
\begin{equation*}
    \frac{2}{b+1}+\frac{4cG_{\mathrm{eff}}}{T(b+1)\eta(T)^2}\leq 1,
\end{equation*}
which is equivalent to
\begin{equation*}
    T\geq\frac{4cG_{\mathrm{eff}}}{(b+1)\left(1-\frac{2}{b+1}\right)\eta(T)^2}=\frac{4cG_{\mathrm{eff}}}{(b-1)\eta(T)^2}=:T_0.
\end{equation*}
\end{proof}
\begin{lemma} \label{lem:M-second-derivative-positive}
There exists a constant $T_1>0$ such that for any $T>T_1$, $M_{\gamma_{t_1,t_2}}''(t)<0$ for all $0<t<t_1$ and $M_{\gamma_{t_1,t_2}}''(t)>0$ for all $t_2<t<T$.
\end{lemma}
\begin{proof}
\Cref{eq:app-m-gamma-derivative} gives
\begin{align}
M_{\gamma_{t_1,t_2}}'(t)
&=\frac{\eta'(t)}{\eta(t)}M_{\gamma_{t_1,t_2}}(t)
    +\frac{bC'\eta(t)^3}{B_{\gamma_{t_1,t_2}}(t)^{b+1}} \notag \\
    &=\eta'(t)\left[
    \frac{aC}{A_\gamma^{a+1}}
    +bC'\int_0^t
        \frac{\eta(s)^2}{B_{\gamma_{t_1,t_2}}(s)^{b+1}}\d s
    \right]
    +\frac{bC'\eta(t)^3}{B_{\gamma_{t_1,t_2}}(t)^{b+1}} \notag \\
    &=-\frac{c}{T}\frac{aC}{A_\gamma^{a+1}}
    +bC'\left[-\frac{c}{T}\int_0^t
        \frac{\eta(s)^2}{B_{\gamma_{t_1,t_2}}(s)^{b+1}}\d s
    +\frac{\eta(t)^3}{B_{\gamma_{t_1,t_2}}(t)^{b+1}}\right]. \label{equ:M-derivative-small-t}
\end{align}
Notice that $B_{\gamma_{t_1,t_2}}(t)\equiv B_{\gamma_{t_1,t_2}}(t_1)$ when $t\in[0, t_1]$, and we denote this constant by $B_1$. We write it explicitly as
\begin{equation*}
    B_1:=B_{\gamma_{t_1,t_2}}(t_1)
        =G_{\mathrm{eff}}+\int_{t_1}^T\eta(\tau)\gamma_{t_1,t_2}(\tau)\d\tau.
\end{equation*}
Then we have
\begin{equation*}
    M_{\gamma_{t_1,t_2}}''(t)
    =-\frac{4bcC'}{T}
      \frac{\eta(t)^2}{B_1^{b+1}}<0,
    \qquad 0<t<t_1.
\end{equation*}
When $t_2<t<T$, $B_\gamma(t)=G_{\mathrm{eff}}+\int_t^T\eta(\tau)\d\tau$
and $\gamma(t)=1$.  Direct differentiation gives
\begin{equation*}
    M_\gamma''(t)
    =\frac{bC'\eta(t)^2}{B_\gamma(t)^{b+2}}
      \left[
        -\frac{4c}{T}B_\gamma(t)+(b+1)\eta(t)^2
      \right].
\end{equation*}
By \Cref{lem:B-upper-bound}, we have
\begin{equation*}
    M_\gamma''(t)
    \geq\frac{bC'\eta(t)^2}{B_\gamma(t)^{b+2}}
      \left[
        -\frac{4c}{T}G_{\mathrm{eff}}+(b-1)\eta(T)^2
      \right]>0
\end{equation*}
when $T>\frac{4cG_{\mathrm{eff}}}{(b-1)\eta(T)^2}$.
\end{proof}
\Cref{thm:opt-ds-multi-power-law-spectra} holds if and only if there exists $(t_1, t_2)$ such that $\gamma_{t_1,t_2}(t)$ satisfies the KKT conditions and the constraints for $\gamma$. For this candidate, the pointwise KKT conditions and the
active budget require
\begin{align*}
    M_{\gamma_{t_1,t_2}}(t)&\leq\kappa_2,
        &&0\leq t<t_1,\\
    M_{\gamma_{t_1,t_2}}(t)&=\kappa_2,
        &&t_1\leq t<t_2,\\
    M_{\gamma_{t_1,t_2}}(t)&\geq\kappa_2,
        &&t_2\leq t\leq T,\\
    0\leq\gamma_{t_1,t_2}(t)&\leq 1,
        &&0\leq t\leq T,    \\
    \int_0^T\gamma_{t_1,t_2}(t)\d t&=\alpha T.
\end{align*}
By \Cref{lem:M-second-derivative-positive} and \Cref{lem:gamma-leq-1}, sufficient conditions for the conditions above are
\begin{align*}
    M_{\gamma_{t_1,t_2}}'(t_1) &= 0, \\
    \int_0^T\gamma_{t_1,t_2}(t)\d t&=\alpha T.
\end{align*}

Before we proceed, we define some additional notation. Let $x=\frac{T-t}{T}$ be the reversed-and-normalized time variable. Then $\eta(t)=\eta_{\max}-c(1-x)$. We define
\begin{equation*}
    \epsilon:=\eta(T)=\eta_{\max}-c>0,
    \quad
    P(x):=\eta(t)=\epsilon+cx,
    \quad
    L(x):=\frac{1}{T}\int_{t}^{T}\eta(s)\d s=\epsilon x+\frac{c}{2}x^2,
\end{equation*}
and write
\begin{equation*}
    x_j:=\frac{T-t_j}{T},
    \qquad j\in\{1,2\}.
\end{equation*}

We first rewrite \Cref{equ:M-derivative-small-t} in terms of the new variables:
\begin{align*}
    M_{\gamma_{t_1,t_2}}'(t)
    &=\frac{bC'}{B_1^{b+1}}\left[
        P(x)^3-c\int_{x}^{1}P(u)^2\d u
    -
    \frac{caC}{bC'T}
    \frac{B_1^{b+1}}{A_{\gamma_{t_1,t_2}}^{a+1}}
      \right].
\end{align*}
Define
\begin{equation*}
    F(x):=P(x)^3-c\int_x^1P(u)^2\d u.
\end{equation*}
Then $M_{\gamma_{t_1,t_2}}'(t_1)=0$ is equivalent to
\begin{equation}
    F(x_1)
    =\frac{caC}{bC'T}
     \frac{B_1^{b+1}}{A_{\gamma_{t_1,t_2}}^{a+1}}.
    \label{eq:M-derivative-zero}
\end{equation}
It remains to solve for $x_2$ using the budget.
From \Cref{equ:gamma-int-expression}, we have
\begin{align*}
    \gamma_{t_1,t_2}\bigl(T(1-x)\bigr)
        =\frac{4cB_2}{T(b+1)}
          P(x_2)^{-4/(b+1)}
          P(x)^{-2(b-1)/(b+1)}.
\end{align*}
Recall that $B_2=B_{\gamma_{t_1,t_2}}(t_2)=G_{\mathrm{eff}}+\int_{t_2}^T\eta(\tau)\d\tau=G_{\mathrm{eff}}+TL(x_2)$. Then we have
\begin{align}
\alpha
&= \frac{1}{T}\int_0^T\gamma_{t_1,t_2}(t)\d t
\notag\\
&= \frac{1}{T}\left(
    \int_{t_1}^{t_2}\gamma_{t_1,t_2}(t)\d t + (T-t_2)
\right)
\notag\\
&= x_2
+\frac{4\bigl(L(x_2)+T^{-1}G_{\mathrm{eff}}\bigr)}{3-b}
P(x_2)^{-4/(b+1)}
\left(
    P(x_1)^{(3-b)/(b+1)}
    -P(x_2)^{(3-b)/(b+1)}
\right).
\label{equ:mpl-budget-constraint}
\end{align}
We denote $f_{\epsilon,b}(x_1,x_2)$ as the right-hand side of the above equation.
The term involving $T^{-1}G_{\mathrm{eff}}$ is of order $T^{-1}$,
uniformly on compact subsets of $0<x_2<x_1<1$.

\begin{lemma}
Under the assumptions of \Cref{thm:opt-ds-multi-power-law-spectra}, the equation
\[
    F(x_1)=0,
    \qquad
\]
has a unique solution $x_1=x_\star$ satisfying
$0<x_\star<1$. This solution is independent of $T$.
\end{lemma}
\begin{proof}
Since $P'(x)=c$ and $P(1)=\eta_{\max}$, direct integration gives
\[
    F(x)
    =P(x)^3-c\int_x^1P(u)^2\d u
    =\frac{4P(x)^3-\eta_{\max}^3}{3}.
\]
Consequently, $F'(x)=4cP(x)^2>0$, so $F$ has at most one zero. Solving
$F(x)=0$ gives
\[
    x=x_\star
    =\frac{4^{-1/3}\eta_{\max}-\epsilon}{c}.
\]
The assumption
$c>(1-4^{-1/3})\eta_{\max}$ implies $x_\star>0$, while
$4^{-1/3}\eta_{\max}-\epsilon<c$ implies $x_\star<1$. Thus the first equation
has exactly one solution in $(0,1)$.

\end{proof}

For some fixed $\frac{\alpha+x_\star}{2}\leq x_1\leq 1$, we have $f_{\epsilon,b}(x_1,\alpha)> \alpha$, and
\begin{align*}
    f_{\epsilon,b}(x_1,x_2)&\leq x_2+\frac{4\bigl(\epsilon x_2+\frac{c}{2}x_2^2+T^{-1}G_{\mathrm{eff}}\bigr)}{3-b}P(x_2)^{-4/(b+1)}P(x_1)^{(3-b)/(b+1)} \\
    &\leq x_2+\frac{4\bigl(\epsilon x_2+\frac{c}{2}x_2+T^{-1}G_{\mathrm{eff}}\bigr)}{3-b}P(0)^{-4/(b+1)}P(1)^{(3-b)/(b+1)} \\
    &\leq \left(1+\frac{4\bigl(\epsilon+\frac{c}{2}\bigr)\eta_{\max}^{(3-b)/(b+1)}}{(3-b)\epsilon^{4/(b+1)}}\right)x_2+\frac{4G_{\mathrm{eff}}\eta_{\max}^{(3-b)/(b+1)}}{T(3-b)\epsilon^{4/(b+1)}}.
\end{align*}
for all $0<x_2<x_\star$.
We denote $D=\left(1+\frac{4\bigl(\epsilon+\frac{c}{2}\bigr)\eta_{\max}^{(3-b)/(b+1)}}{(3-b)\epsilon^{4/(b+1)}}\right)\geq 1$, $E=\frac{4G_{\mathrm{eff}}\eta_{\max}^{(3-b)/(b+1)}}{(3-b)\epsilon^{4/(b+1)}}$, and $T_2=\frac{3E}{\alpha}$. Then when $T>T_2$, we have
\begin{equation*}
    f_{\epsilon,b}(x_1,\frac{\alpha}{3D})\leq D\times \frac{\alpha}{3D}+\frac{E}{T_2}\leq \frac{2\alpha}{3}<\alpha.
\end{equation*}
Hence, for sufficiently large $T$ and any fixed $x_1\in\left[\frac{\alpha+x_\star}{2},1\right]$, there exists a unique solution $x_2(x_1)=\Theta(1)$.

We now return to \Cref{eq:M-derivative-zero}. As $x_2(x_1)=\Theta(1)$ for all $x_1\in\left[\frac{\alpha+x_\star}{2},1\right]$, we have
\begin{align*}
    \int_{0}^{T}\eta(s)\gamma(s)\d s\geq\frac{1}{2}\eta(t_2)(T-t_2) \geq \frac{1}{2}\times x_2 \eta_{\max} \times x_2 T = \Theta(T).
\end{align*}
Therefore,
$A_{\gamma_{t_1,t_2}}=\Theta(T)$ and $B_1=\Theta(T)$.
Since $b-a=\rho_1<1$, the right-hand side of the smooth-pasting equation is
$O(T^{\rho_1-1})=o(1)$, uniformly on $x_1\in\left[\frac{\alpha+x_\star}{2},1\right]$. So there exists $T_3>0$ such that for all $T>T_3$, the right-hand side of \Cref{eq:M-derivative-zero} satisfies
\begin{align*}
    0=F(x^*)<\frac{caC}{bC'T}
     \frac{B_1^{b+1}}{A_{\gamma_{t_1,t_2}}^{a+1}}< F(1).
\end{align*}
As $F(x)$ is increasing, when $T>\max\{T_0, T_1, T_2, T_3\}$, there exists a unique solution $x_1(T)\in(x_\star, 1)$ to \Cref{eq:M-derivative-zero}. This concludes our proof of the existence of a solution $(x_1(T), x_2(T))$ that satisfies the KKT conditions and the constraints.

\end{proof}
\section{Numerical Validation of the Theoretical Model}
\label{app:theory-simulation}

This section provides the details of the simulation and additional experiments supporting the theoretical predictions in
\Cref{thm:informal}. In \Cref{app:sim-setup-detail}, we provide additional details on the simulation setup. In \Cref{app:theory-simulation-sde-check}, we provide a check on the validity of the SDE approximation.
In \Cref{app:theory-numerical}, we
numerically solve the same scheduling problem analyzed in
\Cref{sec:theory}, thereby checking the closed form solution given in \Cref{thm:opt-ds-multi-power-law-spectra}. In \Cref{app:theory-ds-change-lrs}, we intuitively explain how the optimal data schedule changes with the learning rate schedule as shown in \Cref{thm:opt-ds-const-lr,thm:opt-ds-multi-power-law-spectra}. In \Cref{app:theory-simulation-optimization}, we verify that the data schedule solved by numerical methods satisfies the KKT condition.

\subsection{Additional Details on the Simulation Experiments} \label{app:sim-setup-detail}
\paragraph{Simulation setup.} We use the same quadratic model, structure of PT and MT data, power-law spectrum, linearly decaying learning rate and
constrained MT data budget as in the theorem. We use the closed form of the SDE approximation to calculate the expected validation loss. Regarding why this approximation is reasonable, see \Cref{app:theory-simulation-sde-check} for more details.

\paragraph{Calculation of the validation loss.} In our simulation, we use two methods to numerically calculate the expected validation loss of the SDE approximation: one is to sample all the eigenvalues of $\mG$ and $\mH$ according to their corresponding distributions, and the other one is to calculate the expected validation loss with respect to the distributions of eigenvalues directly. For each numerical setup and MT data budget, we first
discretize the interval $[0,T]$ into $M=400$ identical time bins $\{[\tau_{i-1},\tau_i]\}_{i=1}^M$ and approximate
$\gamma(t)$ with ${\gamma(\tau_j)}$ for $\tau_j\leq t \leq \tau_{j+1}$.

\paragraph{Hyperparameter selection.} For the hyperparameters of the MPL spectral model, we take $\alpha_l=-0.5$, $K_l=1$, $\rho_l=0.5$, $F_l=1$, $G_l=1$. The training horizon is set to $T=200$. We take $\eta_{\max}=1$ as the initial learning rate and initialize $\vtheta_i^{l}=1$ for all $l=0,1,2$ and $1\leq i\leq d_l$. We also set the discretization step size $h=0.1$. Note that by the derivations in \Cref{app:terminal-loss-derivation}, the minimization objective is determined only by the contribution of knowledge specific to the MT domain, so the minimization objective depends on $d_1$ and is independent of $d_0$ and $d_2$. In our simulation experiments, we set $d_1=4000$.

\paragraph{Numerical optimization.}
We discretize $[0,T]$ into $K=400$ equal-width bins and minimize
the resulting objective $J(y)$ over the feasible set
\[
\mathcal{Y}
=
\left\{
y\in[0,1]^K:
\sum_{k=1}^{K}y_k=\alpha K
\right\}.
\]
We minimize $J$ over $\mathcal{Y}$ using an accelerated
projected-gradient method based on FISTA \citep{beck2009fast}.
Starting from a feasible point $y^{(0)}\in\mathcal{Y}$, we initialize
an auxiliary point $z^{(0)}=y^{(0)}$. At iteration $n$, we take a
gradient step from $z^{(n)}$ and project the result onto $\mathcal{Y}$:
\[
v^{(n)}
=
z^{(n)}-\eta_n\nabla J\!\left(z^{(n)}\right),
\qquad
y_k^{(n+1)}
=
\operatorname{clip}\!\left(v_k^{(n)}-\tau_n,\,0,\,1\right).
\]
The scalar threshold $\tau_n$ is determined by the bisection method to enforce
$\sum_k y_k^{(n+1)}=\alpha K$. We choose the step size $\eta_n$ by backtracking.
For each trial step size, we compute $y^{(n+1)}$ using the
gradient and projection steps above and check whether
\[
J\!\left(y^{(n+1)}\right)
\le
J\!\left(z^{(n)}\right)
+
\left\langle
\nabla J\!\left(z^{(n)}\right),
y^{(n+1)}-z^{(n)}
\right\rangle
+
\frac{\left\|y^{(n+1)}-z^{(n)}\right\|_2^2}{2\eta_n}.
\]
If the inequality fails, we halve $\eta_n$ and recompute
$y^{(n+1)}$, repeating until the condition is satisfied.
After an accepted step, the next line search is
initialized with $1.2\eta_n$.

The next auxiliary point incorporates momentum along the latest
update direction:
\[
z^{(n+1)}
=
y^{(n+1)}
+
\beta_n\left(y^{(n+1)}-y^{(n)}\right),
\]
where $\beta_n$ is the standard FISTA momentum coefficient.
Whenever $J(y^{(n+1)})>J(y^{(n)})$, we restart the momentum and set
$z^{(n+1)}=y^{(n+1)}$ \citep{odonoghue2015adaptive}.
Iterations terminate when the absolute objective change is below
$10^{-15}\max\{1,|J(y^{(n)})|\}$, or after $20{,}000$ iterations.

\paragraph{\texorpdfstring{Calculation of the closed form schedule predicted by \Cref{thm:opt-ds-multi-power-law-spectra}.}{Calculation of the Closed Form Schedule Predicted by the Theorem.}}
For a fixed parameter setting, the schedule in
\Cref{thm:opt-ds-multi-power-law-spectra} is determined by its two transition
times $t_1$ and $t_2$.  We compute them from the two scalar conditions used in
the proof. The first one, given in \Cref{eq:M-derivative-zero}, is a sufficient condition for satisfying the KKT conditions. The second is the active budget
condition in \Cref{equ:mpl-budget-constraint}, which states that the total amount of MT data should equal the MT data budget.
After solving these two equations, we calculate the interior part of the
schedule using the expression in
\Cref{thm:opt-ds-multi-power-law-spectra}, and set the schedule to zero before
$t_1$ and to one after $t_2$.

\subsection{Checks of the SDE Approximation}
\label{app:theory-simulation-sde-check}
\begin{figure}[t]
    \centering
    \includegraphics[width=\linewidth]{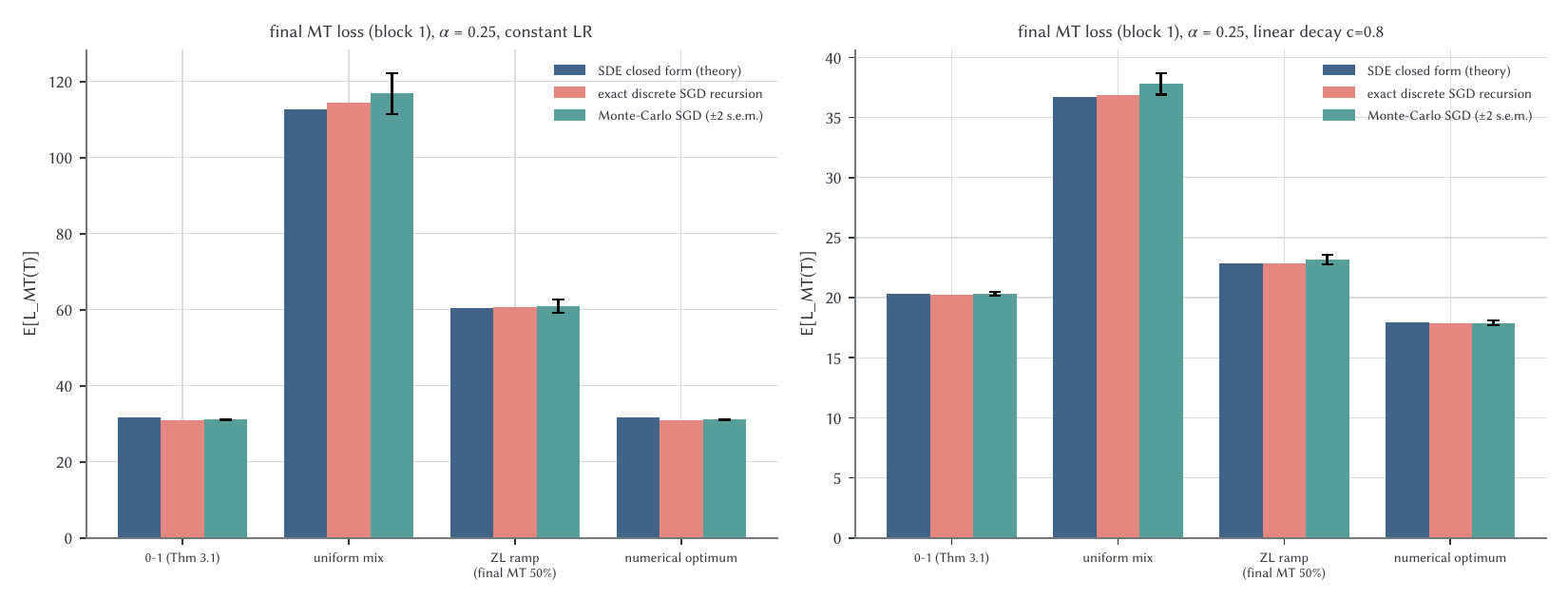}
    \caption{Consistency of the SDE calculation, the exact discrete SGD
    second moment recursion, and Monte Carlo simulation for $\alpha=0.25$.
    Results are shown for a constant learning rate (left) and linear decay with
    $c=0.8$ (right), across four representative data schedules. Error bars give
    two standard errors of the Monte Carlo mean.}
    \label{fig:app-sde-discrete-mc}
\end{figure}
To show the accuracy of the SDE approximation,
for coordinate $i$ in block $l$, we denote $m_{i,k}^{(l)}:=\E\left[\left(\theta_{i,k}^{(l)}\right)^2\right]$, where $k$ indexes the training step.
The discrete SGD update in
\Cref{equ:sgd-dynamics} gives
\begin{equation}
    m_{i,k+1}^{(l)}
    =\left(1-\eta_k a_i^{(l)}(t_k)\right)^2m_{i,k}^{(l)}
     +\eta_k^2\nu_i^{(l)}
    \label{eq:app-sim-discrete-second-moment}
\end{equation}
through a direct calculation, where $t_k=kh$ and $a_i^{(l)}(t)$ is defined in \Cref{def:a_i^{(l)}}.
Iterating \eqref{eq:app-sim-discrete-second-moment} yields the exact expected terminal loss of the discrete
model. As a further
check, we simulate the stochastic recursion itself and average the terminal
loss across all the trajectories. We plot the results of the three methods in \Cref{fig:app-sde-discrete-mc}. We also report the uncertainty of the Monte Carlo estimates using error bars, which indicate \(\pm 2\) standard errors of the mean across simulated trajectories. \Cref{fig:app-sde-discrete-mc} shows that the results of the three methods are almost identical, which supports the accuracy of the SDE approximation.

\subsection{Numerical Validation of Theoretical Predictions}
\label{app:theory-numerical}

In this section, we verify numerically that the closed form solution in
\Cref{thm:opt-ds-multi-power-law-spectra} agrees with the optimum of \Cref{aim:data-schedule-minimization}.

\begin{figure}[t]
    \centering
    \includegraphics[width=\linewidth]{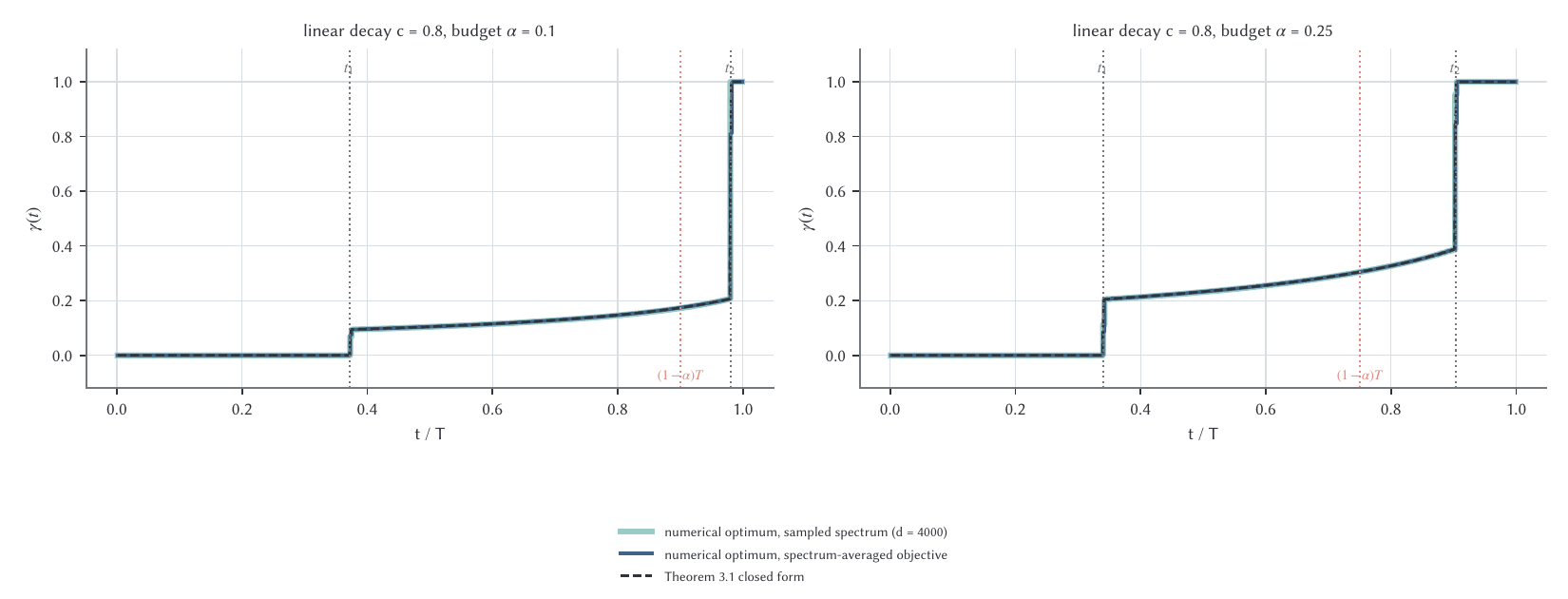}
    \caption{Numerical validation of
    \Cref{thm:opt-ds-multi-power-law-spectra}. Solid curves show the schedules
    predicted by the closed form solution, while markers show the optima
    obtained by directly optimizing the expected validation loss using different numerical methods.
    The two solutions agree for both MT data budgets considered.}
    \label{fig:theory-closed-form}
\end{figure}

Figure~\ref{fig:theory-closed-form} shows that the solutions of the two numerical methods closely agree with the closed form solution given in \Cref{thm:opt-ds-multi-power-law-spectra} under different MT data budgets. In particular,
the numerical solutions recover the three stage structure
predicted by the theorem: an initial PT only phase, an intermediate region in
which the MT ratio increases continuously, and a final MT only phase. The
agreement also extends to the locations of the two transition points. These
results confirm that the schedule structure derived in the theorem
is also observed when the expected validation loss is optimized directly.

\subsection{How the Optimal Data Schedule Changes with the Learning Rate Schedule} \label{app:theory-ds-change-lrs}

\Cref{thm:opt-ds-const-lr,thm:opt-ds-multi-power-law-spectra} together show that the shape of the optimal data schedule differs when the learning rate changes from a constant learning rate to a linearly decaying learning rate schedule. To test the predictions of \Cref{thm:opt-ds-const-lr} and \Cref{thm:opt-ds-multi-power-law-spectra} regarding how the optimal data schedule changes with the learning rate, we numerically optimize the expected validation loss over the data schedule. We use the same setting as in \Cref{thm:opt-ds-multi-power-law-spectra}, and the expectation is taken with respect to the gradient noise and the eigenvalues of $\mG$ and $\mH$. We take two data budgets $\alpha\in\{0.1,0.25\}$ and vary the learning rate decay strength $c$ from a constant learning rate to a learning rate schedule with a strong linear decay.

The numerical results are shown in \Cref{fig:app-optimal-schedule-decay}. With a constant learning rate, the optimum is the two stage schedule from
\Cref{thm:opt-ds-const-lr}: the first stage is to use only PT data, and the second stage is to use only MT data. The conclusion also holds for a learning rate schedule with a weak
decay, namely $c=0.2$, in these instances. Once the decay becomes sufficiently strong, namely $c=0.5$ and $c=0.8$, the optimal data schedule changes to a three stage schedule. In particular, the optimal schedule begins with PT data only, then transitions to a mixture of PT and MT data, and ends with MT data only. During the second stage, the proportion of MT data gradually increases. These observations are consistent with the predictions of \Cref{thm:opt-ds-multi-power-law-spectra}.

\subsection{Verification of the KKT Condition of the Optimal Data Schedule}
\label{app:theory-simulation-optimization}

To further demonstrate the optimality of the data schedule shown in \Cref{thm:opt-ds-const-lr,thm:opt-ds-linear-decay-lr,thm:opt-ds-multi-power-law-spectra}, we additionally evaluate the KKT conditions given by \Cref{eq:app-switching-function} and \Cref{original-kkt}.

\begin{figure}[t]
    \centering
    \includegraphics[width=\linewidth]{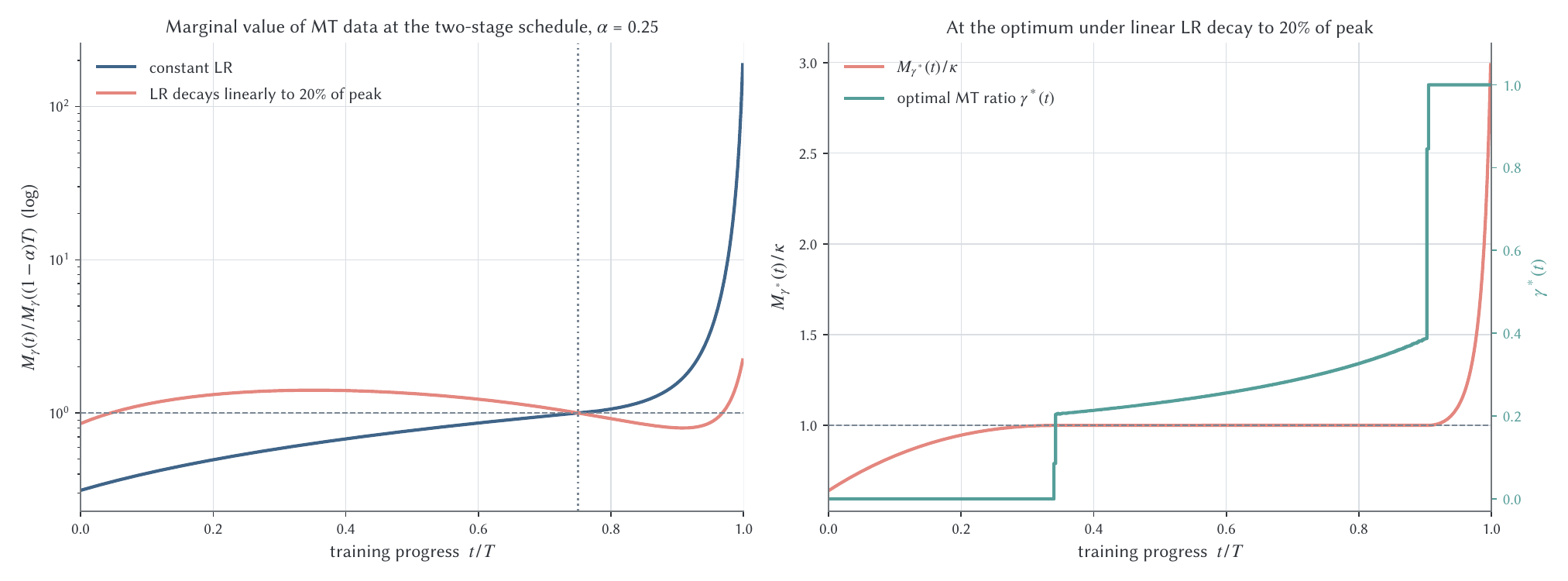}
    \caption{Numerical illustration of the KKT mechanism. \emph{Left:} At the
    two stage schedule with $\alpha=0.25$, the marginal value of MT data is
    increasing under a constant learning rate, whereas linear decay with
    $c=0.8$ violates the switching condition around $(1-\alpha)T$.
    \emph{Right:} At the numerical optimum under linear decay, the normalized
    marginal value equals the KKT multiplier throughout the interior mixing
    phase and satisfies the appropriate inequalities on the two boundary
    phases.}
    \label{fig:app-kkt-marginal-value}
\end{figure}

In the left panel of \Cref{fig:app-kkt-marginal-value}, we fix the data schedule to the two-stage schedule given by \Cref{thm:opt-ds-const-lr}, and select two learning rate schedules: a constant learning rate and a learning rate schedule with linear decay.
Using \Cref{eq:app-switching-function}, we plot the ratio of \(M_{\gamma}(t)\) to \(M_{\gamma}((1-\alpha)T)\) under each of the two learning rate schedules.
In the right panel of \Cref{fig:app-kkt-marginal-value}, we select a learning rate schedule with linear decay and its corresponding optimal data schedule. For easier visual comparison, we rescale \(M_\gamma(t)\) by a positive constant. This rescaling only changes the overall scale of \(M_\gamma(t)\), but does not change its shape over time. We then plot \(M_{\gamma}(t)\) divided by this rescaling factor.

If the 0-1 data schedule given by \Cref{thm:opt-ds-const-lr} is optimal, then by \Cref{original-kkt}, we have \(M_{\gamma}(t)\leq \kappa\) for $0\leq t \leq (1-\alpha)T$, and \(M_{\gamma}(t)\geq \kappa\) for $(1-\alpha)T\leq t \leq T$. The left panel of \Cref{fig:app-kkt-marginal-value} shows that under a constant learning rate, the 0-1 data schedule satisfies this condition, but under a learning rate schedule with linear decay, the 0-1 data schedule violates this condition. This demonstrates that the 0-1 data schedule is optimal for a constant learning rate, but is not optimal for a linear decay learning rate schedule, which aligns with the predictions of \Cref{thm:opt-ds-const-lr,thm:opt-ds-linear-decay-lr}.

The right panel of \Cref{fig:app-kkt-marginal-value} shows that \(M_\gamma(t)\) also has a three-stage structure and is divided into the same three stages as the data schedule. Specifically, in the first stage where only PT data is used, \(M_\gamma(t)\) is monotonically increasing; in the second stage where both PT and MT data are used, \(M_\gamma(t)\) is constant; and in the third stage where only MT data is used, \(M_\gamma(t)\) also monotonically increases. We take the Lagrange multiplier \(\kappa\) and the rescaling factor as the value of \(M_\gamma(t)\) in the second stage. Then by \Cref{original-kkt}, the optimal data schedule given by \Cref{thm:opt-ds-multi-power-law-spectra} satisfies the KKT conditions, which further confirms the optimality of the data schedule.

\clearpage
\section{Optimal Learning Rate Decays after MT Data Enters}
\label{app:lr-schedule}

\subsection{Setup and Formal Statement}
\label{app:lr-formal-statement}
We give the formal version of \Cref{thm:lr-delayed-decay}. Throughout this appendix, we assume \Cref{asp:constrained-cpt-data,asp:multi-power-law-spectra} and use the dynamics in \Cref{equ:sde-dynamics}. The MT-specific block is nonempty, and the initialization, dimensions, noise parameters, and spectral parameters are fixed independently of $T$.
For a fixed MT budget $\alpha\in(0,1)$, set $L=\alpha T$ and
\[
    \gamma(t)=\mathds{1}_{\{t\geq T-L\}}.
\]
Fix $0<m<M$ and $\Delta=M-m$. We optimize over $W\in[0,T]$ in the SD family
\[
    \eta_W(t)=
    \begin{cases}
        M, & 0\leq t<T-W,\\
        m+\Delta(T-t)/W, & T-W\leq t\leq T,
    \end{cases}
    \qquad W>0,
\]
with $\eta_0(t)=M$. Let $J_T(W)$ denote the expected final MT loss, with expectation over both training noise and the spectral distribution.

For $l\in\{0,1\}$, we define
\begin{equation*}
    a_l:=\alpha_l+2\in(1,2),\qquad
    b_l:=\alpha_l+\rho_l+2,\qquad
    g_l:=\frac{K_l+G_l}{2}>0,
\end{equation*}
and
\begin{equation*}
    B_l:=\frac{D_l\Gamma(a_l)}{2^{a_l+1}}
          \sum_{i=1}^{d_l}\theta_i^{(l)}(0)^2,
    \qquad
    V_l:=\frac{h d_l D_l F_l\Gamma(b_l)}{2^{b_l+1}}.
\end{equation*}
The indices $0$ and $1$ refer to shared and MT-specific features, respectively. In particular, $V_1>0$; if the shared block is absent, its bias and variance contributions are zero.

\begin{theorem}[Optimal decay begins after MT data enters training]
\label{thm:lr-delayed-decay-formal}
Under \Cref{asp:constrained-cpt-data,asp:multi-power-law-spectra}, consider the dynamics in \Cref{equ:sde-dynamics} with the TS data schedule $\gamma(t)=\mathds{1}_{\{t\geq(1-\alpha)T\}}$ and the SD LR family $\eta_W$, with fixed $0<m<M$. The initialization and all model parameters are fixed independently of $T$. Assume
\[
    1<b_1<2,\qquad b_0\geq b_1,
    \qquad b_l:=\alpha_l+\rho_l+2,\quad l\in\{0,1\}.
\]
For every fixed $\alpha\in(0,\alpha_c)$, where $\alpha_c$ is given by \Cref{eq:lr-budget-threshold} with $b=b_1$ and $r=m/M$, there exists a finite $T_0$, depending on $\alpha$ and the other fixed parameters, such that for every $T>T_0$, $J_T(W)$ attains its global minimum on $[0,T]$, and every global minimizer $W^*$ satisfies
\[
    0<W^*<\alpha T,
    \qquad
    (1-\alpha)T<T-W^*<T.
\]
\end{theorem}
The second inequality states that LR decay begins strictly after MT entry. The proof is given in \Cref{app:lr-main-proof}.

\subsection{Derivation of the Expected Validation Loss}
In particular, $B_1=C$ and $V_1=C'$ in \Cref{eq:app-mpl-final-loss}.
Note that the contribution from the shared block was omitted from the expression of \Cref{eq:app-mpl-final-loss} because it is independent of the data schedule. Here, we retain this
contribution because the contribution from the shared block depends on the learning rate schedule.
Following a derivation similar to that of
\Cref{eq:app-mpl-final-loss}, we obtain
\begin{equation}
    \begin{aligned}
    J_T(W)
    ={}&C_0'+
    \frac{B_0}
    {\left(K_0/2+\int_0^T\eta(t)\d t\right)^{\alpha_0+2}}\\
    &+V_0\int_0^T
      \frac{\eta(s)^2}
      {\left(
        (K_0+G_0)/2+\int_s^T\eta(t)\d t
      \right)^{\alpha_0+\rho_0+2}}
    \d s \\
    &+
    \frac{B_1}
    {\left(K_1/2+\int_0^T\eta(t)\gamma(t)\d t\right)^{\alpha_1+2}}\\
    &+V_1\int_0^T
      \frac{\eta(s)^2}
      {\left(
        (K_1+G_1)/2+\int_s^T\eta(t)\gamma(t)\d t
      \right)^{\alpha_1+\rho_1+2}}
    \d s.
    \end{aligned}
    \label{eq:app-mpl-final-loss-fix-ds}
\end{equation}
Furthermore, we define some additional notation for simplicity. We write
\begin{equation}
    A_W(u):=\int_0^u\eta_W(T-v)\d v
    =\begin{cases}
        mu+\Delta u^2/(2W), & 0\leq u<W,\\
        Mu-\Delta W/2, & W\leq u\leq T.
    \end{cases}
    \label{eq:lr-cumulative}
\end{equation}
The quantity $A_W(u)$ captures the cumulative effect from time $T-u$ to $T$.
For $0<H\leq T$, $\beta>1$, and $g>0$, write
\begin{equation}
    I_{\beta,g}(H,W):=
        \int_0^H\frac{\eta_W(T-u)^2}{[g+A_W(u)]^\beta}\d u,
    \qquad
    P(W):=\frac{\displaystyle\int_L^T\eta_W(T-u)^2\d u}
                   {[g_1+A_W(L)]^{b_1}}.
    \label{eq:lr-variance-integrals}
\end{equation}
Using this notation, we can rewrite the minimization objective as
\begin{align}
    J_T(W)
    =C_0'+{}&\frac{B_0}{[K_0/2+A_W(T)]^{a_0}}
       +\frac{B_1}{[K_1/2+A_W(L)]^{a_1}} \notag\\
      &+V_0 I_{b_0,g_0}(T,W)
       +V_1 I_{b_1,g_1}(L,W)+V_1 P(W).
    \label{eq:lr-terminal-loss}
\end{align}

\subsection{Auxiliary Lemmas}
\label{app:lr-lemmas}

We first give a lemma on the expression of the budget threshold $\alpha_c$. For $1<b<2$, define $r:=m/M\in(0,1)$ and
$\kappa:=(b-1)/b\in(0,1)$. We define some additional quantities that will be used in the main theorem and its proof:
\begin{align}
    \delta_b(r)
    &:={\min}\left\{
          \left(\frac{1+r}{2}\right)^{-b},
          \frac{1+r+r^2}{3r^b}
        \right\}
        -\left[1-\frac{(1-r)\kappa}{2}\right]^{-b},
        \label{eq:def-delta}\\
    H_b(r)&:=r^{-b}+\frac{b(1-r)(1+r+r^2)}{6r^{b+1}}, \label{eq:def-h}\\
    K_b(r)&:=\frac{(1-r)r^{1-b}\kappa^{1-b}}{(b-1)(2-b)}
               +\frac{r^{2-b}}{b-1},
        \label{eq:def-k}\\
    \chi&:=\begin{cases}
        V_0/V_1, & \text{if the shared block is present and }b_0=b,\\
        0, & \text{otherwise},
    \end{cases}\notag\\
    \alpha_c&:=\min\left\{
        \frac12,\,
        \frac{\delta_b(r)}{2[H_b(r)+(1+\chi)K_b(r)]}
    \right\}.
    \label{eq:lr-budget-threshold}
\end{align}
\begin{lemma}[Positivity of the budget threshold]
For every $1<b<2$ and $0<r<1$, the quantity $\delta_b(r)$ defined in
\Cref{eq:def-delta} is strictly positive. Consequently, the budget
threshold $\alpha_c$ defined in \Cref{eq:lr-budget-threshold} is strictly positive.
\end{lemma}
\begin{proof}
To prove that $\delta_b(r)>0$, we show that $\left(\frac{1+r}{2}\right)^{-b}-\left[1-\frac{(1-r)\kappa}{2}\right]^{-b}>0$ and $\frac{1+r+r^2}{3r^b}-\left[1-\frac{(1-r)\kappa}{2}\right]^{-b}>0$ separately. We first show that $\left(\frac{1+r}{2}\right)^{-b}-\left[1-\frac{(1-r)\kappa}{2}\right]^{-b}>0$. Notice that when $0<r<1$ and $\kappa<1$, we have
\begin{equation*}
    1-\frac{(1-r)\kappa}{2}-\frac{1+r}{2}
       =\frac{(1-r)(1-\kappa)}{2}>0.
\end{equation*}
Since $b>0$, this implies
\begin{equation*}
    \left(\frac{1+r}{2}\right)^{-b}
       >\left[1-\frac{(1-r)\kappa}{2}\right]^{-b}.
\end{equation*}

Next we prove that $\frac{1+r+r^2}{3r^b}-\left[1-\frac{(1-r)\kappa}{2}\right]^{-b}>0$. We use the weighted arithmetic--geometric mean
inequality: for positive numbers $x,y$ and a weight $\lambda\in[0,1]$,
it states that
\begin{equation*}
    (1-\lambda)x+\lambda y\geq x^{1-\lambda}y^\lambda.
\end{equation*}
This inequality follows directly from the concavity of the logarithm:
\begin{equation*}
    \log\bigl((1-\lambda)x+\lambda y\bigr)
       \geq(1-\lambda)\log x+\lambda\log y.
\end{equation*}
We choose $x=1$, $y=r$, and $\lambda=\kappa/2\in(0,1)$, and by the arithmetic--geometric mean
inequality, we have
\begin{equation*}
    1-\frac{(1-r)\kappa}{2}
       =\left(1-\frac{\kappa}{2}\right)\cdot1
          +\frac{\kappa}{2}\cdot r
       \geq 1^{1-\kappa/2}r^{\kappa/2}=r^{\kappa/2}.
\end{equation*}
Since $b\kappa=b-1$, this yields
\begin{equation*}
    \left[1-\frac{(1-r)\kappa}{2}\right]^b
       \geq r^{b\kappa/2}=r^{(b-1)/2}>r^{b-1}.
\end{equation*}
The last inequality holds because $0<r<1$ and
$0<(b-1)/2<b-1$. Consequently,
\begin{equation*}
    \left[1-\frac{(1-r)\kappa}{2}\right]^{-b}<r^{1-b}.
\end{equation*}
On the other hand,
\begin{equation*}
    \frac{1+r+r^2}{3}-r=\frac{(1-r)^2}{3}>0.
\end{equation*}
We combine the two inequalities above and get that
\begin{equation*}
    \frac{1+r+r^2}{3r^b}
       >\frac{r}{r^b}=r^{1-b}
       >\left[1-\frac{(1-r)\kappa}{2}\right]^{-b}.
\end{equation*}
Therefore, we prove that $\delta_b(r)>0$.
Since $H_b(r)>0$, $K_b(r)>0$, and
$\chi\geq0$ by their definitions, $\delta_b(r)>0$ implies that $\alpha_c>0$.
\end{proof}

Next, we give three lemmas on the properties of some terms appearing in \Cref{eq:lr-terminal-loss}, namely $A_W(u)$, $I_{\beta, g}(H,W)$, and $P(W)$.
\begin{lemma}
\label{lem:lr-variance-monotonicity}
    Fix $T>0$, $0<H\leq T$, $\beta>1$, and $u>0$.
    Then $I_{\beta,g}(H,W)$ and $A_W(u)$ are nonincreasing in $W\in(0,T]$.
\end{lemma}
\begin{proof}

We calculate $I_{\beta,g}(H,W)$ by changing the variable to $x=A_W(u)=\int_{0}^{u} \eta_W(T-v) \, dv$. Therefore, we have
\begin{equation*}
    \d x=\eta_W(T-u)\d u,
    \qquad
    \eta_W(T-u)^2\d u
       =\eta_W\bigl(T-A_W^{-1}(x)\bigr)\d x.
\end{equation*}
We substitute the above equations into
\Cref{eq:lr-variance-integrals}, and get
\begin{align*}
    I_{\beta,g}(H,W)
    &=\int_0^{A_W(H)}
       \frac{\eta_W\bigl(T-A_W^{-1}(x)\bigr)^2}{(g+x)^\beta}
       \frac{\d x}{\eta_W\bigl(T-A_W^{-1}(x)\bigr)}\\
    &=\int_0^{A_W(H)}
       \eta_W\bigl(T-A_W^{-1}(x)\bigr)(g+x)^{-\beta}\d x.
\end{align*}
Next, to calculate $I_{\beta,g}(H,W)$, we need to calculate $A_W^{-1}(x)$. When $0\leq u<W$, we have that $x=mu+\Delta u^2/(2W)$, and thus $0\leq x\leq A_W(W)=\frac{(M+m)W}{2}$. A direct calculation gives
\begin{equation*}
    \frac{2\Delta x}{W}=\frac{2m\Delta u}{W}
                              +\frac{\Delta^2u^2}{W^2},
    \qquad
    m^2+\frac{2\Delta x}{W}=\left(m+\frac{\Delta u}{W}\right)^2.
\end{equation*}
Therefore, $u=(W/\Delta)(\sqrt{m^2+2\Delta x/W}-m)$.
When $W\leq u\leq T$, $x=Mu-\Delta W/2$, so $u=(x+\Delta W/2)/M$.
Thus
\begin{equation*}
    A_W^{-1}(x)=
    \begin{cases}
        \displaystyle\frac{W}{\Delta}
            \left(\sqrt{m^2+\frac{2\Delta x}{W}}-m\right),
            & 0\leq x\leq A_W(W),\\[6pt]
        \displaystyle\frac{x+\Delta W/2}{M},
            & A_W(W)\leq x\leq A_W(T).
    \end{cases}
\end{equation*}
We substitute the expression of $A_W^{-1}(x)$ into the expression of the learning rate, and get
\begin{equation*}
\begin{aligned}
    \eta_W\bigl(T-A_W^{-1}(x)\bigr)
    &=
    \begin{cases}
        \displaystyle \sqrt{m^2+\frac{2\Delta x}{W}},
            & 0\leq x\leq A_W(W),\\[4pt]
        M,  & A_W(W)\leq x\leq A_W(T),
    \end{cases}\\[6pt]
    &= \min\left\{M,\sqrt{m^2+\frac{2\Delta x}{W}}\right\}.
\end{aligned}
\end{equation*}

For simplicity, we define
\begin{equation}
    \begin{aligned}
        \xi_W(x)&:=\min\left\{M,\sqrt{m^2+\frac{2\Delta x}{W}}\right\},
    \end{aligned}
    \label{eq:lr-cumulative-integral}
\end{equation}
and rewrite $I_{\beta,g}(H,W)$ as
\begin{equation*}
    I_{\beta,g}(H,W)=\int_0^{A_W(H)}\xi_W(x)(g+x)^{-\beta}\d x.
\end{equation*}
We now prove that $I_{\beta,g}(H,W)$ is nonincreasing in $W$. Given $0<W_1<W_2\leq T$, we need to show that $I_{\beta,g}(H,W_1)\geq I_{\beta,g}(H,W_2)$. We calculate the difference of the two integrals:
\begin{align*}
    I_{\beta,g}(H,W_1)&-I_{\beta,g}(H,W_2)
    =\int_0^{A_{W_1}(H)}\xi_{W_1}(x)(g+x)^{-\beta}\d x
      -\int_0^{A_{W_2}(H)}\xi_{W_2}(x)(g+x)^{-\beta}\d x\\
    &=\int_0^{A_{W_2}(H)}
       [\xi_{W_1}(x)-\xi_{W_2}(x)](g+x)^{-\beta}\d x+\int_{A_{W_2}(H)}^{A_{W_1}(H)}
       \xi_{W_1}(x)(g+x)^{-\beta}\d x.
\end{align*}
For each fixed $x\geq0$, by \Cref{eq:lr-cumulative-integral}, $\xi_W(x)$ is nonincreasing in $W$.
For each fixed $u>0$, direct differentiation of
\Cref{eq:lr-cumulative} gives
\begin{equation*}
    \frac{\partial A_W(u)}{\partial W}=
    \begin{cases}
        -\Delta/2, & 0<W\leq u,\\
        -\Delta u^2/(2W^2), & u\leq W\leq T,
    \end{cases}
\end{equation*}
which is negative. Hence, $A_W(u)$ is nonincreasing in $W$. Combining these two observations, we have $\xi_{W_1}(x)\geq\xi_{W_2}(x)$ for all $x\geq0$ and $A_{W_1}(H)\geq A_{W_2}(H)$. Therefore, both integrals in the last expression are nonnegative, which proves that $I_{\beta,g}(H,W)$ is nonincreasing in $W$.

Finally, we prove that the bias term in \Cref{eq:lr-terminal-loss} is nondecreasing in $W$. This follows directly since
$B_l\geq0$ for all $l$, $a_l>0$, and $A_W(H)$ decreases
with $W$.
\end{proof}

\begin{lemma}
\label{lem:lr-residual-endpoints}
    If $b_1>1$, then
    \begin{equation*}
        \min_{L\leq W\leq T}P(W)=\min\{P(L),P(T)\}.
    \end{equation*}
\end{lemma}

\begin{proof}
For simplicity, we define some additional notation. We define $z=L/W\in[\alpha,1]$, and consider $W$ as a function of $z$. We further define $N(z)=\displaystyle\int_L^T\eta_W(T-u)^2\d u$, $Y(z)=g_1+A_W(L)$ and $b=b_1$. Then by definition,
$P(W)=P(L/z)=N(z)/Y(z)^b$, and $A_W(L)=L(m+\Delta z/2)$.

To prove our claim, we calculate the expressions for $N(z)$ and $Y(z)$. By definition, $Y(z):=g_1+L(m+\Delta z/2)$, and
\begin{align*}
    N(z)
    &:=\int_L^T\eta_{L/z}(T-u)^2\d u\\
    &=\int_L^{L/z}\left(m+\frac{\Delta zu}{L}\right)^2\d u
               +M^2(T-L/z)\\
    &=\int_L^{L/z}\left(m^2+\frac{2m\Delta z}{L}u
                            +\frac{\Delta^2z^2}{L^2}u^2\right)\d u
               +M^2(T-L/z)\\
    &=\left[m^2u+\frac{m\Delta z}{L}u^2
                       +\frac{\Delta^2z^2}{3L^2}u^3\right]_{u=L}^{u=L/z}
               +M^2(T-L/z)\\
    &=m^2L\left(\frac1z-1\right)
       +m\Delta L\left(\frac1z-z\right)\\
    &\quad+\frac{\Delta^2L}{3}\left(\frac1z-z^2\right)
       +M^2\left(T-\frac{L}{z}\right)\\
    &=M^2T+\frac{L}{z}\left(m^2+m\Delta+\frac{\Delta^2}{3}-M^2\right)
       -L\left(m^2+m\Delta z+\frac{\Delta^2z^2}{3}\right).
\end{align*}
We substitute $M=m+\Delta$ into the coefficient of $L/z$ to simplify its expression:
\begin{align*}
    M^2-m^2-m\Delta-\frac{\Delta^2}{3}
       &=(m+\Delta)^2-m^2-m\Delta-\frac{\Delta^2}{3}\\
       &=m\Delta+\frac{2\Delta^2}{3}
        =\frac{\Delta(3m+2\Delta)}{3}
        =\frac{\Delta(2M+m)}{3}.
\end{align*}
We substitute this identity into the expression of $N(z)$ and get
\begin{equation*}
    N(z)=M^2T-L\left[
         \frac{\Delta(2M+m)}{3z}+m^2+m\Delta z+\frac{\Delta^2z^2}{3}
       \right].
\end{equation*}

We observe that both $N(z)$ and $Y(z)$ are nonnegative, since $N(z)=\int_L^T\eta_{L/z}(T-u)^2\d u$, and $Y(z)=g_1+L(m+\Delta z/2)$. Furthermore, $N(z)$ is concave in $z$, because we have
\begin{equation*}
    N''(z)=-\frac{2L\Delta(2M+m)}{3z^3}-\frac{2L\Delta^2}{3}<0
\end{equation*}
by direct calculation. The function $Y(z)$ is positive and affine, so $Y^b$ is convex.
Let $p:=\min\{N(\alpha)/Y(\alpha)^b,N(1)/Y(1)^b\}\geq0$.
For $z=(1-\lambda)\alpha+\lambda \cdot 1$ with $0\leq\lambda\leq1$, by the concavity
of $N$ and convexity of $Y^b$, we have
\begin{align*}
    N(z)&\geq(1-\lambda)N(\alpha)+\lambda N(1)\\
        &\geq p\bigl[(1-\lambda)Y(\alpha)^b+\lambda Y(1)^b\bigr]
         \geq pY(z)^b.
\end{align*}
Thus $P(L/z)\geq p$, and we conclude our proof.
\end{proof}

\begin{lemma}
\label{lem:lr-variance-bound}
    Fix $0<\kappa<1$, let $W^\circ=\kappa L$, and suppose $L\leq H\leq T$.
    For $\beta>1$ and $g>0$,
    \begin{align}
        0\leq{}&I_{\beta,g}(H,W^\circ)-I_{\beta,g}(H,T)\notag\\
        \leq{}&\int_0^\infty\Delta\min\left\{1,\frac{x}{mW^\circ}\right\}(g+x)^{-\beta}\d x+
            \frac{m}{\beta-1}[g+A_T(H)]^{1-\beta}. \label{eq:lr-variance-tail-bound}
     \end{align}
    If $1<\beta<2$, this implies
    \begin{equation}
        I_{\beta,g}(H,W^\circ)-I_{\beta,g}(H,T)
        \leq\frac{\Delta(mW^\circ)^{1-\beta}}{(\beta-1)(2-\beta)}
                 +\frac{m(mH)^{1-\beta}}{\beta-1}.
        \label{eq:lr-variance-power-bound}
    \end{equation}
    When $T\to\infty$, $I_{\beta,g}(H,W^\circ)-I_{\beta,g}(H,T)$ can be bounded by
    \begin{equation}
        \begin{cases}
            O(T^{1-\beta}), & 1<\beta<2,\\
            O(T^{-1}\log T), & \beta=2,\\
            O(T^{-1}), & \beta>2.
        \end{cases}
        \label{eq:lr-variance-rates}
    \end{equation}
\end{lemma}

\begin{proof}
Since $W^\circ=\kappa L<T$, \Cref{lem:lr-variance-monotonicity} gives
both $I_{\beta,g}(H,W^\circ)-I_{\beta,g}(H,T)\geq0$ and
$A_{W^\circ}(H)\geq A_T(H)$. We next derive the upper bound. For either $W=W^\circ$ or $W=T$, we first split $I_{\beta,g}(H,W)$ using
$\xi_W(x)=m+[\xi_W(x)-m]$:
\begin{equation*}
    I_{\beta,g}(H,W)
       =m\int_0^{A_W(H)}(g+x)^{-\beta}\d x
        +\int_0^{A_W(H)}[\xi_W(x)-m](g+x)^{-\beta}\d x.
\end{equation*}
Then we subtract the two decompositions and get
\begin{align*}
    &I_{\beta,g}(H,W^\circ)-I_{\beta,g}(H,T)\\
    &\quad=\int_0^{A_{W^\circ}(H)}
            [\xi_{W^\circ}(x)-m](g+x)^{-\beta}\d x\\
    &\qquad\quad-\int_0^{A_T(H)}
            [\xi_T(x)-m](g+x)^{-\beta}\d x
          +m\int_{A_T(H)}^{A_{W^\circ}(H)}(g+x)^{-\beta}\d x \\
    &\quad\leq \int_0^{A_{W^\circ}(H)}
            [\xi_{W^\circ}(x)-m](g+x)^{-\beta}\d x + m\int_{A_T(H)}^{A_{W^\circ}(H)}(g+x)^{-\beta}\d x.
\end{align*}
The inequality uses the fact that $\xi_{W}(x)-m\geq0$ for all $x\geq0$ and for all $W$, and the fact that $A_{W^\circ}(H)\geq A_T(H)$.

We next bound the two integrals separately. For the second integral,
since $\beta>1$, we have
\begin{align*}
    \int_{A_T(H)}^{A_{W^\circ}(H)}(g+x)^{-\beta}\d x
       &\leq\int_{A_T(H)}^\infty(g+x)^{-\beta}\d x\\
       &=\left[\frac{(g+x)^{1-\beta}}{1-\beta}
          \right]_{x=A_T(H)}^{x=\infty}
        =\frac{[g+A_T(H)]^{1-\beta}}{\beta-1}.
\end{align*}
Here the upper-endpoint value is zero because $1-\beta<0$.
Combining these two bounds proves \Cref{eq:lr-variance-tail-bound}.

We now bound $\int_0^{A_{W^\circ}(H)}
            [\xi_{W^\circ}(x)-m](g+x)^{-\beta}\d x.$
By the definition of $\xi_{W^\circ}$ in \Cref{eq:lr-cumulative-integral}, we have
\begin{equation*}
    \xi_{W^\circ}(x)-m
       =\min\left\{\Delta,
                    \sqrt{m^2+\frac{2\Delta x}{W^\circ}}-m\right\}.
\end{equation*}
Note that
\begin{align*}
    \sqrt{m^2+\frac{2\Delta x}{W^\circ}}-m
       &=\frac{\left(\sqrt{m^2+\frac{2\Delta x}{W^\circ}}-m\right)
                \left(\sqrt{m^2+\frac{2\Delta x}{W^\circ}}+m\right)}
               {\sqrt{m^2+\frac{2\Delta x}{W^\circ}}+m}\\
       &=\frac{m^2+2\Delta x/W^\circ-m^2}
               {\sqrt{m^2+2\Delta x/W^\circ}+m}
        =\frac{2\Delta x/W^\circ}
               {\sqrt{m^2+2\Delta x/W^\circ}+m}\\
       &\leq\frac{2\Delta x/W^\circ}{2m}
        =\frac{\Delta x}{mW^\circ}.
\end{align*}
Therefore
\begin{equation*}
    0\leq\xi_{W^\circ}(x)-m
       \leq\Delta\min\left\{1,\frac{x}{mW^\circ}\right\},
\end{equation*}
and we conclude the first part of the proof.

To conclude the second part of the proof, we further calculate a bound on $\int_0^\infty\Delta\min\left\{1,\frac{x}{mW^\circ}\right\}(g+x)^{-\beta}\d x$. For $1<\beta<2$, we have
\begin{align*}
    &\int_0^\infty\min\{1,x/a\}x^{-\beta}\d x
       =\int_0^a\frac{x}{a}x^{-\beta}\d x
                              +\int_a^\infty x^{-\beta}\d x
       =\frac1a\int_0^a x^{1-\beta}\d x
                              +\int_a^\infty x^{-\beta}\d x \\
       &=a^{1-\beta}\left(\frac1{2-\beta}+\frac1{\beta-1}\right)
       =a^{1-\beta}
           \frac{(\beta-1)+(2-\beta)}{(\beta-1)(2-\beta)}
       =\frac{a^{1-\beta}}{(\beta-1)(2-\beta)}.
\end{align*}
We now apply this identity with $a=mW^\circ$, and get that
\begin{align*}
    \int_0^\infty\Delta\min\left\{1,\frac{x}{mW^\circ}\right\}(g+x)^{-\beta}\d x
       \leq\int_0^\infty\Delta\min\left\{1,\frac{x}{mW^\circ}\right\}x^{-\beta}\d x
       =\frac{\Delta(mW^\circ)^{1-\beta}}{(\beta-1)(2-\beta)}.
\end{align*}
Also,
we have $[g+A_T(H)]^{1-\beta}\leq(mH)^{1-\beta}$ since $g+A_T(H)\geq mH>0$ and $1-\beta<0$. We substitute the above two upper bounds into \Cref{eq:lr-variance-tail-bound} and get \Cref{eq:lr-variance-power-bound}.

Evaluating this last tail, and using $A_T(H)\geq mH$ for the tail in
\Cref{eq:lr-variance-tail-bound}, gives the explicit bound
\begin{align*}
    &I_{\beta,g}(H,W^\circ)-I_{\beta,g}(H,T) \\
       \leq{}&\frac{\Delta}{mW^\circ}
\int_0^{mW^\circ}x(g+x)^{-\beta}\d x+\Delta\int_{mW^\circ}^{\infty}(g+x)^{-\beta}\d x+\frac{m}{\beta-1}(g+mH)^{1-\beta}\\
    \leq{}&\frac{\Delta}{mW^\circ}
\int_0^{mW^\circ}x(g+x)^{-\beta}\d x+\frac{\Delta}{\beta-1}(g+mW^\circ)^{1-\beta}
              +\frac{m}{\beta-1}(g+mH)^{1-\beta}.
\end{align*}
We now analyze the first integral for $\beta=2$ and $\beta>2$ separately.
For $\beta=2$, we have
\begin{align*}
    &\int_0^{mW^\circ}x(g+x)^{-2}\d x
       =\int_g^{g+mW^\circ}(y^{-1}-gy^{-2})\d y
       =\left[\log y+\frac{g}{y}\right]_{y=g}^{y=g+mW^\circ}\\
       &=\log\left(1+\frac{mW^\circ}{g}\right)
                       +\frac{g}{g+mW^\circ}-1
       \leq\log\left(1+\frac{mW^\circ}{g}\right).
\end{align*}
For $\beta>2$, we have instead
\begin{align*}
    \int_0^{mW^\circ}x(g+x)^{-\beta}\d x
       &\leq\int_0^\infty x(g+x)^{-\beta}\d x
       =\int_g^\infty(y^{1-\beta}-gy^{-\beta})\d y\\
       &=\frac{g^{2-\beta}}{\beta-2}
                      -\frac{g^{2-\beta}}{\beta-1}
        =\frac{g^{2-\beta}}{(\beta-1)(\beta-2)}.
\end{align*}
Thus, with $W^\circ=\kappa\alpha T$ and fixed positive $\alpha,\kappa$,
the first term in the explicit bound is $O(T^{-1}\log T)$ when
$\beta=2$ and $O(T^{-1})$ when $\beta>2$.
The second term is $O(T^{1-\beta})$, and the third has the same rate
because $H\geq\alpha T$. For $\beta=2$ these two terms are $O(T^{-1})$;
for $\beta>2$ they decay faster than $T^{-1}$.
We combine these observations with
\Cref{eq:lr-variance-power-bound} and prove \Cref{eq:lr-variance-rates}.
\end{proof}

\subsection{\texorpdfstring{Proof of \Cref{thm:lr-delayed-decay-formal}}{Proof of Theorem \ref*{thm:lr-delayed-decay-formal}}}
\label{app:lr-main-proof}
\begin{proof}[Proof of \Cref{thm:lr-delayed-decay-formal}]
Throughout the proof, $b:=b_1\in(1,2)$, $\kappa=(b-1)/b$, and
$W^\circ=\kappa L$.

\paragraph{Step 1: $J_T(W)$ attains its minimum on $[0,T]$.}
We only need to show that the function $J_T(W)$ is continuous on $[0,T]$. Since
\begin{equation*}
    m\leq \eta_W(T-u)\leq M, \ \text{and} \ 0\leq \frac{\eta_W(T-u)^2}{[g+A_W(u)]^\beta} \leq M^2g^{-\beta},
\end{equation*}
by \Cref{eq:lr-variance-integrals} and \Cref{eq:lr-terminal-loss}, $J_T(W)$ only depends on the integrals of bounded continuous functions of $\eta_W(u)$. Thus we only need to show that all the integrals of bounded continuous functions of $\eta_W(u)$ are continuous in $W$. By the dominated convergence theorem, it suffices to show that for all $W\in[0,T]$, $\eta_W(u)$ is continuous in $W$ for almost every $u\in[0,T]$.

Fix $T$ and let $\{W_n\}_{n\geq 1}$ be any sequence in $[0,T]$. Assume that $\{W_n\}$ has a limit $W\in[0,T]$. Note that when $W\neq 0$, $\eta_{W_n}(T-u)\to\eta_W(T-u)$ for all $u\in[0,T]$, and when $W=0$, $\eta_{W_n}(T-u)\to\eta_0(T-u)$ when $u\neq 0$. This concludes the first part of the proof.

\paragraph{Step 2: $J_T(W)$ cannot attain its minimum on $[L,T]$.}
We take the reference point $W^\circ=\kappa L$ where $\kappa=(b-1)/b$, and show that for all $W\in[L,T]$, $J_T(W)>J_T(W^\circ)$. By \Cref{eq:lr-terminal-loss}, we have
\begin{equation}
\begin{aligned}
    J_T(W)-J_T(W^\circ)
    &= \left(\frac{B_0}{[K_0/2+A_W(T)]^{a_0}}-\frac{B_0}{[K_0/2+A_{W^\circ}(T)]^{a_0}}\right) \\
       &+\left(\frac{B_1}{[K_1/2+A_W(L)]^{a_1}}-\frac{B_1}{[K_1/2+A_{W^\circ}(L)]^{a_1}}\right) \\
       &+V_1\left[P(W)-P(W^\circ)\right]+V_1\left[I_{b_1,g_1}(L,W)-I_{b_1,g_1}(L,W^\circ)\right]\\
    &+V_0\left[I_{b_0,g_0}(T,W)-I_{b_0,g_0}(T,W^\circ)\right].
\end{aligned}
\label{eq:lr-loss-difference}
\end{equation}
Note that $A_W(T)\leq A_{W^\circ}(T)$ and $A_W(L)\leq A_{W^\circ}(L)$ for all $W$ by \Cref{lem:lr-variance-monotonicity}, and by \Cref{lem:lr-residual-endpoints}, we have $P(W)\geq\min\{P(L),P(T)\}$.
So
\begin{equation*}
    J_T(W)-J_T(W^\circ)\geq V_1\left[\min\{P(L),P(T)\}-P(W^\circ)-D_1
                        -\frac{V_0}{V_1}D_0\right],
\end{equation*}
where $D_1:=I_{b_1,g_1}(L,W^\circ)-I_{b_1,g_1}(L,T), D_0:=I_{b_0,g_0}(T,W^\circ)-I_{b_0,g_0}(T,T)
$. Therefore, the second part of the proof is complete if we can show that $\min\{P(L),P(T)\}-P(W^\circ)-D_1
                        -\frac{V_0}{V_1}D_0>0$ for all sufficiently large $T$. To prove this, we calculate $P(L)$, $P(T)$, $P(W^\circ)$, $D_1$ and $D_0$ in turn.
We define
\begin{equation*}
    S_T:=\frac{M^2T}{(ML)^b}=M^{2-b}\alpha^{-b}T^{1-b}.
\end{equation*}
We first compute $P(L)$, $P(T)$ and $P(W^\circ)$.
For $W=L$, we have $\eta_{W^\circ}(T-u)=M$ for all $u\in[L,T]$, so the numerator of $P(L)$ is given by
\begin{equation*}
    \int_L^T\eta_L(T-u)^2\d u=M^2(T-L)=M^2T(1-\alpha),
\end{equation*}
and by \Cref{eq:lr-cumulative}, the denominator of $P(L)$ is
\begin{equation*}
    \left(g_1+A_L(L)\right)^b=\left(g_1+ML-\frac{\Delta L}{2}\right)^b=\left(g_1+ML\frac{1+r}{2}\right)^b.
\end{equation*}
The numerator for $P(W^\circ)$ is the same as that of $P(L)$, since again
$\eta_{W^\circ}(T-u)=M$ for all $u\in[L,T]$. As for the denominator of $P(W^\circ)$, we have
\begin{equation*}
    \left(g_1+A_{W^\circ}(L)\right)^b=\left(g_1+ML-\frac{\Delta W^\circ}{2}\right)^b
       =\left(g_1+ML\left[1-\frac{(1-r)\kappa}{2}\right]\right)^b.
\end{equation*}
For $W=T$, we have
\begin{equation*}
    \int_L^T\eta_T(T-u)^2\d u
       =\int_L^T\left(m+\frac{\Delta u}{T}\right)^2\d u
       =M^2T\int_\alpha^1[r+(1-r)v]^2\d v.
\end{equation*}
We denote $n(\alpha)=\int_\alpha^1[r+(1-r)v]^2\d v$ for simplicity. Through a direct calculation, we have
\begin{align*}
    n(\alpha)
       &=\left[r^2v+r(1-r)v^2+\frac{(1-r)^2v^3}{3}
          \right]_{v=\alpha}^{v=1}\\
       &=r^2(1-\alpha)+r(1-r)(1-\alpha^2)
                          +\frac{(1-r)^2}{3}(1-\alpha^3)\\
       &=\frac{1+r+r^2}{3}-r^2\alpha-r(1-r)\alpha^2
                          -\frac{(1-r)^2\alpha^3}{3}.
\end{align*}
The last equality uses
$r^2+r(1-r)+(1-r)^2/3=(1+r+r^2)/3$.
Also,
\begin{equation*}
    \left(g_1+A_T(L)\right)^b=\left(g_1+ML\left[r+\frac{(1-r)\alpha}{2}\right]\right)^b.
\end{equation*}
Combining the above calculations, we have
\begin{align*}
    \frac{P(L)}{S_T}
       &=\frac{1-\alpha}{[g_1/(ML)+(1+r)/2]^b},\\
    \frac{P(T)}{S_T}
       &=\frac{n(\alpha)}{[g_1/(ML)+r+(1-r)\alpha/2]^b},\\
    \frac{P(W^\circ)}{S_T}
       &=\frac{1-\alpha}{[g_1/(ML)+1-(1-r)\kappa/2]^b},
\end{align*}
and consequently,
\begin{equation}
    \frac{\min\{P(L),P(T)\}-P(W^\circ)}{S_T}
      \longrightarrow G_b(\alpha,r),
      \label{eq:lr-residual-gap-limit}
\end{equation}
where
\begin{align*}
    G_b(\alpha,r):=\min\left\{
          (1-\alpha)\left(\frac{1+r}{2}\right)^{-b}, \frac{n(\alpha)}
                {[r+(1-r)\alpha/2]^b}\right\}-(1-\alpha)\left[1-\frac{(1-r)\kappa}{2}\right]^{-b}.
\end{align*}
A direct observation is that
$G_b(0,r)=\delta_b(r)$. We next control the difference $G_b(\alpha,r)-G_b(0,r)$ for sufficiently small $\alpha$.
Let $f(\alpha)$ denote the second
entry in the minimum:
\begin{equation*}
    f(\alpha)=n(\alpha)[r+(1-r)\alpha/2]^{-b}.
\end{equation*}
Through a direct calculation, we have
\begin{align*}
    n'(\alpha)
       &=-r^2-2r(1-r)\alpha-(1-r)^2\alpha^2\\
       &=-[r+(1-r)\alpha]^2.
\end{align*}
Thus $n(\alpha)$ is nonincreasing on $[0,1]$. Since $n(0)=(1+r+r^2)/3$ and
$n(1)=0$, we have
\begin{equation*}
    0\leq n(\alpha)\leq\frac{1+r+r^2}{3},
    \qquad 0\leq\alpha\leq1.
\end{equation*}
We next differentiate $f(\alpha)$, substitute the derivative of $n(\alpha)$, and get
\begin{align*}
    f'(\alpha)
       &=n'(\alpha)[r+(1-r)\alpha/2]^{-b}-\frac{b(1-r)}{2}n(\alpha)
                    [r+(1-r)\alpha/2]^{-b-1}\\
       &=-\frac{[r+(1-r)\alpha]^2}{[r+(1-r)\alpha/2]^b}
         -\frac{b(1-r)n(\alpha)}{2[r+(1-r)\alpha/2]^{b+1}}.
\end{align*}
Note that both terms in the last line are nonpositive. Consequently,
\begin{align*}
    |f'(\alpha)|
       &=\frac{[r+(1-r)\alpha]^2}{[r+(1-r)\alpha/2]^b}
         +\frac{b(1-r)n(\alpha)}{2[r+(1-r)\alpha/2]^{b+1}}\leq\frac{1}{r^b}
          +\frac{b(1-r)}{2r^{b+1}}\frac{1+r+r^2}{3}\\
       &=r^{-b}+\frac{b(1-r)(1+r+r^2)}{6r^{b+1}}=H_b(r),
\end{align*}
where the inequality holds since
$r\leq r+(1-r)\alpha\leq1$, $r+(1-r)\alpha/2\geq r$, and $0\leq n(\alpha)\leq(1+r+r^2)/3$. Recall that $H_b(r)$ is defined as in \Cref{eq:def-h}.
In particular, $f'(\alpha)\geq-H_b(r)$ for every $\alpha\in[0,1]$. Therefore, by the fundamental theorem of calculus, we have
\begin{equation*}
    f(\alpha)-f(0)=\int_0^\alpha f'(s)\d s
       \geq-\alpha H_b(r).
\end{equation*}
The first entry satisfies the same lower bound, because
$((1+r)/2)^{-b}\leq r^{-b}\leq H_b(r)$:
\begin{equation*}
    (1-\alpha)\left(\frac{1+r}{2}\right)^{-b}
       \geq\left(\frac{1+r}{2}\right)^{-b}-\alpha H_b(r).
\end{equation*}
We apply the lower bounds of
$(1-\alpha)((1+r)/2)^{-b}$ and $f(\alpha)$ and obtain the following
lower bound:
\begin{align*}
    \min\left\{(1-\alpha)\left(\frac{1+r}{2}\right)^{-b},
                    f(\alpha)\right\}\geq\min\left\{\left(\frac{1+r}{2}\right)^{-b},f(0)\right\}
                       -\alpha H_b(r).
\end{align*}
Finally, note that
\begin{equation*}
    -(1-\alpha)\left[1-\frac{(1-r)\kappa}{2}\right]^{-b}
       \geq-\left[1-\frac{(1-r)\kappa}{2}\right]^{-b}.
\end{equation*}
We combine the two inequalities above, and therefore, we have
\begin{equation}
    G_b(\alpha,r)\geq\delta_b(r)-H_b(r)\alpha,
    \qquad 0\leq\alpha\leq1.
    \label{eq:lr-residual-gap-bound}
\end{equation}
Here, we recall that $f(0)=(1+r+r^2)/(3r^b)$, and $\delta_b(r)$ is defined as in \Cref{eq:def-delta}.

We next bound $D_0$ and $D_1$. For $D_1$, we take $H=L$ in \Cref{eq:lr-variance-power-bound}, and get that
\begin{align*}
    D_1
       &\leq\frac{\Delta(mW^\circ)^{1-b}}{(b-1)(2-b)}
                         +\frac{m(mL)^{1-b}}{b-1}\\
       &=\frac{(1-r)M(rM\kappa\alpha T)^{1-b}}{(b-1)(2-b)}
                         +\frac{rM(rM\alpha T)^{1-b}}{b-1}\\
       &=M^{2-b}\alpha^{1-b}T^{1-b}
          \left[\frac{(1-r)r^{1-b}\kappa^{1-b}}{(b-1)(2-b)}
                         +\frac{r^{2-b}}{b-1}\right]\\
       &=\alpha S_T K_b(r).
\end{align*}
Here we used $S_T=M^{2-b}\alpha^{-b}T^{1-b}$, and $K_b(r)$ is defined as in \Cref{eq:def-k}.

For $D_0$, we distinguish
cases according to the value of $b_0$ relative to $b$.
When $b_0=b$, we apply
\Cref{eq:lr-variance-power-bound} with $H=T$ instead:
\begin{align*}
    D_0
       &\leq\frac{(1-r)M(rM\kappa\alpha T)^{1-b}}{(b-1)(2-b)}
                         +\frac{rM(rMT)^{1-b}}{b-1},\\
    \frac{D_0}{S_T}
       &\leq\alpha\frac{(1-r)r^{1-b}\kappa^{1-b}}{(b-1)(2-b)}
                         +\alpha^b\frac{r^{2-b}}{b-1}\\
       &\leq\alpha\left[
           \frac{(1-r)r^{1-b}\kappa^{1-b}}{(b-1)(2-b)}
                         +\frac{r^{2-b}}{b-1}\right]
        =\alpha K_b(r).
\end{align*}
The last step uses $0<\alpha<1$ and $b>1$, which imply
$\alpha^b=\alpha\alpha^{b-1}\leq\alpha$.

If $b_0>b$, \Cref{eq:lr-variance-rates} applies with $\beta=b_0$,
$g=g_0$, $H=T$, and $W^\circ=\kappa\alpha T$. Its constants can
depend on the fixed parameters, including $\alpha$ and $\kappa$, but
not on $T$. To see the normalization explicitly, $1/S_T$ is a fixed
positive multiple of $T^{b-1}$, and hence
\begin{equation*}
    \begin{array}{c|c|c}
        \text{range of }b_0 & D_0 & D_0/S_T\\ \hline
        b<b_0<2 & O(T^{1-b_0}) & O(T^{b-b_0})\\
        b_0=2 & O(T^{-1}\log T) & O(T^{b-2}\log T)\\
        b_0>2 & O(T^{-1}) & O(T^{b-2})
    \end{array}
\end{equation*}
In the first row, $b-b_0<0$; in the last two rows, $b-2<0$.
Moreover, $\log T/T^{2-b}\to0$ because $2-b>0$.
Thus every row gives $D_0/S_T\to0$.

Consequently, combining the bounds for $D_1$ and $D_0$ gives
\begin{equation*}
    \limsup_{T\to\infty}\frac{D_1+(V_0/V_1)D_0}{S_T}
       \leq(1+\chi)\alpha K_b(r).
\end{equation*}
We now combine this estimate
with \Cref{eq:lr-residual-gap-limit,eq:lr-residual-gap-bound} and get
\begin{align*}
    \liminf_{T\to\infty}
    \frac{\min\{P(L),P(T)\}-P(W^\circ)-D_1-(V_0/V_1)D_0}{S_T}
    &\geq\delta_b(r)-\alpha[H_b(r)+(1+\chi)K_b(r)]\\
    &>\frac{\delta_b(r)}{2}>0,
\end{align*}
because $\alpha<\alpha_c$. This finishes the second part of the proof.

\paragraph{Step 3: $J_T(W)$ cannot attain its minimum at $W=0$.}

We prove that $J_T(W)$ cannot attain its minimum at $W=0$ when $T$ is sufficiently
large by showing that $J_T(W^\circ) < J_T(0)$ for sufficiently large $T$. Recall that by \Cref{eq:lr-terminal-loss}, we have
\begin{align*}
    J_T(W)
    =C_0+{}&\frac{B_0}{[K_0/2+A_W(T)]^{a_0}}
       +\frac{B_1}{[K_1/2+A_W(L)]^{a_1}} \notag\\
      &+V_0 I_{b_0,g_0}(T,W)
       +V_1 I_{b_1,g_1}(L,W)+V_1 P(W),
\end{align*}
and we analyze the last five terms separately.
First, we evaluate $I_{\beta,g}(H,W^\circ)$ and $I_{\beta,g}(H,0)$. Here, $H$ can be either $L$ or $T$, and $\beta$ can be either $b_1$ or $b_0$, respectively.
Recall that $W^\circ=\kappa\alpha T$, and for $H=L$ or $H=T$, we have $H\geq\alpha T>\kappa\alpha T=W^\circ$.
So for all $0<W\leq H$:
\begin{align*}
    I_{\beta,g}(H,W)
       =\int_0^W
       \frac{(m+\Delta u/W)^2}
            {[g+mu+\Delta u^2/(2W)]^\beta}\d u
            +\int_W^H\frac{M^2}{[g+Mu-\Delta W/2]^\beta}\d u.
\end{align*}
When $0\leq u \leq W$, $A_W'(u)=m+\Delta u/W$, so
\begin{align*}
    \frac{\d}{\d u}\left[(m+\Delta u/W)
                              [g+A_W(u)]^{1-\beta}\right]
       =\frac{\Delta}{W}[g+A_W(u)]^{1-\beta}
       -(\beta-1)\frac{(m+\Delta u/W)^2}
                              {[g+A_W(u)]^\beta}.
\end{align*}
Therefore, by integration by parts, we have
\begin{align*}
    &I_{\beta,g}(H,W)
       =\frac{mg^{1-\beta}-M[g+(M+m)W/2]^{1-\beta}}{\beta-1}+\frac{\Delta}{(\beta-1)W}
                         \int_0^W[g+A_W(u)]^{1-\beta}\d u\\
       &\qquad \qquad \qquad+\frac{M}{\beta-1}\left[
                  [g+(M+m)W/2]^{1-\beta}
                      -[g+MH-\Delta W/2]^{1-\beta}\right]\\
       &=\frac{mg^{1-\beta}}{\beta-1}
             -\frac{M[g+MH-\Delta W/2]^{1-\beta}}{\beta-1}+\frac{\Delta}{(\beta-1)W}
                 \int_0^W[g+mu+\Delta u^2/(2W)]^{1-\beta}\d u.
\end{align*}
Because $1-\beta<0$, its integral satisfies
\begin{align*}
    0&\leq\frac{1}{W}
          \int_0^W[g+mu+\Delta u^2/(2W)]^{1-\beta}\d u
       \leq\frac{1}{W}\int_0^W(g+mu)^{1-\beta}\d u\\
       &\qquad\leq
       \begin{cases}
           \displaystyle\frac{(g+mW)^{2-\beta}}{m(2-\beta)W}=O(W^{1-\beta}),
                                      &1<\beta<2,\\[6pt]
           \displaystyle\frac{\log(1+mW/g)}{mW}=O(W^{-1}\log W),&\beta=2,\\[6pt]
           \displaystyle\frac{g^{2-\beta}}{m(\beta-2)W}=O(W^{-1}),&\beta>2.
       \end{cases}
\end{align*}
We take $T\to\infty$, and consequently $W=W^\circ=\kappa\alpha T\to\infty$, $H\geq \alpha T\to\infty$. We obtain
\begin{equation*}
    I_{\beta,g}(H,W^\circ)\longrightarrow
                   \frac{mg^{1-\beta}}{\beta-1}.
\end{equation*}
For the constant learning rate $\eta_0(t)=M$, direct integration gives
\begin{equation*}
    I_{\beta,g}(H,0)
       =\int_0^H\frac{M^2}{(g+Mu)^\beta}\d u
       =\frac{M}{\beta-1}
          \left[g^{1-\beta}-(g+MH)^{1-\beta}\right].
\end{equation*}
Consequently,
$I_{\beta,g}(H,0)\to Mg^{1-\beta}/(\beta-1)$ when $T\to\infty$.

For both $W=0$ and $W=W^\circ$, and for all $H\geq\alpha T$, $A_W(H)\geq mH\geq m\alpha T$.
Hence for both $l=0$ and $l=1$, and for all $H\geq\alpha T$, we have
\begin{equation*}
    0\leq\frac{B_l}{[K_l/2+A_W(H)]^{a_l}}
       \leq\frac{B_l}{(m\alpha T)^{a_l}}\longrightarrow0, \quad T\to\infty.
\end{equation*}

We can bound $P(W)$ similarly:
\begin{equation*}
    0\leq P(W)=\frac{\displaystyle\int_L^T\eta_W(T-u)^2\d u}            {[g_1+A_W(L)]^{b}}\leq\frac{M^2T}{(m\alpha T)^b}
       =\frac{M^2}{(m\alpha)^b}T^{1-b}\longrightarrow0, \quad T\to\infty.
\end{equation*}
Substituting these limits into
\Cref{eq:lr-loss-difference} gives
\begin{equation*}
    J_T(0)-J_T(W^\circ)
       \longrightarrow (M-m)\sum_{l=0}^1
           \frac{V_lg_l^{1-b_l}}{b_l-1}>0, \quad T\to\infty.
\end{equation*}
Therefore, there exists a threshold $T_0$ such that for all $T>T_0$, we have
\begin{equation*}
    J_T(0)-J_T(W^\circ)
       >\frac{M-m}{2}\sum_{l=0}^1
          \frac{V_lg_l^{1-b_l}}{b_l-1}>0, \quad T>T_0.
\end{equation*}
This concludes the proof.
\end{proof}

\end{document}